\documentclass[english,final]{elsarticle}
\usepackage[T1]{fontenc}
\usepackage[latin9]{inputenc}
\usepackage{color}
\usepackage{babel}
\usepackage{float}
\usepackage{amsmath}
\usepackage{graphicx}
\usepackage{setspace}
\usepackage[unicode=true,
 bookmarks=false,
 breaklinks=false,pdfborder={0 0 1},backref=section,colorlinks=true]
 {hyperref}
\hypersetup{
 linkcolor=blue,filecolor=magenta,urlcolor=cyan}

\makeatletter

\providecommand{\tabularnewline}{\\}
\newcommand{\lyxdot}{.}

\floatstyle{ruled}
\newfloat{algorithm}{tbp}{loa}
\providecommand{\algorithmname}{Algorithm}
\floatname{algorithm}{\protect\algorithmname}

\usepackage{algorithm}
\usepackage{algpseudocode}
\usepackage{longtable}
\usepackage{pdflscape}
\usepackage{natbib}

\usepackage{babel}
\usepackage{listings}
\addto\captionsbritish{\renewcommand{\algorithmname}{Algorithm}}
\addto\captionsbritish{}
\addto\captionsenglish{}

\@ifundefined{showcaptionsetup}{}{%
 \PassOptionsToPackage{caption=false}{subfig}}
\usepackage{subfig}
\makeatother

\usepackage{listings}

\begin{document}
% draft_revision = 21
% draft state = major_revision_6
\begin{frontmatter}{} 

\title{Kalman Filter-based Heuristic Ensemble (KFHE): A new perspective
on multi-class ensemble classification using Kalman filters}

\author[]{Arjun Pakrashi\corref{cor1}}

\ead{arjun.pakrashi@ucdconnect.ie, arjun.pakrashi@insight-centre.org}

\author[]{Brian Mac\ Namee}

\ead{brian.macnamee@ucd.ie}

\cortext[cor1]{Corresponding author}

\address{Insight Centre for Data Analytics, School of Computer Science, \\
 University College Dublin, Ireland}
\begin{abstract}
This paper introduces
a new perspective on multi-class ensemble classification that considers
training an ensemble as a state estimation problem. The new
perspective considers the final ensemble classifier model
as a static state, which can be estimated using a Kalman filter that
combines noisy estimates made by individual classifier models. A new
algorithm based on this perspective, the Kalman Filter-based Heuristic
Ensemble (KFHE), is also presented in this paper which shows the practical
applicability of the new perspective. Experiments performed on $30$
datasets compare KFHE with state-of-the-art multi-class
ensemble classification algorithms and show the potential and effectiveness
of the new perspective and algorithm. Existing ensemble approaches trade off classification accuracy against robustness
to class label noise, but KFHE is shown to be
significantly better or at least as good as the state-of-the-art algorithms
for datasets both with and without class label noise.

\end{abstract}
\begin{keyword}
Classification \sep Multi-class \sep Ensemble \sep Kalman filter
\sep Heuristic 
\end{keyword}
\end{frontmatter}{}

\section{Introduction}

An ensemble classification model is composed of multiple individual
base classifiers, also known as component classifiers, the outputs
of which are aggregated together into a single prediction. The classification
accuracy of an ensemble model can be expected to exceed that of any
of its individual base classifiers. The main motivation
behind ensemble techniques is that a committee of experts working
together on a problem are more likely to accurately solve it than
a single expert working alone \citep{kelleher2015fundamentals}. Although
many existing ensemble techniques (e.g. \citep{Breiman1996,Friedman00greedyfunction,hastie2009multi,zhu2006multi})
have been repeatedly shown in benchmark experiments to be effective
(see \citep{Narassiguin2016,Opitz:1999:PEM:3013545.3013549}),
current approaches still have limitations. For example, methods based
on \emph{bagging}, although robust, may not lead to models as accurate
as those learned by more sophisticated methods such as those based on \emph{boosting} \citep{Narassiguin2016}. Methods based
on boosting, however, are sensitive to class-label noise and
the presence of outliers in training datasets \citep{Dietterich2000}.

To address the limitations of current multi-class classification ensemble
algorithms, this paper presents a new
perspective on ensemble model training, framing it as a state estimation
problem that can be solved using a Kalman filter \citep{kalman1960,maybeck1982stochastic}. Although Kalman filters
are most commonly used to solve problems associated 
with time series data, this is not the case in this work. Rather,
this work exploits the data fusion property of the Kalman filter to
combine individual multi-class component classifier models to construct
an ensemble.

The new perspective views the ensemble model to be trained as an unknown
static state to be estimated. A Kalman filter can be used to estimate
an unknown static state by combining multiple uncertain measurements
of the state. This exploits the data fusion property of the Kalman
filter. In the new perspective the measurements are the single component
classifiers in the ensemble, and the uncertainties of these measurements
are based on the classification errors of the single component classifiers.
The Kalman filter is used to combine the component classifier models
into an overall ensemble model. This new perspective on ensemble training
provides a framework within which different algorithms can be formulated.
This paper describes one such new algorithm, the Kalman Filter-based
Heuristic Ensemble (KFHE). In an evaluation experiment KFHE is shown to out-perform methods based
on boosting while maintaining the robustness of methods based on bagging. The contributions of this paper are: 

\begin{enumerate}
\item A new perspective on training multi-class ensemble classifiers, which
views it as a state estimation problem and solves it using a Kalman filter \citep{kalman1960,maybeck1982stochastic}. 
\item A new multi-class ensemble classification algorithm, the Kalman Filter-based
Heuristic Ensemble (KFHE).
\item Extensive experiments comparing KFHE with the state-of-the-art ensemble
algorithms that demonstrate the effectiveness of KFHE in both scenarios
of noise free and noisy class-labels.
\end{enumerate}

The remainder of this paper is structured as follows. Section \ref{sec:Background}
discusses previous work on multi-class ensemble classification algorithms
and provides a brief introduction to the Kalman filter. Section \ref{sec:The-new-perspective}
introduces the new Kalman filter-based perspective on building multi-class
classification ensembles. The Kalman Filter-based Heuristic
Ensemble (KFHE) method based on this perspective is described
in Section \ref{sec:Proposed-Method}. The setup of an 
experiment to evaluate the performance of KFHE and the comparison method to state-of-the-art 
approaches on a selection of datasets is described in Section
\ref{sec:Experiment}, and a detailed discussion of the results of this experiment is
presented in Section \ref{sec:Results}. Finally, Section \ref{sec:Discussion-and-Conclusion}
reflects on the newly proposed perspective and explores directions for
future work.

\section{Background\label{sec:Background}}

This section first reviews existing multi-class ensemble classification
methods. Relevant aspects of the Kalman filter approach for state
estimation, which serve as a basis for the
explanation of KFHE, are then introduced.

\subsection{Ensemble methods\label{subsec:Ensemble-methods}}

The advent of ensemble approaches in machine learning
in the early 1990s was due mainly to works by Hansen and Salamon \citep{Hansen:1990:NNE:628297.628429},
and Schapire \citep{Schapire1990}. Hansen and Salamon \citep{Hansen:1990:NNE:628297.628429}
showed that multiple classifiers could be combined to achieve better
 performance than any individual classifier.
Schapire \citep{Schapire1990} proved that the \emph{learnability}
of strong learners and weak learners are equivalent, and then showed
how to boost weak learners to become strong learners. Since then many alternative
and improved approaches to build ensembles have been introduced. Ensemble methods can still,
however, be categorised into three fundamental types:
\emph{bagging}, \emph{boosting}, and \emph{stacking}.

\emph{Bagging} \citep{Breiman1996} or bootstrap aggregation, trains
several base classifiers on bootstrap samples of a training dataset
and combines the outputs of these base classifiers using simple aggregation
such as majority voting. Training models on different samples of the
training set introduces diversity into the ensemble, which is key
to making ensembles work effectively. \emph{UnderBagging} \citep{UnderBagging:Barandela2003}
is a variation of bagging addressing imbalanced datasets that performs
undersampling before every bagging iteration, but also keeps all minority
class instances in every iteration. The \emph{Random Forest} \citep{Breiman2001rf}
is an extension to bagging in which base classifiers (usually decision
trees) are trained using a bootstrap sample of the dataset that has
also been reduced to only a small random sample of the input 
space. The \emph{Rotational Forest}
\citep{Rodriguez:2006:RFN:1159167.1159358} is another extension that attempts to build base classifiers that are simultaneously accurate and diverse. The input dataset is transformed by applying PCA \citep{hastie01statisticallearning}
on different subsets of the attributes of the dataset, and axis rotation is performed by combining the coefficient matrices found by PCA
for each subset. This is repeated multiple times. \emph{Local Linear
Forests} modify random forests by considering random forests as an
adaptive kernel method and combining
it with local linear regression \citep{friedberg2018local}.
% The \emph{Error Correcting Output Codes} (ECOC) approach \citep{Dietterich:1995:SML:1622826.1622834}
% which can be considered to be based on bagging, although it is very
% different to the previously described approaches. ECOC replaces each
% of the unique target classes in a dataset by an error correcting code,
% and then learns a binary classifier for each bit of this code. Unlike
% in other approaches, in ECOC each base classifier learns a different
% function, which may reduce the correlations between the the base classifiers
% and make their combination more effective.

\emph{Boosting} \citep{zhu2006multi} approaches iteratively learn
component classifiers such that each one specialises on specific types
of training examples. Each component classifier is trained using a
weighted sample from a training dataset such that at each iteration
the ensemble emphasises training examples that were misclassified
in the previous iteration.
%A good theoretical overview of boosting can be found in \citep{Schapire:2012:BFA:2207821}.
Since the introduction of the original boosting algorithm, \emph{AdaBoost}
\citep{freund1995desicion}, several new approaches to boosting have
been proposed. In \emph{LogitBoost} \citep{friedman2000additive},
the logistic loss function is minimised while combining the sub-classifiers
in a binary classification context. A linear programming approach
to boosting, \emph{LPBoost} \citep{demiriz2002linear},
was shown to be competitive with AdaBoost. This algorithm minimises
the misclassification error and maximises the soft margin in the feature
space generated by the predictions of the weak hypothesis components
of the ensemble. A multi-class modification for binary class AdaBoost
was introduced in \citep{freund1995desicion}, and an improvement
of it was proposed in \citep{hastie2009multi}. \emph{RotBoost} \citep{ZHANG20081524}
is a direct extension of the rotational forest approach \citep{Rodriguez:2006:RFN:1159167.1159358}
to include boosting. The \emph{Gradient Boosting Machine} (GBM) \citep{Friedman00greedyfunction}
is a sequential tree based ensemble method, where each tree corrects
the errors of the previously trained trees. \emph{Stochastic Gradient
Boosting Machine} (S-GBM) \citep{FRIEDMAN2002367} improves GBM by
training the component trees on bootstrap samples.

AdaBoost is sensitive to noisy class labels and performs poorly as
the level of noise increases \citep{Freund2001}. This is mainly due to the exponential
loss function AdaBoost uses to optimise the ensemble. If a training datapoint
has noisy class-labels AdaBoost will increase its weight for the next
iteration and keep on increasing the weight of the datapoint in a vain
attempt to classify it correctly. Therefore,
given enough such noisy class-labelled datapoints AdaBoost can learn
classifiers with poor generalisation ability. Although
the performance of bagging decreases in the presence of class-label
noise, it does not do so as severely as it does with AdaBoost \citep{Dietterich2000}.

To overcome this problem with noisy class-labeled datasets, \emph{MadaBoost}
\citep{Domingo:2000:MMA:648299.755176_madaboost} was proposed. MadaBoost
changes the standard AdaBoost weight update rule by capping the maximum
value for the weight of a datapoint to be $1$. Similarly \emph{FilterBoost}
\citep{NIPS2007_3321filterboost} optimises the log loss function,
leading to a weight update rule which caps the weight upper bound
of a datapoint to $1$ using a smooth function. \emph{BrownBoost}
\citep{Freund2001} and \emph{Noise Detection Based AdaBoost} \emph{(ND\_AdaBoost)}
\citep{CAO20124451} make AdaBoost more robust to class label noise
by explicitly identifying noisy examples and ignoring them. \emph{Robust
Multi-class AdaBoost} (\emph{Rob\_MulAda}) \citep{SUN201687} is an
extension to \emph{ND\_AdaBoost} for multi-class classification. \emph{Vote-Boosting}
\citep{SABZEVARI2018119}, decides the weights of each datapoint while
training based on the disagreement of the predictions of the component
classifiers that exist at each iteration. For lower levels of class-label
noise, the datapoints with higher disagreement rates are emphasised.
Whereas for higher levels of class-label noise, datapoints which agree
among different component classifiers are highlighted, in an attempt
to achieve robustness to class-label noise.
%AdaBoost was also modified so that it can perform better on datasets with imbalanced classes by performing oversampling \citep{10.1007/978-3-540-39804-2_12} and undersampling of the datapoints \citep{journals/tsmc/SeiffertKHN10,EUS:doi:10.1162/evco.2009.17.3.275,EasyEnsemble:4717268}. Also, a few algorithms to overcome class-imbalance problems based on different sampling methods and using multiple types of weak learners were proposed in \citep{MEBoost:DBLP:journals/corr/abs-1712-06658,DBLP:journals/corr/WangWZMX17}.
A comprehensive review and analysis of the different boosting variations
can be found in \citep{zhou2012ensemble}.

\emph{Stacking} \citep{WOLPERT1992241,Ting97stackedgeneralization:}
is a two stage process in which the outputs of a collection of first
stage base classifiers are combined by a second stage classifier to
produce a final output. %First, different types of classification algorithms at level one are used to generate a second level dataset. The second level dataset consists of the predictions of the classifiers from the first level. Next, the second level dataset is used to train another model with the actual class memberships as the targets. In this way, the first level has the predictions from different level one classifiers for a datapoint, and the second level takes these predictions as input features for the second level classifier to predict the final class membership. Effectively this is the combination of the classifiers in the first level. 
Seewald \citep{Seewald:2002:MSB:645531.656165}, empirically showed
that the extension to stacking by Ting and Witten \citep{Ting97stackedgeneralization:}
does not perform well in the multi-class context, and proposed \emph{StackingC}
to overcome this drawback. In \citep{MENAHEM20094097_troika} the
weaknesses of StackingC were highlighted and were shown to occur due
to increasingly skewed class distributions because of the binarisation
of the multi-class problem. Next, a three layered improved stacked
method for multi-class classification, \emph{Troika} \citep{MENAHEM20094097_troika},
was proposed. The stacking approach to building ensembles has received
much less research attention than approaches based on bagging and
boosting.

\subsection{The Kalman filter\label{subsec:Kalman-filter}}

The Kalman filter \citep{kalman1960} is a mathematical tool for
stochastic estimation of the state of a linear system based on noisy measurements. Let there be a system which evolves linearly over time,
and assume that the state of the system, which is unobservable, has
to be estimated at each time step, $t$. The state may be estimated
in two ways. First, a linear model, which is used to update the state of the
system from step $t$ to step $t+1$, can be used to get an \emph{a
priori} estimate of the state. This estimate will have a degree of
uncertainty as the linear model is unlikely to fully capture the true
nature of the system. Estimating the state using this type of linear
model is commonly known as a \emph{time update} step. Second, an
external sensor can provide a state estimate. This estimate will also
have an associated uncertainty, referred to as \emph{measurement noise},
and introduced because of inaccuracies in the measurement process.

\begin{figure}
\centering{}\includegraphics[width=0.7\textwidth]{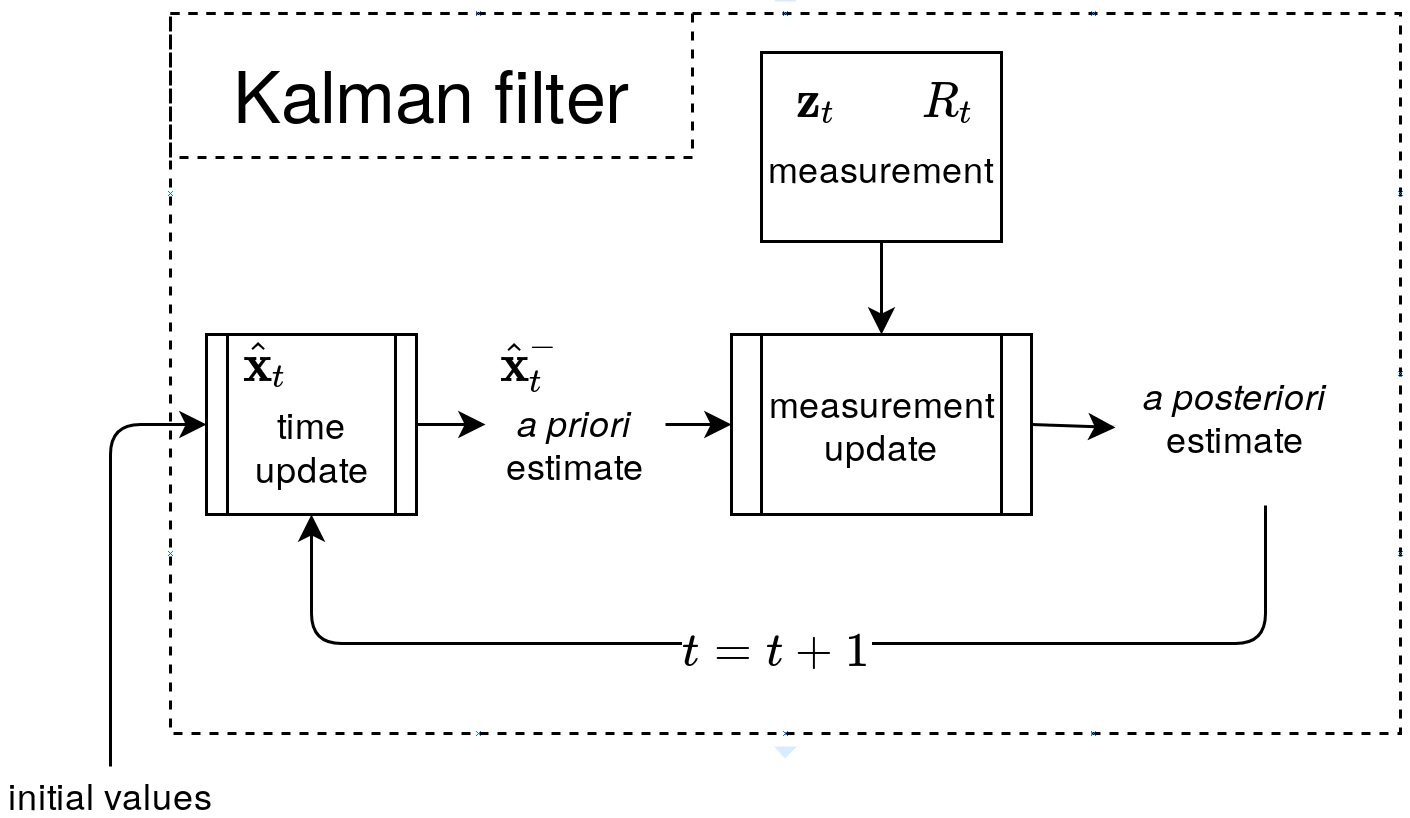} 
\centering{}\caption{A high-level illustration of a Kalman filter\label{fig:kf_high_level_diag}}
\end{figure}

Given these two state estimates, and their related uncertainties, the Kalman filter combines the \emph{a priori}
estimate and the measurement to generate an \emph{a posteriori} state
estimate, such that the uncertainty of the \emph{a posteriori} estimate
is minimised. This combination of a sensor measurement with an \emph{a
priori} estimate is commonly known as the \emph{measurement update}
step.  The process iterates using the \emph{a posteriori} estimate calculated
in a measurement update step as input to the time update step of the
next iteration. A high-level illustration of the Kalman filter is
shown in Figure \ref{fig:kf_high_level_diag}. More formally, the time update step
in a Kalman filter can be defined as:

\begin{equation}
\hat{\mathbf{x}}_{t}^{\text{{-}}}=A_{t}\hat{\mathbf{x}}_{t-1}+B_{t}\mathbf{u}_{t}\label{eq:time_update_state}
\end{equation}
\begin{equation}
P_{t}^{\text{{\ensuremath{^{\text{{-}}}}}}}=A_{t}P_{t-1}A_{t}^{T}+Q_{t}\label{eq:time_update_cov}
\end{equation}

\noindent where: 
\begin{itemize}
\item $\mathbf{\hat{x}}_{t}^{\text{{-}}}$ is the \emph{a priori} estimate
at step $t$ when the knowledge of the state in the previous step
$(t-1)$ is given 
\item $\mathbf{\hat{x}}_{t-1}$ is the \emph{a posteriori }estimate at step
$(t-1)$, which is found through combining the \emph{a priori} estimate
and the measurement 
\item $A_{t}$ is the state transition matrix which defines the linear relationship
between $\hat{\mathbf{x}}_{t-1}$ and $\hat{\mathbf{x}}_{t}$ 
\item $\mathbf{u}_{t}$ is the control input vector, containing inputs which
changes the state based on some external effect 
\item $B_{t}$ is the control input matrix applied to the
control input vector 
\item $P_{t}^{\text{\text{{\ensuremath{^{\text{{-}}}}}}}}$ is the covariance
matrix representing the uncertainty of the \emph{a priori }estimate 
\item $P_{t-1}$ is the covariance matrix representing the uncertainty of
the \emph{a posteriori} estimate at step $t-1$ 
\item $Q_{t}$ is the process noise covariance matrix, induced during the
linear update
\end{itemize}
Similarly, the measurement update step can be defined as:

\begin{equation}
\hat{\mathbf{x}}_{t}=\hat{\mathbf{x}}_{t}^{\text{{-}}}+K_{t}(\mathbf{z}_{t}-H_{t}\hat{\mathbf{x}}_{t}^{\text{{-}}})\label{eq:measure_update_state}
\end{equation}
\begin{equation}
K_{t}=P_{t}^{\text{\text{{\ensuremath{^{\text{-}}}}}}}H_{t}^{T}(H_{t}P_{t}^{\text{\text{{\ensuremath{^{\text{-}}}}}}}H_{t}^{T}+R_{t})^{-1}\label{eq:measure_update_gain}
\end{equation}
\begin{equation}
P_{t}=(I-K_{t}H_{t})P_{t}^{\text{\text{{\ensuremath{^{\text{-}}}}}}}\label{eq:measure_update_cov}
\end{equation}

\noindent where 
\begin{itemize}
\item $\mathbf{z}_{t}$ is the measurement of the system at time $t$ 
\item $R_{t}$ is the measurement noise covariance matrix 
\item $H_{t}$ is a transformation matrix relating the state space
to the measurement space (when they are the same space, then $H_{t}$
can be the identity matrix) 
\item $K_{t}$ is the \emph{Kalman gain} which drives the weighted combination of the measurement and
the \emph{a priori} state
\item $I$ indicates the identity matrix 
\end{itemize}
The Kalman filter iterates through the time update and the measurement
update steps. In this work time steps are considered equidistant and
discreet. Hence, from this point, ``time step'' and ``iteration''
will be used interchangeably. At $t=0$, an initial estimate for $\hat{\mathbf{x}}_{0}$
and $P_{0}$ is used. Next, the time update step is performed using
Eq. \eqref{eq:time_update_state} and \eqref{eq:time_update_cov}
to get $\hat{\mathbf{x}}_{t}^{\text{{-}}}$ and $P_{t}^{\text{\text{{\ensuremath{^{\text{-}}}}}}}$
respectively. The measurement $\mathbf{z}_{t}$ and its related uncertainty
$R_{t}$ are then obtained from a sensor or other appropriate source.
These are combined with the a priori estimate using the measurement
update step to find $\hat{\mathbf{x}}_{t}$ and $P_{t}$ using
Eq \eqref{eq:measure_update_gain}, \eqref{eq:measure_update_state}
and \eqref{eq:measure_update_cov}, which are then used in the next
iteration $(t+1)$. A detailed explanation of Kalman
filters can be found in \citep{kalman1960,maybeck1982stochastic},
and an intuitive description in \citep{Welch:1995:IKF:897831}.

It should be emphasised here that, although a Kalman filter is used and Kalman filters are most commonly used with time series data, the proposed method \emph{does not} perform time series prediction. Rather the focus is on multi-class
classification and the data fusion property of the
Kalman filter is used to combine the individual multi-class classifiers in the ensemble. Also, the term ``ensemble''
in this work relates to multi-class ensemble classifiers, and should not be confused with Ensemble Kalman Filters (EnKF) \citep{evensen2003ensemble}.

Apart from their applications to time series data and sensor fusion, Kalman filters have been used previously in a small
number of supervised and unsupervised machine learning applications.
For example, \citep{SISWANTORO2016112}, improves the predictions
of a neural network using Kalman filters, although this method is
essentially a post-processing of the results of a neural network output. Properties of a Kalman filter were used in
combination with heuristics in population-based metaheuristic optimisation
algorithms \citep{TOSCANO20101955,Monson04thekalman}, and in an unsupervised
context in clustering \citep{PAKRASHI2016704,pakrashikhka_10.1007/978-3-319-20294-5_39}.
To the best of the authors' knowledge this is the first
application of Kalman filters to training multi-class ensemble classifiers.

%To the best of the knowledge of the authors', however, framing the ensemble training process as a state estimation exercise and performing this using a Kalman filter  has not been done before. The next section will introduce this new perspective on ensemble classification and a implementation of the perspective.

\section{Training multi-class ensemble classifiers using a Kalman filter\label{sec:The-new-perspective}}

This section introduces the new perspective on training multi-class
ensemble classifiers using a Kalman filter. First, a toy
example of static state estimation using a Kalman filter is presented, and then the new perspective
is described.

\subsection{A static state estimation problem: Estimating voltage level of a
battery\label{subsec:A-static-estimation}}

Imagine that the exact voltage of a DC battery (which should remain
constant) is unknown and needs to be estimated. A sensor is available
to measure the voltage level of the battery. The measurements made
by this sensor are unfortunately noisy, but the uncertainty associated
with the measurements is known. This is a simple example of a static
state estimation problem that can be solved by taking multiple noisy sensor measurements
of the battery's voltage, and combining these
into a single accurate estimate using a Kalman filter. 

The Kalman filter can be applied in this scenario as follows. As it
is known that the voltage of the battery does not change the state
transition matrix, $A_{t}$, in Eq. \eqref{eq:time_update_state}
is the identity matrix; the control input matrix, $B_{t}$, in Eq.
\eqref{eq:time_update_cov} is non-existent; and the process noise
covariance matrix, $Q_{t}$, in Eq. \eqref{eq:time_update_cov} is
considered to be zero. The voltage read by the sensor at a particular measurement, and the related uncertainty of the value
due to the limited accuracy of the sensor, give $\mathbf{z}_{t}$ and
$R_{t}$ in Eq. \eqref{eq:measure_update_state} and \eqref{eq:measure_update_gain}
respectively. Given this information, the Kalman filter time update
and measurement update steps can be performed to combine the current
estimated voltage, $\hat{\mathbf{x}}_{t}^{\text{{-}}}$, and the measurement,
$\mathbf{z}_{t}$, to get a new and better estimate of the voltage.
The process can be repeated, where at each step, a new voltage measurement
from the sensor is received, which is then combined with the current
estimated voltage value using the measurement update step.

Note that, after $t$ iterations, the estimated voltage
is a combination of the sensor output values, where the Kalman gain,
$K_{t}$ in Eq (\ref{eq:measure_update_gain}) and (\ref{eq:measure_update_cov}),
is controls the influence of each measurement in the combination.
Therefore, after $t$ iterations, the estimated voltage, $\hat{\mathbf{x}}_{t}$,
can be seen as an ensemble of the values received from the sensor,
which are optimally combined. This same idea can be applied to combine
noisy base classifiers into a more accurate ensemble model.

\subsection{Combining multi-class classifiers using the Kalman filter\label{subsec:Combining-multi-class-classifier}}

A machine learning algorithm learns a hypothesis for a specific problem.
Assume that all possible hypotheses make a \textit{hypothesis
space}\footnote{The term hypothesis and hypothesis space is used to introduce the
high level idea in connection with \citep{DietterichHSpace}, but
the terms model and model space will be used synonomoulsy throughout
this text.}, as described in \citep{DietterichHSpace}. Any point in the hypothesis
space represents one hypothesis. For a specific problem, there is at least
one ideal hypothesis within this hypothesis space which the learning
algorithm tries to reach. Different hypotheses within the hypothesis
space differ in their trainable parameters, and the machine learning
algorithm modifies these parameters. Therefore, the training process
can be seen as a search through the hypothesis space toward the ideal
hypothesis.

The perspective presented in this paper views the ideal hypothesis
as the static \textit{state} to be estimated, and the hypothesis space
as a \textit{state space}. When an individual component classifier,
$h_{t}$, is trained, it can be seen as a point in the hypothesis
space. Here, $h_{t}$ can be considered as an attempt to measure the
ideal state with a related uncertainty indicated by the training error
of $h_{t}$. The Kalman filter can be
used to estimate the ideal state by combining these multiple noisy measurements. The combination of these noisy measurements leads to
an estimation of the state that is expected to be more accurate
than the individual measurements, and so an ensemble classification model that is more accurate than its component classifiers. 

\begin{figure}[!t]
\centering{}\includegraphics[width=1\textwidth]{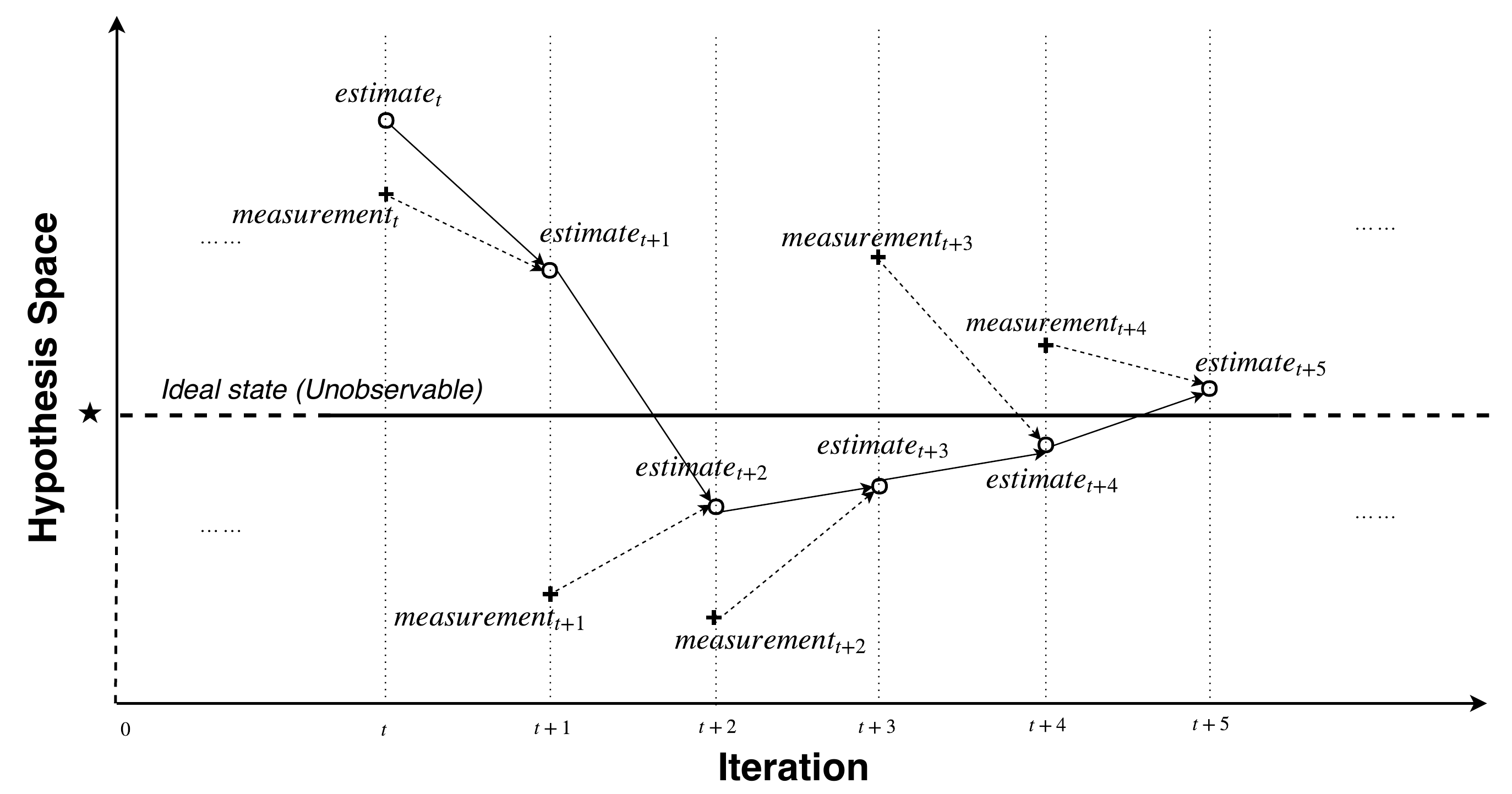}
\caption{An illustration of the state estimation perspective on training a
classification ensemble. The vertical axis indicates the hypothesis
space and the horizontal axis represents iterations. Any point along
the vertical axis represents a hypothesis. The horizontal line marked
with the star indicates the ideal hypothesis. The circles are the
estimates, and the plus symbols are the measurements of the hypothesis.
The absolute distance of the circles and the plus symbols from the
bold horizontal line indicating the ideal hypotheses represents the
uncertainty of the estimates. The target of the training algorithm
is to navigate through the hypothesis space to get as close to the ideal hypothesis as possible. \label{fig:h_space_diag}}
\end{figure}

This is illustrated in Figure \ref{fig:h_space_diag}.
The vertical axis is an abstract representation of the \textit{hypothesis
space} with each point along this axis representing a possible hypothesis.
The star symbol on the vertical axis indicates the ideal hypothesis
for a specific classification problem. The horizontal axis in Figure
\ref{fig:h_space_diag} represents training iterations proceeding
from left to right. The circles are the estimates of the hypothesis
at a time step $t$ (the combination of all models added to the ensemble to this point
in the training process), and the plus symbols represent the measurement
of the hypothesis at a time step $t$ (the last model added to the
ensemble). %The absolute distance from the circles and the plus symbols represent the uncertainty of these states. 
 The dashed and solid arrows connecting the state estimates indicate
the combination of the measurement and the \emph{a priori} estimate
respectively. The goal of the process is to reach a hypothesis as
close as possible to the ideal hypothesis (indicated by the horizontal
line marked with a star) by combining multiple individual hypotheses
using a Kalman filter.

To help with understanding the new perspective, the Kalman filter-based
approach to ensemble training can be directly mapped back to the DC
battery voltage estimation example described in Section \ref{subsec:A-static-estimation}.
The ensemble model capturing the ideal hypothesis is equivalent
to the actual voltage level of the DC battery. An individual component
classifier, $h_{t}$, is analogous to the output from the voltage
sensor. The classification error of the model $h_{t}$ maps to the uncertainty related to the voltage sensor measurements. Just as the
estimated voltage after $t$ iterations can be thought as an \emph{ensemble
of sensor measurements} in the battery voltage estimation case; the trained individual classifiers, $h_{t}$ combined using the Kalman filter leads to an \emph{ensemble of classifier
models}. 
%Therefore, the vertical axis in Figure \ref{fig:h_space_diag} in the case of the toy example in Section \ref{subsec:A-static-estimation} would indicate voltage, and the star symbol will indicate the actual voltage of the battery, which is unknown but needs to be estimated.

\section{Kalman Filter-based Heuristic Ensemble (KFHE)\label{sec:Proposed-Method}}

This section provides a detailed description of the Kalman Filter-based
Heuristic Ensemble (KFHE) algorithm, based on the new perspective
proposed in Section \ref{sec:The-new-perspective}. First,
Section \ref{subsec:Implementation-overview} presents an overview
of the algorithm and connects the high-level concepts from Section
\ref{sec:The-new-perspective}. Sections \ref{subsec:The-model-Kalman},
\ref{subsec:The-weight-Kalman} and \ref{subsec:Making-predictions-using}
then discuss the details of the algorithm.

\subsection{Algorithm overview\label{subsec:Implementation-overview}}

In KFHE the Kalman filter used to estimate an ensemble classifier, as described in Section \ref{sec:The-new-perspective}, is referred to as the \emph{model Kalman filter}, abbreviated to \emph{kf-m}.
To implement \emph{kf-m}, the following questions must be answered: 

\begin{enumerate}
\item What should constitute a state?
\item How should the time update step be defined?
\item What should constitute a measurement?
\item How should measurement uncertainty be evaluated?
\end{enumerate}

\begin{figure}
\includegraphics[width=1\textwidth]{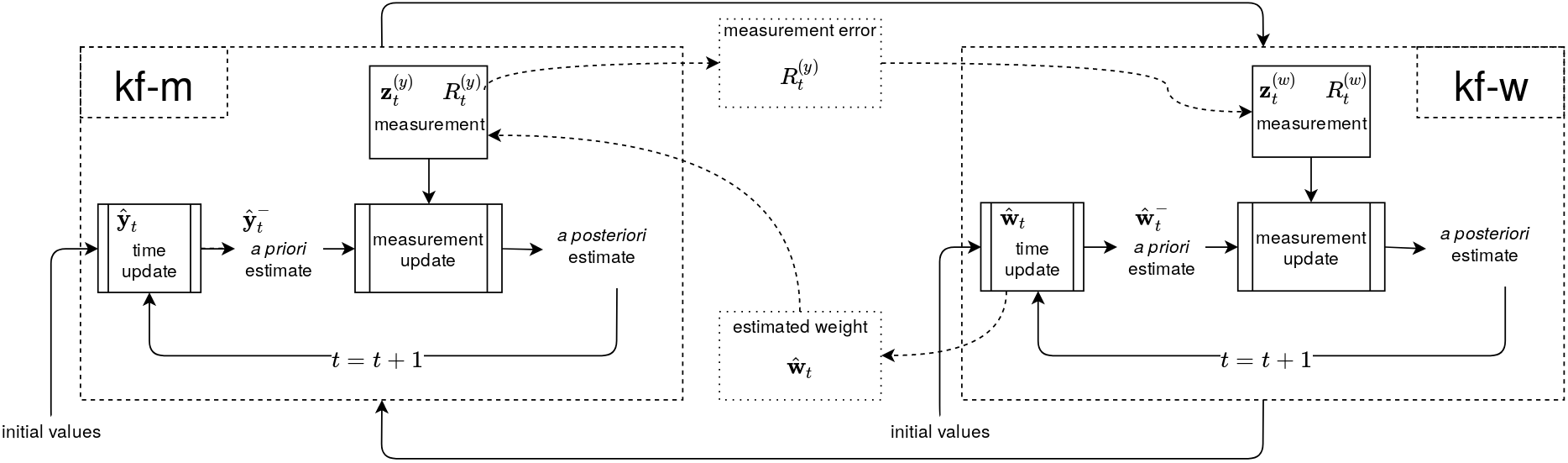}
\caption{Overall dataflow between \emph{kf-m} and \emph{kf-w} \label{fig:Overall-dataflow-between}}
\end{figure}

The \emph{kf-m} state estimates are essentially the trained 
component classifiers. A model specification (for example the
rules encoded in a decision tree or weight values in neural network)
cannot be used directly as a state within the Kalman filter framework.
Instead the predictions made by a component classifier for the instances
in the training dataset are used as the representation of the state,
as shown in Figure \ref{fig:The-representation-of}. This allows states
to be combined using the equations in Section \ref{subsec:Kalman-filter}.
This representation is explained in detail in Section \ref{subsec:The-model-Kalman}.

Heuristics are used to address the remaining
questions. The time update step is implemented as the identity function,
as it can be assumed that the ideal state is static and does not change
over time (as indicated by the horizontal line in Figure \ref{fig:h_space_diag}).
The measurement is a function of the output of the multi-class classifier trained at the $t$th iteration. This model is trained using a weighted sample from the overall training
dataset. The classification error of the model trained at the $t$th iteration, measured against its predictions for the full training set, is used as the uncertainty of the measurement. 

A Kalman filter is then used to combine
a measurement, which is the classification model at step $t$ represented as shown in Figure \ref{fig:The-representation-of}, and
the \emph{a priori} estimate to get an \emph{a posteriori} estimate.
The \emph{a posteriori} state estimate at the $t$th iteration is
considered the ensemble classifier up to the $t$th iteration.
This \emph{a posteriori} estimate is used in the next iteration, and
the process continues until a stopping condition is met. As the uncertainties
of the estimates are represented as the classification errors, the
process continues towards estimating states expected to yield lower classification
errors.

The use of weighted samples from the training set to train component classifiers at each step of the \emph{kf-m} process gives rise to another question: how should the weights for the weighted sampling of the training dataset be decided? In KFHE the answer is through another Kalman filter, which is referred to as the \emph{weight Kalman filter} and abbreviated to \emph{kf-w}. The \emph{kf-w} Kalman filter works very similarly to \emph{kf-m}, but estimates sampling weights for the training dataset instead of the overall model state. This is described in detail in Section \ref{subsec:The-weight-Kalman}.

The interactions between the model Kalman filter,
\emph{kf-m}, and the weight Kalman filter, \emph{kf-w}, are illustrated in
Figure \ref{fig:Overall-dataflow-between}. Essentially \emph{kf-w}
provides weights for the measurement step in \emph{kf-m}, and
\emph{kf-m} provides measurement errors back to \emph{kf-w} for
its measurement step. The training
process is summarised in Algorithm \ref{alg:Train} and the following subsections describe the workings of \emph{kf-m}
and \emph{kf-w} in detail.

\subsection{The model Kalman filter: \emph{kf-m\label{subsec:The-model-Kalman}}}

The model Kalman filter, \emph{kf-m,} estimates the ensemble classifier
by combining component classifiers
%\footnote{When the Gaussian condition is relaxed, the Kalman filter can be shown to be the best minimum error variance filter among all linear unbiased filters \citep{maybeck1982stochastic}.}
into a single ensemble classification
model. This is a static estimation problem as the state
to be estimated, the ideal ensemble classifier, does not change over time. For this reason
the time update step for \emph{kf-m} is the identity function and
the \emph{a posteriori} estimate of iteration $t-1$ is directly transferred
to the \emph{a priori} estimate at iteration $t$.

\begin{figure}
\includegraphics[width=1\textwidth]{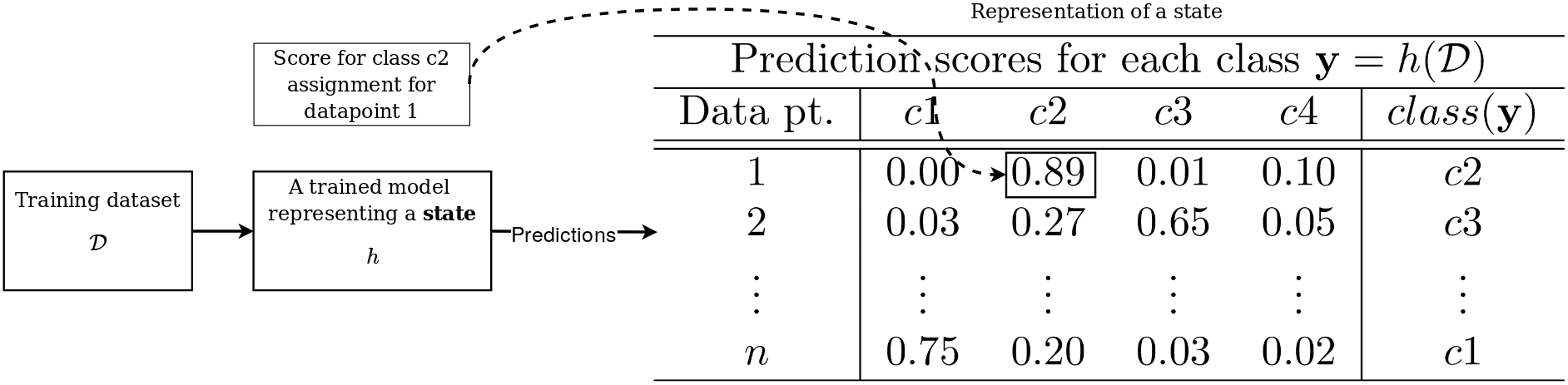} 
\caption{The representation of a state for \emph{kf-m}. \label{fig:The-representation-of}}
\end{figure}

% BMN to fix

The trained base classifiers of the ensemble (the measurements) or
the \emph{a posteriori} state estimate (ensemble classifier) themselves
are not directly usable as a state in the Kalman filter framework.
Therefore a proxy numerical representation is required to perform
the computations. The proxy representation of the state is shown in
Figure \ref{fig:The-representation-of} where each row represents a datapoint from the training set and the estimate scores for the classes for	
the corresponding datapoint. The class membership is determined by
taking the class with the maximum score, and this membership is expressed
as $class(\mathbf{y})$. For example in Figure \ref{fig:The-representation-of},
the first datapoint has the highest prediction score assigned to class-label
$c2$, and thus the first datapoint is considered as a member of class
$c2$. This representation of a model is used as the state in the
Kalman filter framework.

Hence, the time update equations for \emph{kf-m} are very simply defined
as:

\begin{equation}
\hat{\mathbf{y}}_{t}^{\text{{-}}}=\hat{\mathbf{y}}_{t-1}\label{eq:time_update_state_kfhe}
\end{equation}
\begin{equation}
P_{t}^{(y)\text{{\ensuremath{^{\text{{-}}}}}}}=P_{t-1}^{(y)}\label{eq:time_update_cov_kfhe}
\end{equation}

\noindent where 
\begin{itemize}
\item $\hat{\mathbf{y}}_{t-1}$ is the \emph{a posteriori} estimate from
the previous iteration and $\hat{\mathbf{y}}_{t}^{\text{{-}}}$ represents
the \emph{a priori} estimate at the present iteration. These are the
predictions of the ensemble model at the $t$th iteration in the representation
 shown in Figure \ref{fig:The-representation-of}. So, for example, $\mathbf{y}_{t}=[y_{t1};y_{t2};\ldots;y_{tn}]$ where $y_{ti}$ denotes the prediction for the $i^{th}$ datapoint; and each $y_{ti}$ is a vector of $c$ prediction scores where $c$ is the number of classes in the prediction problem. 
\item $P_{t}^{(y)\text{{\ensuremath{^{\text{{-}}}}}}}$ and  $P_{t-1}^{(y)}$ are the uncertainties
related to $\hat{\mathbf{y}}_{t}^{\text{{-}}}$ and $\hat{\mathbf{y}}_{t-1}$ respectively. 
\end{itemize}

Eq. \eqref{eq:time_update_state_kfhe} is derived directly from
Eq. \eqref{eq:time_update_state} by setting $A_{k}$ to the identity
matrix, and assuming that $\mathbf{u}_{k}$ is non-existent (there
is no control process involved in KFHE). $H_{t}$ in Eq. \eqref{eq:measure_update_gain},
\eqref{eq:measure_update_state}, and \eqref{eq:measure_update_cov}
is set to the identity function. Also, it is assumed that no process
noise is induced, hence $Q_{t}$ in Eq. \eqref{eq:time_update_cov}
is set to $0$ to get Eq. \eqref{eq:time_update_cov_kfhe}. The superscript
$(y)$ throughout indicates that these parameters are related to the
model Kalman filter, \emph{kf-m}, estimating the state $\mathbf{y}_{t}$.

The \emph{kf-m} measurement step is more interesting. At every $t$th
iteration a new classification model $h_{t}$ is trained with a weighted
sample of the training dataset. The sampling is done with replacement,
with the same number of datapoints as in the original training dataset.
The weights are designed to highlight the points which were misclassified
previously, as is common in boosting algorithms (although the weight
updates are performed using the other Kalman filter\emph{ kf-w}).
The measurement is taken as the average of the previous prediction,
$\hat{\mathbf{y}}_{t-1}$, and the prediction of this $t$th model,
$h_{t}$, as in Eq. \eqref{eq:measurement_kfm}. This effectively
attempts to capture how much the trained model of the present iteration
impacts the ensemble predictions until iteration $t-1$. Therefore
the measurement step and its related error for \emph{kf-m} becomes:

\begin{equation}
\mathbf{z}_{t}^{(y)}=\frac{1}{2}(\mathbf{\hat{y}}_{t-1}+h_{t}(\mathcal{D}))\label{eq:measurement_kfm}
\end{equation}
\begin{equation}
R_{t}^{(y)}=\frac{1}{n}\sum_{i}^{n}(\mathcal{D}_{Y_{i}}\ne class(z_{ti}^{(y)}))\label{eq:measure_err_kfhe}
\end{equation}
\begin{equation}
\hat{\mathbf{y}}_{t}=\hat{\mathbf{y}}_{t}^{\text{{-}}}+K_{t}^{(y)}(\mathbf{z}_{t}^{(y)}-\hat{\mathbf{y}}_{t}^{\text{{-}}})\label{eq:measure_update_state_kfhe}
\end{equation}
\begin{equation}
K_{t}^{(y)}=P_{t}^{(y)\text{{\ensuremath{^{\text{{-}}}}}}}(P_{t}^{(y)\text{{\ensuremath{^{\text{{-}}}}}}}+R_{t}^{(y)})^{-1}\label{eq:measure_update_gain_kfhe}
\end{equation}
\begin{equation}
P_{t}^{(y)}=(I-K_{t}^{(y)})P_{t}^{(y)\text{{\ensuremath{^{\text{{-}}}}}}}\label{eq:measure_update_cov_kfhe}
\end{equation}

\noindent where:
\begin{itemize} 
\item $h_{t}=\mathcal{L}(\mathcal{D},\hat{\mathbf{w}}_{t-1})$ is a
model trained on dataset $\mathcal{D}$, using the learning algorithm
$\mathcal{L}$, where the dataset is sampled using the weights $\hat{\mathbf{w}}_{t-1}$.
\item $h_{t}(\mathcal{D})$ indicates the predictions made by the trained model $h_{t}$ for the datapoints in the dataset $\mathcal{D}$.
\item $\mathbf{z}_{t}^{(y)}=[z_{t1};z_{t2};\ldots;z_{tn}]$ represents the measurement heuristic,
the representation of which is as explained in Figure \ref{fig:The-representation-of}.

\item $R_{t}^{(y)}$ is the uncertainty related to $\mathbf{z}_{t}^{(y)}$ and is a misclassification rate calculated by comparing the class predictions made by the current ensemble, $class(z_{ti}^{(y)})$, with the ground truth classes, $\mathcal{D}_{Y}$. 
\end{itemize}

The remaining steps of the Kalman filter process to compute the Kalman gain, the \emph{a
posteriori} state estimate, and the variance are as described for the standard Kalman filter framework but are repeated in Eq. \eqref{eq:measure_update_gain_kfhe},
\eqref{eq:measure_update_state_kfhe} and \eqref{eq:measure_update_cov_kfhe}
for completeness. Note that, the uncertainty and the Kalman gain are
scalars in the KFHE implementation, as the state to be estimated is
one model and only one measurement is taken per iteration.

To initialise the \emph{kf-m} processthe initial learner is trained
as $h_{0}=\mathcal{L}(\mathcal{D},\hat{\mathbf{w}}_{0})$ and $\hat{\mathbf{y}}_{0}=h_{0}(\mathcal{D})$,
where $\hat{\mathbf{w}}_{0}$ is a uniform distribution. Also, $P_{0}^{(y)}$ is set to $1$, indicating
that the initial \emph{a priori} estimates are uncertain. After initialisation,
the iteration starts at $t=1$. The goal of the training phase is
to compute and store the learned $h_{t}$ models and the Kalman gain
$K_{t}^{(y)}$ values for all $t$.

To avoid measurements with large errors, if the measurement error
is more than $(1-\frac{1}{c})$, where $c$ is the number of classes,
then the sampling weights, $\mathbf{w}_{t}$, are reset to a uniform distribution,
which is a similar modification to that used in the AdaBoost implementation
in \citep{adabag}. 

\subsection{The weight Kalman filter: \emph{kf-w\label{subsec:The-weight-Kalman}}}

The previous description mentioned how a component learner $h_{t}$
depends on a vector of sampling weights, $\hat{\mathbf{w}}_{t-1}$,
which is estimated using \emph{kf-w}. The purpose of $\hat{\mathbf{w}}_{t-1}$
is to give more weight to the datapoints which were not classified
correctly in the previous iteration to encourage specialisation. The
implementation of \emph{kf-w} is very similar to the previous Kalman
filter implementation. In this case the state estimated by the Kalman
filter is a vector of real numbers representing weights. The time
update step in this case is also the identity function:

\begin{equation}
\hat{\mathbf{w}}_{t}^{\text{{-}}}=\hat{\mathbf{w}}_{t-1}\label{eq:time_update_state_kfhe-1}
\end{equation}
\begin{equation}
P_{t}^{(w){^{\text{{-}}}}}=P_{t-1}^{(w)}\label{eq:time_update_cov_kfhe-1}
\end{equation}

\noindent To estimate the measurement of the weights the following
equations are used:

\begin{equation}
\mathbf{\boldsymbol{z}}_{t}^{(w)}=[z_{ti}^{(w)}|z_{ti}^{(w)}=\hat{w}_{ti}\times f(\mathcal{D}_{Y_{i}}\ne class(z_{ti}^{(y)}))\,\,\,1\le i\le n]\label{eq:measurement_kfm-1}
\end{equation}
\begin{equation}
R_{t}^{(w)}=R_{t}^{(y)}\label{eq:measure_err_kfhe-1}
\end{equation}

This heuristic derives the measurement $\mathbf{z}_{t}^{(w)}$ of
\emph{kf-w}, from the classification error, $R_{t}^{(y)}$, of the
measurement $\mathbf{z}_{t}^{(y)}$ of \emph{kf-m}, as shown in Figure
\ref{fig:Overall-dataflow-between}. In Eq. \eqref{eq:measurement_kfm-1},
the function $f$ can adjust the impact of
misclassified datapoints on the weight vector. In the present work on KFHE, two options are explored: $f(x)=x$ and $f(x)=exp(x)$, where the second option
places more emphasis on misclassified datapoints. We refer to the variant of KFHE using the first, linear definition for $f(x)$ as KFHE-l and the variant using the second, exponential definition as KFHE-e.

A trivial heuristic is used in this step to compute the measurement
error, $R_{t}^{(w)}$, by setting it to $R_{t}^{(y)}$ (Eq. \eqref{eq:measure_err_kfhe-1}).
This assumes the measurement weight, $\mathbf{z}_{t}^{(w)}$,
has an error at most equal to the last measurement error for
\emph{kf-m}, which assumes that the weights $\hat{\mathbf{w}}_{t}^{-}$
will lead to a model with an error no more than the last measurement
by \emph{kf-m}. The measurement update of \emph{kf-w} becomes:

\begin{equation}
\hat{\mathbf{w}}_{t}=\hat{\mathbf{w}}_{t}^{\text{{-}}}+K_{t}^{(w)}(\mathbf{z}_{t}^{(w)}-\hat{\mathbf{w}}_{t}^{\text{{-}}})\label{eq:measure_update_state_kfhe-1}
\end{equation}
\begin{equation}
K_{t}^{(w)}=P_{t}^{(w)\text{{-}}}(P_{t}^{(w)\text{{-}}}+R_{t}^{(w)})^{-1}\label{eq:measure_update_gain_kfhe-1}
\end{equation}
\begin{equation}
P_{t}^{(w)}=(I-K_{t}^{(w)})P_{t}^{(w)\text{{-}}}\label{eq:measure_update_cov_kfhe-1}
\end{equation}

\noindent The superscript $(w)$ indicates that these parameters are related
to \emph{kf-w}. Here $\mathbf{w}_{t}=[w_{t1,}w_{t2,\ldots,}w_{tn}]$
and $\mathbf{\boldsymbol{z}}_{t}^{(w)}=[z_{t1,}z_{t2,\ldots,}z_{tn}]$
are vectors, with $w_{ti}$ and $z_{ti}$ representing the weight
estimate and the weight measurement of the $t$th iteration for
the $i^{th}$ datapoint.

The equations for \emph{kf-w} to compute the Kalman gain, $K_{t}^{(w)}$;
the \emph{a posteriori} state estimate for the weights, $\hat{\mathbf{w}}_{t}$;
and the variance, $P_{t}^{(w)}$, are shown in Eq. \eqref{eq:measure_update_gain_kfhe-1},
\eqref{eq:measure_update_state_kfhe-1} and \eqref{eq:measure_update_cov_kfhe-1}.
These are identical to those presented for \emph{kf-m} in
Section \ref{subsec:The-model-Kalman} (except for the superscripts),
but are included here for completeness.

Initially, $\mathbf{w}_{0}$ is set to have equal weights for every
datapoint in the training set, and $P_{0}^{(w)}$ is initialised to
$1$. Note that under this implementation the calculation of the measurement
error for \emph{kf-w} and the initialisation of $P_{0}^{(w)}$, makes
the Kalman gain $K_{t}^{(w)}$ the same as $K_{t}^{(y)}$. No information from the \emph{kf-w} process needs to be stored to support predictions from the ensmble. 

\begin{algorithm}[!t]
\caption{KFHE training \label{alg:Train}}

\begin{algorithmic}[1]
\Procedure{kfhe\_train}{$\mathcal{D}_{train}, T$}\Comment{\parbox[t]{.5\linewidth}{$\mathcal{D}_{train}$: training dataset,\\ $T$: Max iterations}}
\State \textbf{Initialise} \textit{kf-m}: $h_{0}$, $\hat{\textbf{y}}_{0}$ and $P_{0}^{(y)}$ following Section \ref{subsec:The-model-Kalman}
\State \textbf{Initialise} \textit{kf-w}: $\hat{\mathbf{w}}_{0}$ and $P_{0}^{(d)}$ following Section \ref{subsec:The-weight-Kalman}
\State $t=1$

\While{$t \le T$}
	\State \rule{0.6\textwidth}{1px}\Comment{\textit{kf-m} Section}
	\State \textbf{\textit{kf-m} time update}: \parbox[t]{.5\linewidth}{Find $\hat{\mathbf{y}}^{\text{-}}$, Eq. (\ref{eq:time_update_state_kfhe}) and \\related uncertainty $P_{t}^{(y)^{\text{-}}}$, Eq. (\ref{eq:time_update_cov_kfhe})}
    \State \textbf{\textit{kf-m} measurement}: \parbox[t]{.5\linewidth}{Train $h_{t}=\mathcal{L}(\mathcal{D},\hat{\mathbf{w}}_{t-1})$, compute $\mathbf{z}_{t}^{(y)}$ and $R_{t}^{(y)}$, Eq. (\ref{eq:measurement_kfm}) and (\ref{eq:measure_err_kfhe})}
    \If{(misclassification rate of $h_{t}$ more than $(1-\frac{1}{c})$)}
    	\State Reset $\hat{\mathbf{w}}_{t-1}$ and $P_{t-1}^{(w)}$ to initial values
		\State $t=t-1$
        \State \textbf{continue}\Comment{Repeat measurement step}
    \EndIf
    \State \textbf{\textit{kf-m} measurement update}: \parbox[t]{.5\linewidth}{Compute $\hat{\mathbf{y}}_{t}$, $K_{t}^{(y)}$ and $P_{t}^{(y)}$, using Eq. (\ref{eq:measure_update_state_kfhe}), (\ref{eq:measure_update_gain_kfhe}) and (\ref{eq:measure_update_cov_kfhe})}
	\State \rule{0.6\textwidth}{1px}\Comment{\textit{kf-w} Section}
    \\
    \State \textbf{\textit{kf-w} time update}: \parbox[t]{.5\linewidth}{Find $\hat{\mathbf{w}}^{\text{-}}$ and related uncertainty $P_{t}^{(w)^{\text{-}}}$ using Eq. (\ref{eq:time_update_state_kfhe-1}) and (\ref{eq:time_update_cov_kfhe-1})}
    \State \textbf{\textit{kf-w} measurement}: \parbox[t]{.5\linewidth}{Compute $\mathbf{z}_{t}^{(w)}$ and $R_{t}^{(w)}$ using Eq. (\ref{eq:measurement_kfm-1}) and (\ref{eq:measure_err_kfhe-1})}
    \State \textbf{\textit{kf-w} measurement update}: \parbox[t]{.5\linewidth}{Compute $\hat{\mathbf{w}}_{t}$, $K_{t}^{(w)}$ and $P_{t}^{(w)}$ using Eq. (\ref{eq:measure_update_state_kfhe-1}), (\ref{eq:measure_update_gain_kfhe-1}) and (\ref{eq:measure_update_cov_kfhe-1})}
     \State \rule{0.6\textwidth}{1px}
    \State $t=t+1$
\EndWhile
\State \textbf{return}$(\{h_{t};\forall t\},\{K_{t}^{(y)};\forall t\})$\Comment{\parbox[t]{.5\linewidth}{The learned classifier models, and the \textit{kf-m} Kalman gain values}}
\EndProcedure
\end{algorithmic}
\end{algorithm}

\subsection{Making predictions using KFHE\label{subsec:Making-predictions-using}}

The goal of KFHE training is to calculate and store the
trained base model, $h_{t}$, and Kalman gain, $K_{t}^{(y)}$, for
each iteration, $t$, of the model Kalman filter, \emph{kf-m}, process for $t = 0$ to $t = T$ (the total number of component classifiers trained). Once
this is done, generating predictions is straight-forward.  Given a new datapoint,
$\mathbf{d}$, $\hat{\boldsymbol{y}}_{0}$ is
found using the initial model $h_{0}$. Then Eq. \eqref{eq:measurement_kfm} and \eqref{eq:measure_update_state_kfhe} are iteratively applied to generate predictions from each model, $h_{t}$, which are combined using the appropriate Kalman gain values, $K_{t}^{(y)}$. The final $\hat{\boldsymbol{y}}_{T}$
value is taken as the ensemble prediction,
and is a vector containing a prediction score for each class. Datapoints as classified as belonging to the class with the maximum score. Algorithm \ref{alg:KFHE-Predict}
summarises the prediction process for KFHE.

\begin{algorithm}[!t]
\caption{KFHE prediction \label{alg:KFHE-Predict}}

\begin{algorithmic}[1]
\Procedure{kfhe\_prediction}{$\mathbf{d},\{h_{t};\forall t\},\{K_{t}^{(y)};\forall t\}, T$}\Comment{\parbox[t]{.5\linewidth}{$\mathbf{d}$: test datapoint, \\$h_{t}$: The $t$th sub-classifier, \\$K_{t}^{(y)}$: The $t$th Kalman gain, \\$T$: max training iterations $0\le t\le T$}}
\State $\hat{\mathbf{y}}_{0}=h_{0}(\mathbf{d})$\Comment{Initial \textit{a posteriori} estimate}
%\State $\hat{\mathbf{y}}_{0}^{-}=h_{0}(\mathbf{d})$\Comment{Initial \textit{a posteriori} estimate}
\State $t=1$
\While{$t \le T$}
	\State \parbox[t]{.7\linewidth}{Compute $\hat{\mathbf{y}}_{t}^{\text{-}}$, the time update for $\textit{kf-m}$ using Eq. (\ref{eq:time_update_state_kfhe})}
    
    \State \parbox[t]{.7\linewidth}{Compute $\mathbf{z}_{t}^{(y)}$, the measurement for $\textit{kf-m}$, using the $h_{t}$ using Eq. (\ref{eq:measurement_kfm})}
    \State \parbox[t]{.7\linewidth}{Compute the $\textit{a posteriori}$ estimate $\hat{\mathbf{y}}_{t}$ using Eq. (\ref{eq:measure_update_state_kfhe})}
    \State $t=t+1$
\EndWhile
\State \textbf{return} $(\hat{\mathbf{y}}_{T})$\Comment{Return class-wise prediction scores}
\EndProcedure
\end{algorithmic} 
\end{algorithm}

\section{Experiments\label{sec:Experiment}}

This section describes the datasets, algorithms, experimental setup,
and evaluation processes used in a set of experiments designed to
evaluate the effectiveness of the KFHE algorithm. Two variants of
KFHE, KFHE-e and KFHE-l (as described in Section \ref{subsec:The-weight-Kalman}),
are evaluated and a set of state-of-the-art ensemble methods are used
as benchmarks.

\subsection{Datasets \& performance measure}

$30$ multi-class datasets (described in Table \ref{tab:Dataset-descriptions-used}) from the UCI Machine
Learning repository \citep{Lichman:2013} are used. These datasets are frequently used in classifier benchmark experiments \citep{Dietterich2000,ZHANG20081524,CAO20124451,SUN201687}, cover diverse domains, have numbers
of classes ranging from $2$ to $15$, and exhibit varying amounts of
class imbalance.

\begin{table}[!t]
\caption{The datasets used in this paper \label{tab:Dataset-descriptions-used}}
\centering %
\begin{tabular}{rrrr}
\hline 
dataset names & \#datapoints  & \#dimensions  & \#classes \tabularnewline
\hline 
mushroom  & 8124  & 22  & 2 \tabularnewline
iris  & 150  & 5  & 3 \tabularnewline
glass  & 214  & 10  & 6 \tabularnewline
car\_eval  & 1728  & 7  & 4 \tabularnewline
cmc  & 1473  & 10  & 3 \tabularnewline
tvowel  & 871  & 4  & 6 \tabularnewline
balance\_scale  & 625  & 5  & 3 \tabularnewline
breasttissue  & 106  & 10  & 6 \tabularnewline
german  & 1000  & 21  & 2 \tabularnewline
ilpd  & 579  & 11  & 2 \tabularnewline
ionosphere  & 351  & 34  & 2 \tabularnewline
knowledge  & 403  & 6  & 4 \tabularnewline
vertebral  & 310  & 7  & 2 \tabularnewline
sonar  & 208  & 61  & 2 \tabularnewline
diabetes  & 145  & 4  & 3 \tabularnewline
skulls  & 150  & 5  & 5 \tabularnewline
physio  & 464  & 37  & 3 \tabularnewline
flags  & 194  & 30  & 8 \tabularnewline
bupa  & 345  & 7  & 2 \tabularnewline
cleveland  & 303  & 14  & 5 \tabularnewline
haberman  & 306  & 4  & 2 \tabularnewline
hayes-roth  & 132  & 6  & 3 \tabularnewline
monks  & 432  & 7  & 2 \tabularnewline
newthyroid  & 432  & 7  & 3 \tabularnewline
yeast  & 1484  & 9  & 10 \tabularnewline
spam  & 4601  & 58  & 2 \tabularnewline
lymphography  & 148  & 19  & 4 \tabularnewline
movement\_libras  & 360  & 91  & 15 \tabularnewline
SAheart  & 462  & 10  & 2 \tabularnewline
zoo  & 101  & 17  & 7 \tabularnewline
\hline 
\end{tabular}
\end{table}

To evaluate the performance of each model, the macro-averaged $F_{1}$-score
\citep{kelleher2015fundamentals} was used. The $F_{1}$-score in
a binary classifier context indicates how precise as well as how robust
a classifier model is, and it can be easily extended to a multi-class
scenario. The macro-averaged $F_{1}$-score will be denoted as $F_{1}^{(macro)}$,
and is defined as:

\[
F_{1}^{(macro)}=\frac{1}{c}\sum_{i=1}^{c}F_{1}^{(i)}=\frac{1}{c}\sum_{i=1}^{c}2\times\frac{precision^{(i)}\times recall^{(i)}}{precision^{(i)}+recall^{(i)}}
\]

\noindent where $precision^{(i)}$ and $recall^{(i)}$ are the precision
and recall values for the $i^{th}$ class, where $c$ is the number of classes. $F_{1}^{(macro)}$ is appropriate for this experiment because the datasets used exhibit different levels of class imbalance. 

\subsection{Experimental setup}

The state-of-the-art methods used as benchmarks are AdaBoost
\citep{zhu2006multi}, Bagging \citep{Breiman1996}, Gradient Boosting
Machine (GBM) \citep{Friedman00greedyfunction} and Stochastic Gradient
Boosting Machine (S-GBM) \citep{FRIEDMAN2002367}. This set covers
the different fundamental ensemble classifier types described in Section
\ref{sec:Background}. For all algorithms, including KFHE-e and KFHE-l,
the component learners are CART models \citep{kelleher2015fundamentals}.
The performance of a single CART model is also included as a baseline
to compare against the ensemble methods. The number of ensemble components
is set to $100$ for all algorithms (initial experiments showed that
for all datasets there were no significant improvements in performance
beyond 100 components).

All implementations and evaluations were performed in R\footnote{A version of KFHE is available at \href{https://github.com/phoxis/kfhe}{https://github.com/phoxis/kfhe}}. The
AdaBoost and Bagging implementations were from the package \lstinline!adabag!
\citep{adabag}, and the GBM and S-GBM implementations were from the
package \lstinline!gbm! \citep{gbm}. 
%The CART implementation was from the \lstinline!RPart! \citep{rpart} package. 
As multi-class datasets were used in this experiment, the multi-class variant of
AdaBoost, AdaBoost.SAMME \citep{zhu2006multi}, was used for this
experiment (this will be described just as AdaBoost in the remainder
of the paper). For the KFHE experiments the training was stopped when
the value of $K_{t}^{(y)}$ reached $0$, which can be interpreted
as an indication that the state estimated by \emph{kf-m} has no uncertainty. 

The experiments were divided into two parts. First, to evaluate the
effectiveness of KFHE-e and KFHE-l and
to compare these to the state-of-the-art methods, the performance of all
algorithms is assessed using the datasets listed in Table \ref{tab:Dataset-descriptions-used}.
Second, the robustness of the different algorithms to class-label noise is compared. For both sets of experiments,
for each algorithm-dataset pair, a $20$ times $4$-fold cross-validation
experiment was performed, and the mean of the $F_{1}^{(macro)}$-scores
across the folds are measured.

For the second set of experiments, class-label noise was introduced
synthetically into each of the datasets in Table \ref{tab:Dataset-descriptions-used}.
To induce noise a fraction of the datapoints
from the training set was sampled randomly  and the class of each selected datapoint was randomly changed, following a uniform distribution, to a different one.
For each dataset in Table \ref{tab:Dataset-descriptions-used}, datasets with $5\%$, $10\%$,
$15\%$ and $20\%$ noise were generated. For each of
these noisy datasets, a $20$ times $4$-fold cross-validation
experiment was performed. For each fold, the noisy class labels were
used in training, but the $F_{1}^{(macro)}$-scores were computed
with respect to the original unchanged dataset labels. 

\section{Results\label{sec:Results}}

The experiment results comparing the performance of KFHE-e and KFHE-l
to the other methods are shown first. Next, the results of the experiments
comparing the performance of the different methods in the presence
of noisy class-labels are presented. Statistical
significance tests that analyse the differences between
the proposed and other methods are also presented.

\subsection{Performance comparison of the methods\label{subsec:Performance-comparison-of}}

\begin{figure}[!t]
\centering{}\includegraphics[width=0.65\textwidth]{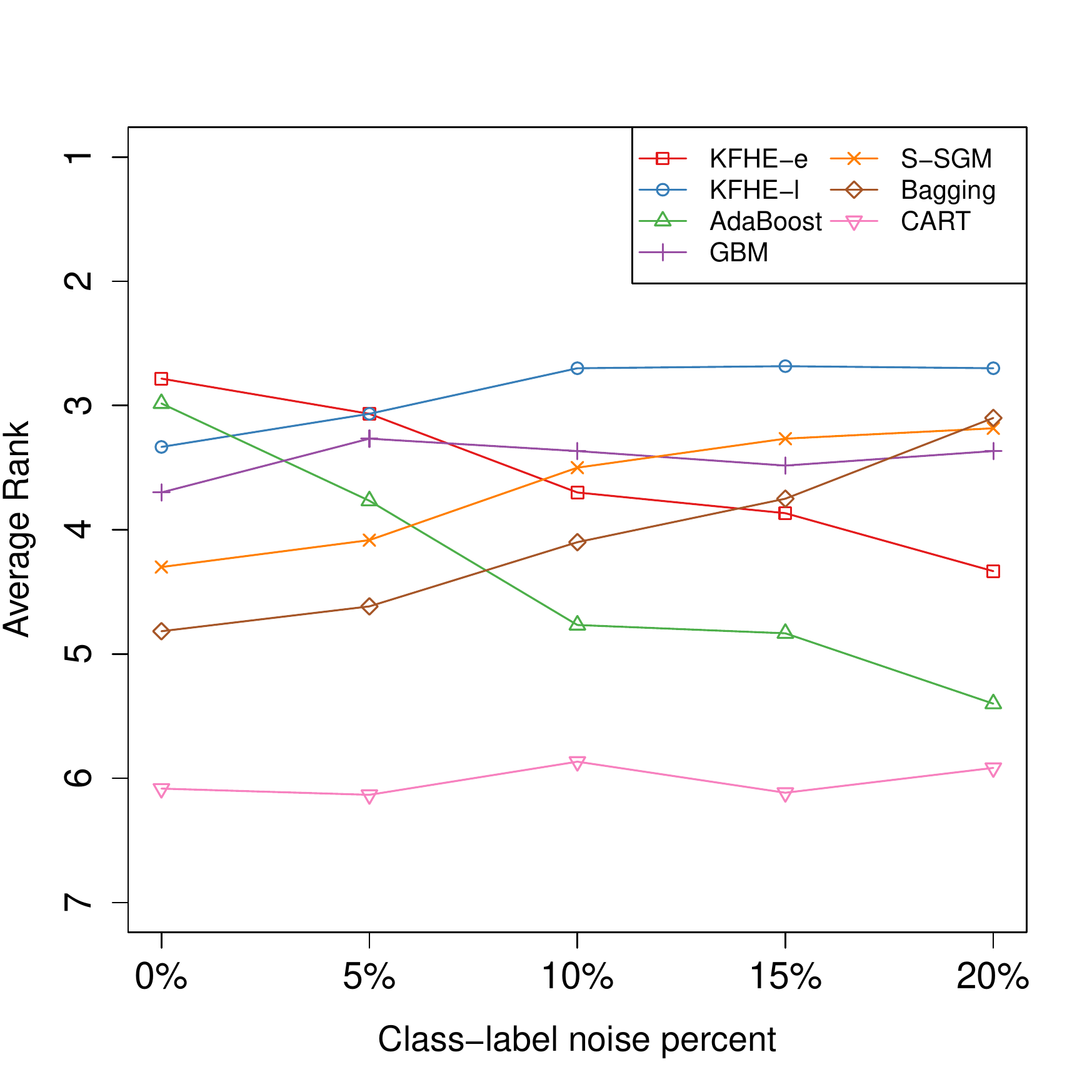}
\caption{Changes in average rank with noisy class-labels, over the datasets
used (y-axis is inverted to highlight lower average ranks are better).
\label{fig:Changes-in-average}}
\end{figure}

The relative performance of each algorithm, based on the average $F_{1}^{(macro)}$-scores achieved in the cross validation experiments, on each of the $30$ datasets was ranked (from $1$ to $7$, where $1$ implies best performance). The first row of Table \ref{tab:summary-rank-tab} (labelled $0\%$) shows the average rank of each algorithm across the datasets (detailed performance results for each algorithm on each dataset are shown in Table \ref{tab:Average-F-Scores-00} in \ref{sec:resultsTables}). These average ranks are also visualised in the first column of Figure \ref{fig:Changes-in-average} (also labelled $0\%$).

The average ranking shows that KFHE-e was able to attain the best
average rank $2.78$, AdaBoost was very close with average rank $2.98$,
followed by KFHE-l with the average rank $3.33$. It is clear that
KFHE-e outperformed GBM, S-GBM, Bagging and CART. Also, KFHE-l performs
better overall than GBM, S-GBM, Bagging and CART. It was concluded
that KFHE-l performed slightly less well than KHFE-e and AdaBoost
due to the lack of emphasis on misclassified points in the weight
measurement step. In Section \ref{subsec:Statistical-significance-test},
a statistical significance test will be performed to uncover significant
differences between methods on datasets with non-noisy class-labels.

\begin{table}[!t]
\caption{Average ranks over all datasets for different levels of class label noise summarised from Tables \ref{tab:Average-F-Scores-00} to \ref{tab:noise20} in \ref{sec:resultsTables}. Lower ranks are better, best ranks are highlighted in boldface. Percent values in the rows indicate class-label noise.\label{tab:summary-rank-tab}}
\centering
\begin{tabular}{rrrrrrrr}
  \hline
 & KFHE-e & KFHE-l & AdaBoost & GBM & S-GBM & Bagging & CART \\ 
  \hline
  0\% & \textbf{2.78} & 3.33 & 2.98 & 3.70 & 4.30 & 4.82 & 6.08 \\ 
  5\% & \textbf{3.07} & \textbf{3.07} & 3.77 & 3.27 & 4.08 & 4.62 & 6.13 \\ 
  10\% & 3.70 & \textbf{2.70} & 4.77 & 3.37 & 3.50 & 4.10 & 5.87 \\ 
  15\% & 3.87 & \textbf{2.68} & 4.83 & 3.48 & 3.27 & 3.75 & 6.12 \\ 
  20\% & 4.33 & \textbf{2.70} & 5.40 & 3.37 & 3.18 & 3.10 & 5.92 \\ 
   \hline
\end{tabular}
\end{table}

%The mean $F_{1}^{(macro)}$-scores from the $20$ times $4$-fold cross-validation experiments and their standard deviations for each algorithm-dataset pair are shown in Table \ref{tab:Average-F-Scores-00}. A higher $F_{1}^{(macro)}$-score indicates better performance. In each cell of Table \ref{tab:Average-F-Scores-00}, the values in parenthesis show the relative ranking of the algorithm on the dataset in the corresponding row (lower ranks are better). The last row shows the average ranking of each algorithm over the datasets.

\begin{figure}[!t]
\subfloat[\emph{knowledge} \label{fig:knowledge_params_intext}]{\includegraphics[width=0.5\textwidth]{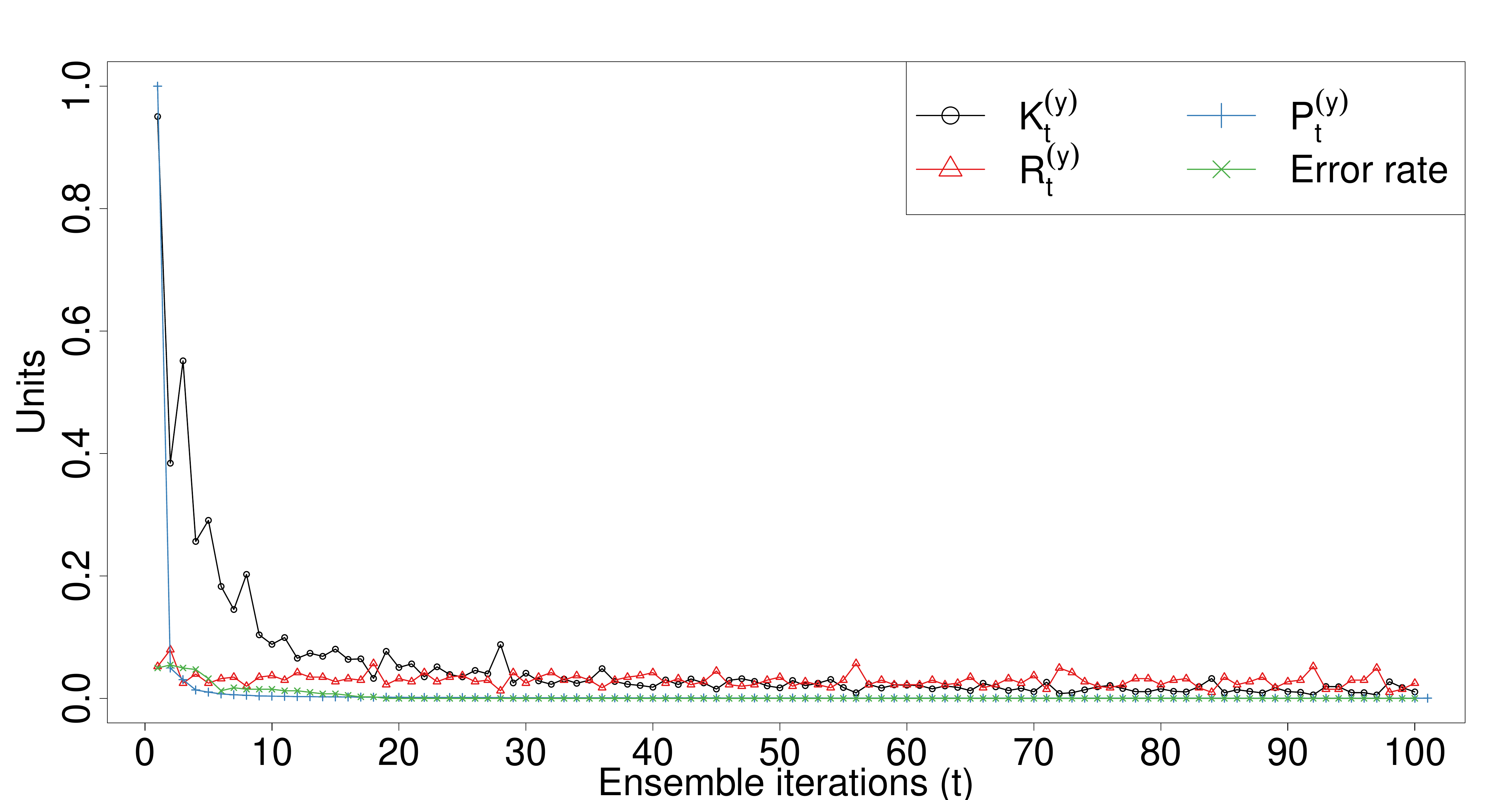}}
\subfloat[\emph{diabetes} \label{fig:german_params_intext}]{\includegraphics[width=0.5\textwidth]{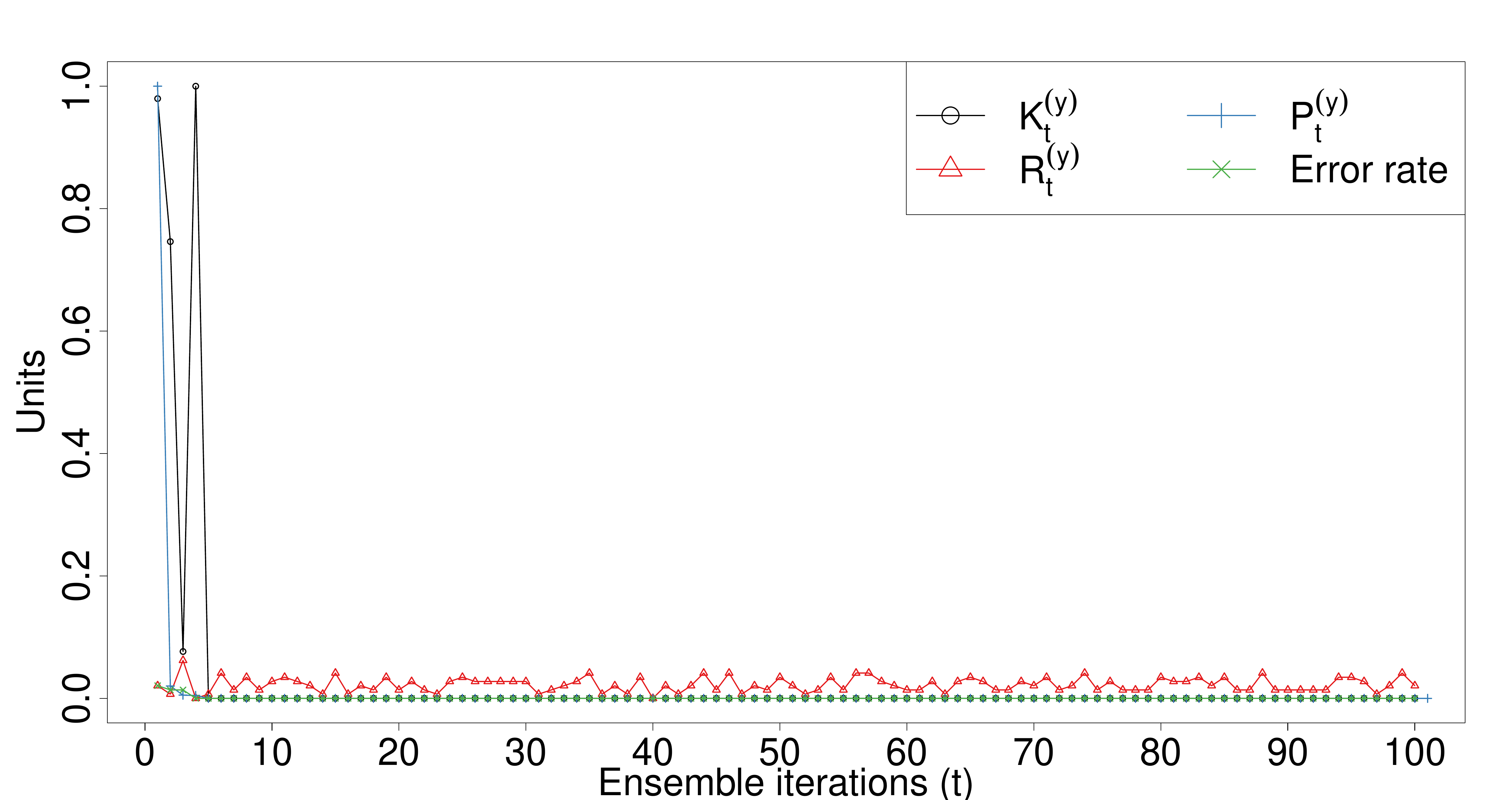}}

\subfloat[\emph{car\_eval} \label{fig:ilpd_params_intext}]{\includegraphics[width=0.5\textwidth]{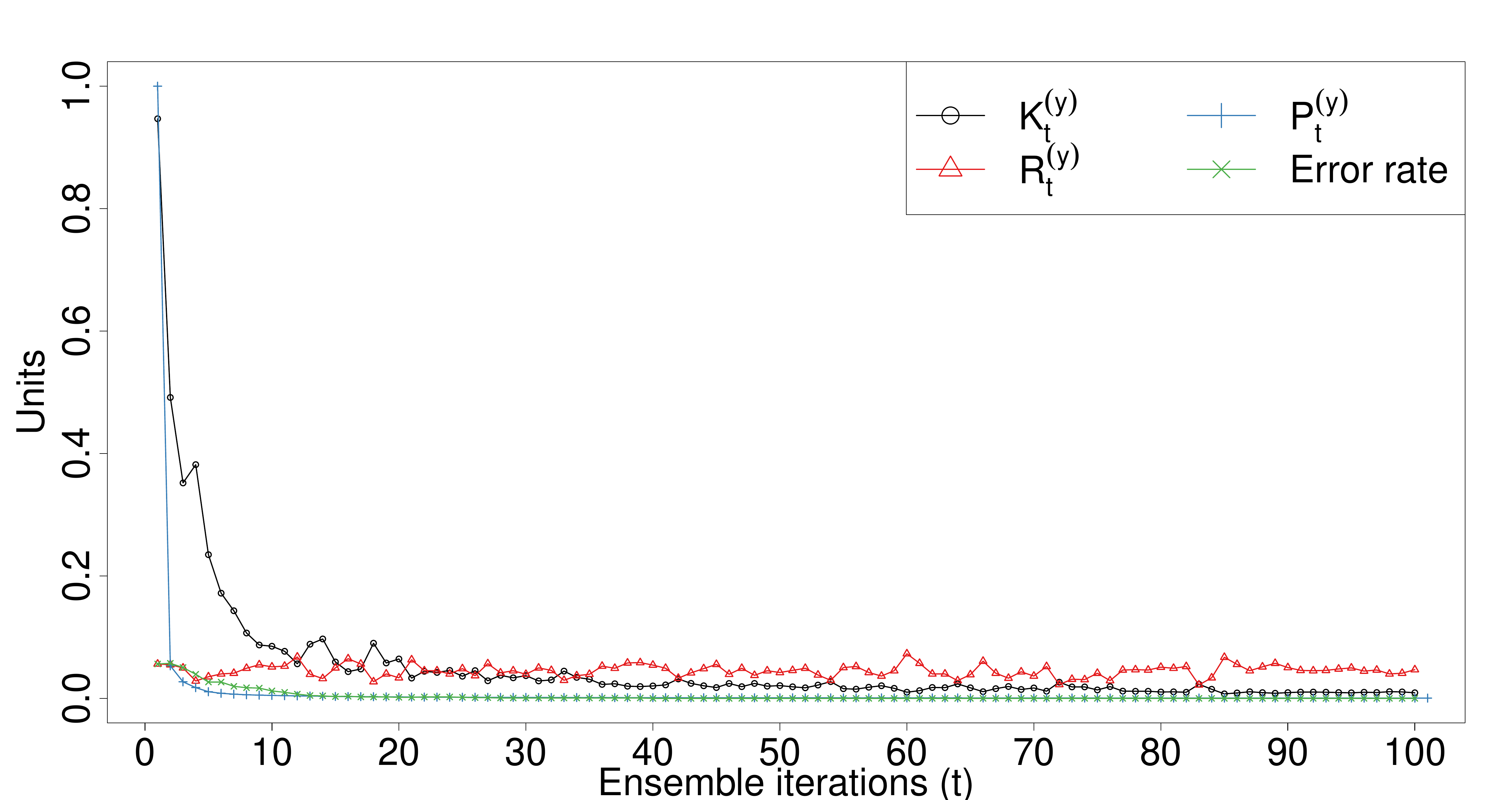}}
\subfloat[\emph{lymphography} \label{fig:glass_params_intext}]{\includegraphics[width=0.5\textwidth]{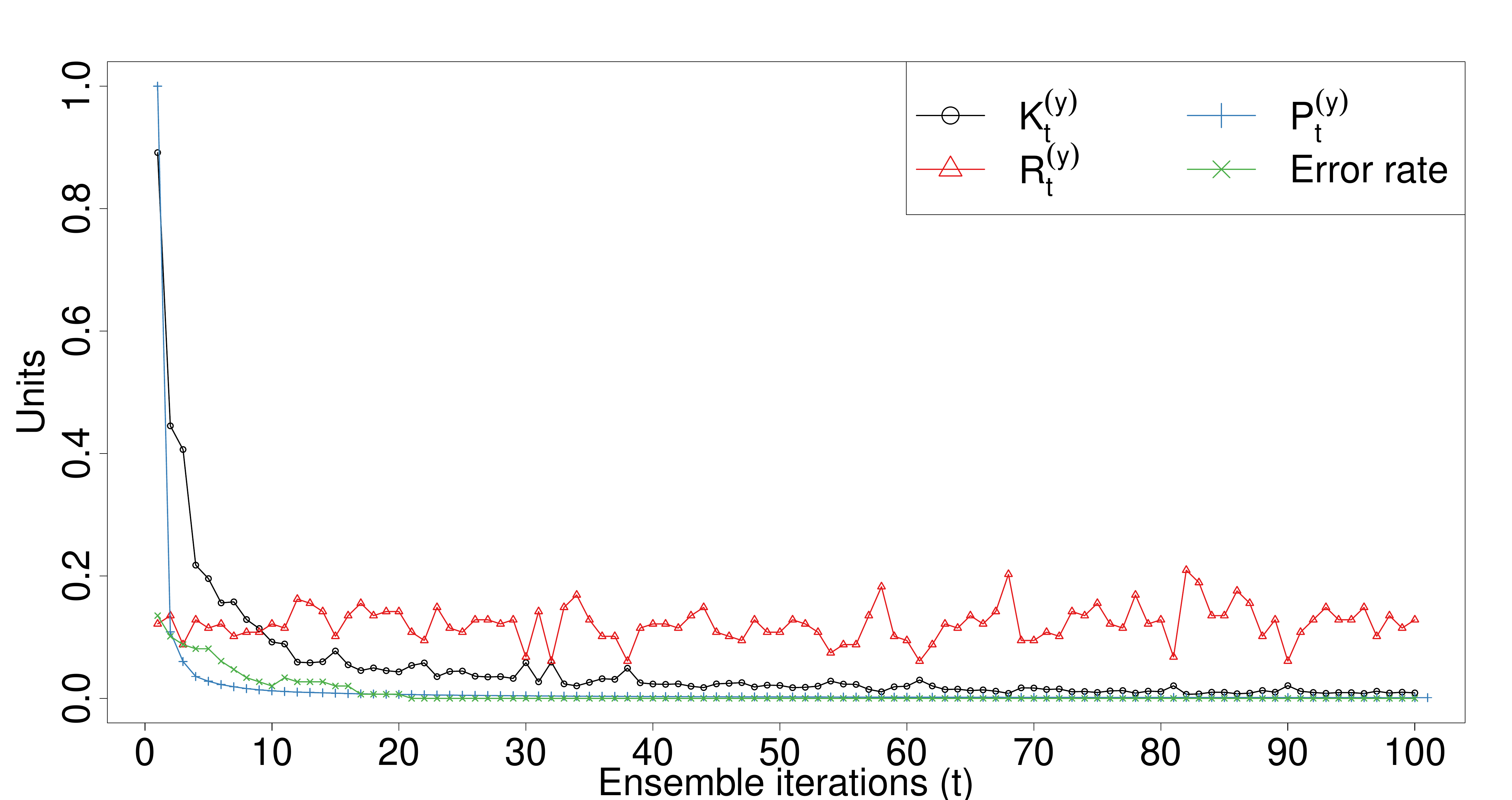}}
\caption{Changes in the parameters and the misclassification rate for the training
sets for KFHE-e, for the \emph{knowledge}, \emph{diabetes},  \emph{car\_eval}, and \emph{lymphography} datasets.\label{fig:param_changes_intext}}
\end{figure}

The evolution of the key parameters of KFHE (the measurement error,
$R_{t}^{(y)}$; the posteriori variance, $P_{t}^{(y)}$; the Kalman
gain, $K_{t}^{(y)}$; and the training misclassification rate of the
\emph{kf-m} component) with respect to the number of ensemble iterations
$t$, for a selection of datasets (\emph{knowledge},
\emph{diabetes}, \emph{car\_eval} and \emph{lymphography}) are plotted in Figure \ref{fig:param_changes_intext}. The plots show the results of the first $100$ iterations after
which, for most of the datasets, the error reduces to $0$. The plots
for all of the datasets are given in \ref{sec:parameterPlots}.

The plots in Figure \ref{fig:param_changes_intext} show that in all
cases the value of $P_{t}^{(y)}$ decreases monotonically, which can
be interpreted as the system becoming more confident on the \emph{a
posteriori} estimate, and therefore that the values of $K_{t}^{(y)}$
reduce and stabilise, implying less impact for subsequent measurements.
This is because of the way the \emph{time step} update was formulated
in Section \ref{subsec:The-model-Kalman}: no uncertainty induced
during the time update step, and no process noise is assumed. Therefore,
in effect the steepness of $P_{t}^{(y)}$ controls how much of the
measurement is combined through Eq. \eqref{eq:measure_update_state_kfhe}
and \eqref{eq:measure_update_gain_kfhe}. Also, it is interesting
to note the similarity and the rate of change of the error rate of
the ensemble with the $P_{t}^{(y)}$ value. For most of the datasets
they show a similar trend.  The value of $K_{t}^{(y)}$ indicates
the fraction of the measurement which will be incorporated into the
ensemble. A measurement with less error is incorporated more into
the final model.

\subsection{Performance for the noisy class-label case\label{subsec:Performance-for-the}}

The relative performance of each algorithm, based on the average $F_{1}^{(macro)}$-scores achieved in the cross validation experiments, on each of the $30$ datasets was ranked (from $1$ to $7$, where $1$ implies best performance). this was performed separately for datasets with $5\%$, $10\%$, $15\%$ and $20\%$ induced class label noise. Table \ref{tab:summary-rank-tab} shows the average rank of each algorithm for each level of noise (detailed performance results for each algorithm on each dataset are shown in Tables \ref{tab:noise05} to \ref{tab:noise20} in \ref{sec:resultsTables}). These average ranks are also visualised in Figure \ref{fig:Changes-in-average}. For ease of reading, the vertical axis in Figure \ref{fig:Changes-in-average} is inverted to highlight that the lower average ranks are better.

Out of the algorithms tested the KFHE-l algorithm performs most consistently in the presence of class-label noise. At the $5\%$ noise level KFHE-e and KFHE-l had the same rank, and as the
class-noise level increases to $10\%$, $15\%$ and $20\%$, KFHE-l
attains the best average rank over the datasets. Along with KFHE-l,
S-GBM and Bagging also improve their relative ranking. As the fraction
of mislabelled datapoints increased in the training set, the average
rank of AdaBoost degrades sharply. The performance of AdaBoost and
Bagging in the presence of noisy class labels is studied in \citep{Dietterich2000},
where a similar result was found. On the other hand the change in
the relative rank for GBM, and CART was consistently stable.

It should be noted that the degradation of performance with respect
to class-noise in AdaBoost is more severe than KFHE-e, although both
of them use the $exp$ function to highlight the weights of the misclassified
datapoints. This is due to the smoothing effect in the KFHE algorithm,
which makes KFHE-e less sensitive to noise than AdaBoost. On the other
hand, KFHE-l does not use $exp$ in Eq. \eqref{eq:measurement_kfm-1}
for the weight measurement step, which makes it more robust to noise
and allows it achieve high performance across all noise levels.

\begin{figure}[!t]
\centering\subfloat[\emph{knowledge} \label{fig:knowledge_noise_intext}]{\includegraphics[width=0.35\textwidth]{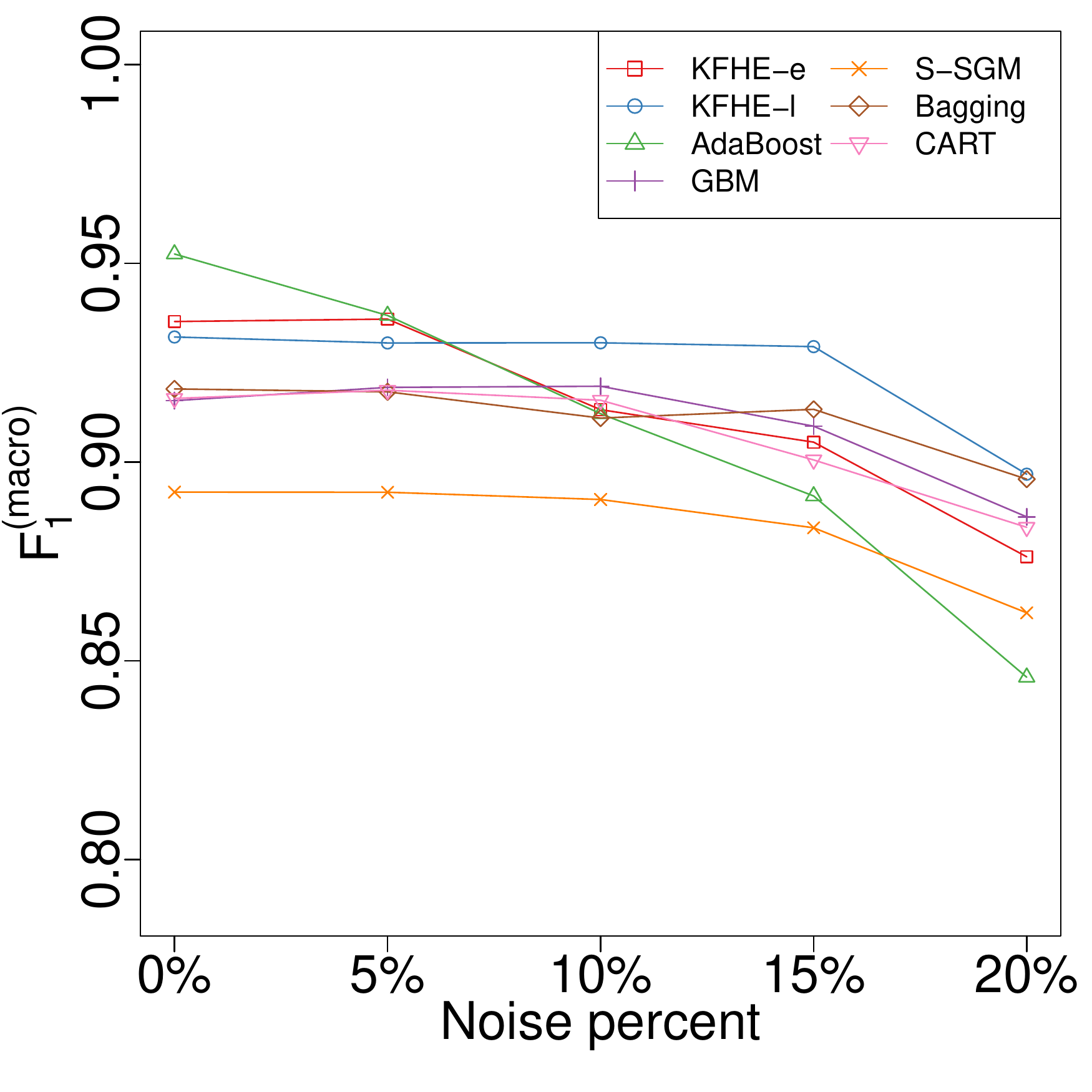}}
\subfloat[\emph{diabetes }\label{fig:german_noise_intext}]{\includegraphics[width=0.35\textwidth]{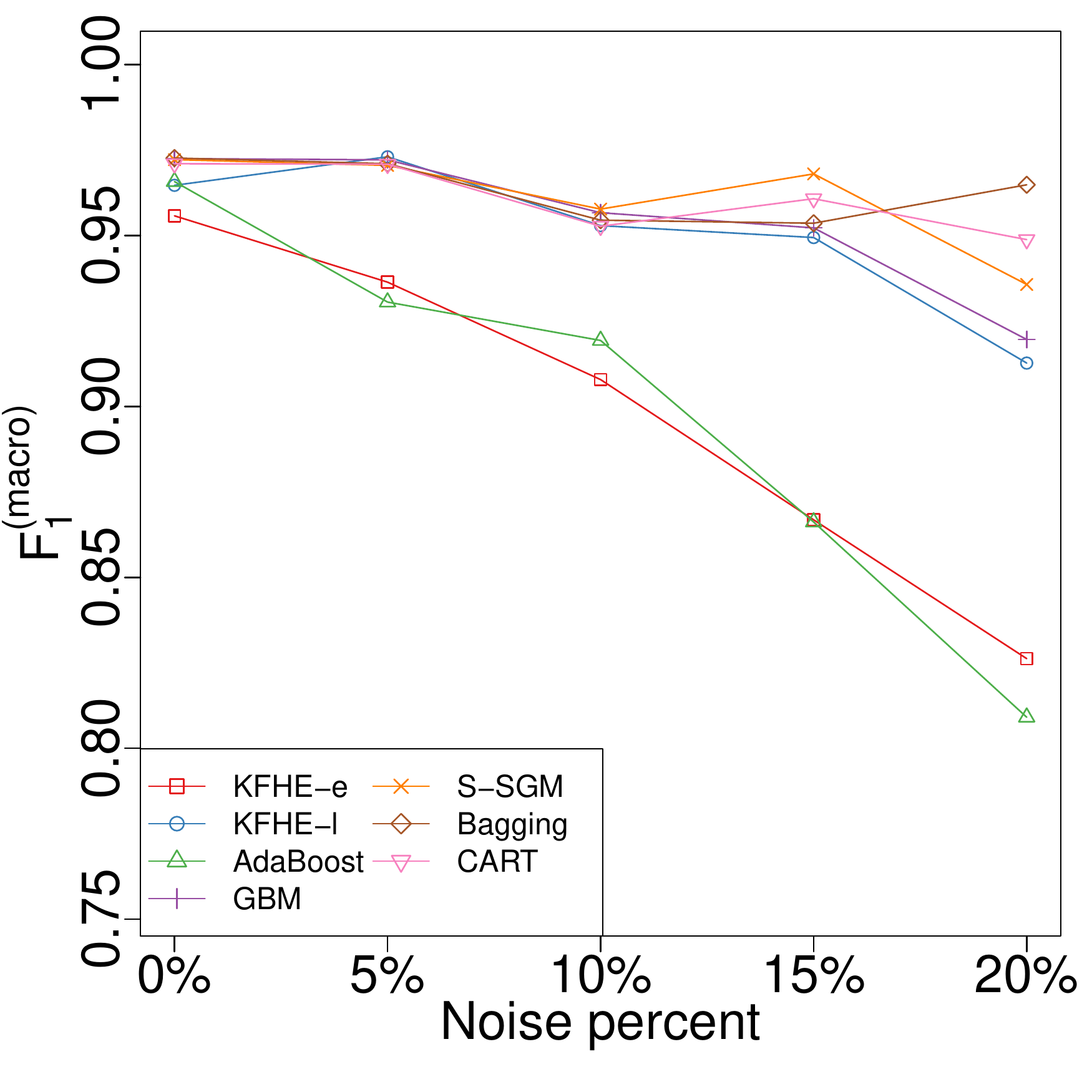}}

\centering\subfloat[\emph{car\_eval} \label{fig:ilpd_noise_intext}]{\includegraphics[width=0.35\textwidth]{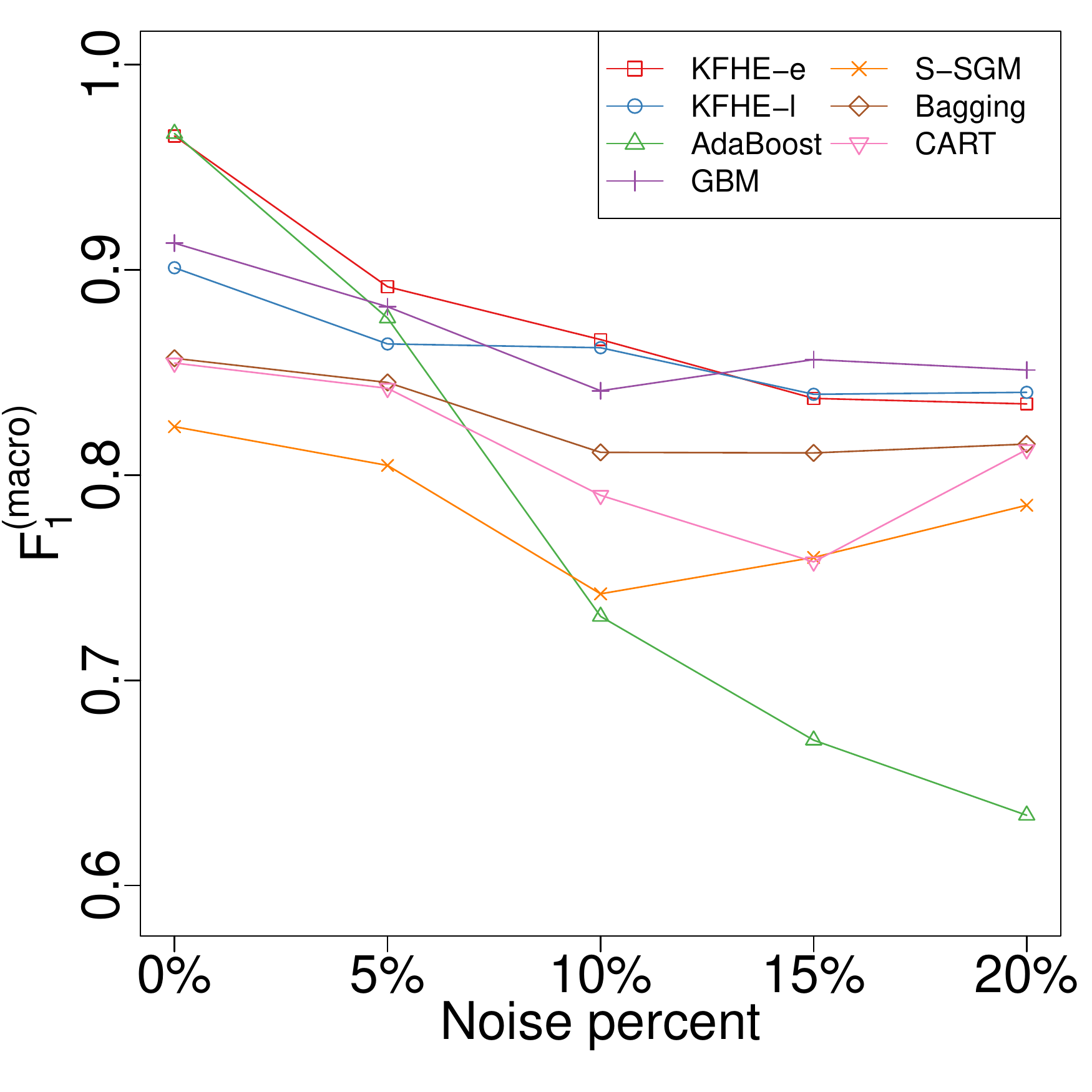}}
\subfloat[\emph{lymphography} \label{fig:glass_noise_intext}]{\includegraphics[width=0.35\textwidth]{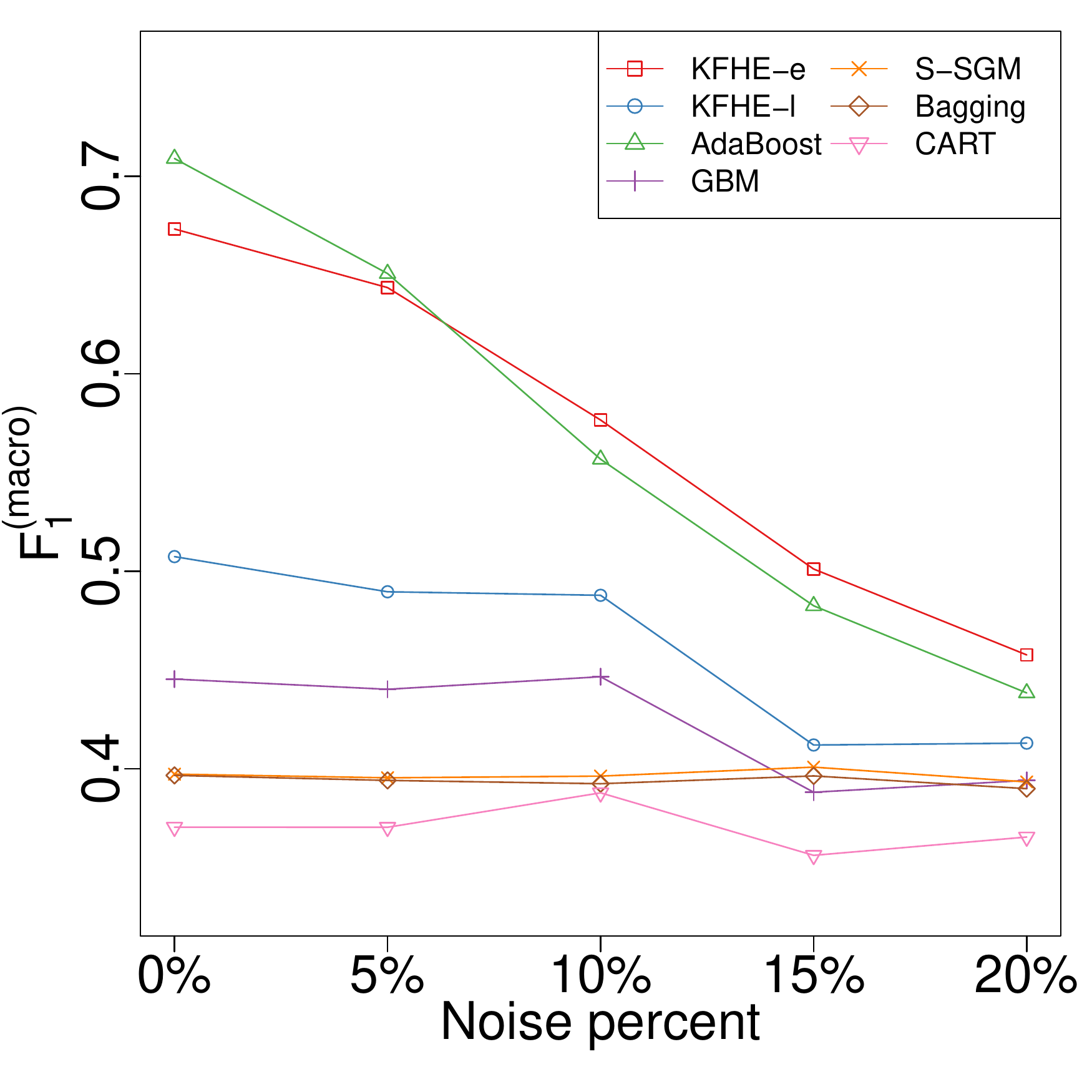}}
\caption{Changes in $F_{1}^{(macro)}$-score with increased induced class-label noise 
for the \emph{knowledge}, \emph{diabetes}, \emph{car\_eval}, and \emph{lymphography} datasets.
\label{fig:Changes-in-fscore-with-noise_intext}}
\end{figure}

Figure \ref{fig:Changes-in-fscore-with-noise_intext} shows the
change in $F_{1}^{(macro)}$-score for each algorithm on the
\emph{knowledge}, \emph{diabetes}, \emph{car\_eval}, and \emph{lymphography} datasets (similar plots for all datasets are given in \ref{sec:performancePlots}), as the amount of class-label noise increases (note that to highlight changes in performance the vertical axes in these charts are scaled to narrow ranges of possible $F_{1}^{(macro)}$-scores). These plots are derived from
Tables \ref{tab:Average-F-Scores-00}-\ref{tab:noise20}. With few
exceptions the performances of KFHE-l,
GBM, S-GBM and Bagging are not impacted as much as the other approaches
by noise. Although KFHE-e is generally better than
the other approaches when there is no class-label noise present, as the induced noise
increases, the $F_{1}^{(macro)}$-score for KFHE-e decreases---albeit less
severely than in the case of AdaBoost. 

\subsection{Statistical significance testing\label{subsec:Statistical-significance-test}}

This section presents two types of statistical significance tests that compare the performance
of the different algorithms tested. First, to assess
the overall differences in performance a multiple classifier comparison
test was performed following the recommendations of \citep{GARCIA20102044}.
Second, a comparison of each pair of algorithms
in isolation is performed using the Wilcoxon's
Signed Rank Sum test \citep{GARCIA20102044}. 
%Unlike the multiple classifier comparison tests, this test \emph{only} uncovers the statistical differences between a single pair in an isolated setting.

\subsubsection{Multiple classifier comparison\label{subsec:Multiple-classifier-comparison}}

\begin{figure}[!t]
\centering\subfloat[Rank chart for no induced class-label noise]{

\includegraphics[width=0.55\textwidth]{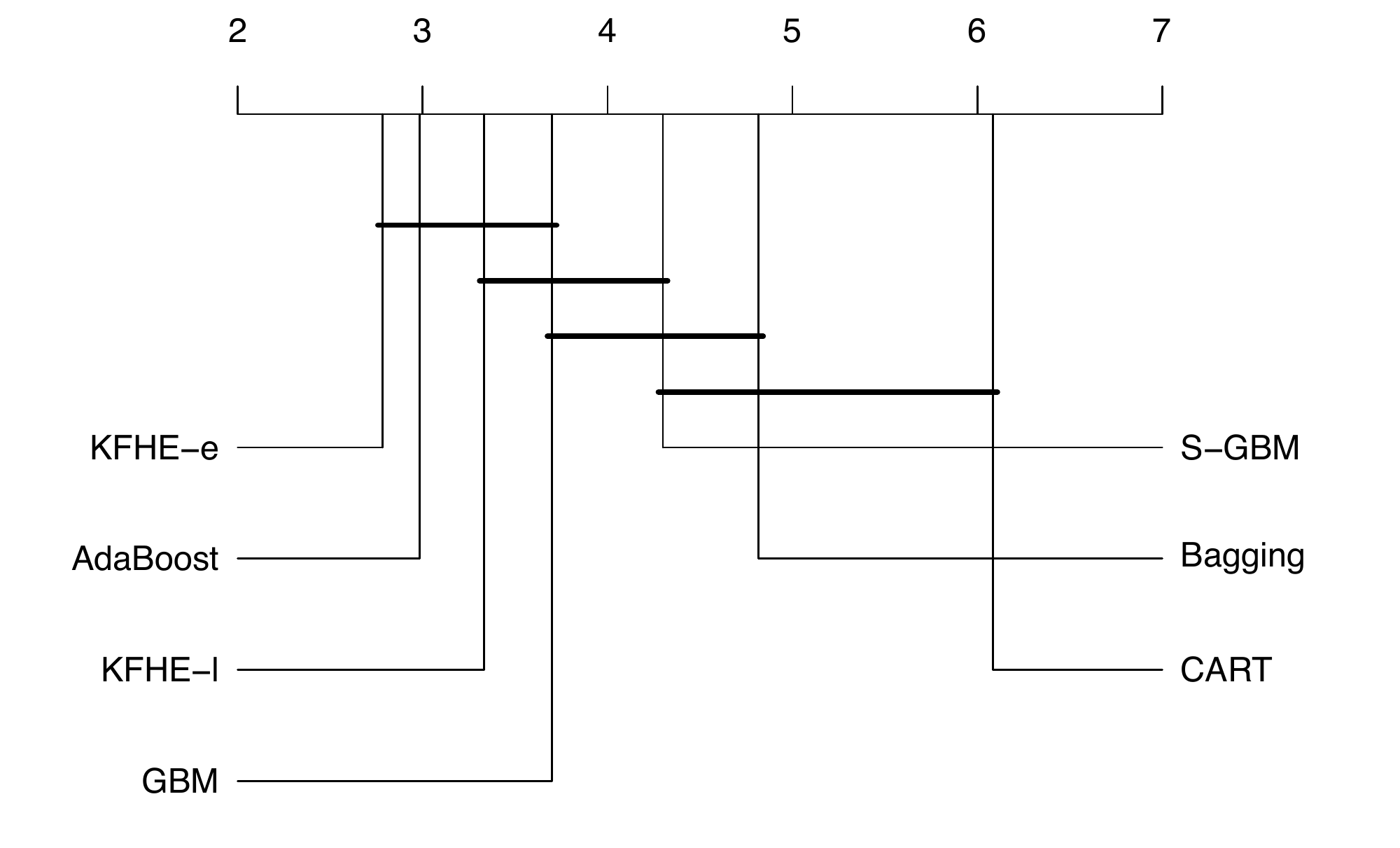}}

\subfloat[Rank chart for $5\%$ induced class-label noise]{

\includegraphics[width=0.55\textwidth]{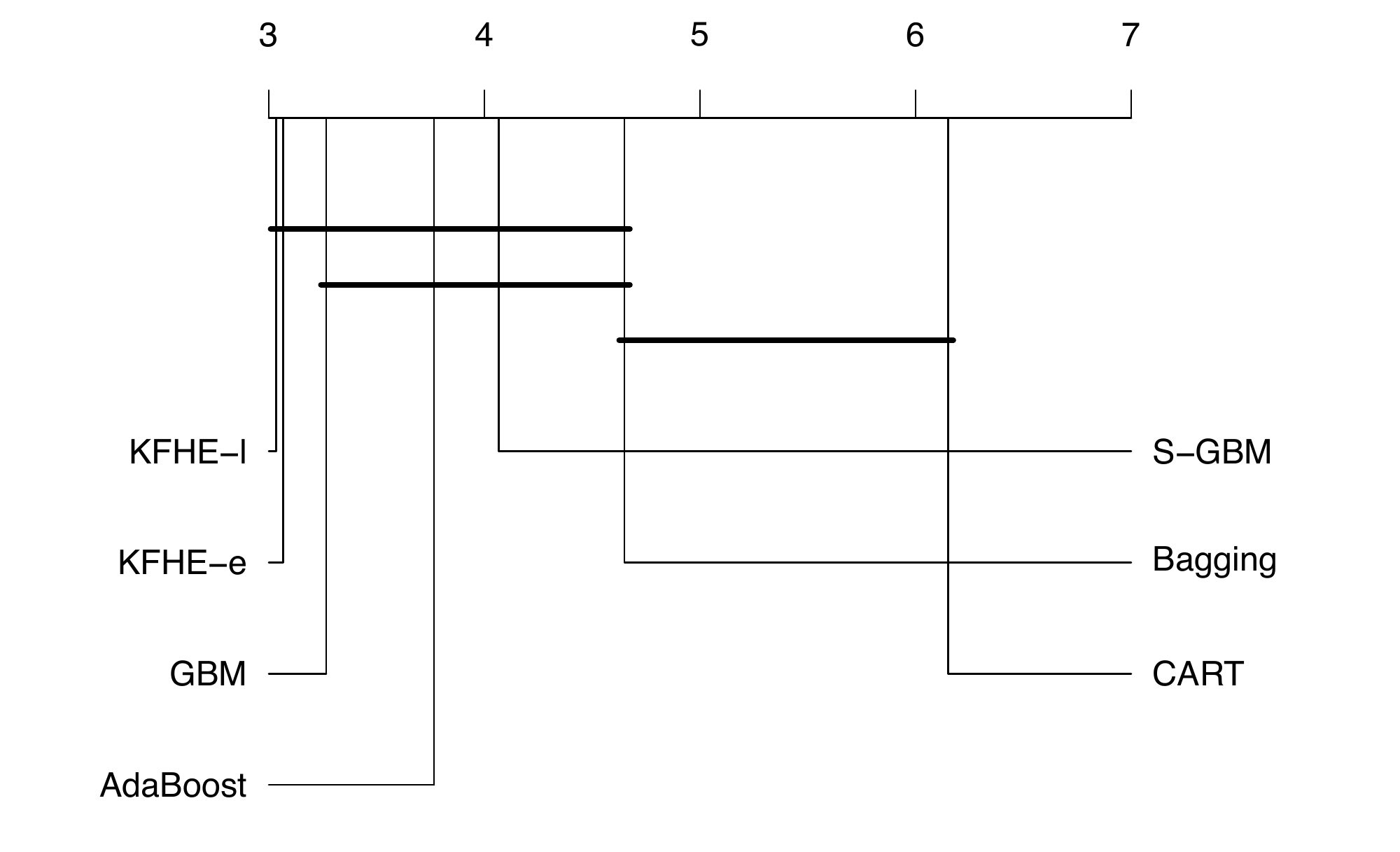}}\subfloat[Rank chart for $10\%$ induced class-label noise]{

\includegraphics[width=0.55\textwidth]{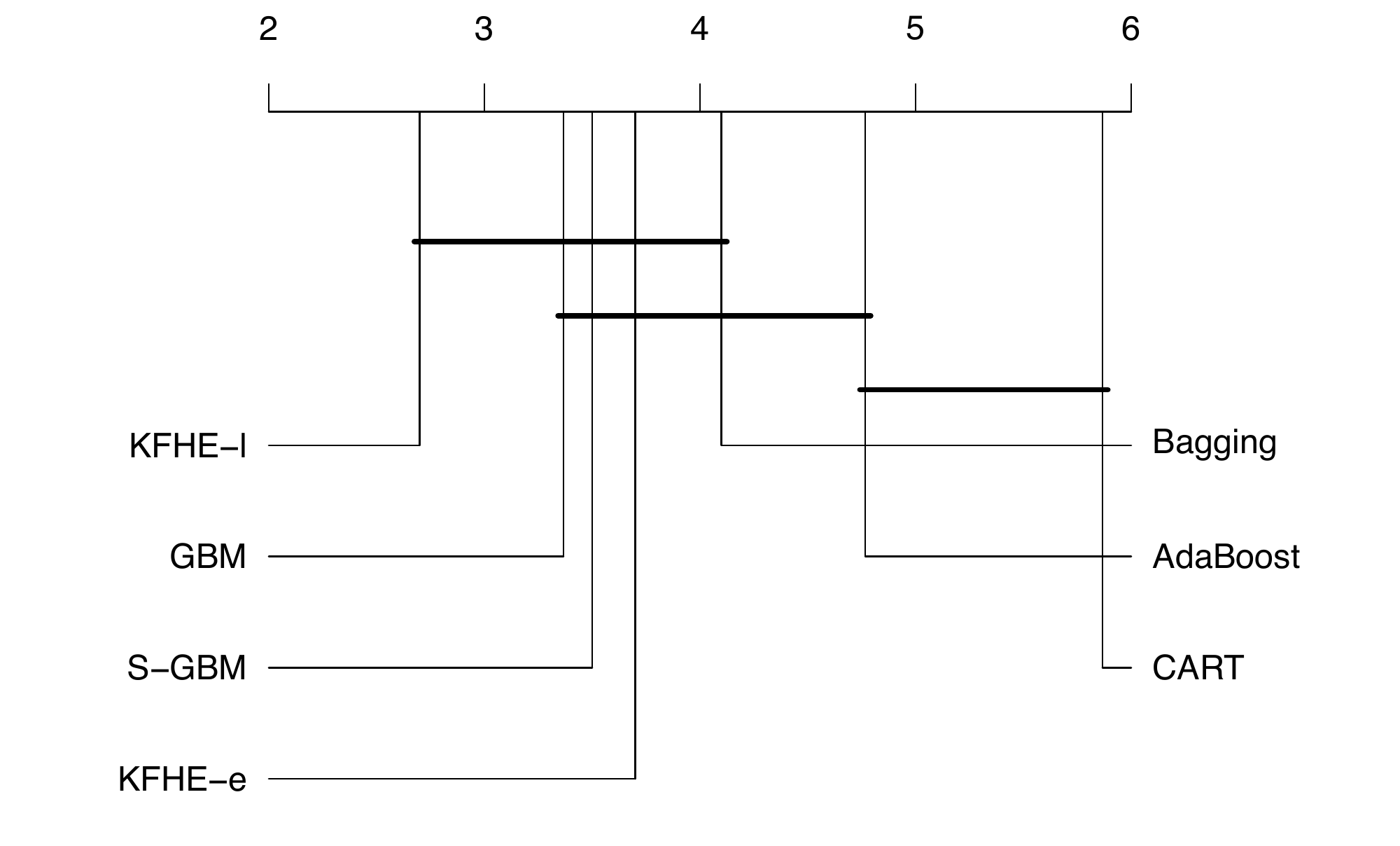}}

\subfloat[Rank chart for $15\%$ induced class-label noise]{

\includegraphics[width=0.55\textwidth]{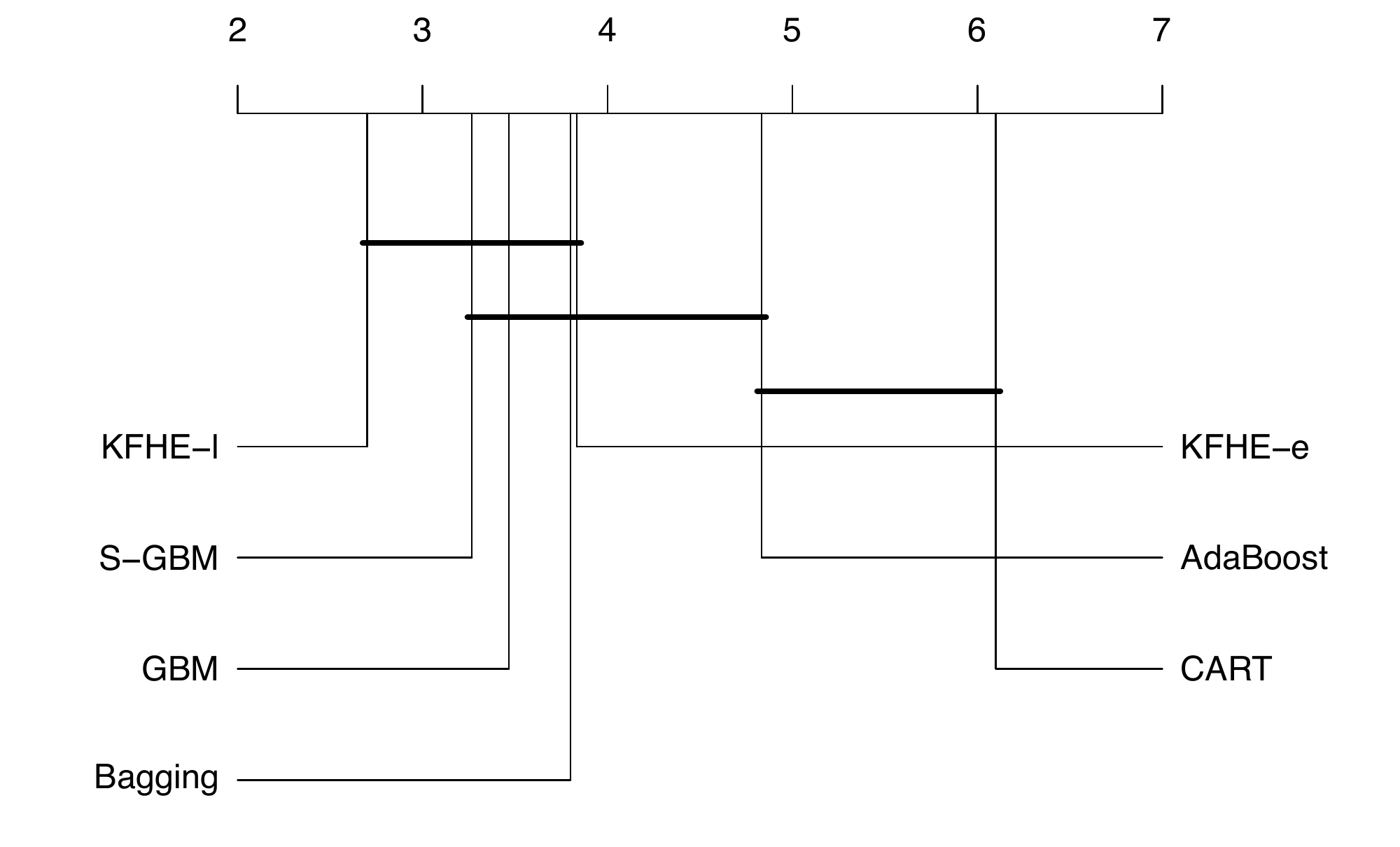}}\subfloat[Rank chart for $20\%$ induced class-label noise]{

\includegraphics[width=0.55\textwidth]{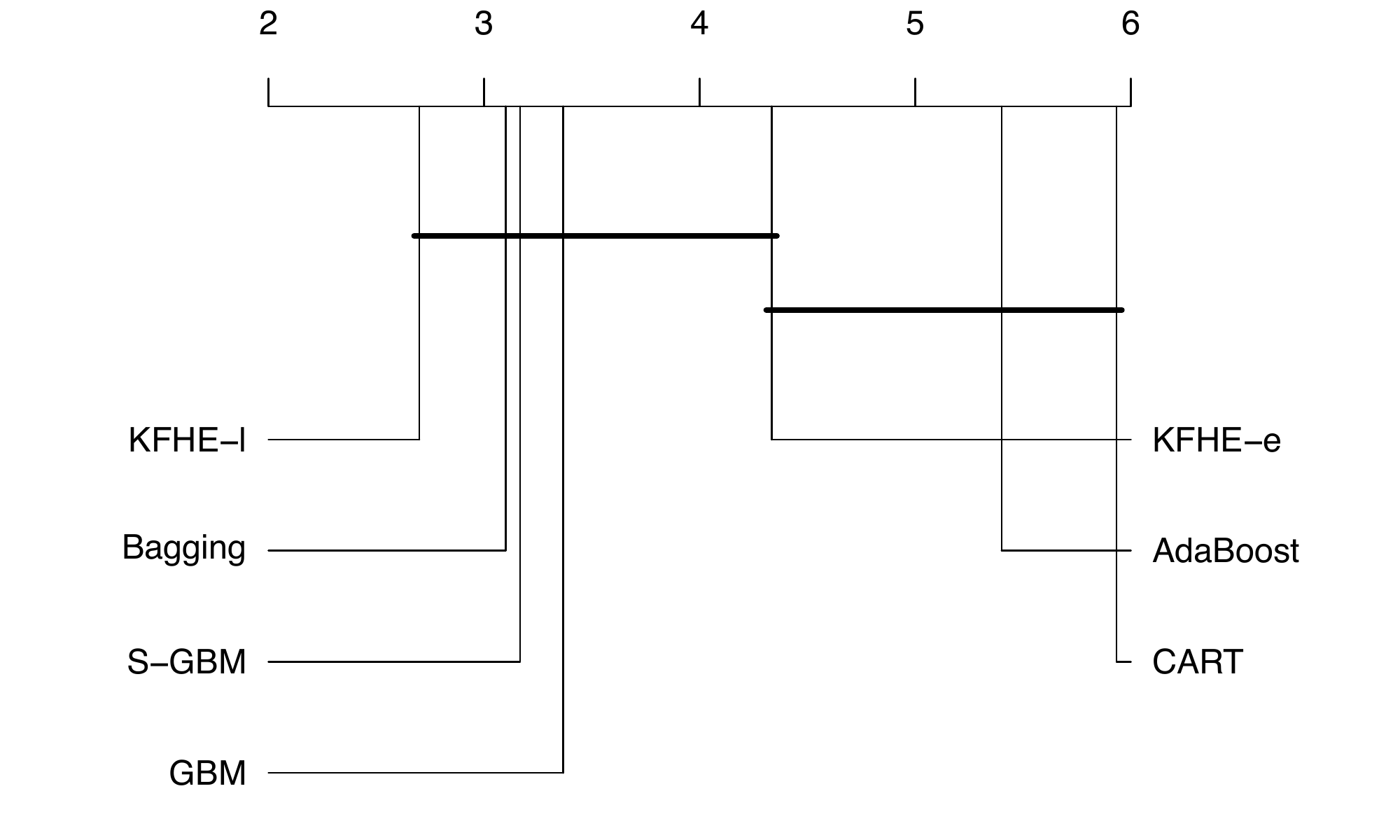}}

\caption{Rank plots from post-hoc Friedman Aligned Rank test with the Finner
$p$-value adjustment, using a significance level of $0.05$. Algorithms
connected with the horizontal bars in the sub-plots are similar based on this test. \label{fig:Rank-plots}}
\end{figure}

To understand the overall effectiveness of the variants of KFHE (KFHE-e
and KFHE-l), following the recommendations of García et. al. \citep{GARCIA20102044}, a multiple classifier comparison
significance test was performed (separate tests were performed on the
performance of algorithms at each noise level). First, a Friedman
Aligned Rank test was performed. This indicated that, at all noise levels, the null hypothesis that the performance of all algorithms
is similar can be rejected, with $p < 0.001$. To further investigate
these differences, post-hoc pairwise Friedman Aligned Rank tests along
with the Finner $p$-value adjustment \citep{GARCIA20102044} were performed. Rank plots describing the results of the post-hoc tests (with a significance level of $\alpha=0.05$) are shown in Figure \ref{fig:Rank-plots}.

When no class-label noise is present, the results indicate that KFHE-e
(avg. rank $2.78$) was significantly better than S-GBM (avg. rank
$4.3$), Bagging (avg. rank $4.82$) and CART (avg. rank $6.08$)
with $\alpha=0.01$;
%, and marginally better than GBM (avg. rank $3.7$)
and that KFHE-l (avg. rank $3.33$) was significantly
better than S-GBM, Bagging and CART with $\alpha=0.01$. Although
KFHE-e attained a better average rank, $2.78$, than AdaBoost, $2.98$, the null-hypothesis could not be rejected, and so it cannot be determined that the performances of KFHE-e and AdaBoost are significantly
different. Similarly, KFHE-l attained a worse average rank, $3.33$, 
than AdaBoost, but tests did not identify a statistically significant difference.
%The difference in performance between KFHE-e and KFHE-l is due to the different heuristics they use (described in Section \ref{subsec:The-weight-Kalman}). 

%Overall this experiment shows that when no class-label noise is present, KFHE-e performs at least as well as AdaBoost and better than GBM, S-GBM, Bagging and CART, whereas KFHE-l performs as well as AdaBoost and GBM and better than S-GBM, Bagging and CART.

The results of the experiment for the datasets with class-label noise indicate
that, as the noise continues to increase, the relative performance
of KFHE-l improves, but the relative performance of KFHE-e decreases.
This is as expected, because of the chosen weight measurement heuristic
for the two variants of KFHE as explained in Section \ref{subsec:The-weight-Kalman}.
KFHE-l was found to be statistically significantly better than S-GBM,
and Bagging at all noise levels except $20\%$.
KFHE-l was also found to be statistically significantly better than AdaBoost at the $10\%$, $15\%$ and $20\%$ noise levels.
Although the performance of KFHE-e decreases with increasing class-label
noise, it does not decrease as sharply as AdaBoost.
The complete details of the tests are given in Table \ref{tab:Friedman-pairwise} in \ref{sec:statsTables}.

Overall these tests confirm that when no class-label noise is present KFHE-e performs as well as AdaBoost and GBM, but significantly better than
S-GBM, Bagging and CART. KFHE-e, however, is not as robust to class
label noise as the other approaches. KFHE-l, on the other hand, is
robust to noise and performs very well in all class-label noise
settings.

\subsubsection{Isolated algorithm pairs comparison}

To further understand how individual algorithm pairs compare with each other,
ignoring other algorithms, a two tailed Wilcoxon's Signed Rank Sum
test for each pair of algorithms was performed. It must be emphasised that the Wilcoxon's rank sum test \emph{cannot}
be used to perform multiple classifier comparison without introducing
Type I error (rejecting the null hypothesis when it cannot be rejected),
as it does not control the Family Wise Error Rate (FWER) \citep{GARCIA20102044}.
Therefore, the $p$-values for each pair from this experiment should
\emph{only} be interpreted in isolation from any other algorithms. Table \ref{tab:Wilcox-pairwise-intext} shows the results of these tests for the datasets without any class-label noise (Tables \ref{tab:Wilcox-pairwise-00} to \ref{tab:Wilcox-pairwise-04}
in \ref{sec:statsTables} show the results for the noisy cases).  The cells in the lower diagonal show the $p$-values of the Wilcoxon's Signed Rank Sum test for corresponding algorithm pair  and the cells in the upper diagonal show the pairwise win/lost/tie counts.

\begin{table}
\caption{Result of pairwise Wilcoxon's Signed Rank Sum test over the different
datasets when no class-label noise is present. Upper diagonal: win/lose/tie.
Lower diagonal: Wilcoxon's Signed Rank Sum Test $p$-values. {*} $\alpha=0.1$, {*}{*} $\alpha=0.05$
and {*}{*}{*} $\alpha=0.01$ \label{tab:Wilcoxon-pairwise}. \label{tab:Wilcox-pairwise-intext}}

\centering \resizebox{\textwidth}{!}{%
\begin{tabular}{rlllllll}
\hline 
 & KFHE-e  & KFHE-l  & AdaBoost  & GBM  & S-GBM  & Bagging  & CART \tabularnewline
\hline 
KFHE-e  &  & (19/11/0)  & (13/16/1)  & (23/7/0)  & (21/9/0)  & (24/6/0)  & (26/4/0) \tabularnewline
KFHE-l  & 0.009519 {*}{*}{*} &  & (12/18/0)  & (16/14/0)  & (20/10/0)  & (25/5/0)  & (26/4/0) \tabularnewline
AdaBoost  & 0.491795  & 0.028548 {*}{*} &  & (19/11/0)  & (20/10/0)  & (22/8/0)  & (25/5/0) \tabularnewline
GBM  & 0.003018 {*}{*}{*} & 0.144739  & 0.013515 {*}{*} &  & (22/8/0)  & (20/10/0)  & (25/5/0) \tabularnewline
S-GBM  & 0.001128 {*}{*}{*} & 0.004921 {*}{*}{*} & 0.002834 {*}{*}{*} & 0.003418 {*}{*}{*} &  & (19/11/0)  & (25/5/0) \tabularnewline
Bagging  & 0.000210 {*}{*}{*} & 0.000034 {*}{*}{*} & 0.000415 {*}{*}{*} & 0.004108 {*}{*}{*} & 0.336640 &  & (25/4/1) \tabularnewline
CART  & 0.000010 {*}{*}{*} & 0.000006 {*}{*}{*} & 0.000019 {*}{*}{*} & 0.000055 {*}{*}{*} & 0.002057 {*}{*}{*}  & 0.000016 {*}{*}{*} & \tabularnewline
\hline 
\end{tabular}}
\end{table}

The results in Table \ref{tab:Wilcox-pairwise-intext} show that without class-label noise when compared in isolation KFHE-e performs significantly better than any other method, except AdaBoost. In the noise free case KFHE-l performs significantly better than S-GBM, Bagging and CART. Similarly, the test results at different noise levels (described in  \ref{sec:statsTables}) show that as class-label noise increases, the performance
of KFHE-e starts to become significantly better than AdaBoost, although it is worse than other methods. When compared in isolation KFHE-l performs significantly better than almost all other methods at all noise levels.

\section{Conclusion and future work\label{sec:Discussion-and-Conclusion}}

This paper introduces a new perspective on training multi-class ensemble
classification models. The ensemble classifier model is viewed as
a state to be estimated, and this state is estimated using a Kalman
filter. Unlike more common applications of Kalman filters to time
series data, this work exploits the sensor fusion property of the
Kalman filter to combine multiple individual multi-class classifiers
to build a multi-class ensemble classifier algorithm. Based on this
new perspective a new multi-class ensemble classification algorithm,
the Kalman Filter-based Heuristic Ensemble (KFHE), is proposed.

Detailed experiments on two slight variants of KFHE, KFHE-e and KFHE-l,
were performed. KFHE-e is more effective on non-noisy class-labels,
as it emphasises the misclassified training datapoints from one iteration
of the training algorithm to the next, and KFHE-l is more effective
on noisy class-labels as it does not emphasise misclassified training
datapoints as much. Experiments show that KFHE-e and KFHE-l perform
at least as well as, and in many cases, better than Bagging, S-GBM,
GBM and AdaBoost. For datasets with noisy class labels, KFHE-l performed
significantly better than all other methods across different levels
of class-label noise. For these datasets KFHE-e performed more poorly
than KFHE-l, GBM, and S-GBM, but better than AdaBoost.

KFHE can be seen as a hybrid ensemble approach mixing the benefits
of both bagging and a boosting. Bagging weighs each of the component
learner's votes equally, whereas boosting finds the optimum weights,
using which the component learners are combined. KFHE does not find
the optimum weights analytically as AdaBoost does, but attempts to
combine the classifiers based on how well the measurement is in a
given iteration.

Given the new perspective, other implementations that expand upon
KFHE can also be designed following the framework and methods described
in Sections \ref{sec:The-new-perspective} and \ref{sec:Proposed-Method}.
In future, it would be interesting to pursue the following studies:
\begin{itemize}
\item The effect when process noise and a linear time update step
are introduced. 

\item Multiple and different types of measurements can also be performed.
That is, instead of having one component classifier model per iteration,
more than one classifier model could be used. This is analogous to
having multiple noisy sensors measuring the DC voltage level of the
toy example presented in Section \ref{subsec:A-static-estimation}.

\item To further study the effects of other types of noise (class-wise label
noise, noise in input space, etc.), higher levels of noise induced
in the class-label assignments, and performance on imbalanced class
datasets.
\end{itemize}

\section*{Acknowledgements}

This research was supported by Science Foundation Ireland (SFI) under
Grant number SFI/12/RC/2289. The authors would like to thank Gevorg
Poghosyan, PhD Research Student at Insight Centre for Data Analytics,
School of Computer Science, University College Dublin, for feedback
and discussions which led to the state space representation in Figure
\ref{fig:h_space_diag}. Also, the authors would like to thank the
unnamed reviewers for their detailed and constructive comments which
helped to significantly improve the quality of the paper.

\pagebreak 

\appendix

\section{Complete experiment results\label{sec:resultsTables}}

\begin{landscape}

\begin{table*}
\caption{Each cell in the table shows the mean and standard deviation of the
$F_{1}^{(macro)}$-score (higher value is better) for the $20$ times
$4$-fold cross-validation experiment for each algorithm and each
of the datasets listed in Table \ref{tab:Dataset-descriptions-used}.
The values in parenthesis are the relative rankings of the algorithms
on the dataset in the corresponding row (lower ranks are better).
\label{tab:Average-F-Scores-00}}
\centering{}\centering\resizebox{\paperwidth}{!}{ %
\begin{tabular}{rrrrrrrr}
\hline 
 & KFHE-e  & KFHE-l  & AdaBoost  & GBM  & S-GBM  & Bagging  & CART \tabularnewline
\hline 
mushroom  & 1.0000 $\pm$ 0.00 (1.5)  & 0.9968 $\pm$ 0.00 (5)  & 1.0000 $\pm$ 0.00 (1.5)  & 0.9997 $\pm$ 0.00 (3)  & 0.9990 $\pm$ 0.00 (4)  & 0.9941 $\pm$ 0.00 (6.5)  & 0.9941 $\pm$ 0.00 (6.5) \tabularnewline
iris  & 0.9433 $\pm$ 0.03 (4)  & 0.9487 $\pm$ 0.03 (1)  & 0.9448 $\pm$ 0.03 (2)  & 0.9403 $\pm$ 0.04 (5)  & 0.9437 $\pm$ 0.03 (3)  & 0.9376 $\pm$ 0.03 (6)  & 0.9298 $\pm$ 0.03 (7) \tabularnewline
glass  & 0.7125 $\pm$ 0.08 (3)  & 0.7153 $\pm$ 0.07 (1)  & 0.7144 $\pm$ 0.07 (2)  & 0.6666 $\pm$ 0.08 (4)  & 0.5695 $\pm$ 0.09 (6)  & 0.5965 $\pm$ 0.09 (5)  & 0.5466 $\pm$ 0.06 (7) \tabularnewline
car\_eval  & 0.9653 $\pm$ 0.02 (2)  & 0.9011 $\pm$ 0.03 (4)  & 0.9665 $\pm$ 0.02 (1)  & 0.9131 $\pm$ 0.04 (3)  & 0.8236 $\pm$ 0.04 (7)  & 0.8569 $\pm$ 0.04 (5)  & 0.8546 $\pm$ 0.04 (6) \tabularnewline
cmc  & 0.5222 $\pm$ 0.02 (5)  & 0.5280 $\pm$ 0.03 (2)  & 0.5038 $\pm$ 0.02 (7)  & 0.5270 $\pm$ 0.02 (4)  & 0.5275 $\pm$ 0.02 (3)  & 0.5291 $\pm$ 0.03 (1)  & 0.5187 $\pm$ 0.03 (6) \tabularnewline
tvowel  & 0.8451 $\pm$ 0.02 (2)  & 0.8283 $\pm$ 0.03 (3)  & 0.8279 $\pm$ 0.02 (4)  & 0.8458 $\pm$ 0.03 (1)  & 0.8236 $\pm$ 0.03 (5)  & 0.8004 $\pm$ 0.03 (6)  & 0.7855 $\pm$ 0.03 (7) \tabularnewline
balance\_scale  & 0.6345 $\pm$ 0.03 (1)  & 0.5984 $\pm$ 0.01 (4)  & 0.6186 $\pm$ 0.03 (2)  & 0.5935 $\pm$ 0.02 (5)  & 0.6045 $\pm$ 0.01 (3)  & 0.5861 $\pm$ 0.02 (6)  & 0.5412 $\pm$ 0.02 (7) \tabularnewline
flags  & 0.3059 $\pm$ 0.05 (3)  & 0.2771 $\pm$ 0.05 (4)  & 0.3187 $\pm$ 0.06 (2)  & 0.3236 $\pm$ 0.06 (1)  & 0.2602 $\pm$ 0.03 (5)  & 0.2525 $\pm$ 0.03 (6)  & 0.2439 $\pm$ 0.03 (7) \tabularnewline
german  & 0.6907 $\pm$ 0.03 (2)  & 0.6960 $\pm$ 0.02 (1)  & 0.6837 $\pm$ 0.03 (5)  & 0.6860 $\pm$ 0.03 (3)  & 0.6852 $\pm$ 0.03 (4)  & 0.6826 $\pm$ 0.02 (6)  & 0.6550 $\pm$ 0.03 (7) \tabularnewline
ilpd  & 0.6126 $\pm$ 0.04 (2)  & 0.5797 $\pm$ 0.04 (5)  & 0.6153 $\pm$ 0.04 (1)  & 0.5809 $\pm$ 0.04 (4)  & 0.5733 $\pm$ 0.04 (6)  & 0.5675 $\pm$ 0.04 (7)  & 0.5865 $\pm$ 0.04 (3) \tabularnewline
ionosphere  & 0.9238 $\pm$ 0.03 (2)  & 0.9157 $\pm$ 0.03 (4)  & 0.9298 $\pm$ 0.02 (1)  & 0.9179 $\pm$ 0.03 (3)  & 0.9105 $\pm$ 0.03 (5)  & 0.9004 $\pm$ 0.03 (6)  & 0.8617 $\pm$ 0.03 (7) \tabularnewline
knowledge  & 0.9354 $\pm$ 0.03 (2)  & 0.9315 $\pm$ 0.03 (3)  & 0.9524 $\pm$ 0.02 (1)  & 0.9155 $\pm$ 0.03 (6)  & 0.8925 $\pm$ 0.04 (7)  & 0.9184 $\pm$ 0.03 (4)  & 0.9160 $\pm$ 0.03 (5) \tabularnewline
vertebral  & 0.8036 $\pm$ 0.04 (4)  & 0.8090 $\pm$ 0.04 (2)  & 0.8001 $\pm$ 0.04 (6)  & 0.8030 $\pm$ 0.04 (5)  & 0.8130 $\pm$ 0.05 (1)  & 0.8042 $\pm$ 0.04 (3)  & 0.7857 $\pm$ 0.04 (7) \tabularnewline
sonar  & 0.8072 $\pm$ 0.05 (2)  & 0.7836 $\pm$ 0.06 (4)  & 0.8371 $\pm$ 0.05 (1)  & 0.7893 $\pm$ 0.06 (3)  & 0.7797 $\pm$ 0.06 (5)  & 0.7766 $\pm$ 0.06 (6)  & 0.7021 $\pm$ 0.05 (7) \tabularnewline
skulls  & 0.2358 $\pm$ 0.06 (5)  & 0.2380 $\pm$ 0.06 (3)  & 0.2362 $\pm$ 0.06 (4)  & 0.2514 $\pm$ 0.08 (1)  & 0.2436 $\pm$ 0.06 (2)  & 0.2300 $\pm$ 0.06 (7)  & 0.2309 $\pm$ 0.07 (6) \tabularnewline
diabetes  & 0.9558 $\pm$ 0.03 (7)  & 0.9647 $\pm$ 0.03 (6)  & 0.9658 $\pm$ 0.03 (5)  & 0.9725 $\pm$ 0.02 (2)  & 0.9722 $\pm$ 0.02 (3)  & 0.9727 $\pm$ 0.03 (1)  & 0.9710 $\pm$ 0.03 (4) \tabularnewline
physio  & 0.9069 $\pm$ 0.02 (4)  & 0.9109 $\pm$ 0.03 (2)  & 0.9079 $\pm$ 0.02 (3)  & 0.9046 $\pm$ 0.02 (5)  & 0.9136 $\pm$ 0.03 (1)  & 0.8959 $\pm$ 0.03 (6)  & 0.8847 $\pm$ 0.03 (7) \tabularnewline
breasttissue  & 0.6766 $\pm$ 0.08 (1)  & 0.6711 $\pm$ 0.08 (2)  & 0.6606 $\pm$ 0.08 (4)  & 0.6605 $\pm$ 0.07 (5)  & 0.6347 $\pm$ 0.08 (6)  & 0.6653 $\pm$ 0.08 (3)  & 0.6338 $\pm$ 0.08 (7) \tabularnewline
bupa  & 0.7027 $\pm$ 0.04 (2)  & 0.7114 $\pm$ 0.04 (1)  & 0.6926 $\pm$ 0.04 (6)  & 0.7018 $\pm$ 0.04 (3)  & 0.6944 $\pm$ 0.04 (5)  & 0.6954 $\pm$ 0.04 (4)  & 0.6433 $\pm$ 0.05 (7) \tabularnewline
cleveland  & 0.2975 $\pm$ 0.05 (2)  & 0.2845 $\pm$ 0.04 (5)  & 0.3058 $\pm$ 0.04 (1)  & 0.2938 $\pm$ 0.04 (3)  & 0.2865 $\pm$ 0.04 (4)  & 0.2736 $\pm$ 0.04 (7)  & 0.2766 $\pm$ 0.04 (6) \tabularnewline
haberman  & 0.5504 $\pm$ 0.05 (6)  & 0.5743 $\pm$ 0.05 (5)  & 0.5465 $\pm$ 0.05 (7)  & 0.5751 $\pm$ 0.05 (4)  & 0.5996 $\pm$ 0.04 (1)  & 0.5757 $\pm$ 0.05 (3)  & 0.5772 $\pm$ 0.05 (2) \tabularnewline
hayes\_roth  & 0.8602 $\pm$ 0.05 (1)  & 0.8491 $\pm$ 0.05 (3)  & 0.8510 $\pm$ 0.04 (2)  & 0.6094 $\pm$ 0.08 (6)  & 0.5683 $\pm$ 0.10 (7)  & 0.7418 $\pm$ 0.10 (4)  & 0.6691 $\pm$ 0.10 (5) \tabularnewline
monks  & 0.9997 $\pm$ 0.00 (2)  & 0.9981 $\pm$ 0.01 (3)  & 1.0000 $\pm$ 0.00 (1)  & 0.9671 $\pm$ 0.06 (4)  & 0.9114 $\pm$ 0.06 (5)  & 0.9002 $\pm$ 0.06 (6)  & 0.8178 $\pm$ 0.09 (7) \tabularnewline
newthyroid  & 0.3973 $\pm$ 0.04 (4)  & 0.3867 $\pm$ 0.04 (6)  & 0.3972 $\pm$ 0.04 (5)  & 0.3742 $\pm$ 0.04 (7)  & 0.4162 $\pm$ 0.04 (2)  & 0.4275 $\pm$ 0.04 (1)  & 0.4087 $\pm$ 0.11 (3) \tabularnewline
yeast  & 0.5339 $\pm$ 0.05 (1)  & 0.4701 $\pm$ 0.03 (4)  & 0.5209 $\pm$ 0.05 (3)  & 0.5225 $\pm$ 0.05 (2)  & 0.4359 $\pm$ 0.02 (5)  & 0.4187 $\pm$ 0.03 (6)  & 0.4069 $\pm$ 0.03 (7) \tabularnewline
spam  & 0.9477 $\pm$ 0.01 (2)  & 0.9256 $\pm$ 0.01 (4)  & 0.9508 $\pm$ 0.01 (1)  & 0.9309 $\pm$ 0.01 (3)  & 0.9225 $\pm$ 0.01 (5)  & 0.9029 $\pm$ 0.01 (6)  & 0.8870 $\pm$ 0.01 (7) \tabularnewline
lymphography  & 0.6733 $\pm$ 0.19 (2)  & 0.5074 $\pm$ 0.13 (3)  & 0.7089 $\pm$ 0.18 (1)  & 0.4454 $\pm$ 0.10 (4)  & 0.3973 $\pm$ 0.03 (5)  & 0.3966 $\pm$ 0.03 (6)  & 0.3704 $\pm$ 0.04 (7) \tabularnewline
movement\_libras  & 0.7772 $\pm$ 0.04 (1)  & 0.7488 $\pm$ 0.05 (3)  & 0.7679 $\pm$ 0.04 (2)  & 0.6434 $\pm$ 0.05 (5)  & 0.6190 $\pm$ 0.05 (6)  & 0.6715 $\pm$ 0.05 (4)  & 0.5176 $\pm$ 0.05 (7) \tabularnewline
SAheart  & 0.6214 $\pm$ 0.04 (6)  & 0.6408 $\pm$ 0.04 (4)  & 0.6090 $\pm$ 0.04 (7)  & 0.6436 $\pm$ 0.04 (3)  & 0.6509 $\pm$ 0.03 (1)  & 0.6466 $\pm$ 0.04 (2)  & 0.6237 $\pm$ 0.05 (5) \tabularnewline
zoo  & 0.8548 $\pm$ 0.12 (2)  & 0.8490 $\pm$ 0.11 (3)  & 0.8740 $\pm$ 0.11 (1)  & 0.8450 $\pm$ 0.10 (4)  & 0.5455 $\pm$ 0.11 (7)  & 0.5922 $\pm$ 0.09 (5)  & 0.5840 $\pm$ 0.05 (6) \tabularnewline
\hline 
Average rank  & \textbf{2.78}  & 3.33  & 2.98  & 3.7  & 4.3  & 4.82  & 6.08 \tabularnewline
\hline 
\end{tabular}} 
\end{table*}

\end{landscape}

\begin{landscape}

\begin{table}[p]
\caption{Noise Level: $5\%$ . Each cell in the table shows the $F_{1}^{(macro)}$-measure
(higher value is better) from the $20$ times $4$-fold cross-validation
experiment with $5\%$ noise induced on the class labels. The values
in parenthesis is the relative ranking of the algorithm on the dataset
in the corresponding row (lower ranks are better). \label{tab:noise05}}
\centering{}\centering\resizebox{\paperwidth}{!}{ %
\begin{tabular}{rrrrrrrr}
\hline 
 & KFHE-e  & KFHE-l  & AdaBoost  & GBM  & S-GBM  & Bagging  & CART \tabularnewline
\hline 
mushroom  & 0.9972 $\pm$ 0.00 (4)  & 0.9941 $\pm$ 0.00 (6)  & 0.9993 $\pm$ 0.00 (2)  & 0.9997 $\pm$ 0.00 (1)  & 0.9990 $\pm$ 0.00 (3)  & 0.9941 $\pm$ 0.00 (6)  & 0.9941 $\pm$ 0.00 (6) \tabularnewline
iris  & 0.9205 $\pm$ 0.05 (7)  & 0.9413 $\pm$ 0.03 (3)  & 0.9282 $\pm$ 0.04 (6)  & 0.9359 $\pm$ 0.03 (4)  & 0.9486 $\pm$ 0.03 (1)  & 0.9436 $\pm$ 0.03 (2)  & 0.9335 $\pm$ 0.03 (5) \tabularnewline
glass  & 0.6818 $\pm$ 0.09 (2)  & 0.6969 $\pm$ 0.07 (1)  & 0.6597 $\pm$ 0.08 (3)  & 0.6498 $\pm$ 0.08 (4)  & 0.5532 $\pm$ 0.08 (6)  & 0.5876 $\pm$ 0.10 (5)  & 0.5367 $\pm$ 0.06 (7) \tabularnewline
car\_eval  & 0.8918 $\pm$ 0.03 (1)  & 0.8639 $\pm$ 0.03 (4)  & 0.8765 $\pm$ 0.03 (3)  & 0.8820 $\pm$ 0.04 (2)  & 0.8047 $\pm$ 0.04 (7)  & 0.8451 $\pm$ 0.03 (5)  & 0.8423 $\pm$ 0.03 (6) \tabularnewline
cmc  & 0.5223 $\pm$ 0.02 (5)  & 0.5285 $\pm$ 0.02 (3)  & 0.5047 $\pm$ 0.03 (7)  & 0.5303 $\pm$ 0.02 (2)  & 0.5274 $\pm$ 0.02 (4)  & 0.5330 $\pm$ 0.02 (1)  & 0.5197 $\pm$ 0.03 (6) \tabularnewline
tvowel  & 0.8346 $\pm$ 0.03 (2)  & 0.8275 $\pm$ 0.03 (4)  & 0.7903 $\pm$ 0.03 (6)  & 0.8435 $\pm$ 0.03 (1)  & 0.8329 $\pm$ 0.03 (3)  & 0.7961 $\pm$ 0.03 (5)  & 0.7805 $\pm$ 0.04 (7) \tabularnewline
balance\_scale  & 0.5989 $\pm$ 0.03 (1)  & 0.5940 $\pm$ 0.02 (3)  & 0.5917 $\pm$ 0.03 (4)  & 0.5912 $\pm$ 0.02 (5)  & 0.5982 $\pm$ 0.02 (2)  & 0.5799 $\pm$ 0.02 (6)  & 0.5418 $\pm$ 0.02 (7) \tabularnewline
flags  & 0.3113 $\pm$ 0.06 (3)  & 0.2988 $\pm$ 0.05 (4)  & 0.3193 $\pm$ 0.06 (1)  & 0.3185 $\pm$ 0.06 (2)  & 0.2678 $\pm$ 0.04 (5)  & 0.2518 $\pm$ 0.03 (6)  & 0.2420 $\pm$ 0.04 (7) \tabularnewline
german  & 0.6732 $\pm$ 0.03 (2)  & 0.6765 $\pm$ 0.03 (1)  & 0.6690 $\pm$ 0.03 (4)  & 0.6695 $\pm$ 0.03 (3)  & 0.6655 $\pm$ 0.03 (5)  & 0.6608 $\pm$ 0.03 (6)  & 0.6348 $\pm$ 0.04 (7) \tabularnewline
ilpd  & 0.6130 $\pm$ 0.04 (2)  & 0.5874 $\pm$ 0.04 (3)  & 0.6220 $\pm$ 0.04 (1)  & 0.5815 $\pm$ 0.04 (4)  & 0.5755 $\pm$ 0.04 (6)  & 0.5724 $\pm$ 0.04 (7)  & 0.5759 $\pm$ 0.04 (5) \tabularnewline
ionosphere  & 0.9093 $\pm$ 0.03 (2)  & 0.9087 $\pm$ 0.03 (4)  & 0.9090 $\pm$ 0.03 (3)  & 0.9018 $\pm$ 0.03 (6)  & 0.9103 $\pm$ 0.03 (1)  & 0.9023 $\pm$ 0.03 (5)  & 0.8507 $\pm$ 0.04 (7) \tabularnewline
knowledge  & 0.9360 $\pm$ 0.02 (2)  & 0.9300 $\pm$ 0.02 (3)  & 0.9369 $\pm$ 0.02 (1)  & 0.9188 $\pm$ 0.03 (4)  & 0.8924 $\pm$ 0.03 (7)  & 0.9177 $\pm$ 0.03 (6)  & 0.9181 $\pm$ 0.03 (5) \tabularnewline
vertebral  & 0.7928 $\pm$ 0.05 (5)  & 0.8073 $\pm$ 0.05 (3)  & 0.7740 $\pm$ 0.05 (6)  & 0.8024 $\pm$ 0.05 (4)  & 0.8163 $\pm$ 0.04 (1.5)  & 0.8163 $\pm$ 0.05 (1.5)  & 0.7736 $\pm$ 0.05 (7) \tabularnewline
sonar  & 0.7900 $\pm$ 0.05 (2)  & 0.7759 $\pm$ 0.05 (3)  & 0.8116 $\pm$ 0.05 (1)  & 0.7719 $\pm$ 0.06 (4)  & 0.7708 $\pm$ 0.05 (5)  & 0.7610 $\pm$ 0.05 (6)  & 0.6867 $\pm$ 0.06 (7) \tabularnewline
skulls  & 0.2462 $\pm$ 0.07 (3)  & 0.2226 $\pm$ 0.06 (6)  & 0.2275 $\pm$ 0.06 (5)  & 0.2550 $\pm$ 0.07 (1)  & 0.2510 $\pm$ 0.07 (2)  & 0.2295 $\pm$ 0.06 (4)  & 0.1935 $\pm$ 0.06 (7) \tabularnewline
diabetes  & 0.9364 $\pm$ 0.04 (6)  & 0.9731 $\pm$ 0.03 (1)  & 0.9305 $\pm$ 0.04 (7)  & 0.9722 $\pm$ 0.02 (2)  & 0.9705 $\pm$ 0.02 (5)  & 0.9710 $\pm$ 0.03 (3)  & 0.9709 $\pm$ 0.03 (4) \tabularnewline
physio  & 0.8781 $\pm$ 0.03 (5)  & 0.9092 $\pm$ 0.02 (2)  & 0.8712 $\pm$ 0.03 (6)  & 0.8995 $\pm$ 0.02 (3)  & 0.9113 $\pm$ 0.02 (1)  & 0.8955 $\pm$ 0.03 (4)  & 0.8658 $\pm$ 0.03 (7) \tabularnewline
breasttissue  & 0.6557 $\pm$ 0.07 (2)  & 0.6709 $\pm$ 0.07 (1)  & 0.6511 $\pm$ 0.08 (3)  & 0.6502 $\pm$ 0.08 (4)  & 0.6285 $\pm$ 0.08 (6)  & 0.6416 $\pm$ 0.08 (5)  & 0.5927 $\pm$ 0.08 (7) \tabularnewline
bupa  & 0.6852 $\pm$ 0.04 (2)  & 0.6962 $\pm$ 0.04 (1)  & 0.6625 $\pm$ 0.05 (6)  & 0.6839 $\pm$ 0.04 (4)  & 0.6846 $\pm$ 0.04 (3)  & 0.6795 $\pm$ 0.05 (5)  & 0.6309 $\pm$ 0.05 (7) \tabularnewline
cleveland  & 0.2895 $\pm$ 0.05 (4)  & 0.2906 $\pm$ 0.05 (3)  & 0.3076 $\pm$ 0.05 (1)  & 0.2922 $\pm$ 0.05 (2)  & 0.2883 $\pm$ 0.05 (5)  & 0.2793 $\pm$ 0.04 (7)  & 0.2864 $\pm$ 0.05 (6) \tabularnewline
haberman  & 0.5429 $\pm$ 0.05 (6)  & 0.5554 $\pm$ 0.06 (5)  & 0.5342 $\pm$ 0.05 (7)  & 0.5665 $\pm$ 0.06 (3)  & 0.5793 $\pm$ 0.06 (1)  & 0.5643 $\pm$ 0.06 (4)  & 0.5738 $\pm$ 0.06 (2) \tabularnewline
hayes\_roth  & 0.8022 $\pm$ 0.07 (2)  & 0.8289 $\pm$ 0.06 (1)  & 0.7815 $\pm$ 0.07 (3)  & 0.5869 $\pm$ 0.09 (6)  & 0.5208 $\pm$ 0.09 (7)  & 0.7145 $\pm$ 0.10 (4)  & 0.6695 $\pm$ 0.10 (5) \tabularnewline
monks  & 0.9644 $\pm$ 0.02 (2)  & 0.9985 $\pm$ 0.00 (1)  & 0.9311 $\pm$ 0.03 (5)  & 0.9473 $\pm$ 0.06 (3)  & 0.9268 $\pm$ 0.06 (6)  & 0.9379 $\pm$ 0.06 (4)  & 0.8498 $\pm$ 0.09 (7) \tabularnewline
newthyroid  & 0.3972 $\pm$ 0.04 (5)  & 0.3962 $\pm$ 0.04 (6)  & 0.4039 $\pm$ 0.04 (4)  & 0.3960 $\pm$ 0.04 (7)  & 0.4475 $\pm$ 0.04 (1)  & 0.4395 $\pm$ 0.03 (2)  & 0.4086 $\pm$ 0.11 (3) \tabularnewline
yeast  & 0.4797 $\pm$ 0.06 (2)  & 0.4354 $\pm$ 0.03 (4)  & 0.4521 $\pm$ 0.07 (3)  & 0.4829 $\pm$ 0.05 (1)  & 0.4312 $\pm$ 0.03 (5)  & 0.4176 $\pm$ 0.02 (6)  & 0.4021 $\pm$ 0.03 (7) \tabularnewline
spam  & 0.9325 $\pm$ 0.01 (1)  & 0.9265 $\pm$ 0.01 (4)  & 0.9311 $\pm$ 0.01 (2)  & 0.9294 $\pm$ 0.01 (3)  & 0.9219 $\pm$ 0.01 (5)  & 0.9039 $\pm$ 0.01 (6)  & 0.8866 $\pm$ 0.01 (7) \tabularnewline
lymphography  & 0.6436 $\pm$ 0.16 (2)  & 0.4896 $\pm$ 0.12 (3)  & 0.6507 $\pm$ 0.15 (1)  & 0.4402 $\pm$ 0.09 (4)  & 0.3954 $\pm$ 0.04 (5)  & 0.3941 $\pm$ 0.04 (6)  & 0.3704 $\pm$ 0.05 (7) \tabularnewline
movement\_libras  & 0.7365 $\pm$ 0.04 (1)  & 0.7138 $\pm$ 0.05 (3)  & 0.7362 $\pm$ 0.04 (2)  & 0.6064 $\pm$ 0.06 (5)  & 0.5868 $\pm$ 0.06 (6)  & 0.6517 $\pm$ 0.06 (4)  & 0.5008 $\pm$ 0.05 (7) \tabularnewline
SAheart  & 0.6120 $\pm$ 0.04 (5)  & 0.6252 $\pm$ 0.04 (4)  & 0.6020 $\pm$ 0.04 (7)  & 0.6255 $\pm$ 0.04 (3)  & 0.6446 $\pm$ 0.03 (1)  & 0.6405 $\pm$ 0.04 (2)  & 0.6106 $\pm$ 0.05 (6) \tabularnewline
zoo  & 0.7765 $\pm$ 0.11 (4)  & 0.8025 $\pm$ 0.13 (2)  & 0.7841 $\pm$ 0.12 (3)  & 0.8058 $\pm$ 0.12 (1)  & 0.5566 $\pm$ 0.11 (7)  & 0.5823 $\pm$ 0.09 (5)  & 0.5704 $\pm$ 0.07 (6) \tabularnewline
\hline 
Average rank  & \textbf{3.07}  & \textbf{3.07}  & 3.77  & 3.27  & 4.08  & 4.62  & 6.13 \tabularnewline
\hline 
\end{tabular}} 
\end{table}

\end{landscape}

\begin{landscape}

\begin{table}[p]
\caption{Noise Level: $10\%$ . Each cell in the table shows the $F_{1}^{(macro)}$-measure
(higher value is better) from the $20$ times $4$-fold cross-validation
experiment with $10\%$ noise induced on the class labels. The values
in parenthesis is the relative ranking of the algorithm on the dataset
in the corresponding row (lower ranks are better). \label{tab:noise10}}
\centering{}\centering\resizebox{\paperwidth}{!}{ %
\begin{tabular}{rrrrrrrr}
\hline 
 & KFHE-e  & KFHE-l  & AdaBoost  & GBM  & S-GBM  & Bagging  & CART \tabularnewline
\hline 
mushroom  & 0.9942 $\pm$ 0.00 (4)  & 0.9941 $\pm$ 0.00 (6)  & 0.9970 $\pm$ 0.00 (3)  & 0.9988 $\pm$ 0.00 (1)  & 0.9987 $\pm$ 0.00 (2)  & 0.9941 $\pm$ 0.00 (6)  & 0.9941 $\pm$ 0.00 (6) \tabularnewline
iris  & 0.8749 $\pm$ 0.06 (6)  & 0.9384 $\pm$ 0.04 (3)  & 0.8569 $\pm$ 0.06 (7)  & 0.9319 $\pm$ 0.05 (5)  & 0.9487 $\pm$ 0.03 (1)  & 0.9433 $\pm$ 0.03 (2)  & 0.9380 $\pm$ 0.03 (4) \tabularnewline
glass  & 0.6850 $\pm$ 0.09 (2)  & 0.6990 $\pm$ 0.08 (1)  & 0.6801 $\pm$ 0.07 (3)  & 0.6253 $\pm$ 0.08 (4)  & 0.5658 $\pm$ 0.08 (6)  & 0.6189 $\pm$ 0.09 (5)  & 0.5641 $\pm$ 0.10 (7) \tabularnewline
car\_eval  & 0.8660 $\pm$ 0.04 (1)  & 0.8621 $\pm$ 0.03 (2)  & 0.7311 $\pm$ 0.04 (7)  & 0.8411 $\pm$ 0.05 (3)  & 0.7421 $\pm$ 0.05 (6)  & 0.8111 $\pm$ 0.04 (4)  & 0.7901 $\pm$ 0.04 (5) \tabularnewline
cmc  & 0.5133 $\pm$ 0.02 (6)  & 0.5232 $\pm$ 0.02 (1)  & 0.4886 $\pm$ 0.02 (7)  & 0.5220 $\pm$ 0.02 (2)  & 0.5212 $\pm$ 0.02 (4)  & 0.5217 $\pm$ 0.02 (3)  & 0.5175 $\pm$ 0.03 (5) \tabularnewline
tvowel  & 0.8264 $\pm$ 0.03 (2)  & 0.8187 $\pm$ 0.03 (4)  & 0.7747 $\pm$ 0.04 (7)  & 0.8383 $\pm$ 0.03 (1)  & 0.8238 $\pm$ 0.03 (3)  & 0.7961 $\pm$ 0.03 (5)  & 0.7791 $\pm$ 0.03 (6) \tabularnewline
balance\_scale  & 0.6016 $\pm$ 0.03 (2)  & 0.5939 $\pm$ 0.02 (3)  & 0.5929 $\pm$ 0.03 (4)  & 0.5912 $\pm$ 0.02 (5)  & 0.6024 $\pm$ 0.02 (1)  & 0.5757 $\pm$ 0.02 (6)  & 0.5332 $\pm$ 0.02 (7) \tabularnewline
flags  & 0.2716 $\pm$ 0.05 (4)  & 0.2544 $\pm$ 0.03 (6)  & 0.2860 $\pm$ 0.04 (2)  & 0.2885 $\pm$ 0.06 (1)  & 0.2763 $\pm$ 0.03 (3)  & 0.2577 $\pm$ 0.03 (5)  & 0.2484 $\pm$ 0.04 (7) \tabularnewline
german  & 0.6698 $\pm$ 0.03 (4)  & 0.6786 $\pm$ 0.03 (1)  & 0.6611 $\pm$ 0.03 (6)  & 0.6695 $\pm$ 0.03 (5)  & 0.6756 $\pm$ 0.03 (2)  & 0.6706 $\pm$ 0.03 (3)  & 0.6348 $\pm$ 0.04 (7) \tabularnewline
ilpd  & 0.5836 $\pm$ 0.04 (2)  & 0.5782 $\pm$ 0.04 (3)  & 0.5849 $\pm$ 0.05 (1)  & 0.5737 $\pm$ 0.04 (4)  & 0.5696 $\pm$ 0.04 (5)  & 0.5650 $\pm$ 0.04 (7)  & 0.5694 $\pm$ 0.05 (6) \tabularnewline
ionosphere  & 0.8922 $\pm$ 0.04 (5)  & 0.9098 $\pm$ 0.03 (1)  & 0.8867 $\pm$ 0.04 (6)  & 0.9048 $\pm$ 0.03 (4)  & 0.9095 $\pm$ 0.03 (2)  & 0.9093 $\pm$ 0.03 (3)  & 0.8486 $\pm$ 0.04 (7) \tabularnewline
knowledge  & 0.9132 $\pm$ 0.03 (4)  & 0.9300 $\pm$ 0.02 (1)  & 0.9122 $\pm$ 0.03 (5)  & 0.9191 $\pm$ 0.03 (2)  & 0.8906 $\pm$ 0.03 (7)  & 0.9111 $\pm$ 0.02 (6)  & 0.9156 $\pm$ 0.03 (3) \tabularnewline
vertebral  & 0.7904 $\pm$ 0.05 (5)  & 0.7987 $\pm$ 0.04 (3)  & 0.7749 $\pm$ 0.04 (7)  & 0.7949 $\pm$ 0.05 (4)  & 0.8099 $\pm$ 0.05 (1)  & 0.8059 $\pm$ 0.05 (2)  & 0.7838 $\pm$ 0.05 (6) \tabularnewline
sonar  & 0.7818 $\pm$ 0.05 (2)  & 0.7719 $\pm$ 0.06 (3)  & 0.7947 $\pm$ 0.05 (1)  & 0.7588 $\pm$ 0.06 (5)  & 0.7702 $\pm$ 0.06 (4)  & 0.7505 $\pm$ 0.06 (6)  & 0.6677 $\pm$ 0.06 (7) \tabularnewline
skulls  & 0.2396 $\pm$ 0.06 (4)  & 0.2362 $\pm$ 0.07 (5)  & 0.2275 $\pm$ 0.05 (6)  & 0.2547 $\pm$ 0.06 (3)  & 0.2586 $\pm$ 0.06 (1)  & 0.2557 $\pm$ 0.07 (2)  & 0.2247 $\pm$ 0.07 (7) \tabularnewline
diabetes  & 0.9079 $\pm$ 0.05 (7)  & 0.9529 $\pm$ 0.04 (4)  & 0.9193 $\pm$ 0.05 (6)  & 0.9567 $\pm$ 0.04 (2)  & 0.9578 $\pm$ 0.04 (1)  & 0.9545 $\pm$ 0.04 (3)  & 0.9527 $\pm$ 0.04 (5) \tabularnewline
physio  & 0.8736 $\pm$ 0.03 (6)  & 0.9114 $\pm$ 0.03 (1)  & 0.8458 $\pm$ 0.04 (7)  & 0.8977 $\pm$ 0.03 (4)  & 0.9100 $\pm$ 0.03 (2)  & 0.8988 $\pm$ 0.03 (3)  & 0.8822 $\pm$ 0.03 (5) \tabularnewline
breasttissue  & 0.6136 $\pm$ 0.08 (5)  & 0.6489 $\pm$ 0.10 (2)  & 0.6150 $\pm$ 0.09 (4)  & 0.6721 $\pm$ 0.09 (1)  & 0.5737 $\pm$ 0.10 (7)  & 0.6470 $\pm$ 0.09 (3)  & 0.5890 $\pm$ 0.07 (6) \tabularnewline
bupa  & 0.6737 $\pm$ 0.05 (5)  & 0.6894 $\pm$ 0.04 (3)  & 0.6670 $\pm$ 0.05 (6)  & 0.6823 $\pm$ 0.05 (4)  & 0.6901 $\pm$ 0.04 (1)  & 0.6898 $\pm$ 0.05 (2)  & 0.6241 $\pm$ 0.05 (7) \tabularnewline
cleveland  & 0.2805 $\pm$ 0.05 (4)  & 0.2849 $\pm$ 0.05 (2)  & 0.2828 $\pm$ 0.05 (3)  & 0.2889 $\pm$ 0.06 (1)  & 0.2654 $\pm$ 0.05 (6)  & 0.2608 $\pm$ 0.04 (7)  & 0.2700 $\pm$ 0.04 (5) \tabularnewline
haberman  & 0.5436 $\pm$ 0.06 (6)  & 0.5729 $\pm$ 0.06 (5)  & 0.5326 $\pm$ 0.05 (7)  & 0.5819 $\pm$ 0.05 (3)  & 0.5975 $\pm$ 0.05 (1)  & 0.5858 $\pm$ 0.06 (2)  & 0.5796 $\pm$ 0.07 (4) \tabularnewline
hayes\_roth  & 0.7349 $\pm$ 0.07 (2)  & 0.7971 $\pm$ 0.07 (1)  & 0.7291 $\pm$ 0.09 (3)  & 0.5641 $\pm$ 0.09 (6)  & 0.5330 $\pm$ 0.11 (7)  & 0.7036 $\pm$ 0.09 (4)  & 0.6459 $\pm$ 0.08 (5) \tabularnewline
monks  & 0.9336 $\pm$ 0.02 (2)  & 0.9835 $\pm$ 0.02 (1)  & 0.8884 $\pm$ 0.03 (6)  & 0.9058 $\pm$ 0.08 (5)  & 0.9069 $\pm$ 0.05 (4)  & 0.9162 $\pm$ 0.05 (3)  & 0.8280 $\pm$ 0.09 (7) \tabularnewline
newthyroid  & 0.3922 $\pm$ 0.04 (7)  & 0.4255 $\pm$ 0.04 (4)  & 0.4015 $\pm$ 0.04 (6)  & 0.4178 $\pm$ 0.04 (5)  & 0.4601 $\pm$ 0.04 (3)  & 0.4641 $\pm$ 0.04 (2)  & 0.4903 $\pm$ 0.08 (1) \tabularnewline
yeast  & 0.5061 $\pm$ 0.05 (1)  & 0.4508 $\pm$ 0.03 (3)  & 0.4499 $\pm$ 0.05 (4)  & 0.4756 $\pm$ 0.05 (2)  & 0.4382 $\pm$ 0.03 (5)  & 0.4099 $\pm$ 0.03 (6)  & 0.3958 $\pm$ 0.03 (7) \tabularnewline
spam  & 0.9215 $\pm$ 0.01 (4)  & 0.9251 $\pm$ 0.01 (2)  & 0.9128 $\pm$ 0.01 (5)  & 0.9279 $\pm$ 0.01 (1)  & 0.9231 $\pm$ 0.01 (3)  & 0.8993 $\pm$ 0.01 (6)  & 0.8848 $\pm$ 0.01 (7) \tabularnewline
lymphography  & 0.5765 $\pm$ 0.15 (1)  & 0.4878 $\pm$ 0.12 (3)  & 0.5567 $\pm$ 0.12 (2)  & 0.4466 $\pm$ 0.10 (4)  & 0.3963 $\pm$ 0.04 (5)  & 0.3924 $\pm$ 0.03 (6)  & 0.3878 $\pm$ 0.05 (7) \tabularnewline
movement\_libras  & 0.7025 $\pm$ 0.06 (1)  & 0.6890 $\pm$ 0.05 (3)  & 0.6957 $\pm$ 0.05 (2)  & 0.5678 $\pm$ 0.05 (6)  & 0.5818 $\pm$ 0.05 (5)  & 0.6224 $\pm$ 0.05 (4)  & 0.4789 $\pm$ 0.05 (7) \tabularnewline
SAheart  & 0.6259 $\pm$ 0.04 (5)  & 0.6410 $\pm$ 0.04 (3)  & 0.6137 $\pm$ 0.05 (7)  & 0.6359 $\pm$ 0.05 (4)  & 0.6503 $\pm$ 0.04 (1)  & 0.6436 $\pm$ 0.04 (2)  & 0.6141 $\pm$ 0.05 (6) \tabularnewline
zoo  & 0.7423 $\pm$ 0.11 (2)  & 0.7612 $\pm$ 0.11 (1)  & 0.7238 $\pm$ 0.10 (3)  & 0.6636 $\pm$ 0.10 (4)  & 0.5182 $\pm$ 0.08 (6)  & 0.5183 $\pm$ 0.08 (5)  & 0.5104 $\pm$ 0.06 (7) \tabularnewline
\hline 
Average rank  & 3.70  & \textbf{2.70}  & 4.77  & 3.37  & 3.50  & 4.10  & 5.87 \tabularnewline
\hline 
\end{tabular}} 
\end{table}

\end{landscape}

\begin{landscape}

\begin{table}[p]
\caption{Noise Level: $15\%$ . Each cell in the table shows the $F_{1}^{(macro)}$-measure
(higher value is better) from the $20$ times $4$-fold cross-validation
experiment with $15\%$ noise induced on the class labels. The values
in parenthesis is the relative ranking of the algorithm on the dataset
in the corresponding row (lower ranks are better). \label{tab:noise15}}
\centering{}\centering\resizebox{\paperwidth}{!}{ %
\begin{tabular}{rrrrrrrr}
\hline 
 & KFHE-e  & KFHE-l  & AdaBoost  & GBM  & S-GBM  & Bagging  & CART \tabularnewline
\hline 
mushroom  & 0.9941 $\pm$ 0.00 (4.5)  & 0.9941 $\pm$ 0.00 (4.5)  & 0.9967 $\pm$ 0.00 (3)  & 0.9992 $\pm$ 0.00 (1)  & 0.9990 $\pm$ 0.00 (2)  & 0.9934 $\pm$ 0.00 (6.5)  & 0.9934 $\pm$ 0.00 (6.5) \tabularnewline
iris  & 0.8619 $\pm$ 0.06 (6)  & 0.9379 $\pm$ 0.04 (3)  & 0.8438 $\pm$ 0.06 (7)  & 0.9317 $\pm$ 0.05 (4)  & 0.9499 $\pm$ 0.03 (1)  & 0.9463 $\pm$ 0.04 (2)  & 0.9299 $\pm$ 0.03 (5) \tabularnewline
glass  & 0.5976 $\pm$ 0.09 (2)  & 0.6240 $\pm$ 0.08 (1)  & 0.5901 $\pm$ 0.09 (4)  & 0.5909 $\pm$ 0.09 (3)  & 0.5076 $\pm$ 0.08 (6)  & 0.5345 $\pm$ 0.08 (5)  & 0.4721 $\pm$ 0.08 (7) \tabularnewline
car\_eval  & 0.8374 $\pm$ 0.04 (3)  & 0.8394 $\pm$ 0.04 (2)  & 0.6708 $\pm$ 0.04 (7)  & 0.8563 $\pm$ 0.04 (1)  & 0.7599 $\pm$ 0.05 (5)  & 0.8108 $\pm$ 0.05 (4)  & 0.7577 $\pm$ 0.07 (6) \tabularnewline
cmc  & 0.5182 $\pm$ 0.03 (5)  & 0.5199 $\pm$ 0.02 (4)  & 0.4889 $\pm$ 0.03 (7)  & 0.5221 $\pm$ 0.02 (3)  & 0.5245 $\pm$ 0.03 (2)  & 0.5258 $\pm$ 0.03 (1)  & 0.4949 $\pm$ 0.04 (6) \tabularnewline
tvowel  & 0.8261 $\pm$ 0.03 (3)  & 0.8208 $\pm$ 0.03 (4)  & 0.7552 $\pm$ 0.03 (7)  & 0.8274 $\pm$ 0.03 (2)  & 0.8311 $\pm$ 0.03 (1)  & 0.7924 $\pm$ 0.03 (5)  & 0.7758 $\pm$ 0.03 (6) \tabularnewline
balance\_scale  & 0.5908 $\pm$ 0.03 (3)  & 0.5948 $\pm$ 0.03 (2)  & 0.5808 $\pm$ 0.04 (5)  & 0.5871 $\pm$ 0.02 (4)  & 0.5949 $\pm$ 0.02 (1)  & 0.5674 $\pm$ 0.02 (6)  & 0.5326 $\pm$ 0.03 (7) \tabularnewline
flags  & 0.3032 $\pm$ 0.06 (1)  & 0.2998 $\pm$ 0.05 (3)  & 0.2958 $\pm$ 0.06 (4)  & 0.3021 $\pm$ 0.06 (2)  & 0.2463 $\pm$ 0.03 (5)  & 0.2451 $\pm$ 0.03 (6)  & 0.2352 $\pm$ 0.04 (7) \tabularnewline
german  & 0.6488 $\pm$ 0.03 (3)  & 0.6522 $\pm$ 0.03 (1)  & 0.6376 $\pm$ 0.03 (5)  & 0.6491 $\pm$ 0.03 (2)  & 0.6432 $\pm$ 0.03 (4)  & 0.6275 $\pm$ 0.03 (6)  & 0.6202 $\pm$ 0.04 (7) \tabularnewline
ilpd  & 0.5645 $\pm$ 0.04 (3)  & 0.5699 $\pm$ 0.04 (1)  & 0.5698 $\pm$ 0.04 (2)  & 0.5592 $\pm$ 0.04 (4)  & 0.5564 $\pm$ 0.04 (5)  & 0.5557 $\pm$ 0.04 (6)  & 0.5517 $\pm$ 0.04 (7) \tabularnewline
ionosphere  & 0.8572 $\pm$ 0.04 (5)  & 0.8892 $\pm$ 0.04 (3)  & 0.8416 $\pm$ 0.05 (6)  & 0.8714 $\pm$ 0.04 (4)  & 0.9025 $\pm$ 0.04 (1)  & 0.8995 $\pm$ 0.04 (2)  & 0.8043 $\pm$ 0.06 (7) \tabularnewline
knowledge  & 0.9050 $\pm$ 0.03 (4)  & 0.9291 $\pm$ 0.03 (1)  & 0.8915 $\pm$ 0.03 (6)  & 0.9090 $\pm$ 0.03 (3)  & 0.8835 $\pm$ 0.03 (7)  & 0.9133 $\pm$ 0.03 (2)  & 0.9006 $\pm$ 0.04 (5) \tabularnewline
vertebral  & 0.7463 $\pm$ 0.04 (5)  & 0.7790 $\pm$ 0.05 (3)  & 0.7275 $\pm$ 0.05 (6)  & 0.7641 $\pm$ 0.05 (4)  & 0.7997 $\pm$ 0.05 (1)  & 0.7866 $\pm$ 0.05 (2)  & 0.7267 $\pm$ 0.05 (7) \tabularnewline
sonar  & 0.7548 $\pm$ 0.06 (2)  & 0.7451 $\pm$ 0.06 (5)  & 0.7462 $\pm$ 0.06 (4)  & 0.7330 $\pm$ 0.06 (6)  & 0.7643 $\pm$ 0.06 (1)  & 0.7485 $\pm$ 0.07 (3)  & 0.6356 $\pm$ 0.08 (7) \tabularnewline
skulls  & 0.2545 $\pm$ 0.06 (3)  & 0.2635 $\pm$ 0.06 (1)  & 0.2587 $\pm$ 0.06 (2)  & 0.2530 $\pm$ 0.05 (4.5)  & 0.2408 $\pm$ 0.08 (6)  & 0.2530 $\pm$ 0.06 (4.5)  & 0.2183 $\pm$ 0.06 (7) \tabularnewline
diabetes  & 0.8668 $\pm$ 0.06 (6)  & 0.9494 $\pm$ 0.04 (5)  & 0.8662 $\pm$ 0.08 (7)  & 0.9523 $\pm$ 0.04 (4)  & 0.9680 $\pm$ 0.03 (1)  & 0.9536 $\pm$ 0.03 (3)  & 0.9608 $\pm$ 0.03 (2) \tabularnewline
physio  & 0.8496 $\pm$ 0.03 (6)  & 0.8964 $\pm$ 0.02 (3)  & 0.8155 $\pm$ 0.04 (7)  & 0.8893 $\pm$ 0.03 (4)  & 0.9052 $\pm$ 0.02 (1)  & 0.8995 $\pm$ 0.02 (2)  & 0.8734 $\pm$ 0.03 (5) \tabularnewline
breasttissue  & 0.6072 $\pm$ 0.08 (6)  & 0.6283 $\pm$ 0.08 (2)  & 0.6202 $\pm$ 0.09 (3)  & 0.6139 $\pm$ 0.07 (5)  & 0.6161 $\pm$ 0.08 (4)  & 0.6500 $\pm$ 0.07 (1)  & 0.5652 $\pm$ 0.08 (7) \tabularnewline
bupa  & 0.6553 $\pm$ 0.04 (6)  & 0.6779 $\pm$ 0.05 (2)  & 0.6565 $\pm$ 0.05 (5)  & 0.6640 $\pm$ 0.05 (4)  & 0.6828 $\pm$ 0.05 (1)  & 0.6776 $\pm$ 0.05 (3)  & 0.6089 $\pm$ 0.05 (7) \tabularnewline
cleveland  & 0.2917 $\pm$ 0.05 (3.5)  & 0.2869 $\pm$ 0.05 (5)  & 0.2989 $\pm$ 0.05 (2)  & 0.2998 $\pm$ 0.05 (1)  & 0.2770 $\pm$ 0.04 (6)  & 0.2917 $\pm$ 0.05 (3.5)  & 0.2758 $\pm$ 0.05 (7) \tabularnewline
haberman  & 0.5448 $\pm$ 0.05 (6)  & 0.5606 $\pm$ 0.06 (5)  & 0.5404 $\pm$ 0.05 (7)  & 0.5625 $\pm$ 0.06 (4)  & 0.5819 $\pm$ 0.07 (1)  & 0.5645 $\pm$ 0.06 (3)  & 0.5677 $\pm$ 0.07 (2) \tabularnewline
hayes\_roth  & 0.7335 $\pm$ 0.08 (2)  & 0.7653 $\pm$ 0.08 (1)  & 0.6615 $\pm$ 0.09 (4)  & 0.5390 $\pm$ 0.08 (6)  & 0.4691 $\pm$ 0.10 (7)  & 0.6970 $\pm$ 0.09 (3)  & 0.6558 $\pm$ 0.10 (5) \tabularnewline
monks  & 0.8757 $\pm$ 0.04 (4)  & 0.9623 $\pm$ 0.03 (1)  & 0.8265 $\pm$ 0.04 (6)  & 0.8671 $\pm$ 0.07 (5)  & 0.9045 $\pm$ 0.05 (3)  & 0.9134 $\pm$ 0.06 (2)  & 0.7957 $\pm$ 0.08 (7) \tabularnewline
newthyroid  & 0.3816 $\pm$ 0.04 (7)  & 0.4305 $\pm$ 0.04 (3)  & 0.3825 $\pm$ 0.04 (6)  & 0.4224 $\pm$ 0.04 (4)  & 0.4700 $\pm$ 0.04 (1)  & 0.4646 $\pm$ 0.04 (2)  & 0.4175 $\pm$ 0.10 (5) \tabularnewline
yeast  & 0.4700 $\pm$ 0.05 (1)  & 0.4462 $\pm$ 0.04 (3)  & 0.4198 $\pm$ 0.05 (4)  & 0.4695 $\pm$ 0.05 (2)  & 0.4194 $\pm$ 0.03 (5)  & 0.4115 $\pm$ 0.03 (6)  & 0.4018 $\pm$ 0.03 (7) \tabularnewline
spam  & 0.9154 $\pm$ 0.01 (4)  & 0.9256 $\pm$ 0.01 (1)  & 0.8968 $\pm$ 0.02 (6)  & 0.9236 $\pm$ 0.01 (2)  & 0.9206 $\pm$ 0.01 (3)  & 0.9013 $\pm$ 0.01 (5)  & 0.8808 $\pm$ 0.01 (7) \tabularnewline
lymphography  & 0.5011 $\pm$ 0.13 (1)  & 0.4120 $\pm$ 0.08 (3)  & 0.4825 $\pm$ 0.13 (2)  & 0.3881 $\pm$ 0.03 (6)  & 0.4008 $\pm$ 0.03 (4)  & 0.3964 $\pm$ 0.03 (5)  & 0.3561 $\pm$ 0.04 (7) \tabularnewline
movement\_libras  & 0.6955 $\pm$ 0.05 (2)  & 0.6914 $\pm$ 0.06 (3)  & 0.6995 $\pm$ 0.05 (1)  & 0.5758 $\pm$ 0.05 (6)  & 0.5934 $\pm$ 0.05 (5)  & 0.6325 $\pm$ 0.06 (4)  & 0.4584 $\pm$ 0.06 (7) \tabularnewline
SAheart  & 0.6105 $\pm$ 0.04 (5)  & 0.6279 $\pm$ 0.05 (4)  & 0.5929 $\pm$ 0.04 (7)  & 0.6322 $\pm$ 0.04 (2)  & 0.6368 $\pm$ 0.04 (1)  & 0.6290 $\pm$ 0.05 (3)  & 0.6081 $\pm$ 0.04 (6) \tabularnewline
zoo  & 0.6541 $\pm$ 0.12 (4)  & 0.7949 $\pm$ 0.12 (1)  & 0.6618 $\pm$ 0.12 (3)  & 0.7736 $\pm$ 0.11 (2)  & 0.4358 $\pm$ 0.10 (7)  & 0.5625 $\pm$ 0.07 (6)  & 0.5627 $\pm$ 0.07 (5) \tabularnewline
\hline 
Average rank  & 3.87  & \textbf{2.68}  & 4.83  & 3.48  & 3.27  & 3.75  & 6.12 \tabularnewline
\hline 
\end{tabular}} 
\end{table}

\end{landscape}

\begin{landscape}

\begin{table}[p]
\caption{Noise Level: $20\%$ . Each cell in the table shows the $F_{1}^{(macro)}$-measure
(higher value is better) from the $20$ times $4$-fold cross-validation
experiment with $20\%$ noise induced on the class labels. The values
in parenthesis is the relative ranking of the algorithm on the dataset
in the corresponding row (lower ranks are better). \label{tab:noise20}}
\centering{}\centering\resizebox{\paperwidth}{!}{ %
\begin{tabular}{rrrrrrrr}
\hline 
 & KFHE-e  & KFHE-l  & AdaBoost  & GBM  & S-GBM  & Bagging  & CART \tabularnewline
\hline 
mushroom  & 0.9939 $\pm$ 0.00 (5)  & 0.9943 $\pm$ 0.00 (4)  & 0.9964 $\pm$ 0.00 (3)  & 0.9981 $\pm$ 0.00 (2)  & 0.9984 $\pm$ 0.00 (1)  & 0.9912 $\pm$ 0.01 (7)  & 0.9914 $\pm$ 0.00 (6) \tabularnewline
iris  & 0.8457 $\pm$ 0.07 (6)  & 0.9208 $\pm$ 0.05 (4)  & 0.7999 $\pm$ 0.06 (7)  & 0.9199 $\pm$ 0.05 (5)  & 0.9560 $\pm$ 0.03 (1)  & 0.9537 $\pm$ 0.03 (2)  & 0.9369 $\pm$ 0.04 (3) \tabularnewline
glass  & 0.6253 $\pm$ 0.08 (2)  & 0.6242 $\pm$ 0.08 (3)  & 0.5804 $\pm$ 0.09 (5)  & 0.5985 $\pm$ 0.09 (4)  & 0.5692 $\pm$ 0.07 (6)  & 0.6284 $\pm$ 0.08 (1)  & 0.5329 $\pm$ 0.09 (7) \tabularnewline
car\_eval  & 0.8347 $\pm$ 0.03 (3)  & 0.8403 $\pm$ 0.04 (2)  & 0.6342 $\pm$ 0.04 (7)  & 0.8512 $\pm$ 0.04 (1)  & 0.7853 $\pm$ 0.04 (6)  & 0.8151 $\pm$ 0.05 (4)  & 0.8124 $\pm$ 0.04 (5) \tabularnewline
cmc  & 0.5183 $\pm$ 0.02 (4)  & 0.5167 $\pm$ 0.02 (5)  & 0.4847 $\pm$ 0.03 (7)  & 0.5202 $\pm$ 0.02 (3)  & 0.5284 $\pm$ 0.02 (2)  & 0.5285 $\pm$ 0.02 (1)  & 0.5006 $\pm$ 0.03 (6) \tabularnewline
tvowel  & 0.8244 $\pm$ 0.03 (1)  & 0.8198 $\pm$ 0.03 (4)  & 0.7136 $\pm$ 0.04 (7)  & 0.8226 $\pm$ 0.03 (3)  & 0.8237 $\pm$ 0.03 (2)  & 0.7897 $\pm$ 0.03 (5)  & 0.7669 $\pm$ 0.03 (6) \tabularnewline
balance\_scale  & 0.5856 $\pm$ 0.04 (4)  & 0.5951 $\pm$ 0.03 (2)  & 0.5442 $\pm$ 0.03 (7)  & 0.5960 $\pm$ 0.03 (1)  & 0.5892 $\pm$ 0.03 (3)  & 0.5660 $\pm$ 0.03 (5)  & 0.5471 $\pm$ 0.04 (6) \tabularnewline
flags  & 0.2893 $\pm$ 0.06 (3)  & 0.2824 $\pm$ 0.05 (4)  & 0.2934 $\pm$ 0.05 (2)  & 0.2964 $\pm$ 0.05 (1)  & 0.2609 $\pm$ 0.04 (5)  & 0.2598 $\pm$ 0.05 (6)  & 0.2243 $\pm$ 0.04 (7) \tabularnewline
german  & 0.6311 $\pm$ 0.03 (5)  & 0.6613 $\pm$ 0.03 (2)  & 0.6158 $\pm$ 0.03 (7)  & 0.6496 $\pm$ 0.03 (4)  & 0.6671 $\pm$ 0.03 (1)  & 0.6535 $\pm$ 0.03 (3)  & 0.6267 $\pm$ 0.04 (6) \tabularnewline
ilpd  & 0.5794 $\pm$ 0.04 (5)  & 0.5836 $\pm$ 0.04 (4)  & 0.5667 $\pm$ 0.04 (7)  & 0.5843 $\pm$ 0.05 (3)  & 0.5909 $\pm$ 0.04 (1)  & 0.5848 $\pm$ 0.04 (2)  & 0.5713 $\pm$ 0.04 (6) \tabularnewline
ionosphere  & 0.8458 $\pm$ 0.04 (4)  & 0.8682 $\pm$ 0.04 (3)  & 0.8260 $\pm$ 0.04 (6)  & 0.8397 $\pm$ 0.04 (5)  & 0.8731 $\pm$ 0.04 (2)  & 0.8786 $\pm$ 0.04 (1)  & 0.8053 $\pm$ 0.06 (7) \tabularnewline
knowledge  & 0.8762 $\pm$ 0.04 (5)  & 0.8970 $\pm$ 0.03 (1)  & 0.8459 $\pm$ 0.04 (7)  & 0.8862 $\pm$ 0.03 (3)  & 0.8621 $\pm$ 0.03 (6)  & 0.8957 $\pm$ 0.04 (2)  & 0.8836 $\pm$ 0.04 (4) \tabularnewline
vertebral  & 0.7763 $\pm$ 0.05 (5)  & 0.8039 $\pm$ 0.05 (3)  & 0.7563 $\pm$ 0.05 (7)  & 0.7847 $\pm$ 0.04 (4)  & 0.8211 $\pm$ 0.05 (1)  & 0.8104 $\pm$ 0.05 (2)  & 0.7608 $\pm$ 0.05 (6) \tabularnewline
sonar  & 0.7274 $\pm$ 0.07 (5)  & 0.7423 $\pm$ 0.07 (1)  & 0.7340 $\pm$ 0.07 (4)  & 0.7104 $\pm$ 0.07 (6)  & 0.7410 $\pm$ 0.07 (2)  & 0.7384 $\pm$ 0.07 (3)  & 0.6257 $\pm$ 0.08 (7) \tabularnewline
skulls  & 0.2173 $\pm$ 0.06 (6)  & 0.2390 $\pm$ 0.07 (2)  & 0.2315 $\pm$ 0.06 (4)  & 0.2385 $\pm$ 0.06 (3)  & 0.2299 $\pm$ 0.06 (5)  & 0.2402 $\pm$ 0.06 (1)  & 0.1998 $\pm$ 0.05 (7) \tabularnewline
diabetes  & 0.8262 $\pm$ 0.07 (6)  & 0.9127 $\pm$ 0.05 (5)  & 0.8091 $\pm$ 0.07 (7)  & 0.9196 $\pm$ 0.05 (4)  & 0.9357 $\pm$ 0.04 (3)  & 0.9649 $\pm$ 0.04 (1)  & 0.9488 $\pm$ 0.06 (2) \tabularnewline
physio  & 0.7971 $\pm$ 0.05 (7)  & 0.9012 $\pm$ 0.03 (2)  & 0.8001 $\pm$ 0.05 (6)  & 0.8879 $\pm$ 0.03 (4)  & 0.9142 $\pm$ 0.03 (1)  & 0.8983 $\pm$ 0.03 (3)  & 0.8655 $\pm$ 0.04 (5) \tabularnewline
breasttissue  & 0.5644 $\pm$ 0.09 (7)  & 0.6337 $\pm$ 0.09 (1)  & 0.5745 $\pm$ 0.10 (5)  & 0.6187 $\pm$ 0.08 (2)  & 0.6051 $\pm$ 0.08 (4)  & 0.6183 $\pm$ 0.07 (3)  & 0.5737 $\pm$ 0.09 (6) \tabularnewline
bupa  & 0.6388 $\pm$ 0.05 (5)  & 0.6600 $\pm$ 0.05 (3)  & 0.6236 $\pm$ 0.05 (6)  & 0.6418 $\pm$ 0.05 (4)  & 0.6710 $\pm$ 0.04 (1)  & 0.6606 $\pm$ 0.05 (2)  & 0.5981 $\pm$ 0.05 (7) \tabularnewline
cleveland  & 0.2819 $\pm$ 0.05 (6)  & 0.2958 $\pm$ 0.05 (2)  & 0.3033 $\pm$ 0.05 (1)  & 0.2858 $\pm$ 0.05 (5)  & 0.2913 $\pm$ 0.05 (4)  & 0.2941 $\pm$ 0.05 (3)  & 0.2761 $\pm$ 0.05 (7) \tabularnewline
haberman  & 0.5216 $\pm$ 0.06 (6)  & 0.5520 $\pm$ 0.06 (4)  & 0.5172 $\pm$ 0.05 (7)  & 0.5619 $\pm$ 0.05 (3)  & 0.5956 $\pm$ 0.05 (1)  & 0.5742 $\pm$ 0.05 (2)  & 0.5442 $\pm$ 0.06 (5) \tabularnewline
hayes\_roth  & 0.6835 $\pm$ 0.10 (2)  & 0.6840 $\pm$ 0.09 (1)  & 0.6451 $\pm$ 0.09 (3)  & 0.5240 $\pm$ 0.08 (6)  & 0.4190 $\pm$ 0.09 (7)  & 0.6130 $\pm$ 0.09 (4)  & 0.5557 $\pm$ 0.10 (5) \tabularnewline
monks  & 0.8150 $\pm$ 0.04 (5)  & 0.8836 $\pm$ 0.04 (1)  & 0.7689 $\pm$ 0.04 (7)  & 0.8402 $\pm$ 0.05 (3)  & 0.8253 $\pm$ 0.04 (4)  & 0.8428 $\pm$ 0.05 (2)  & 0.8036 $\pm$ 0.06 (6) \tabularnewline
newthyroid  & 0.3845 $\pm$ 0.05 (6)  & 0.4459 $\pm$ 0.04 (4)  & 0.3749 $\pm$ 0.05 (7)  & 0.4467 $\pm$ 0.05 (3)  & 0.4724 $\pm$ 0.05 (2)  & 0.4835 $\pm$ 0.04 (1)  & 0.4137 $\pm$ 0.10 (5) \tabularnewline
yeast  & 0.4512 $\pm$ 0.05 (2)  & 0.4216 $\pm$ 0.03 (3)  & 0.3847 $\pm$ 0.04 (6)  & 0.4534 $\pm$ 0.05 (1)  & 0.4164 $\pm$ 0.03 (4)  & 0.4004 $\pm$ 0.03 (5)  & 0.3846 $\pm$ 0.03 (7) \tabularnewline
spam  & 0.9089 $\pm$ 0.01 (4)  & 0.9244 $\pm$ 0.01 (1)  & 0.8884 $\pm$ 0.02 (6)  & 0.9211 $\pm$ 0.01 (2)  & 0.9206 $\pm$ 0.01 (3)  & 0.8989 $\pm$ 0.01 (5)  & 0.8681 $\pm$ 0.01 (7) \tabularnewline
lymphography  & 0.4576 $\pm$ 0.13 (1)  & 0.4130 $\pm$ 0.07 (3)  & 0.4384 $\pm$ 0.11 (2)  & 0.3941 $\pm$ 0.04 (4)  & 0.3933 $\pm$ 0.04 (5)  & 0.3899 $\pm$ 0.03 (6)  & 0.3654 $\pm$ 0.04 (7) \tabularnewline
movement\_libras  & 0.6627 $\pm$ 0.05 (1)  & 0.6585 $\pm$ 0.05 (3)  & 0.6612 $\pm$ 0.05 (2)  & 0.5284 $\pm$ 0.05 (6)  & 0.5559 $\pm$ 0.06 (5)  & 0.6263 $\pm$ 0.05 (4)  & 0.4527 $\pm$ 0.05 (7) \tabularnewline
SAheart  & 0.5972 $\pm$ 0.05 (5)  & 0.6205 $\pm$ 0.04 (3)  & 0.5762 $\pm$ 0.05 (7)  & 0.6039 $\pm$ 0.04 (4)  & 0.6368 $\pm$ 0.04 (1)  & 0.6349 $\pm$ 0.05 (2)  & 0.5806 $\pm$ 0.05 (6) \tabularnewline
zoo  & 0.6653 $\pm$ 0.13 (4)  & 0.7603 $\pm$ 0.12 (1)  & 0.6714 $\pm$ 0.12 (3)  & 0.7389 $\pm$ 0.12 (2)  & 0.5562 $\pm$ 0.10 (6.5)  & 0.5573 $\pm$ 0.08 (5)  & 0.5562 $\pm$ 0.07 (6.5) \tabularnewline
\hline 
Average rank  & 4.33  & \textbf{2.70}  & 5.40  & 3.37  & 3.18  & 3.10  & 5.92 \tabularnewline
\hline 
\end{tabular}} 
\end{table}

\end{landscape}

\section{Complete statistical test results\label{sec:statsTables}}

\begin{table}[H]
\caption{Result of post-hoc Friedman Aligned Rank test with Finner $p$-value
adjustment over the different datasets using with different induced
noise in the class-labels. Lower diagonal:
post-hoc Friedman Aligned Rank Test $p$-values after the Finner adjustment.
{*} $\alpha=0.1$, {*}{*} $\alpha=0.05$ and {*}{*}{*} $\alpha=0.01$ \label{tab:Friedman-pairwise}}

\subfloat[No induced class-label noise\label{tab:Friedman-pairwise-00} ]{\centering\resizebox{\textwidth}{!}{ %
\begin{tabular}{rllllll}
\hline 
 & KFHE-e  & KFHE-l  & AdaBoost  & GBM  & S-GBM  & Bagging \tabularnewline
\hline 
KFHE-l  & 0.296371  &  &  &  &  & \tabularnewline
AdaBoost  & 0.717099  & 0.485252  &  &  &  & \tabularnewline
GBM  & 0.101786  & 0.573588  & 0.201563  &  &  & \tabularnewline
S-GBM  & 0.000174 {*}{*}{*} & 0.009166 {*}{*}{*} & 0.000696 {*}{*}{*} & 0.042330 {*}{*} &  & \tabularnewline
Bagging  & 0.000015 {*}{*}{*} & 0.001233 {*}{*}{*} & 0.000072 {*}{*}{*} & 0.009166 {*}{*}{*} & 0.573588  & \tabularnewline
CART  & 0.000000 {*}{*}{*} & 0.000001 {*}{*}{*} & 0.000000 {*}{*}{*} & 0.000013 {*}{*}{*} & 0.022192 {*}{*} & 0.089728 {*}\tabularnewline
\hline 
\end{tabular}}

}

\subfloat[$5\%$ induced class-label noise\label{tab:Friedman-pairwise-01}]{\centering\resizebox{\textwidth}{!}{ %
\begin{tabular}{rllllll}
\hline 
 & KFHE-e  & KFHE-l  & AdaBoost  & GBM  & S-GBM  & Bagging \tabularnewline
\hline 
KFHE-l  & 0.783059  &  &  &  &  & \tabularnewline
AdaBoost  & 0.286251  & 0.412077  &  &  &  & \tabularnewline
GBM  & 0.300240  & 0.430539  & 0.952742  &  &  & \tabularnewline
S-GBM  & 0.020310 {*}{*} & 0.042946 {*}{*} & 0.247823  & 0.236744  &  & \tabularnewline
Bagging  & 0.004712 {*}{*}{*} & 0.011182 {*}{*} & 0.087277 {*} & 0.082096 {*} & 0.601616  & \tabularnewline
CART  & 0.000000 {*}{*}{*} & 0.000000 {*}{*}{*} & 0.000011 {*}{*}{*} & 0.000011 {*}{*}{*} & 0.003113 {*}{*}{*} & 0.013687 {*}{*}\tabularnewline
\hline 
\end{tabular}}

}

\subfloat[$10\%$ induced class-label noise\label{tab:Friedman-pairwise-02}]{\centering\resizebox{\textwidth}{!}{ %
\begin{tabular}{rllllll}
\hline 
 & KFHE-e  & KFHE-l  & AdaBoost  & GBM  & S-GBM  & Bagging \tabularnewline
\hline 
KFHE-l  & 0.195755  &  &  &  &  & \tabularnewline
AdaBoost  & 0.195755  & 0.007206 {*}{*}{*} &  &  &  & \tabularnewline
GBM  & 0.963993  & 0.195755  & 0.194619  &  &  & \tabularnewline
S-GBM  & 0.505287  & 0.043617 {*}{*} & 0.505287  & 0.505287  &  & \tabularnewline
Bagging  & 0.505287  & 0.042152 {*}{*} & 0.505287  & 0.498988  & 0.959589  & \tabularnewline
CART  & 0.000621 {*}{*}{*} & 0.000001 {*}{*}{*} & 0.043617 {*}{*} & 0.000621 {*}{*}{*} & 0.007206 {*}{*}{*} & 0.007206 {*}{*}{*}\tabularnewline
\hline 
\end{tabular}}

}

\subfloat[$15\%$ induced class-label noise\label{tab:Friedman-pairwise-03}]{\centering\resizebox{\textwidth}{!}{ %
\begin{tabular}{rllllll}
\hline 
 & KFHE-e  & KFHE-l  & AdaBoost  & GBM  & S-GBM  & Bagging \tabularnewline
\hline 
KFHE-l  & 0.065548 {*} &  &  &  &  & \tabularnewline
AdaBoost  & 0.099521 {*} & 0.000256 {*}{*}{*} &  &  &  & \tabularnewline
GBM  & 0.813436  & 0.085647 {*} & 0.074584 {*} &  &  & \tabularnewline
S-GBM  & 0.925528  & 0.056191 {*} & 0.118441  & 0.77674  &  & \tabularnewline
Bagging  & 0.854408  & 0.076564 {*} & 0.074584 {*} & 0.938200 & 0.811715 & \tabularnewline
CART  & 0.000334 {*}{*}{*} & 0.000000 {*}{*}{*} & 0.074584 {*} & 0.000256 {*}{*}{*} & 0.000436 {*}{*}{*}  & 0.000256 {*}{*}{*}\tabularnewline
\hline 
\end{tabular}}

}

\subfloat[$20\%$ induced class-label noise\label{tab:Friedman-pairwise-04}]{\centering\resizebox{\textwidth}{!}{ %
\begin{tabular}{rllllll}
\hline 
 & KFHE-e  & KFHE-l  & AdaBoost  & GBM  & S-GBM  & Bagging \tabularnewline
\hline
KFHE-l  & 0.009541 {*}{*}{*} &  &  &  &  & \tabularnewline
AdaBoost  & 0.057691 {*} & 0.000007 {*}{*}{*} &  &  &  & \tabularnewline
GBM  & 0.384210  & 0.113135  & 0.003655 {*}{*}{*} &  &  & \tabularnewline
S-GBM  & 0.354782  & 0.127197  & 0.003080 {*}{*}{*} & 0.933339  &  & \tabularnewline
Bagging  & 0.109001  & 0.395131  & 0.000261 {*}{*}{*}  & 0.461039  & 0.479407  & \tabularnewline
CART  & 0.007492 {*}{*}{*} & 0.000000 {*}{*}{*} & 0.474921  & 0.000261 {*}{*}{*} & 0.000261 {*}{*}{*} & 0.000011 {*}{*}{*}\tabularnewline
\hline 
\end{tabular}}

}

\end{table}

\begin{table}
\caption{Result of pairwise Wilcoxon's Signed Rank Sum test over the different
datasets using with different induced noise in the class-labels, to
compare individual isolated pair of algorithms. Upper diagonal: win/lose/tie.
Lower diagonal: Wilcoxon's Signed Rank Sum Test $p$-values. {*}: $\alpha=0.1$, {*}{*}: $\alpha=0.05$
and {*}{*}{*}: $\alpha=0.01$ \label{tab:Wilcoxon-pairwise}. The
$p$-values for a pair of algorithms are to be interpreted in isolation
and not to be combined with more than one methods.\label{tab:Wilcox-pairwise}}

\subfloat[No induced class-label noise\label{tab:Wilcox-pairwise-00}]{\centering \resizebox{\textwidth}{!}{%
\begin{tabular}{rlllllll}
\hline 
 & KFHE-e  & KFHE-l  & AdaBoost  & GBM  & S-GBM  & Bagging  & CART \tabularnewline
\hline 
KFHE-e  &  & (19/11/0)  & (13/16/1)  & (23/7/0)  & (21/9/0)  & (24/6/0)  & (26/4/0) \tabularnewline
KFHE-l  & 0.009519 {*}{*}{*} &  & (12/18/0)  & (16/14/0)  & (20/10/0)  & (25/5/0)  & (26/4/0) \tabularnewline
AdaBoost  & 0.491795  & 0.028548 {*}{*} &  & (19/11/0)  & (20/10/0)  & (22/8/0)  & (25/5/0) \tabularnewline
GBM  & 0.003018 {*}{*}{*} & 0.144739  & 0.013515 {*}{*} &  & (22/8/0)  & (20/10/0)  & (25/5/0) \tabularnewline
S-GBM  & 0.001128 {*}{*}{*} & 0.004921 {*}{*}{*} & 0.002834 {*}{*}{*} & 0.003418 {*}{*}{*} &  & (19/11/0)  & (25/5/0) \tabularnewline
Bagging  & 0.000210 {*}{*}{*} & 0.000034 {*}{*}{*} & 0.000415 {*}{*}{*} & 0.004108 {*}{*}{*} & 0.336640 &  & (25/4/1) \tabularnewline
CART  & 0.000010 {*}{*}{*} & 0.000006 {*}{*}{*} & 0.000019 {*}{*}{*} & 0.000055 {*}{*}{*} & 0.002057 {*}{*}{*}  & 0.000016 {*}{*}{*} & \tabularnewline
\hline 
\end{tabular}}

}

\subfloat[$5\%$ induced class-label noise\label{tab:Wilcox-pairwise-01}]{\centering \resizebox{\textwidth}{!}{%
\begin{tabular}{rlllllll}
\hline 
 & KFHE-e  & KFHE-l  & AdaBoost  & GBM  & S-GBM  & Bagging  & CART \tabularnewline
\hline 
KFHE-e  &  & (15/15/0)  & (20/10/0)  & (16/14/0)  & (19/11/0)  & (22/8/0)  & (26/4/0) \tabularnewline
KFHE-l  & 0.344180 &  & (16/14/0)  & (18/12/0)  & (19/11/0)  & (22/7/1)  & (27/2/1) \tabularnewline
AdaBoost  & 0.020351 {*}{*} & 0.085688 {*} &  & (14/16/0)  & (17/13/0)  & (18/12/0)  & (24/6/0) \tabularnewline
GBM  & 0.299968  & 0.092311 {*} & 0.314422  &  & (21/9/0)  & (22/8/0)  & (27/3/0) \tabularnewline
S-GBM  & 0.028548 {*}{*} & 0.020862 {*}{*} & 0.118468  & 0.016213 {*}{*} &  & (19/10/1)  & (24/6/0) \tabularnewline
Bagging  & 0.005874 {*}{*}{*} & 0.000320 {*}{*}{*} & 0.049937 {*}{*} & 0.012819 {*}{*} & 0.289329  &  & (25/4/1) \tabularnewline
CART  & 0.000020 {*}{*}{*} & 0.000005 {*}{*}{*} & 0.000095 {*}{*}{*} & 0.000041 {*}{*}{*} & 0.000386 {*}{*}{*} & 0.000007 {*}{*}{*} & \tabularnewline
\hline 
\end{tabular}}

}

\subfloat[$10\%$ induced class-label noise\label{tab:Wilcox-pairwise-02}]{\centering\resizebox{\textwidth}{!}{ %
\begin{tabular}{rlllllll}
\hline 
 & KFHE-e  & KFHE-l  & AdaBoost  & GBM  & S-GBM  & Bagging  & CART \tabularnewline
\hline 
KFHE-e  &  & (11/19/0)  & (22/8/0)  & (12/18/0)  & (14/16/0)  & (17/13/0)  & (23/7/0) \tabularnewline
KFHE-l  & 0.015399 {*}{*} &  & (24/6/0)  & (20/10/0)  & (18/12/0)  & (20/9/1)  & (27/2/1) \tabularnewline
AdaBoost  & 0.001053 {*}{*}{*} & 0.000518 {*}{*}{*} &  & (8/22/0)  & (12/18/0)  & (14/16/0)  & (19/11/0) \tabularnewline
GBM  & 0.471305  & 0.034357 {*}{*} & 0.065296 {*} &  & (15/15/0)  & (17/13/0)  & (27/3/0) \tabularnewline
S-GBM  & 0.229510 & 0.067954 {*} & 0.329165  & 0.180019  &  & (20/10/0)  & (24/6/0) \tabularnewline
Bagging  & 0.122595  & 0.002125 {*}{*}{*} & 0.258524  & 0.106679  & 0.19383  &  & (25/4/1) \tabularnewline
CART  & 0.000916 {*}{*}{*} & 0.000020 {*}{*}{*} & 0.029918 {*}{*} & 0.000142 {*}{*}{*} & 0.002341 {*}{*}{*} & 0.000030 {*}{*}{*} & \tabularnewline
\hline 
\end{tabular}}

}

\subfloat[$15\%$ induced class-label noise\label{tab:Wilcox-pairwise-03}]{\centering\resizebox{\textwidth}{!}{ %
\begin{tabular}{rlllllll}
\hline 
 & KFHE-e  & KFHE-l  & AdaBoost  & GBM  & S-GBM  & Bagging  & CART \tabularnewline
\hline 
KFHE-e  &  & (7/22/1)  & (21/9/0)  & (11/19/0)  & (13/17/0)  & (16/13/1)  & (25/5/0) \tabularnewline
KFHE-l  & 0.001529 {*}{*}{*} &  & (25/5/0)  & (20/10/0)  & (16/14/0)  & (18/12/0)  & (28/2/0) \tabularnewline
AdaBoost  & 0.000285 {*}{*}{*} & 0.000047 {*}{*}{*} &  & (7/23/0)  & (12/18/0)  & (12/18/0)  & (20/10/0) \tabularnewline
GBM  & 0.069314 {*}{*}{*} & 0.003418 {*}{*}{*} & 0.006987 {*}{*}{*} &  & (13/17/0)  & (13/16/1)  & (27/3/0) \tabularnewline
S-GBM  & 0.406508  & 0.046838 {*}{*} & 0.133351  & 0.479495  &  & (19/11/0)  & (27/3/0) \tabularnewline
Bagging  & 0.386690 & 0.007829 {*}{*}{*} & 0.051005 {*} & 0.459044  & 0.495897  &  & (26/3/1) \tabularnewline
CART  & 0.000518 {*}{*}{*} & 0.000002 {*}{*}{*} & 0.039323 {*}{*} & 0.000029 {*}{*}{*} & 0.000285 {*}{*}{*} & 0.000002 {*}{*}{*} & \tabularnewline
\hline 
\end{tabular}}

}

\subfloat[$20\%$ induced class-label noise\label{tab:Wilcox-pairwise-04}]{\centering\resizebox{\textwidth}{!}{ %
\begin{tabular}{rlllllll}
\hline 
 & KFHE-e  & KFHE-l  & AdaBoost  & GBM  & S-GBM  & Bagging  & CART \tabularnewline
\hline 
KFHE-e  &  & (7/23/0)  & (22/8/0)  & (7/23/0)  & (10/20/0)  & (11/19/0)  & (23/7/0) \tabularnewline
KFHE-l  & 0.000464 {*}{*}{*} &  & (25/5/0)  & (19/11/0)  & (16/14/0)  & (18/12/0)  & (28/2/0) \tabularnewline
AdaBoost  & 0.000227 {*}{*}{*} & 0.000007 {*}{*}{*} &  & (5/25/0)  & (8/22/0)  & (7/23/0)  & (15/15/0) \tabularnewline
GBM  & 0.015802 {*}{*} & 0.006226 {*}{*}{*} & 0.001479 {*}{*}{*} &  & (13/17/0)  & (10/20/0)  & (27/3/0) \tabularnewline
S-GBM  & 0.159245  & 0.174673  & 0.007398 {*}{*}{*} & 0.344180 &  & (16/14/0)  & (25/4/1) \tabularnewline
Bagging  & 0.116442  & 0.067954 {*} & 0.000854 {*}{*}{*} & 0.092311 {*} & 0.406508  &  & (29/1/0) \tabularnewline
CART  & 0.005874 {*}{*}{*} & 0.000005 {*}{*}{*} & 0.430606  & 0.000019 {*}{*}{*} & 0.000084 {*}{*}{*} & 0.000001 {*}{*}{*} & \tabularnewline
\hline 
\end{tabular}}

}
\end{table}

\section{Plots of how model performance changes with noise\label{sec:performancePlots}}

\begin{figure}[H]
\subfloat[\emph{yeast} \label{fig:yeast_noise}]{\includegraphics[width=0.33\textwidth]{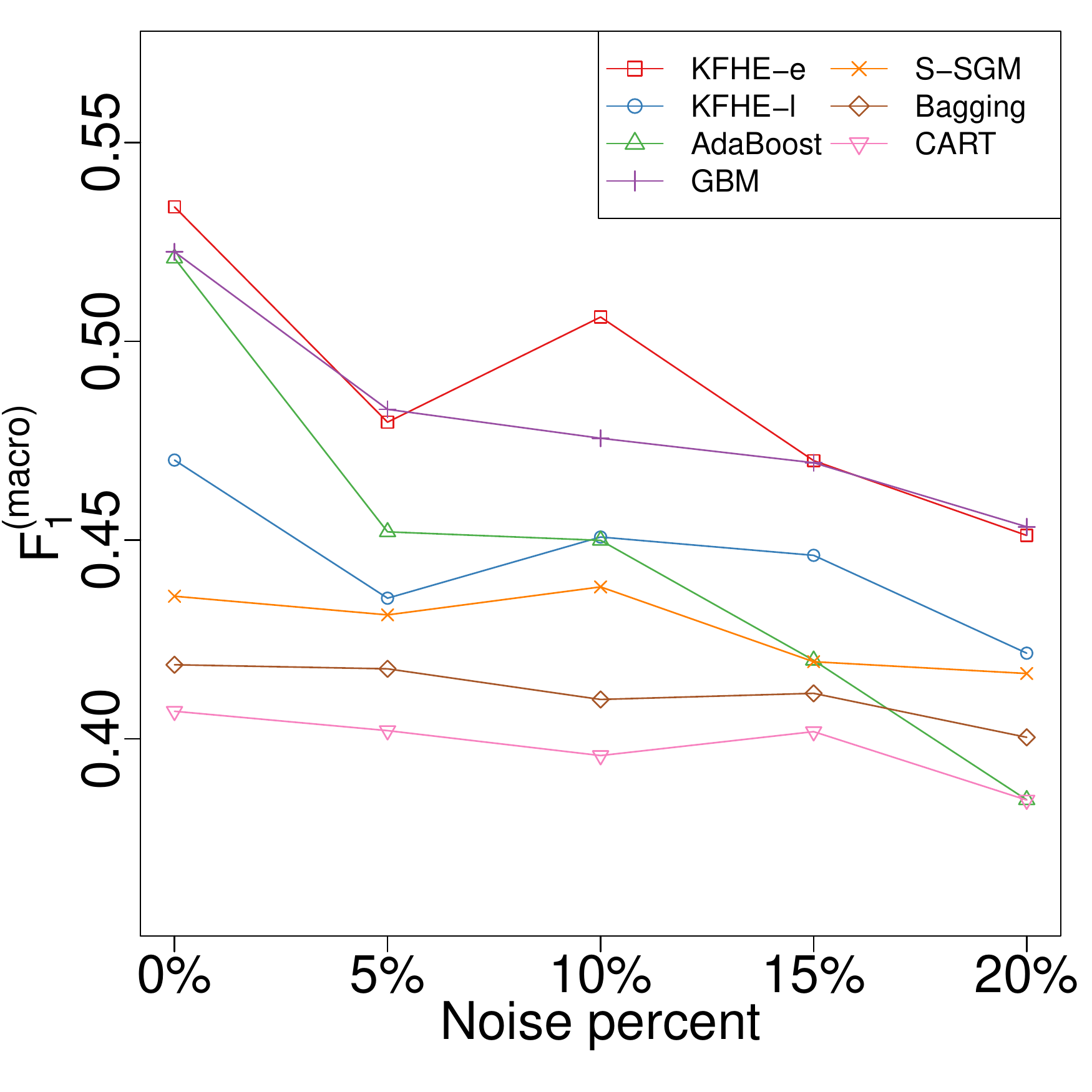}

}\subfloat[\emph{spam} \label{fig:spam_noise}]{\includegraphics[width=0.33\textwidth]{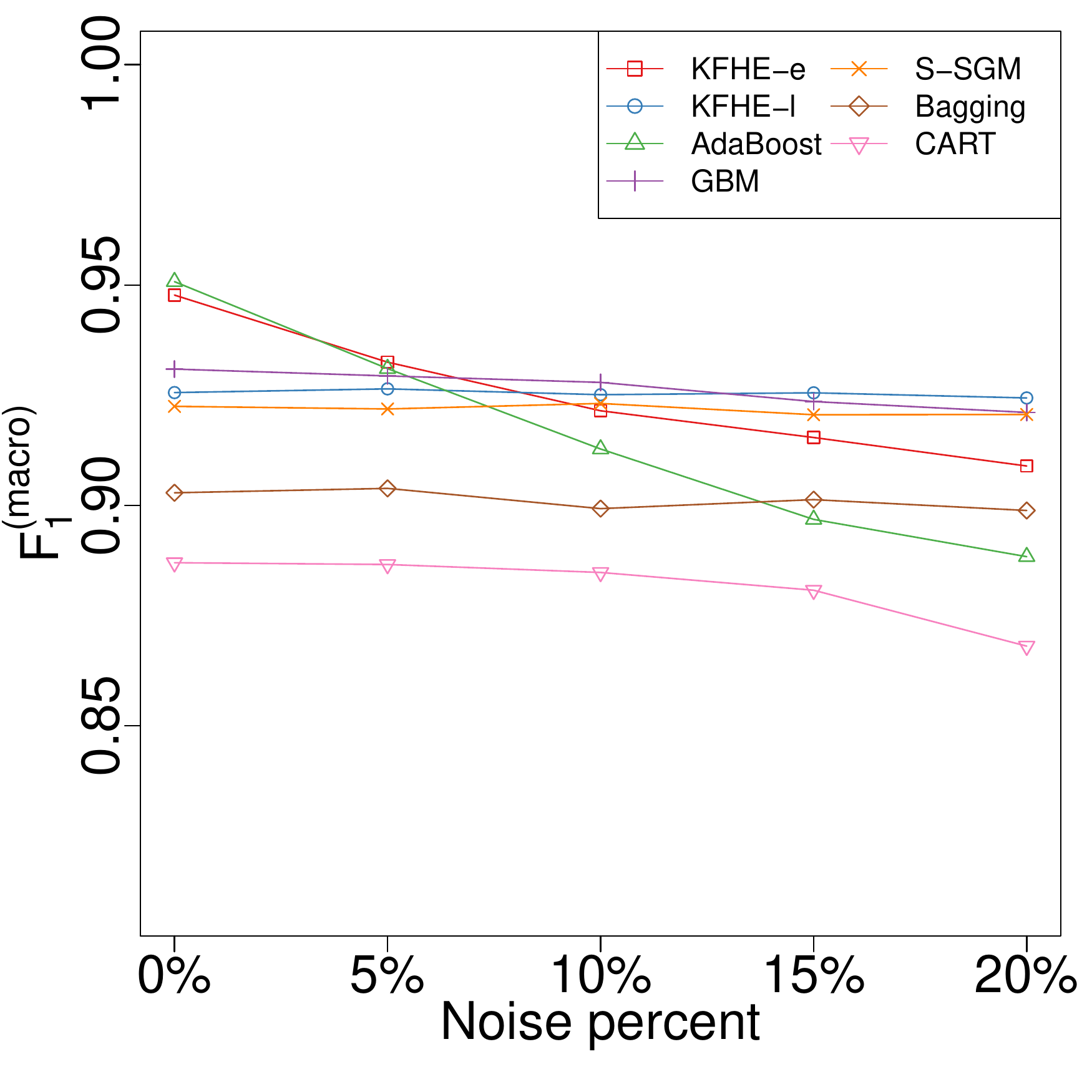}

}\subfloat[\emph{lymphography} \label{fig:lymphography_noise}]{\includegraphics[width=0.33\textwidth]{plots/plot_by_data_vs_noise_lymphography.pdf}

}

\subfloat[\emph{movement\_libras} \label{fig:movement_libras_noise}]{\includegraphics[width=0.33\textwidth]{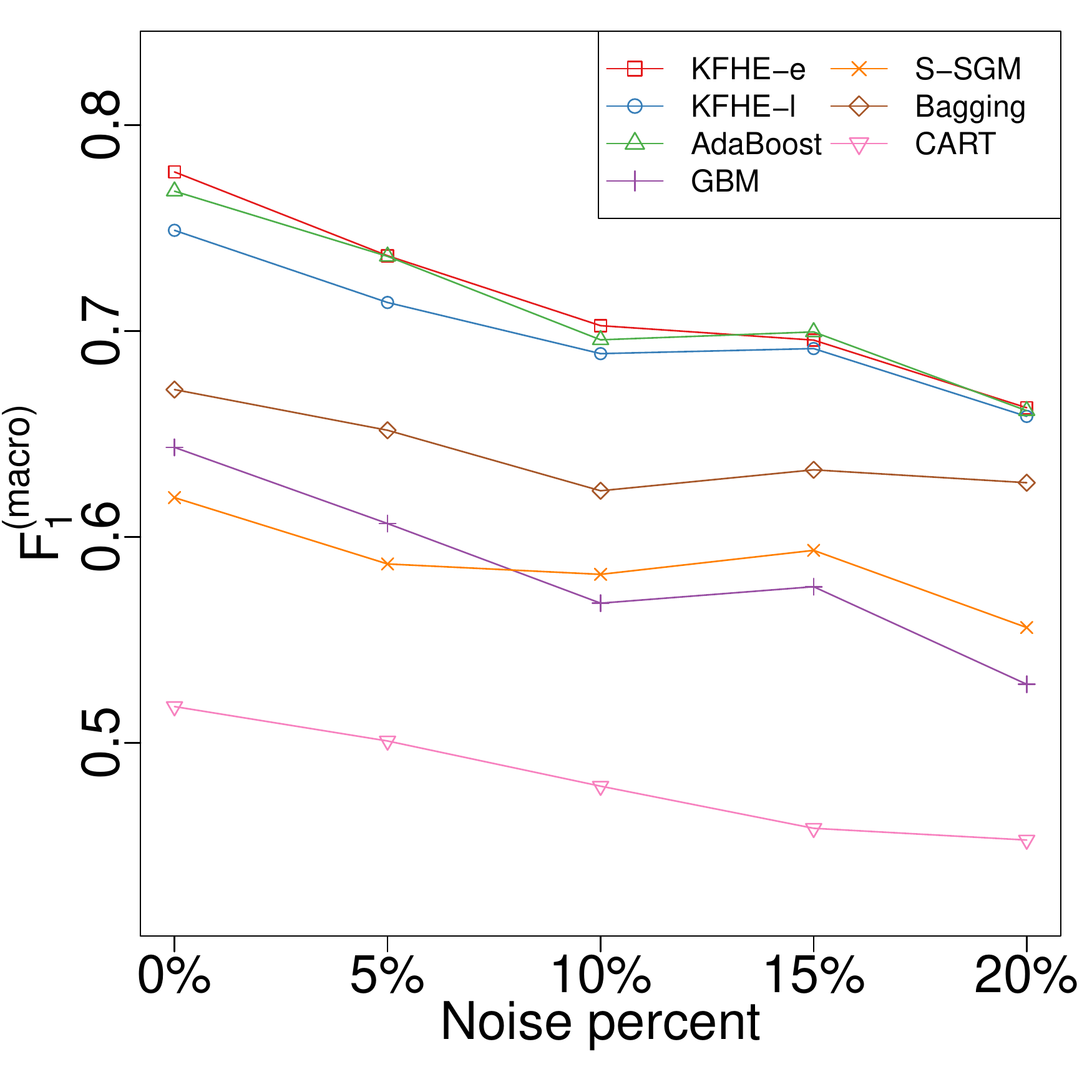}

}\subfloat[\emph{SAheart} \label{fig:SAheart_noise}]{\includegraphics[width=0.33\textwidth]{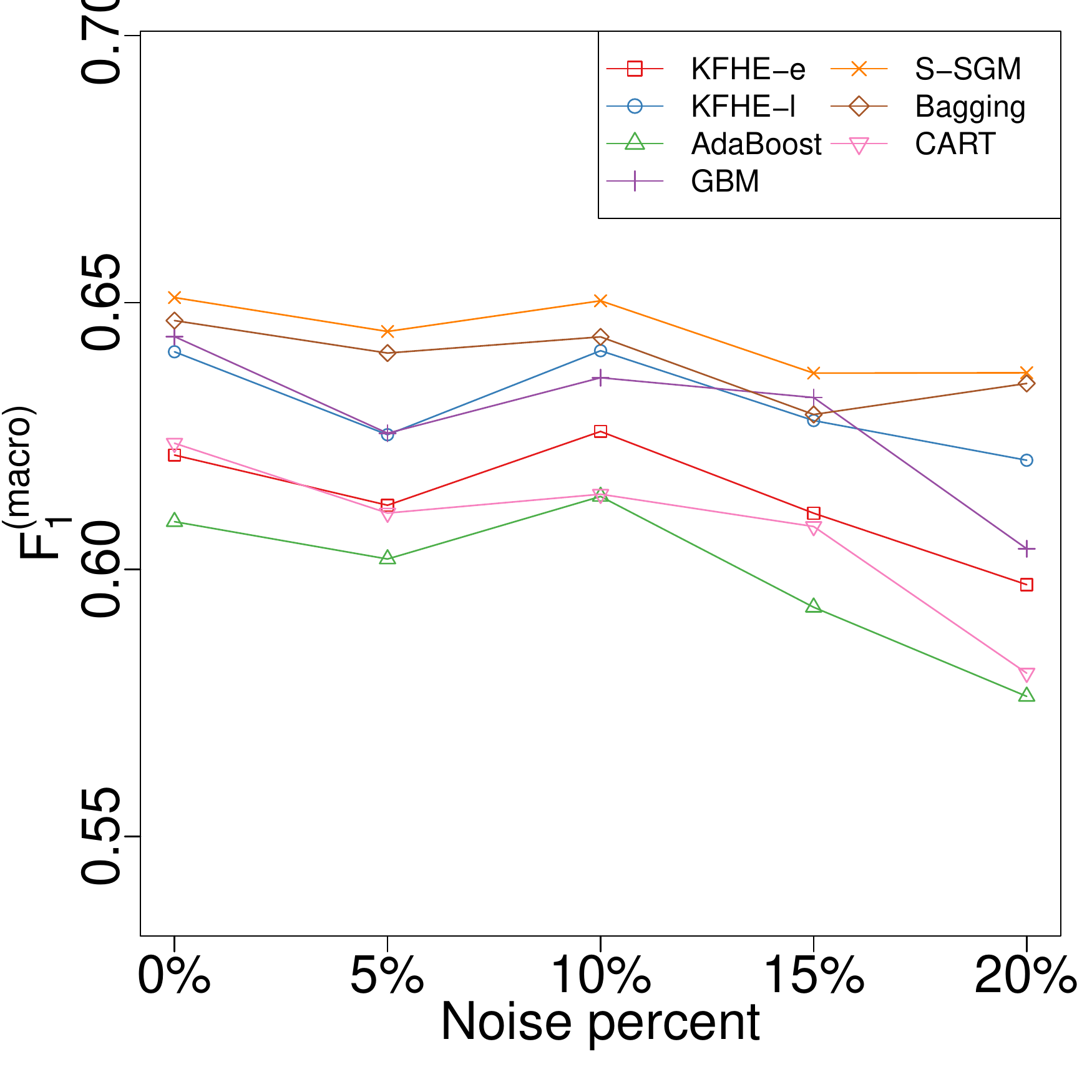}

}\subfloat[\emph{zoo} \label{fig:zoo_noise}]{\includegraphics[width=0.33\textwidth]{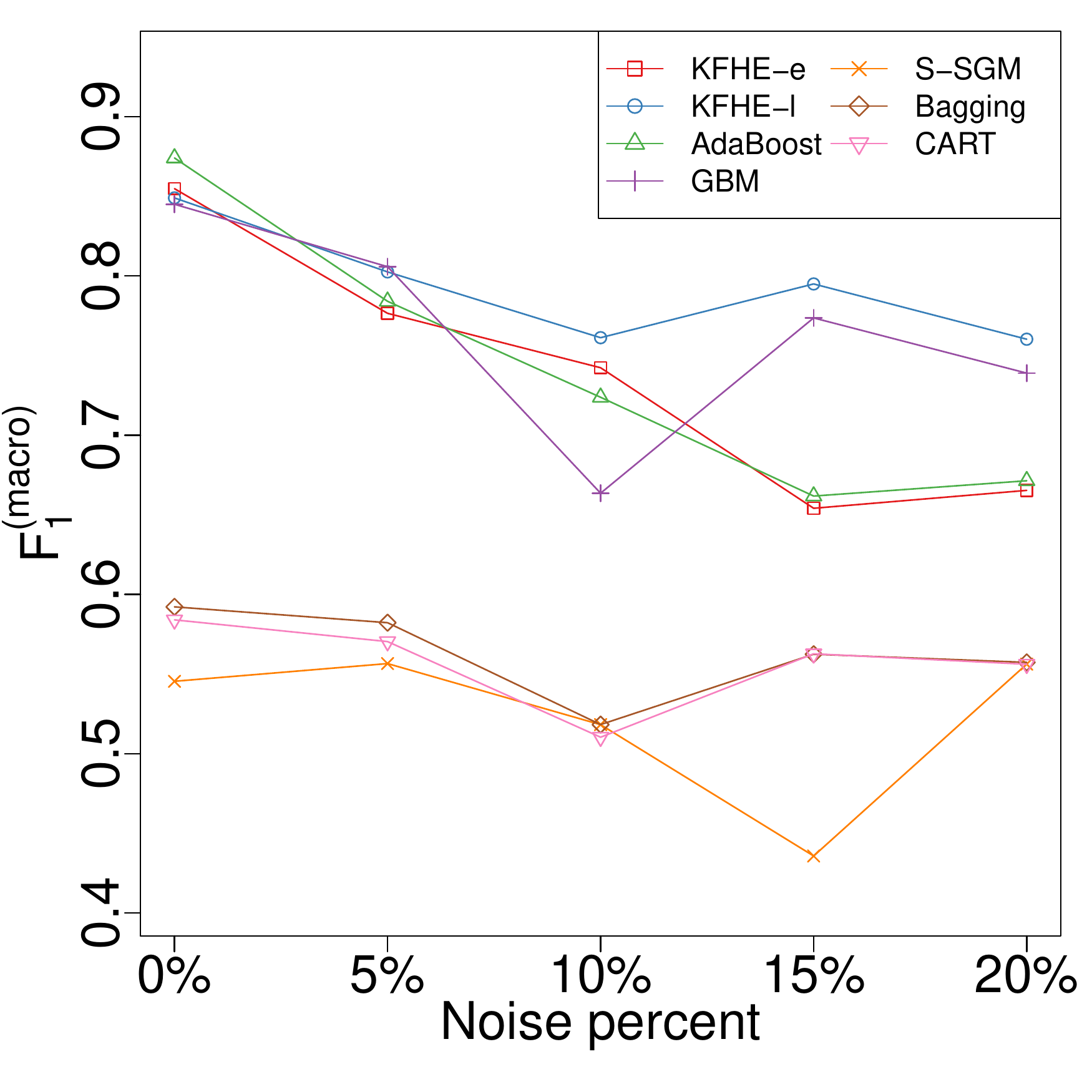}

}

\caption{Changes in $F_{1}^{(macro)}$-score with the induced noise in class
from datasets \emph{yeast} to \emph{zoo} labels\label{fig:Changes-in-fscore-with-noise-02}}
\end{figure}

\begin{figure}[p]
\subfloat[\emph{mushroom} \label{fig:mushroom_noise}]{\includegraphics[width=0.33\textwidth]{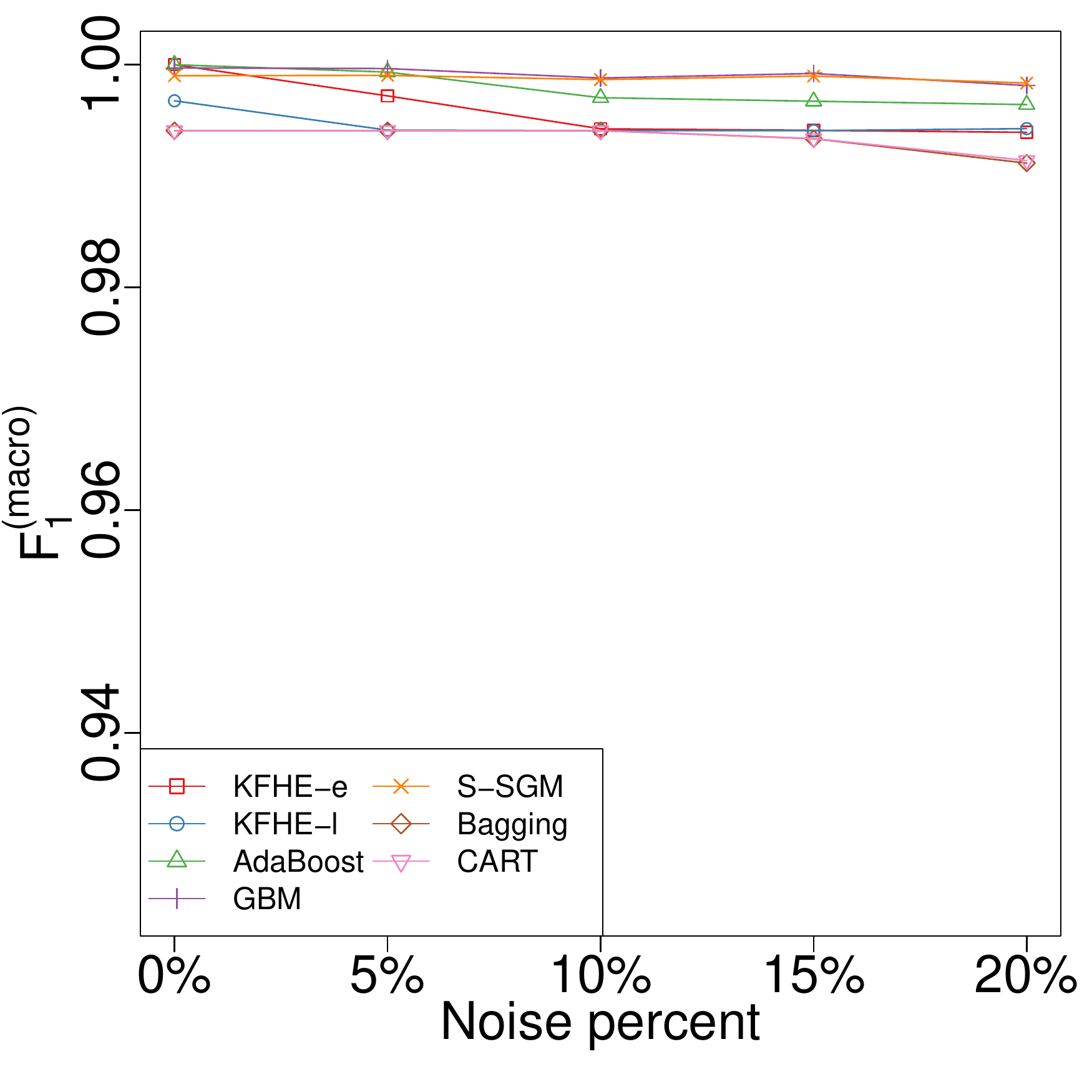}

}\subfloat[\emph{iris} \label{fig:iris_noise}]{\includegraphics[width=0.33\textwidth]{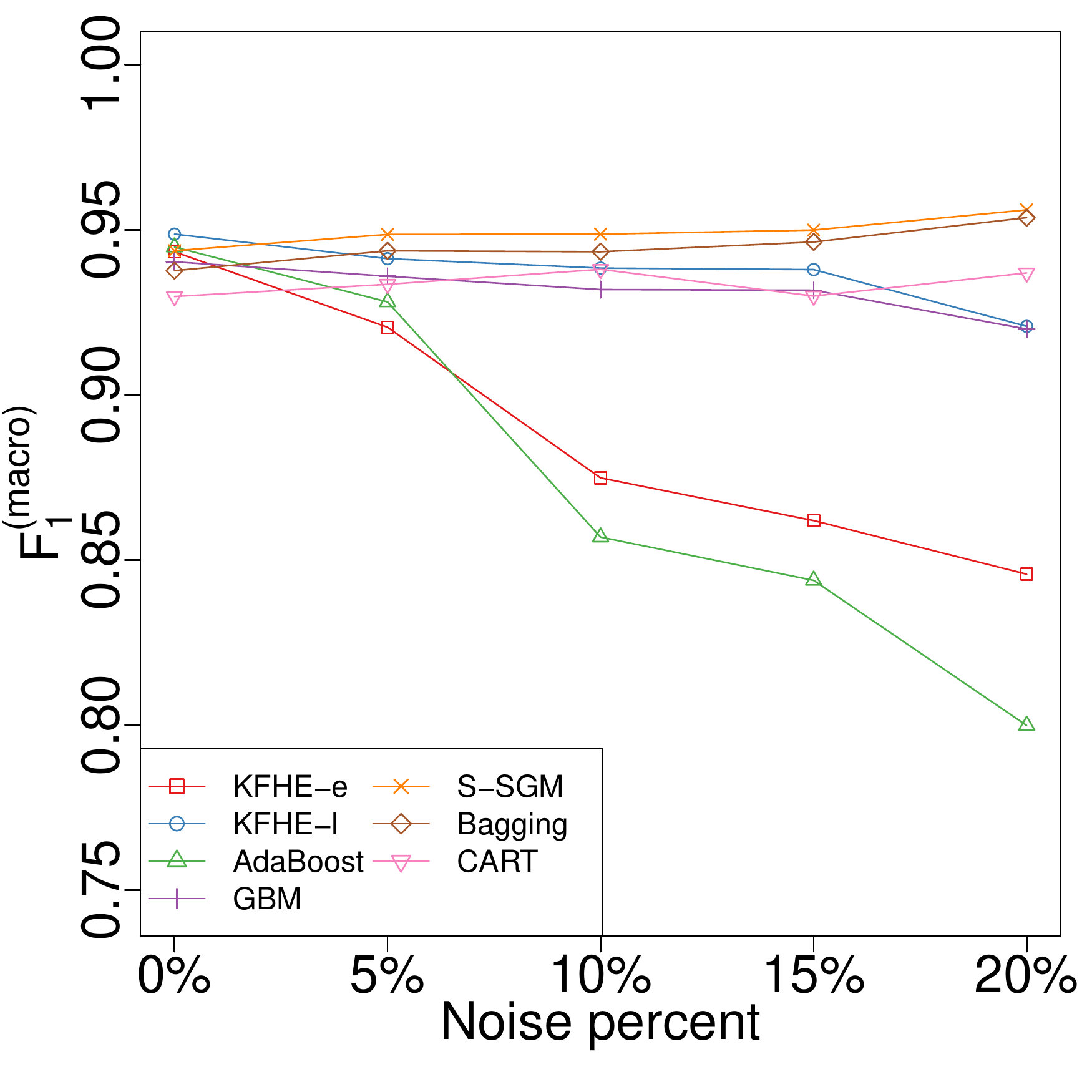}

}\subfloat[\emph{glass} \label{fig:glass_noise}]{\includegraphics[width=0.33\textwidth]{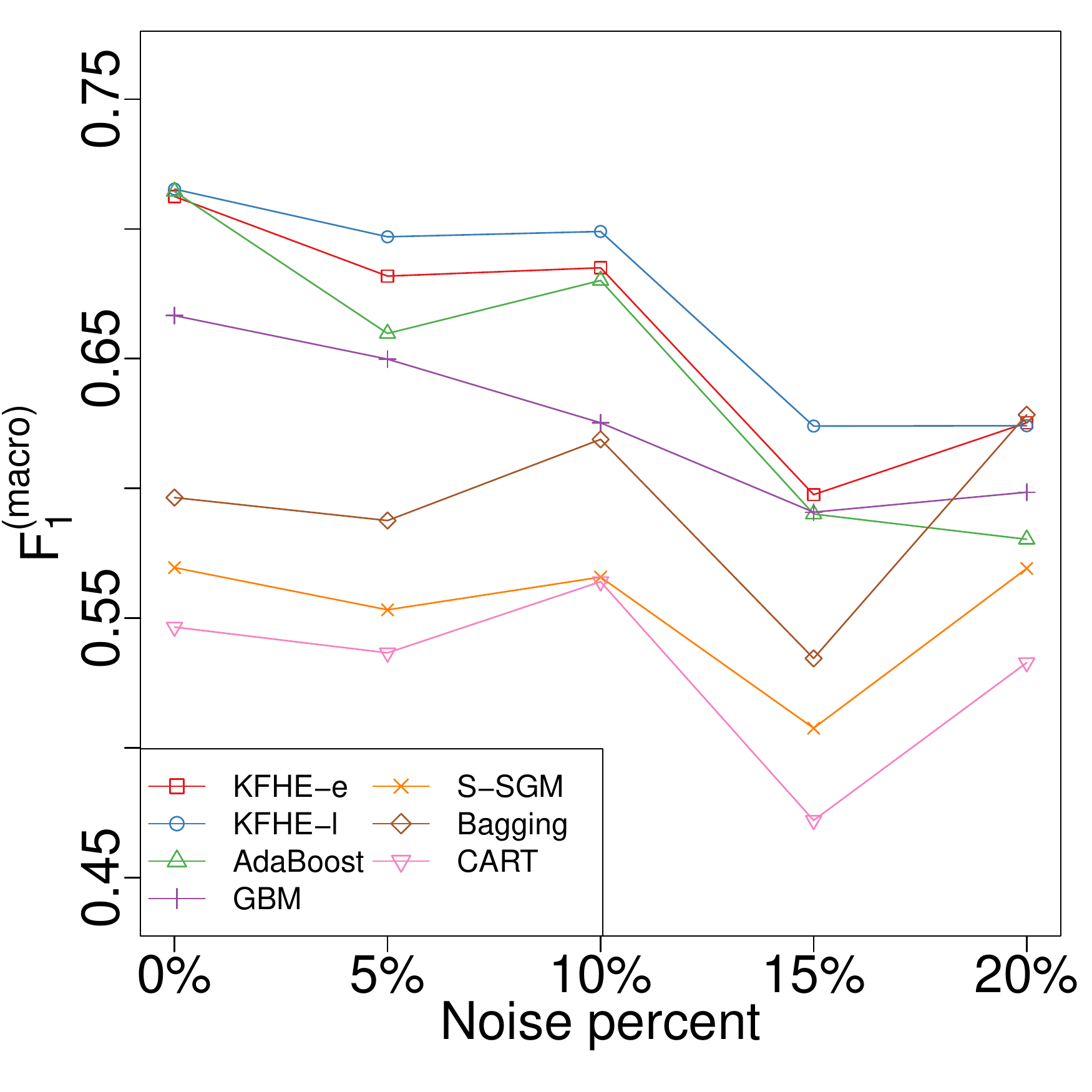}

}

\subfloat[\emph{car\_eval} \label{fig:car_eval_noise}]{\includegraphics[width=0.33\textwidth]{plots/plot_by_data_vs_noise_car_eval.pdf}

}\subfloat[\emph{cmc} \label{fig:cmc_noise}]{\includegraphics[width=0.33\textwidth]{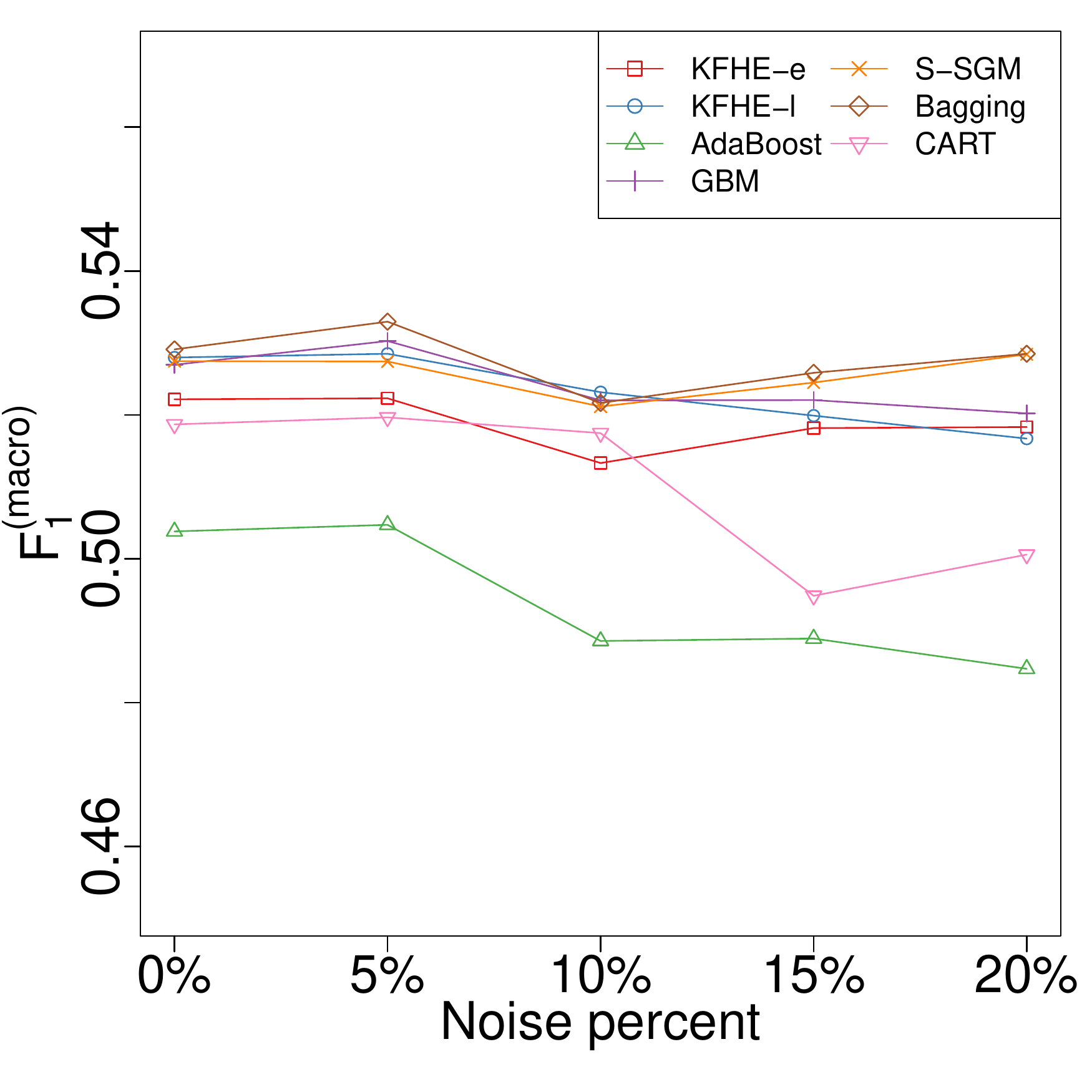}

}\subfloat[\emph{tvowel} \label{fig:tvowel_noise}]{\includegraphics[width=0.33\textwidth]{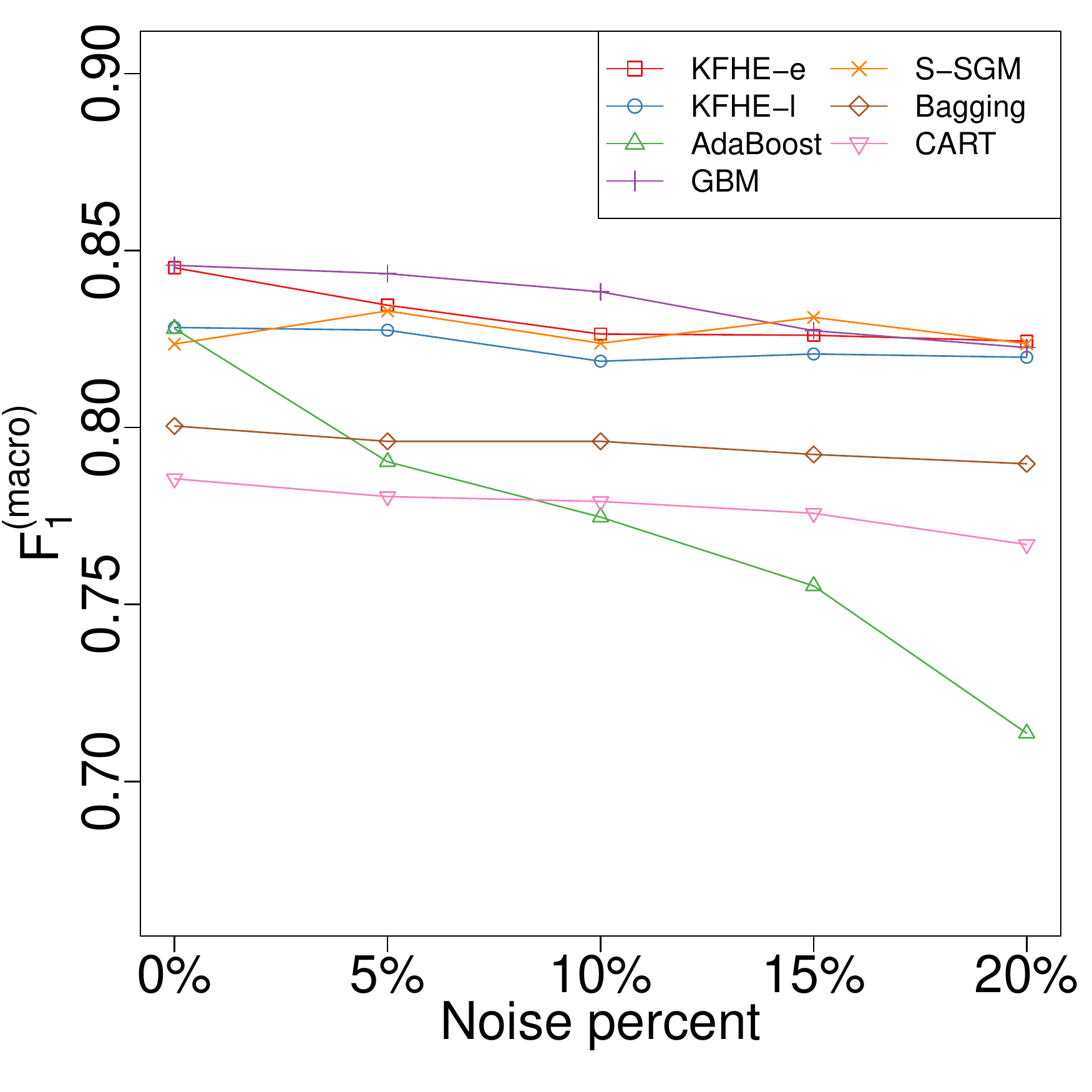}

}

\subfloat[\emph{balance\_scale} \label{fig:balance_scale_noise}]{\includegraphics[width=0.33\textwidth]{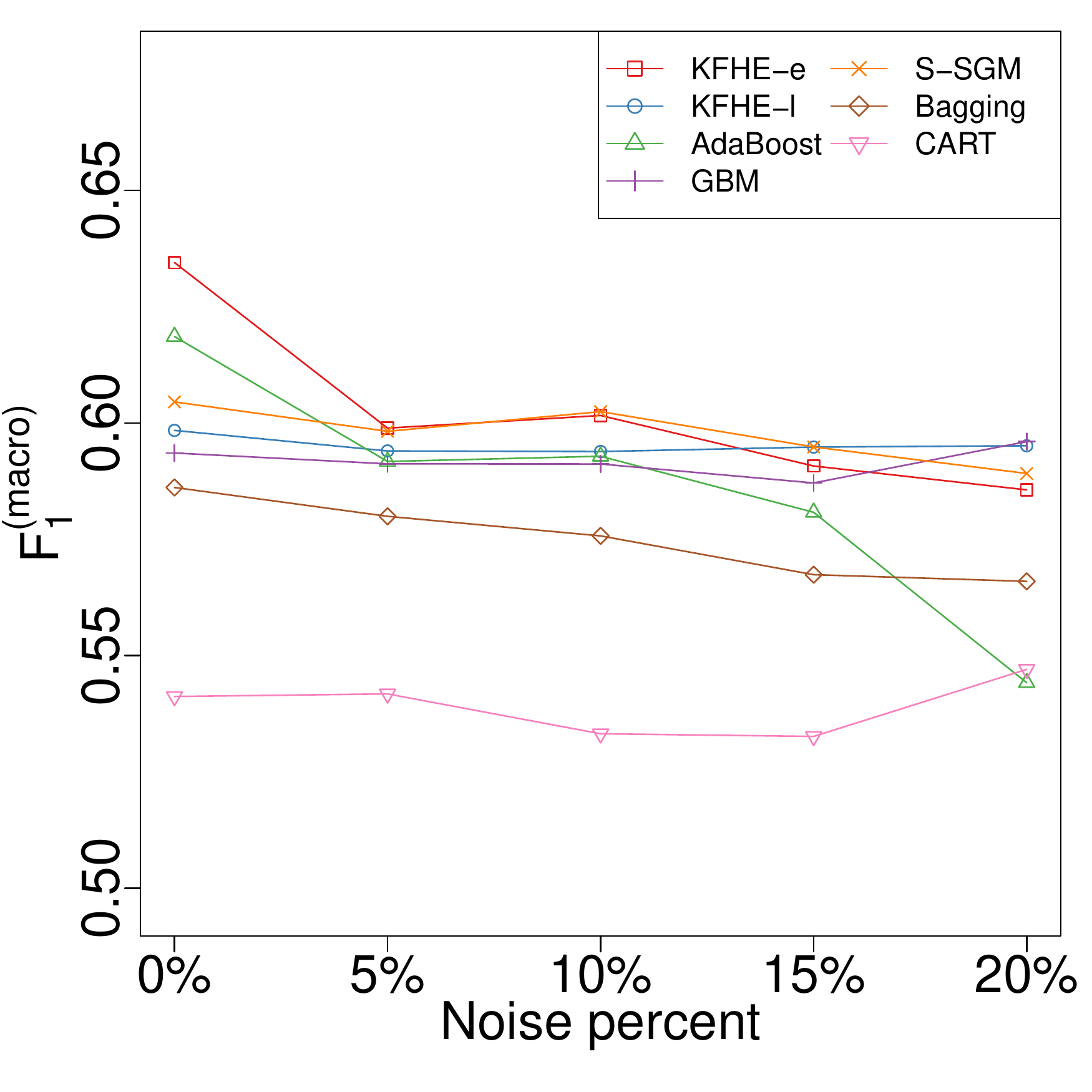}

}\subfloat[\emph{flags} \label{fig:flags_noise}]{\includegraphics[width=0.33\textwidth]{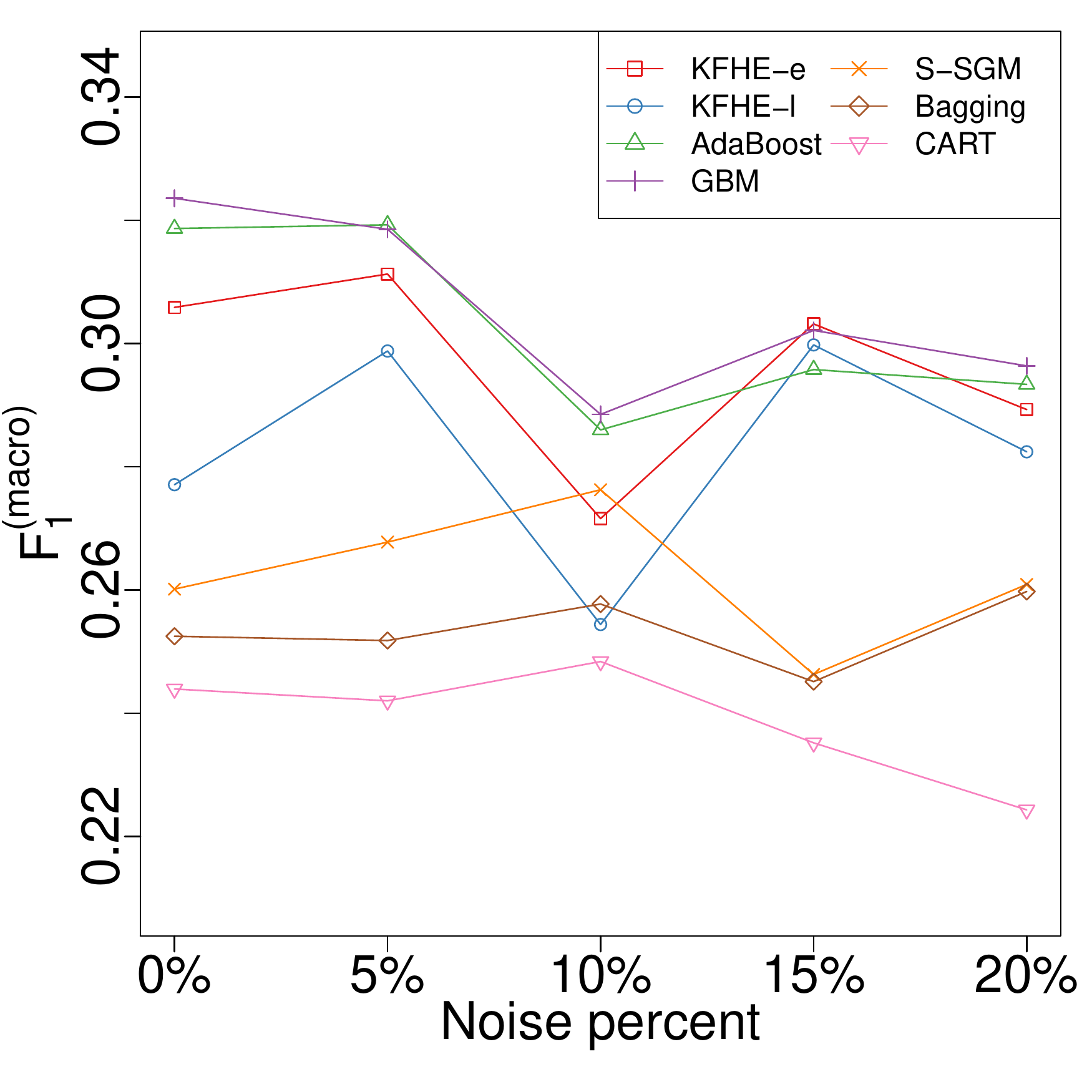}

}\subfloat[\emph{german} \label{fig:german_noise}]{\includegraphics[width=0.33\textwidth]{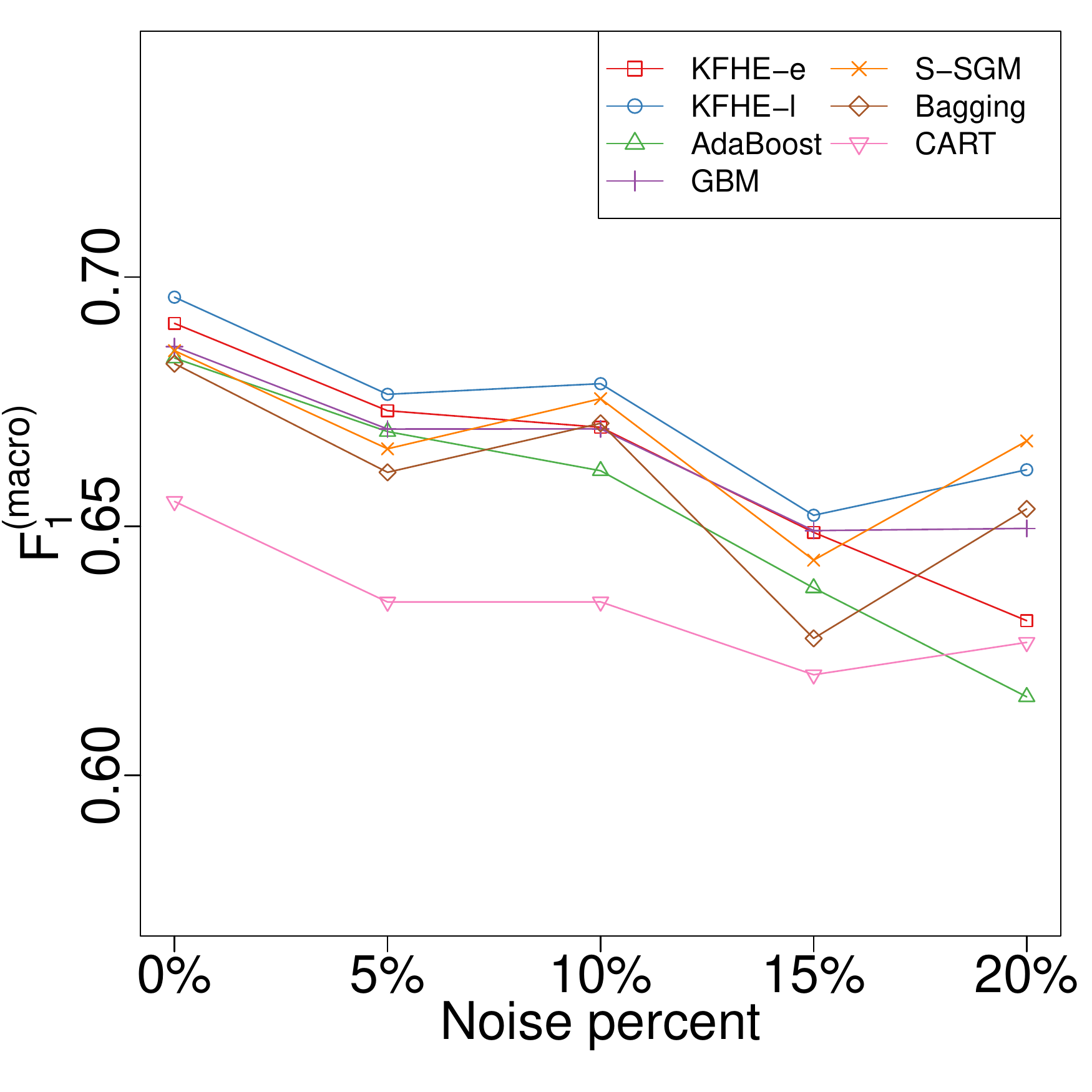}

}

%\subfloat[\emph{dermatology} \label{fig:dermatology_noise}]{\includegraphics[width=0.25\textwidth]{plots/plot_by_data_vs_noise_dermatology}}%\subfloat[\emph{breastcancer} \label{fig:breastcancer_noise}]{\includegraphics[width=0.25\textwidth]{plots/plot_by_data_vs_noise_breastcancer}}%\subfloat[\emph{wine} \label{fig:wine_noise}]{\includegraphics[width=0.25\textwidth]{plots/plot_by_data_vs_noise_wine}}

\subfloat[\emph{ilpd} \label{fig:ilpd_noise}]{\includegraphics[width=0.33\textwidth]{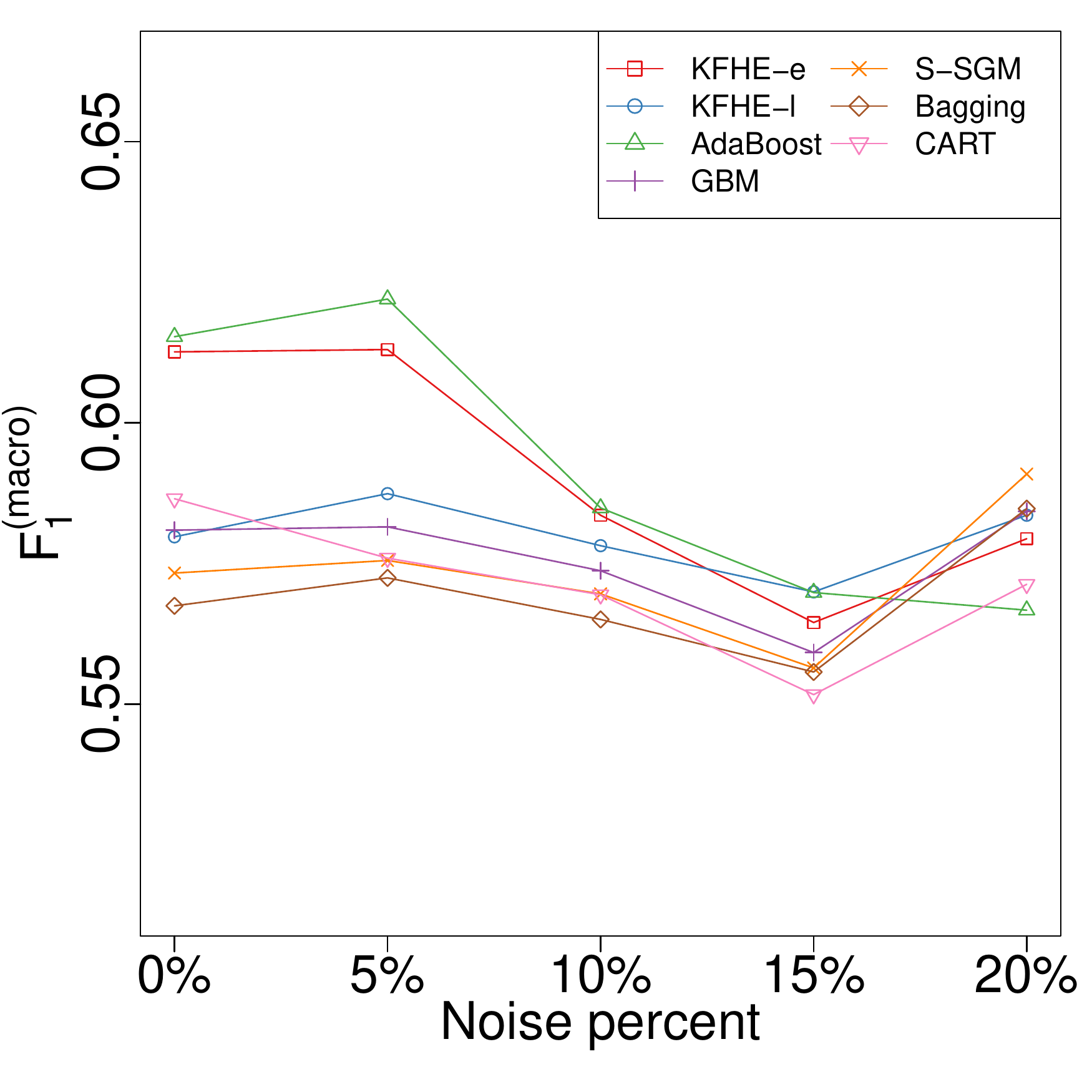}

}\subfloat[\emph{ionosphere} \label{fig:ionosphere_noise}]{\includegraphics[width=0.33\textwidth]{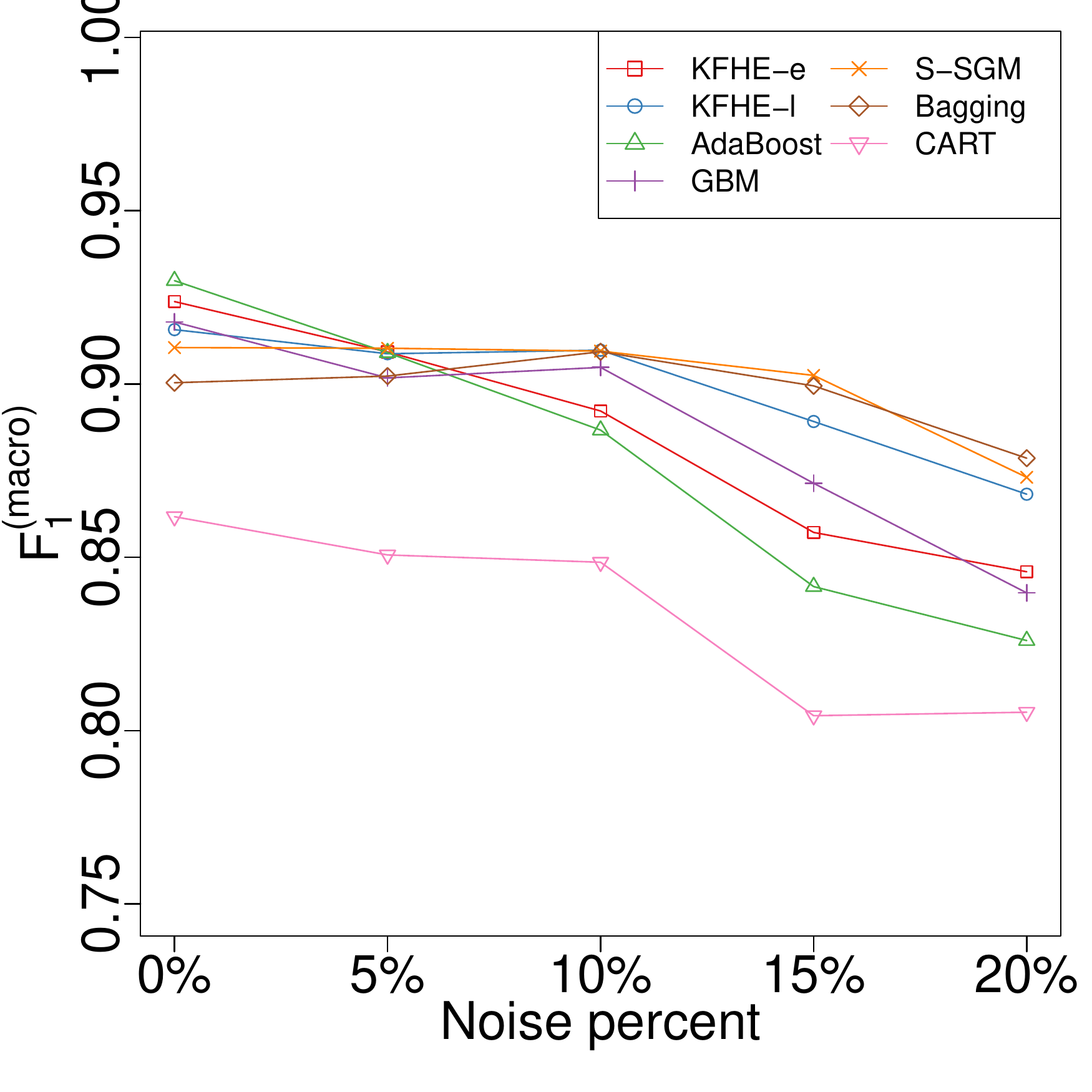}

}\subfloat[\emph{knowledge} \label{fig:knowledge_noise}]{\includegraphics[width=0.33\textwidth]{plots/plot_by_data_vs_noise_knowledge.pdf}

}

\caption{Changes in $F_{1}^{(macro)}$-score with the induced noise in class
from datasets \emph{mushroom} to \emph{knowledge} labels\label{fig:Changes-in-fscore-with-noise-00}}
\end{figure}

%\subfloat[\emph{ecoli} \label{fig:ecoli_noise}]{\includegraphics[width=0.33\textwidth]{plots/plot_by_data_vs_noise_ecoli}}

\begin{figure}[p]
\subfloat[\emph{vertebral} \label{fig:vertebral_noise}]{\includegraphics[width=0.33\textwidth]{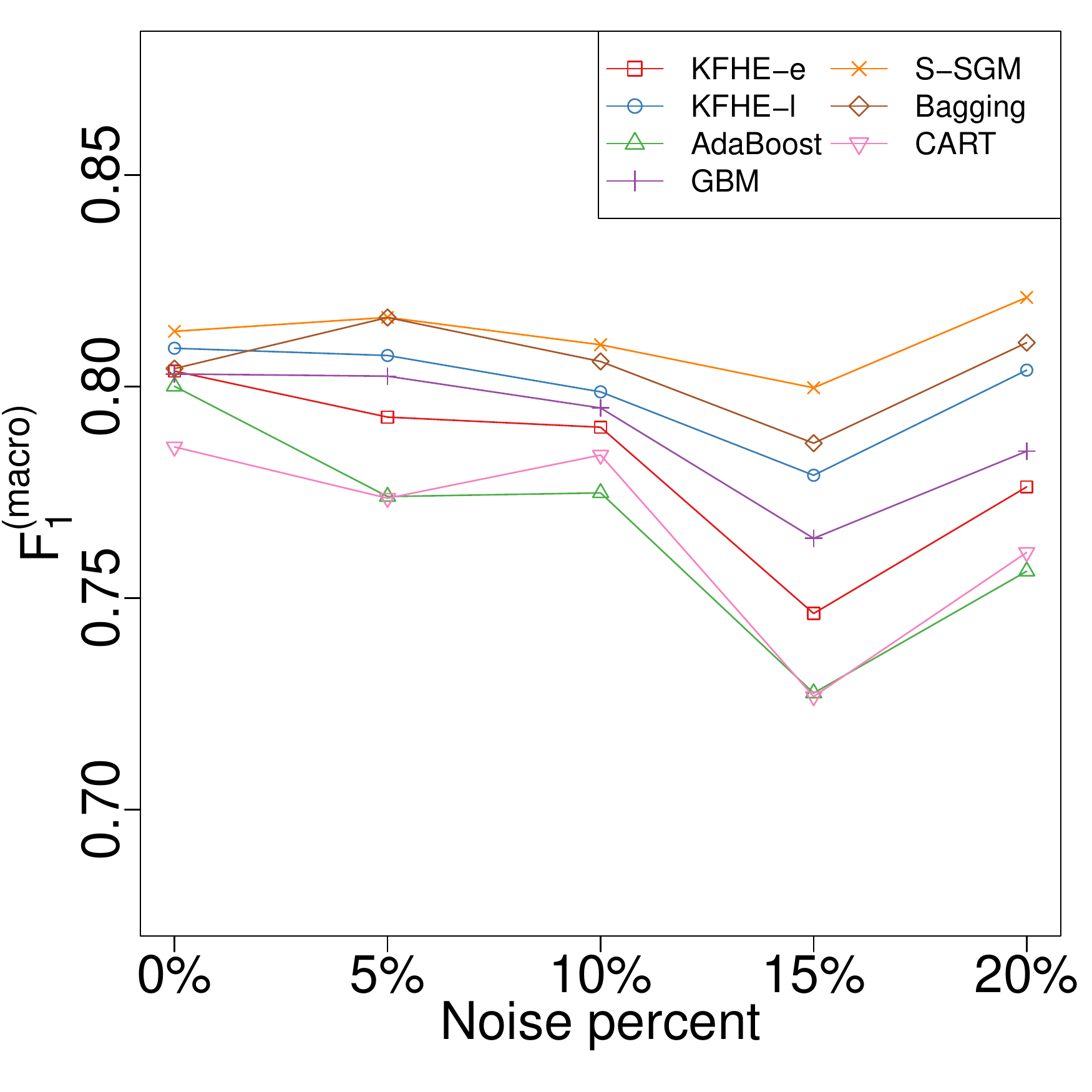}

}\subfloat[\emph{sonar} \label{fig:sonar_noise}]{\includegraphics[width=0.33\textwidth]{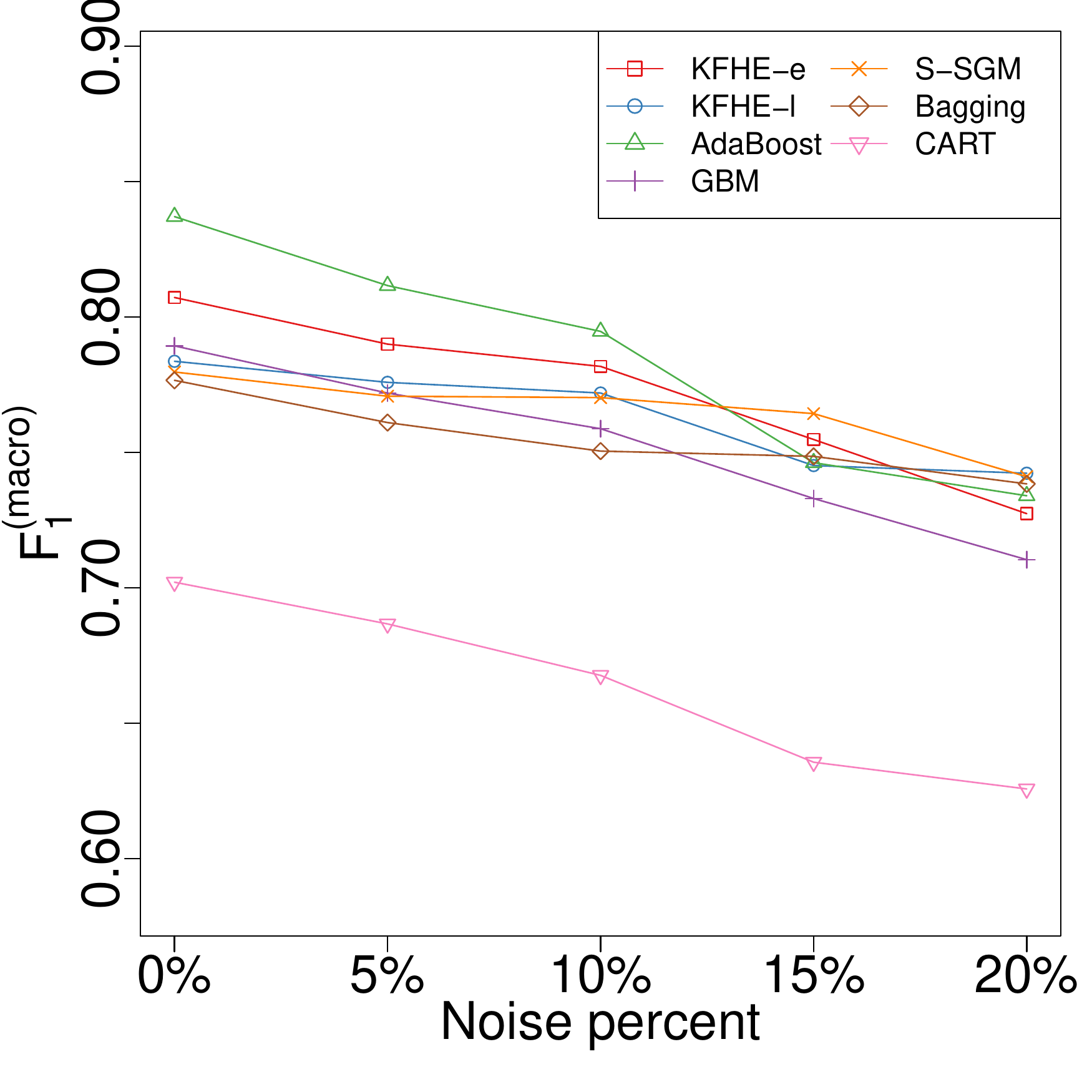}

}\subfloat[\emph{skulls} \label{fig:skulls_noise}]{\includegraphics[width=0.33\textwidth]{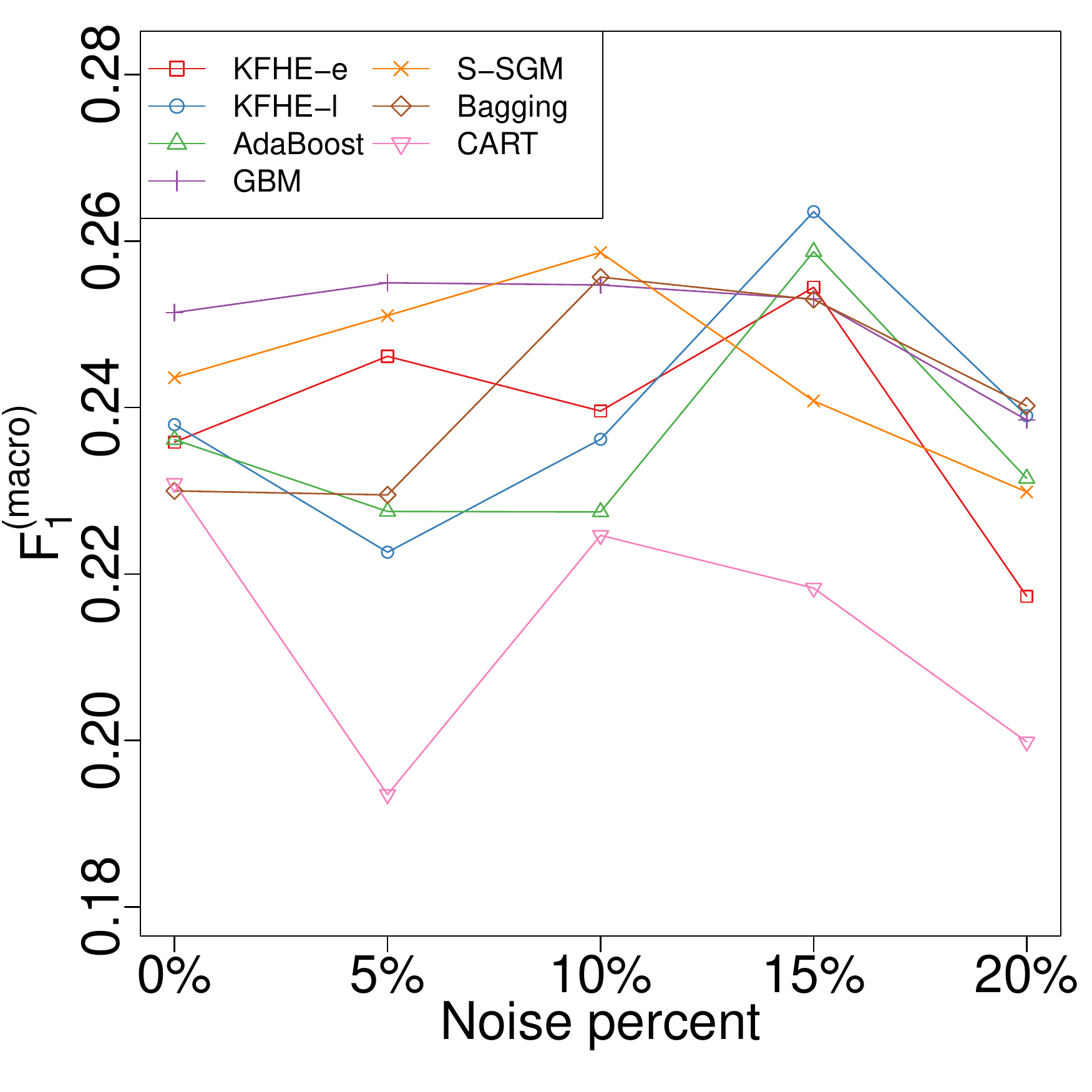}

}

\subfloat[\emph{diabetes} \label{fig:diabetes_noise}]{\includegraphics[width=0.33\textwidth]{plots/plot_by_data_vs_noise_diabetes.pdf}

}\subfloat[\emph{physio} \label{fig:physio_noise}]{\includegraphics[width=0.33\textwidth]{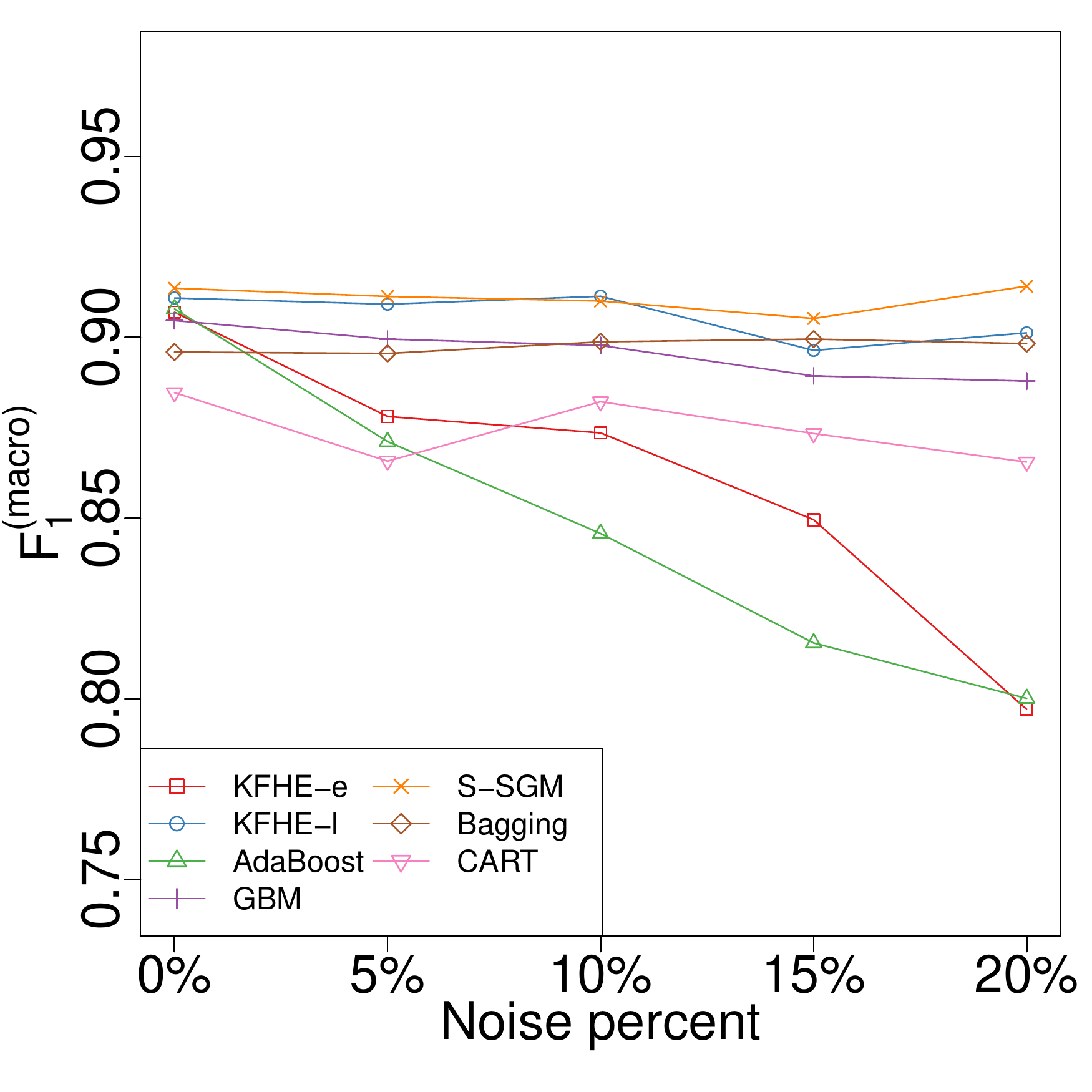}

}\subfloat[\emph{breasttissue} \label{fig:breasttissue_noise}]{\includegraphics[width=0.33\textwidth]{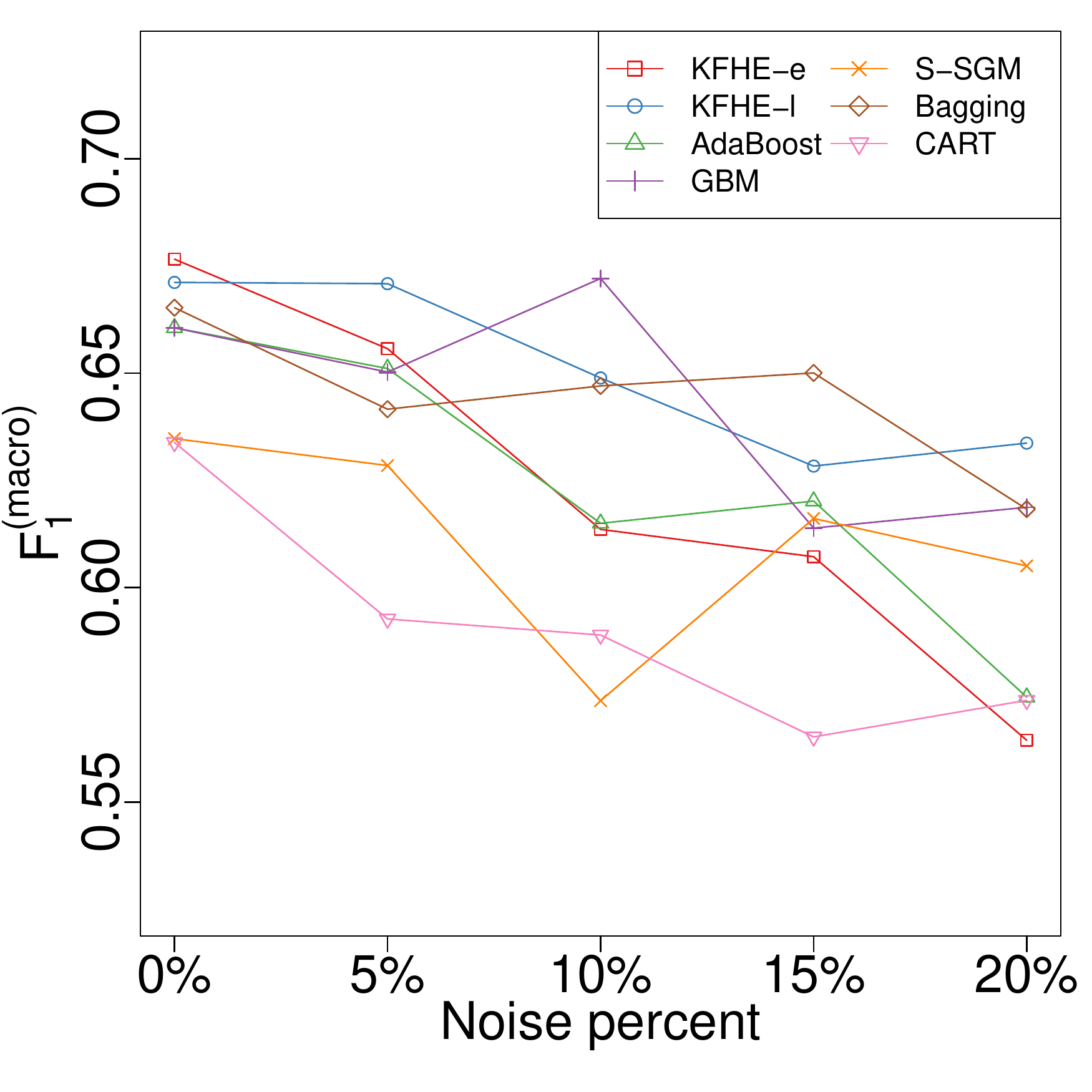}

}

\subfloat[\emph{bupa} \label{fig:bupa_noise}]{\includegraphics[width=0.33\textwidth]{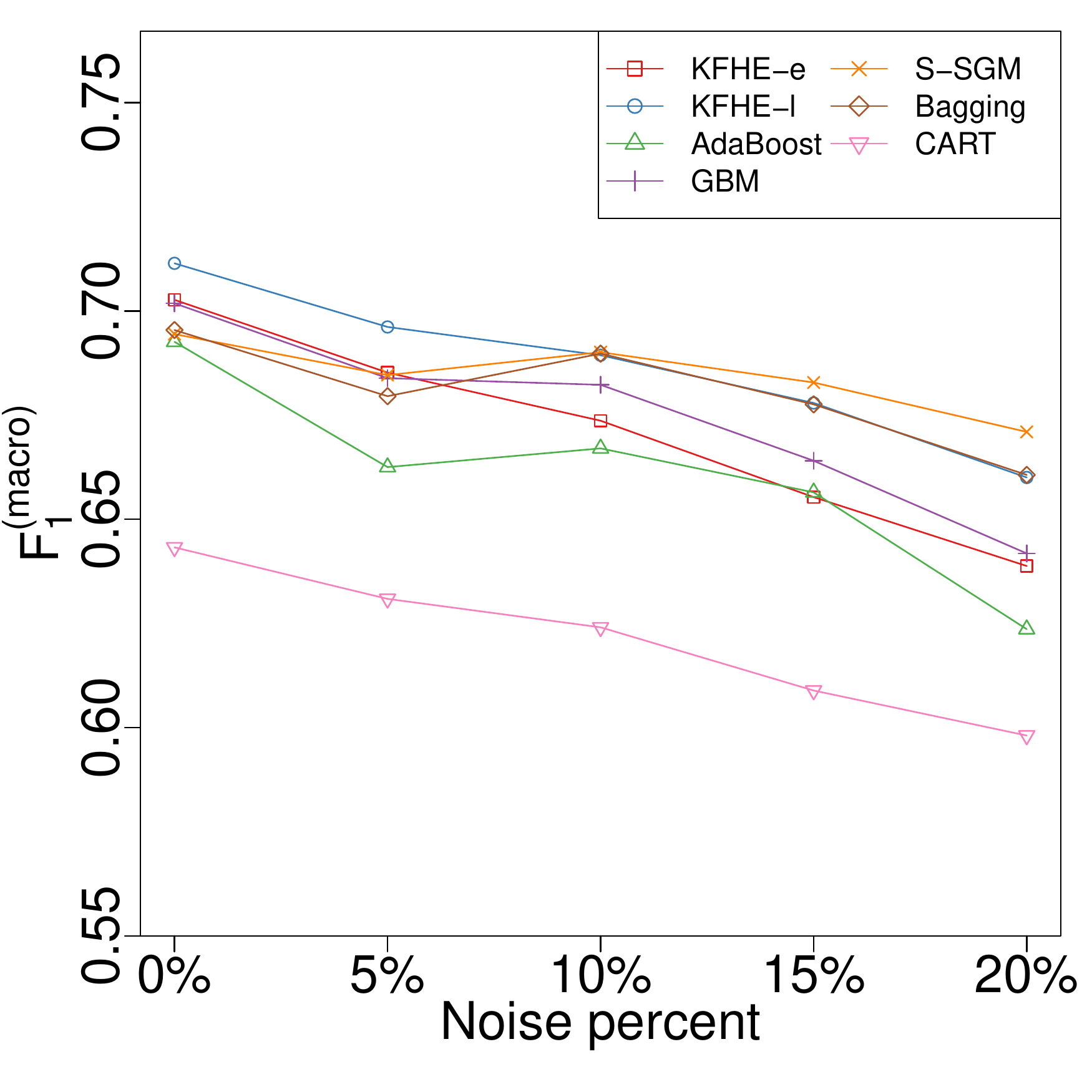}

}\subfloat[\emph{cleveland} \label{fig:cleveland_noise}]{\includegraphics[width=0.33\textwidth]{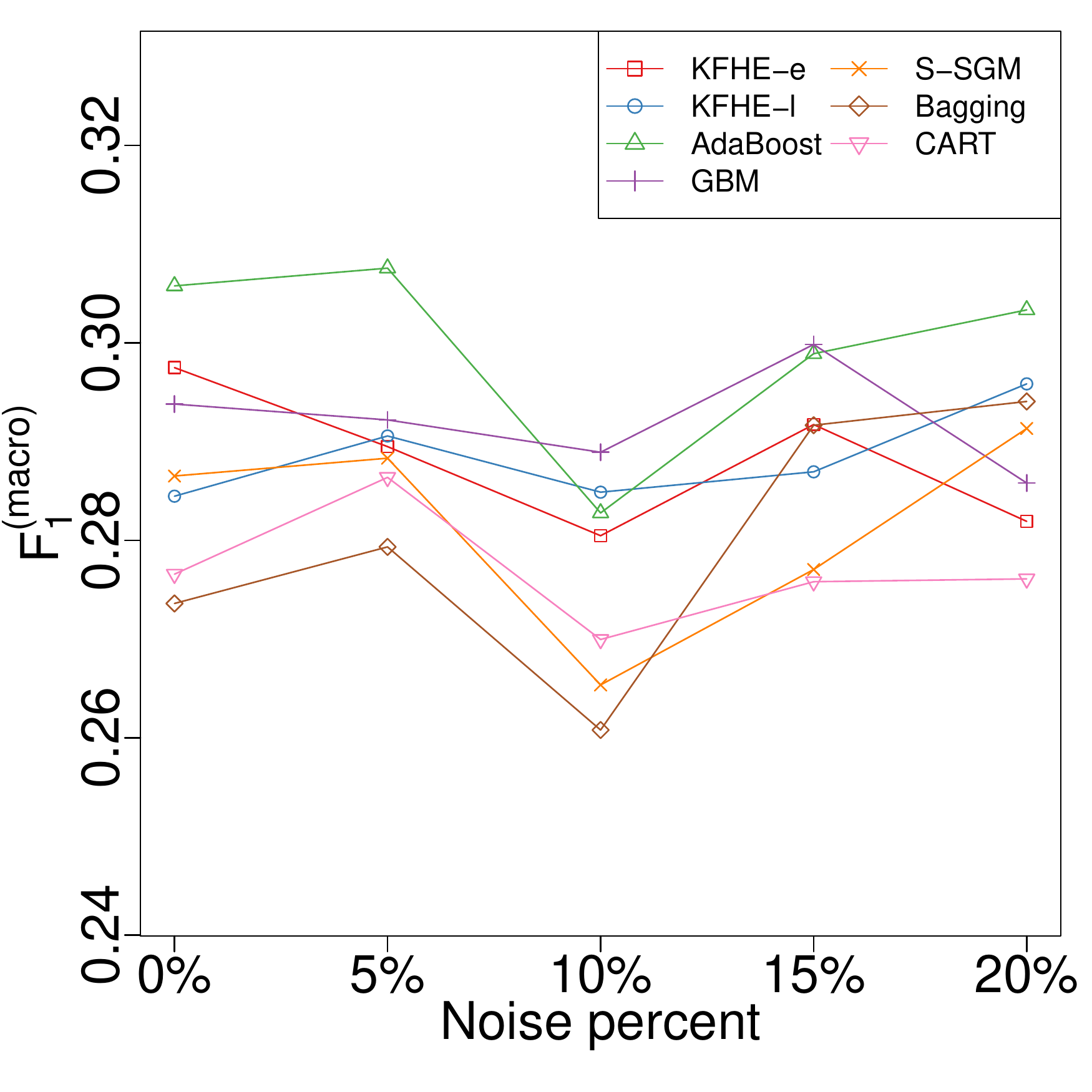}

}\subfloat[\emph{haberman} \label{fig:haberman_noise}]{\includegraphics[width=0.33\textwidth]{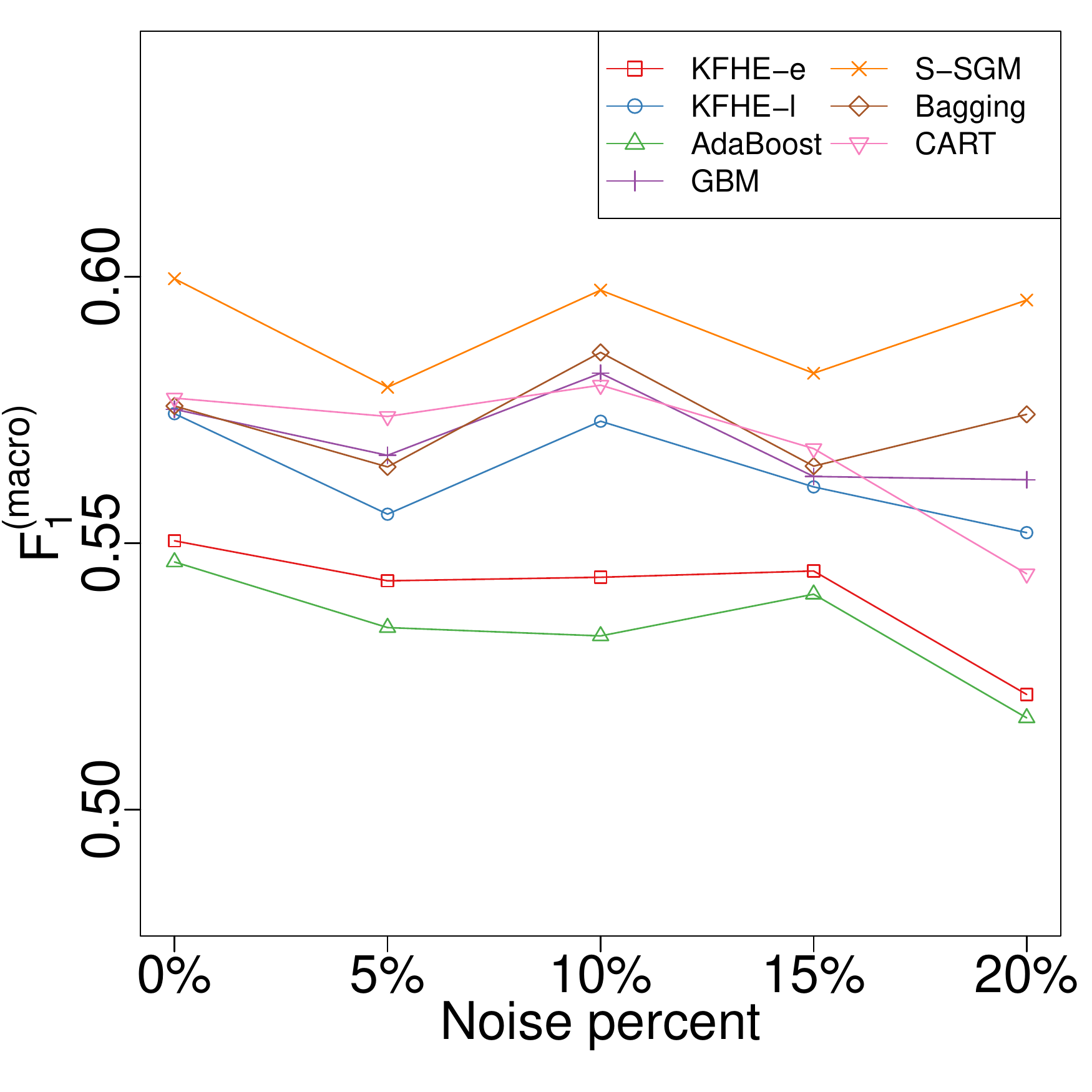}

}

\subfloat[\emph{hayes\_roth} \label{fig:hayes_roth_noise}]{\includegraphics[width=0.33\textwidth]{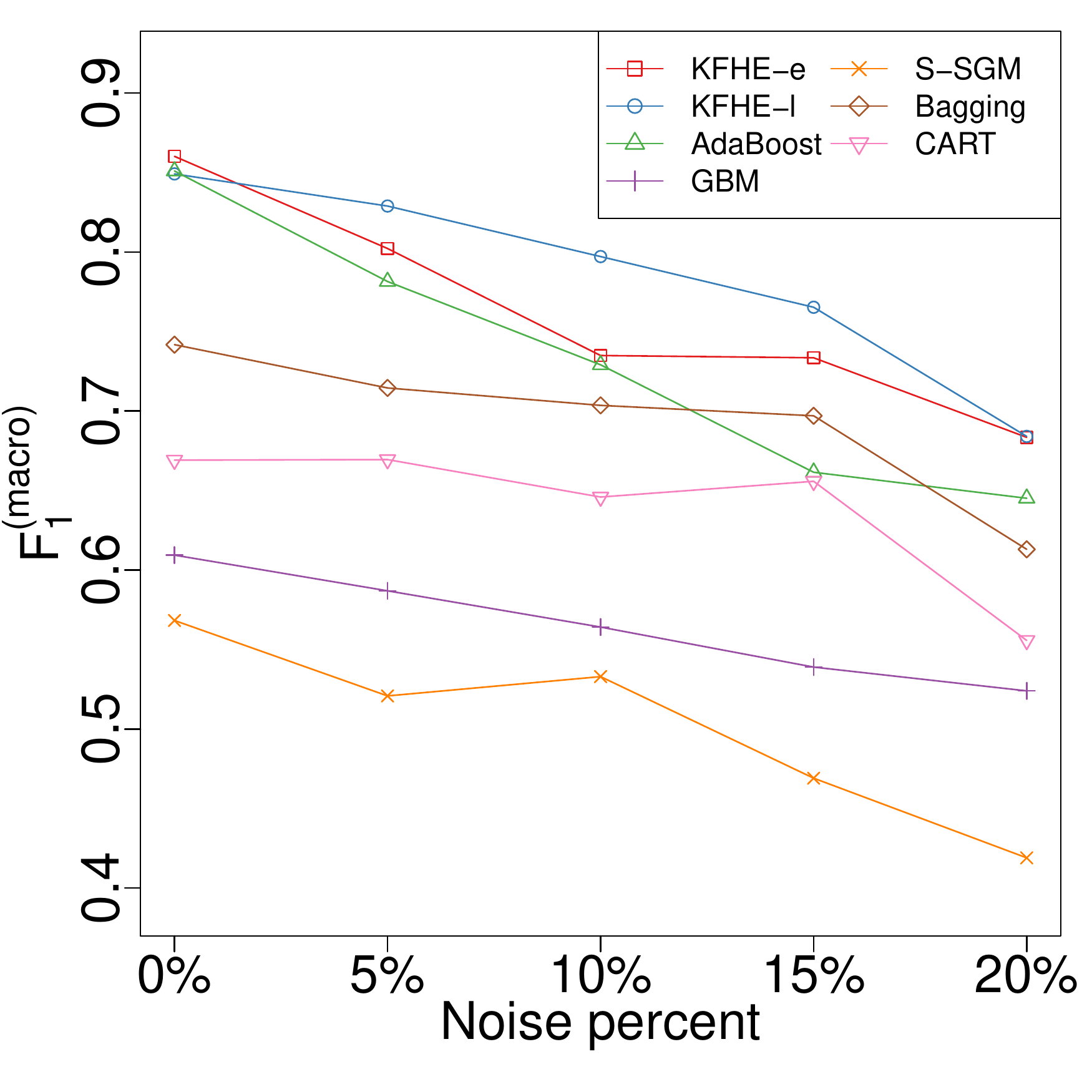}

}\subfloat[\emph{monks} \label{fig:monks_noise}]{\includegraphics[width=0.33\textwidth]{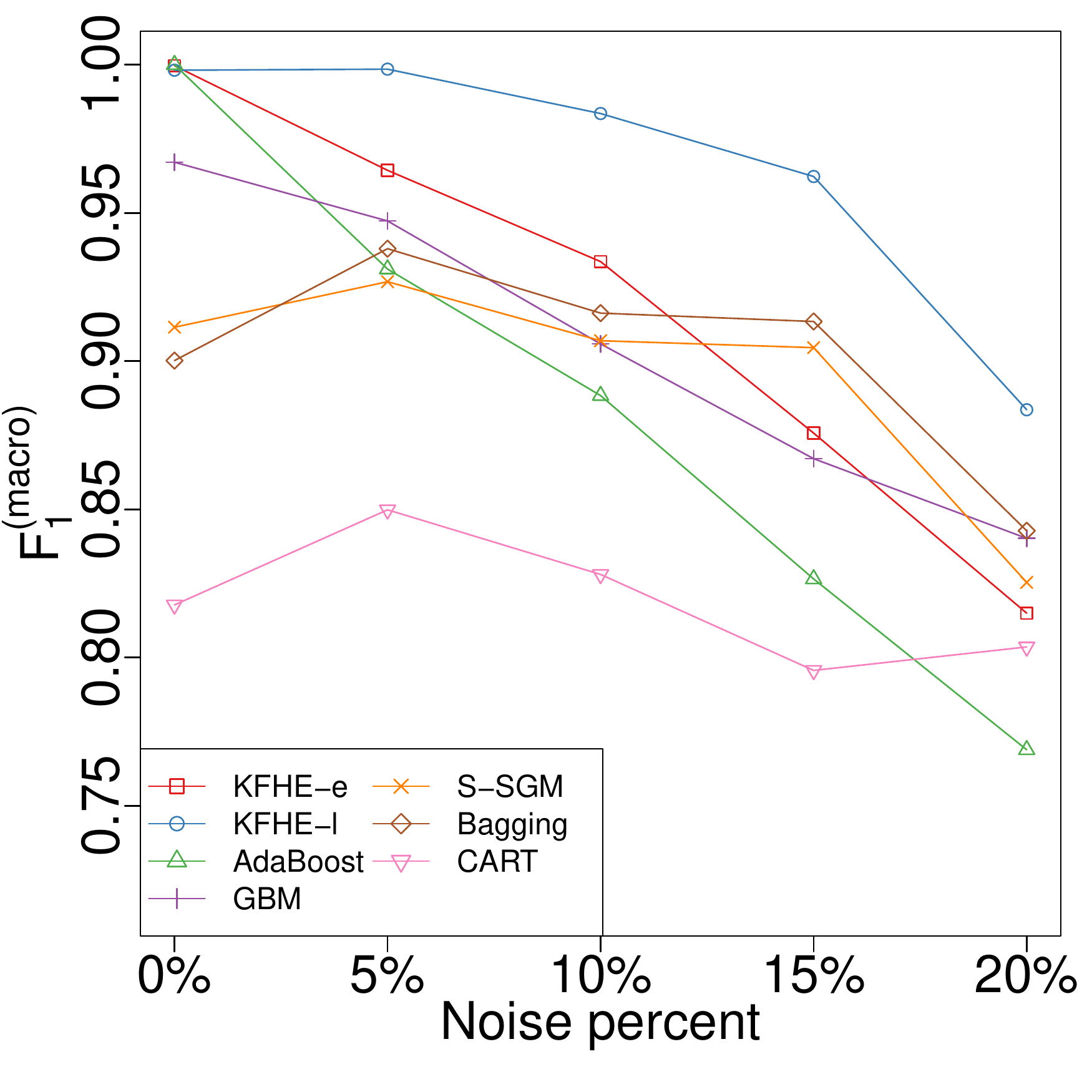}

}\subfloat[\emph{newthyroid} \label{fig:newthyroid_noise}]{\includegraphics[width=0.33\textwidth]{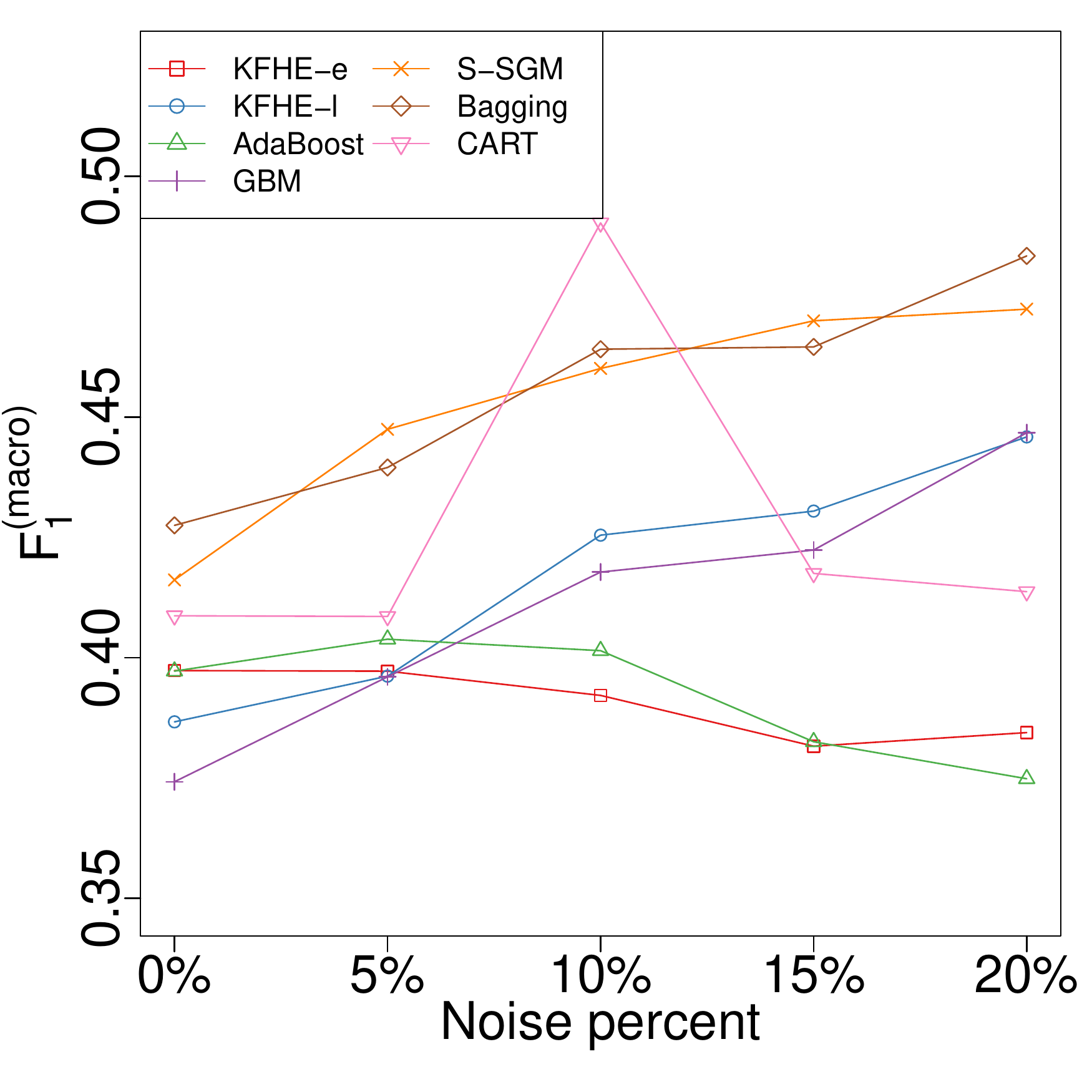}

}

\caption{Changes in $F_{1}^{(macro)}$-score with the induced noise in class
from datasets \emph{vertebral} to \emph{newthyroid} labels\label{fig:Changes-in-fscore-with-noise-01}}
\end{figure}

\clearpage
\section{Plots of how model parameters change during training\label{sec:parameterPlots}}

\begin{figure}[ph]
\subfloat[\emph{mushroom} \label{fig:mushroom_params}]{\includegraphics[width=0.5\textwidth]{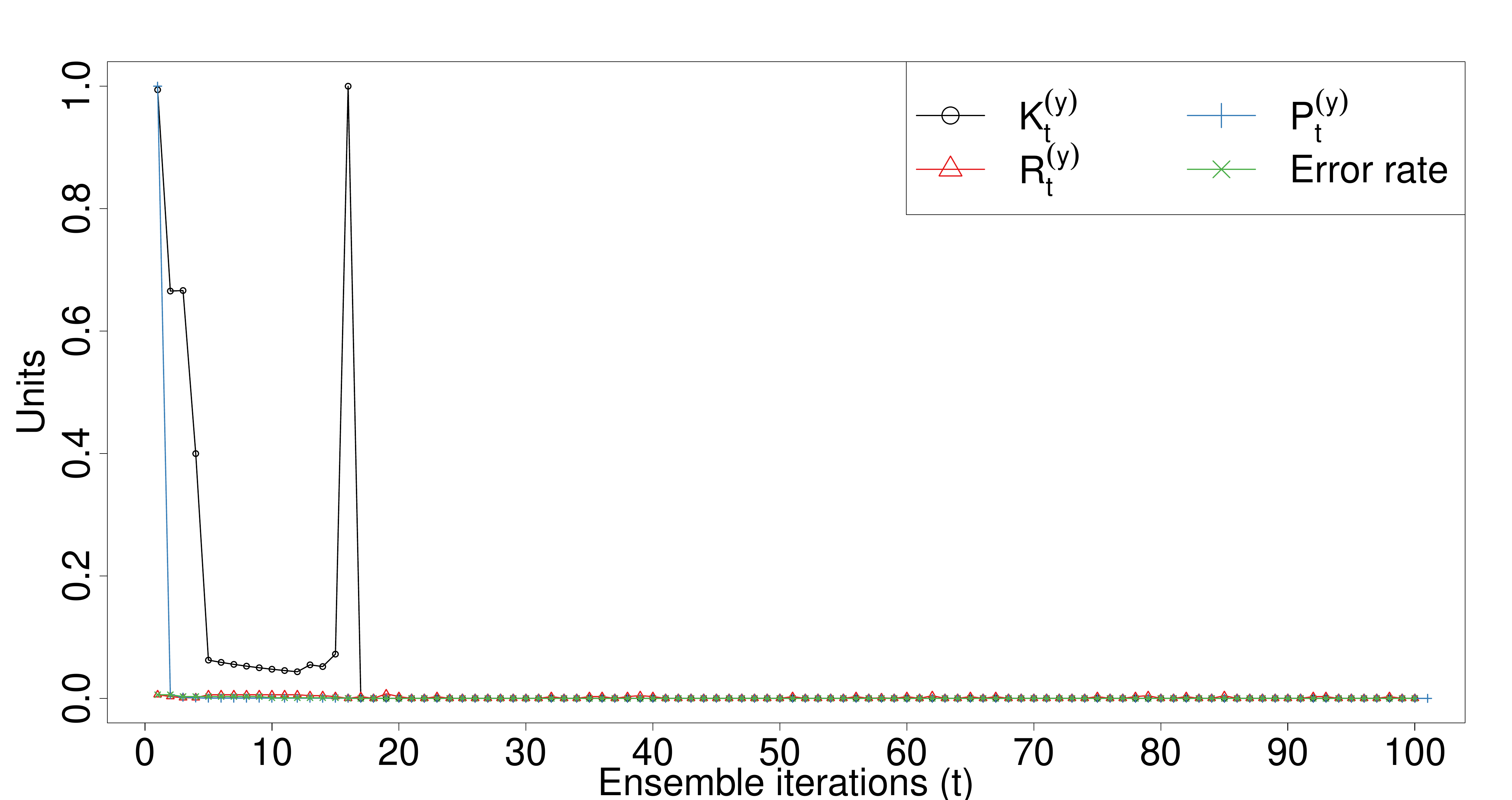}

}\subfloat[\emph{iris} \label{fig:iris_params}]{\includegraphics[width=0.5\textwidth]{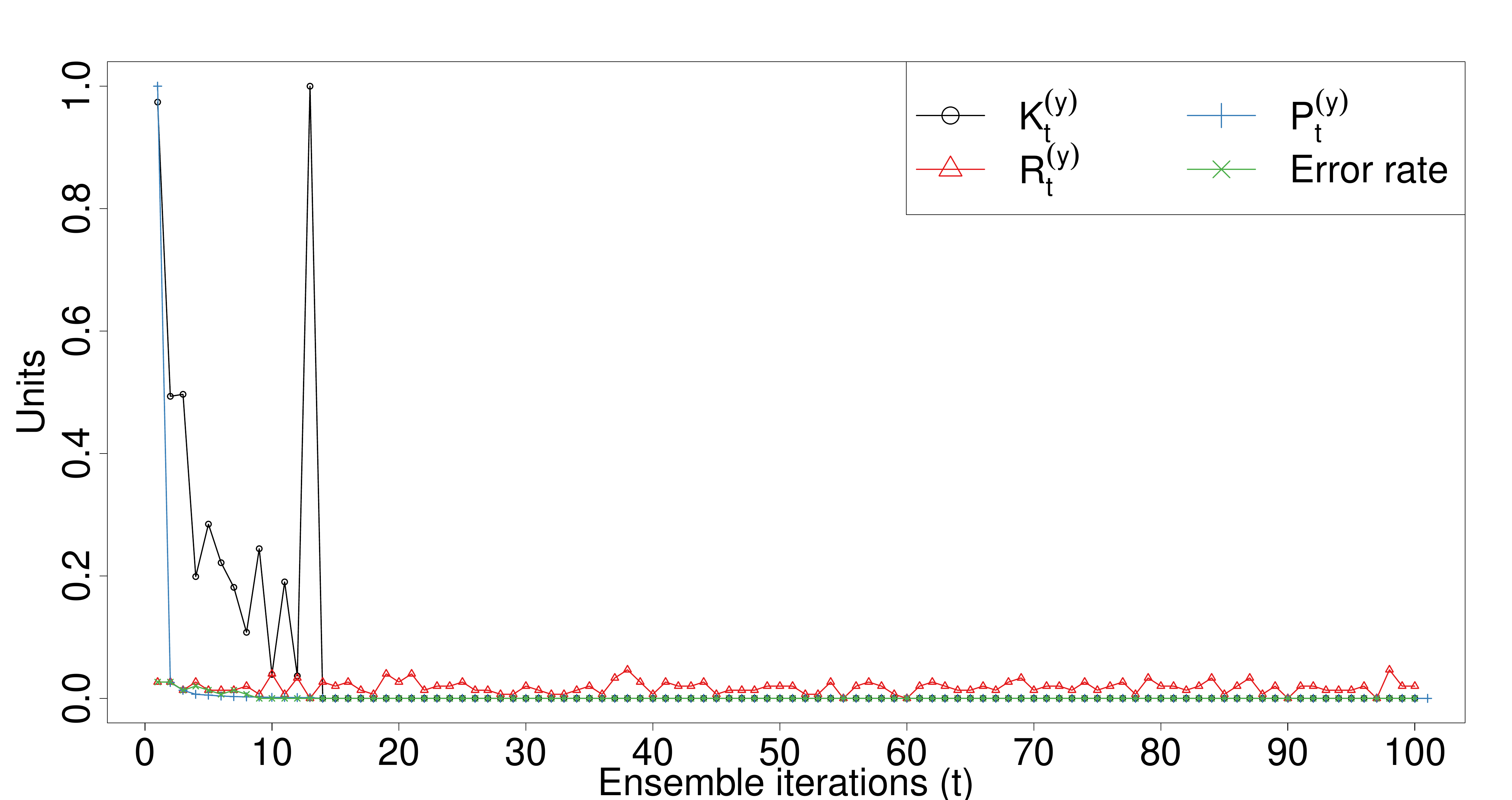}

}

\subfloat[\emph{glass} \label{fig:glass_params}]{\includegraphics[width=0.5\textwidth]{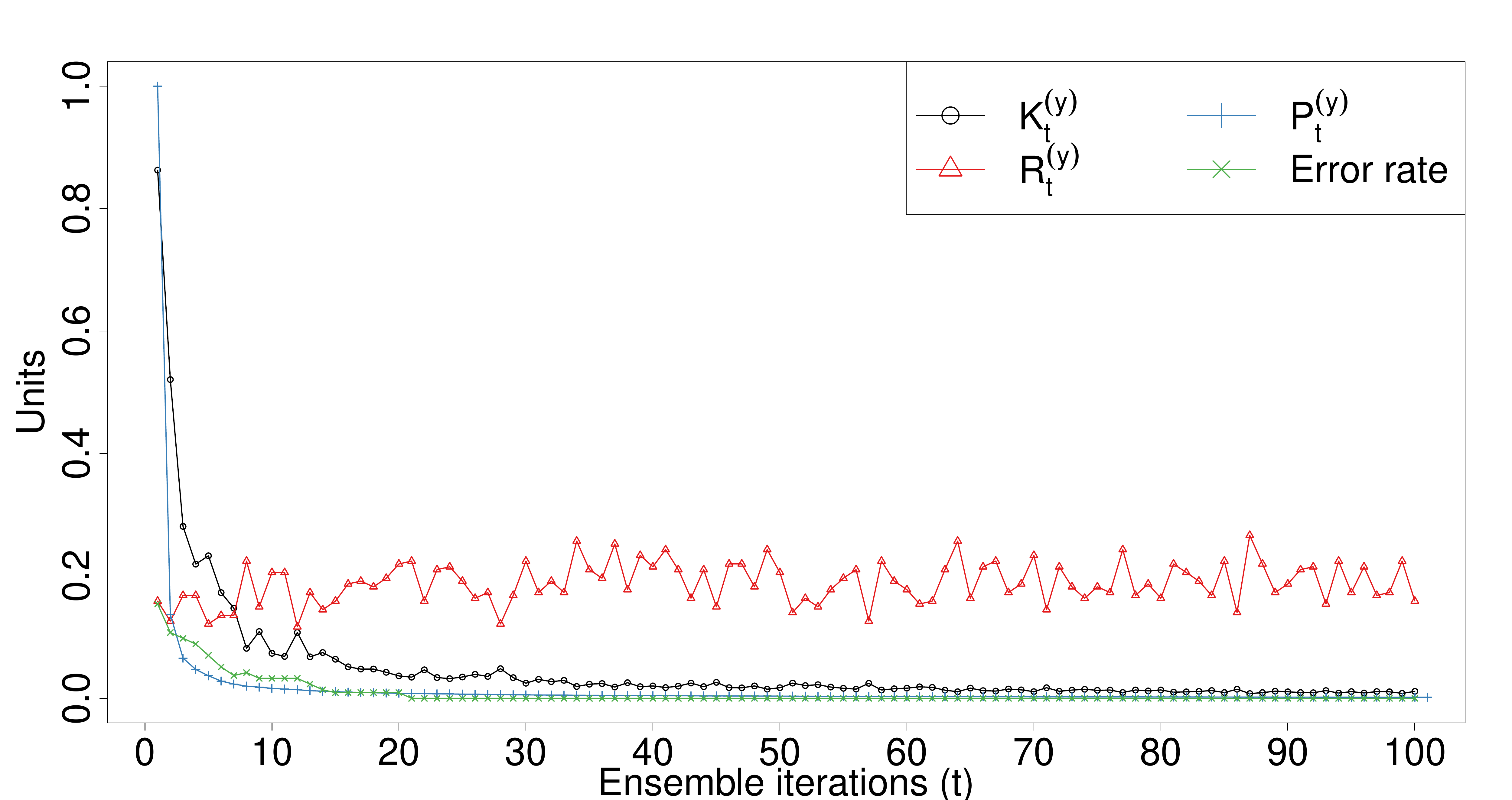}

}\subfloat[\emph{car\_eval} \label{fig:car_eval_params}]{\includegraphics[width=0.5\textwidth]{plots/car_eval_plot.pdf}

}

\subfloat[\emph{cmc} \label{fig:cmc_params}]{\includegraphics[width=0.5\textwidth]{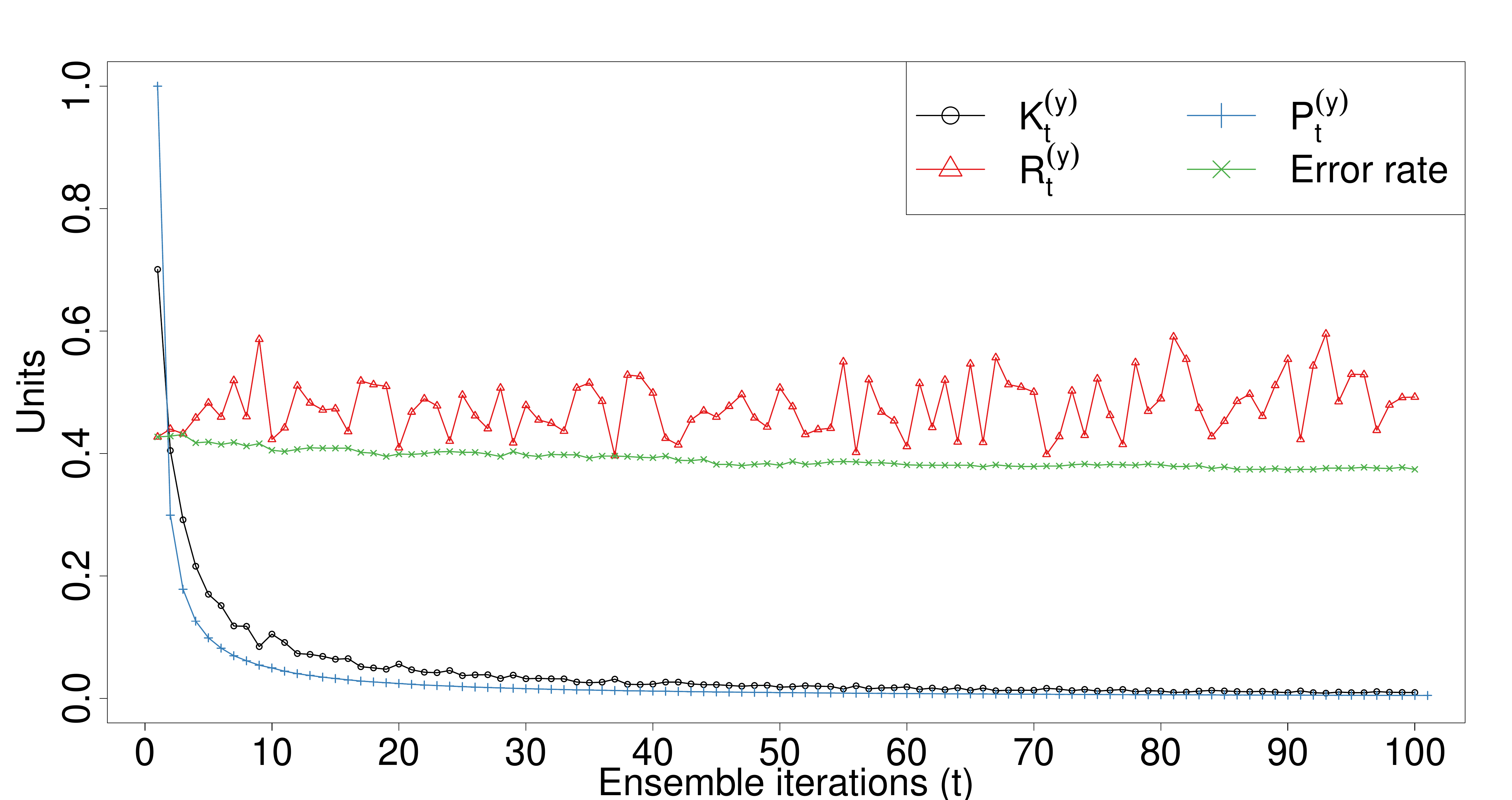}

}\subfloat[\emph{tvowel} \label{fig:tvowel_params}]{\includegraphics[width=0.5\textwidth]{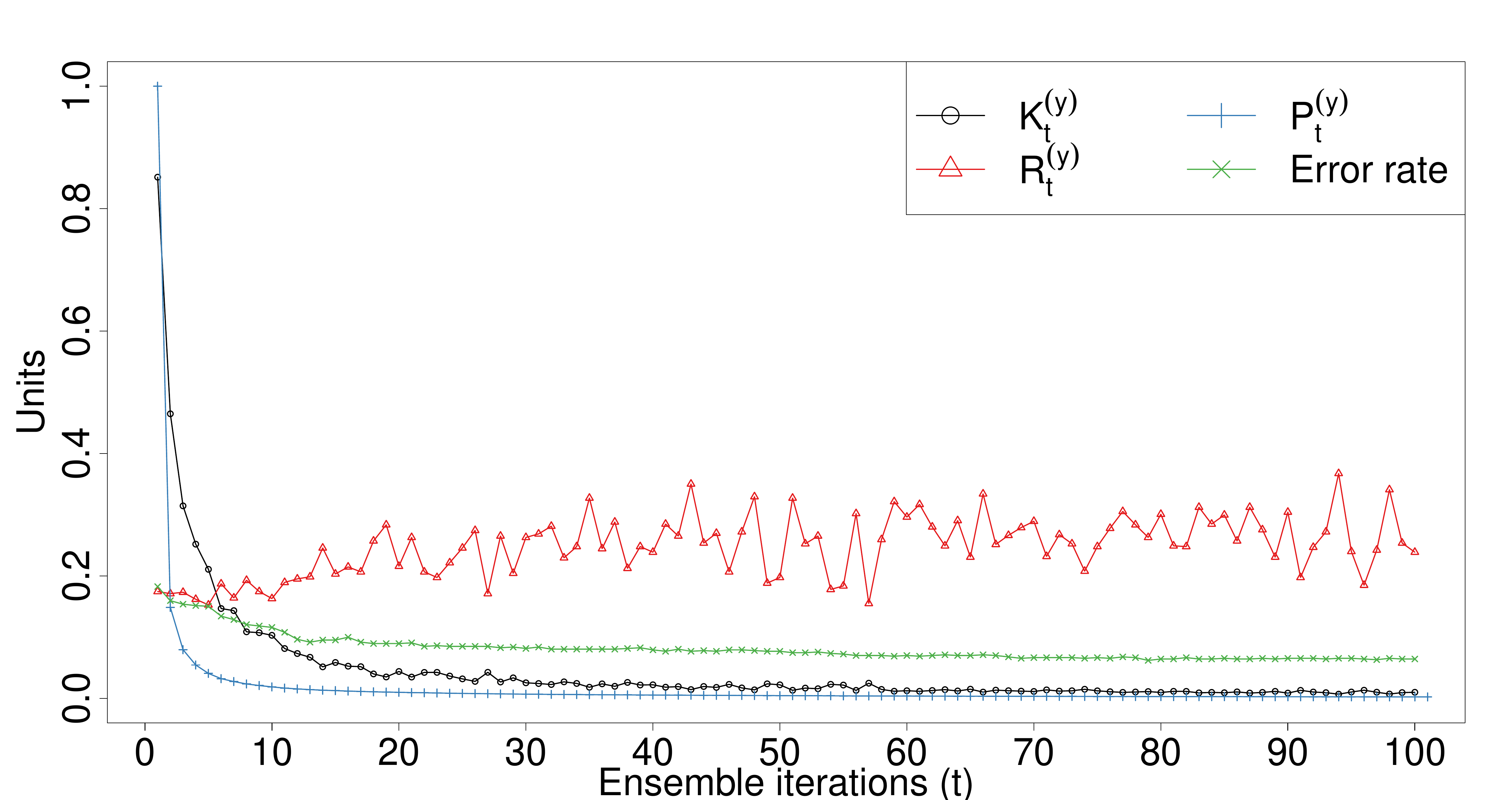}

}

\subfloat[\emph{balance\_scale} \label{fig:balance_scale_params}]{\includegraphics[width=0.5\textwidth]{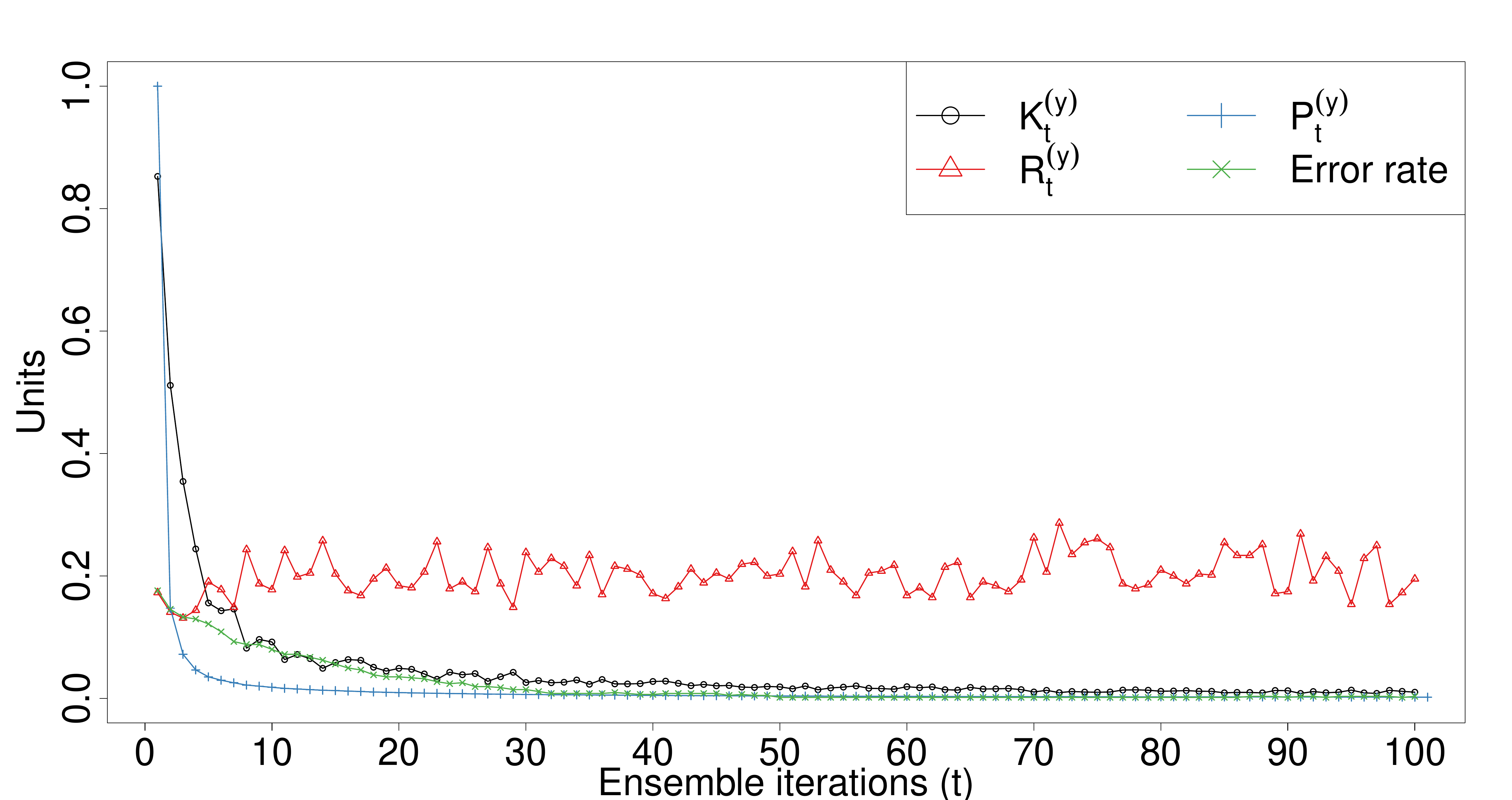}

}\subfloat[\emph{flags} \label{fig:flags_params}]{\includegraphics[width=0.5\textwidth]{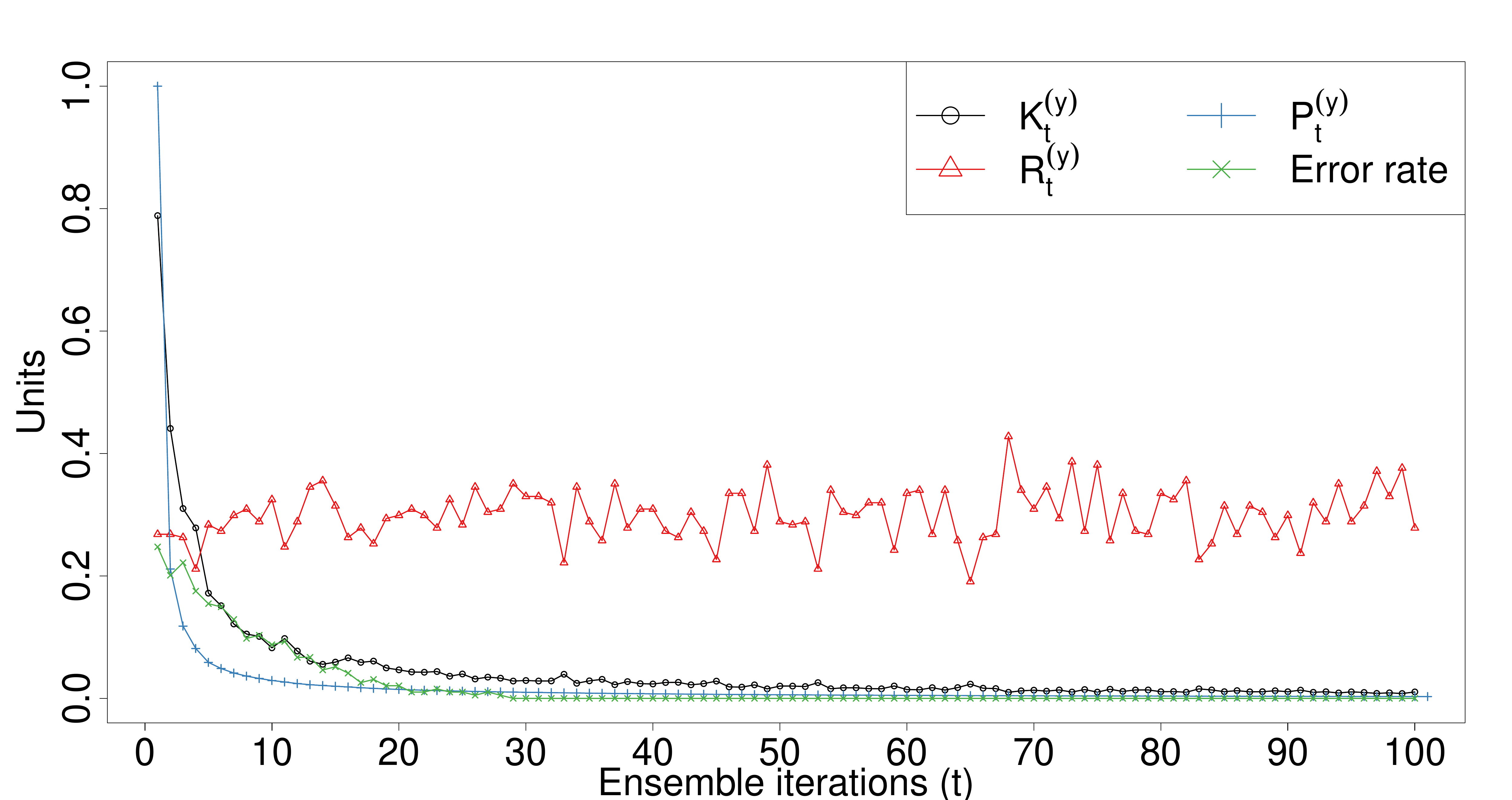}

}

\subfloat[\emph{german} \label{fig:german_params}]{\includegraphics[width=0.5\textwidth]{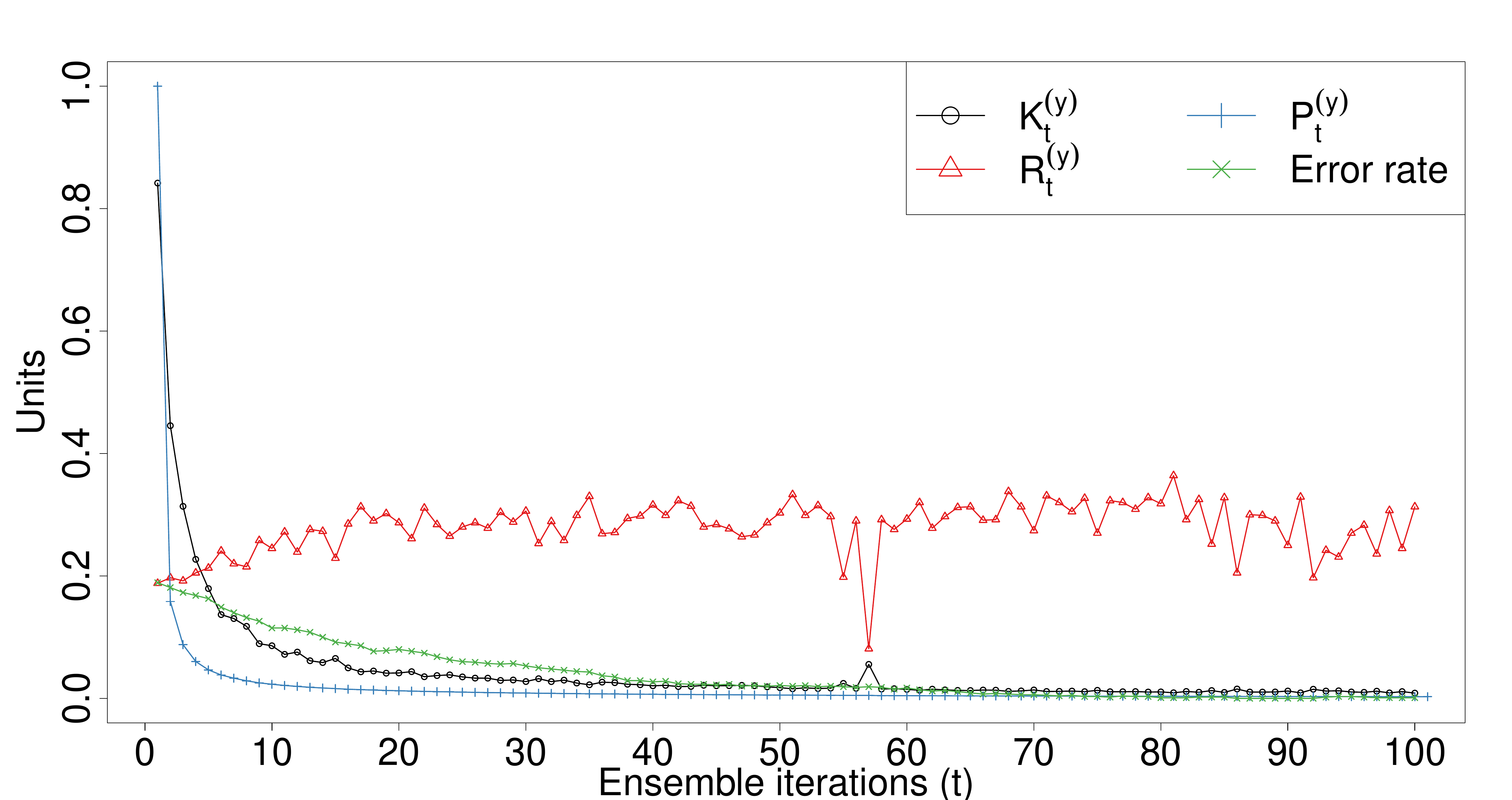}

}\subfloat[\emph{ilpd} \label{fig:ilpd_params}]{\includegraphics[width=0.5\textwidth]{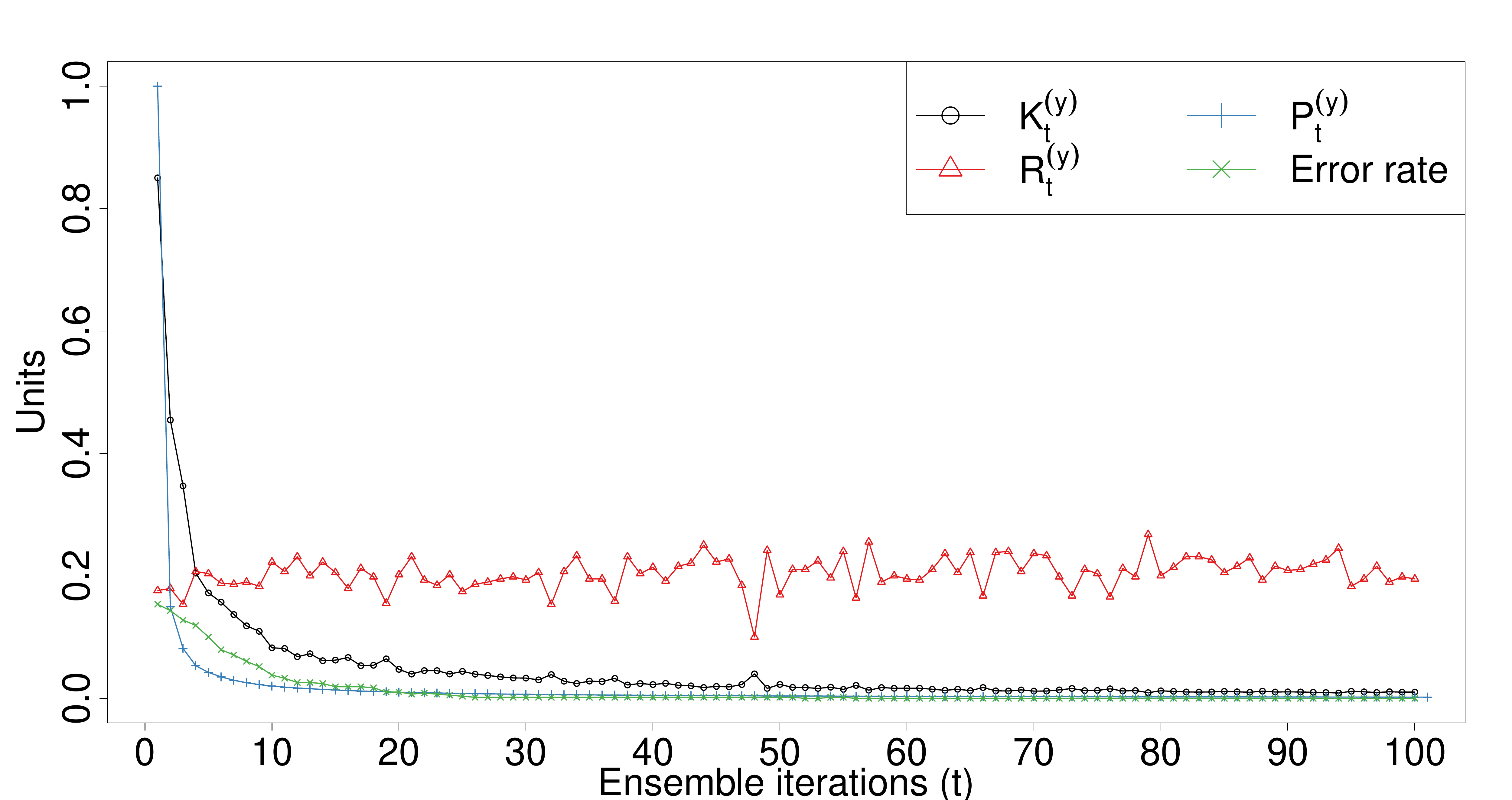}

}

\caption{Changes in the parameters and the misclassification rate for the training
sets for KFHE. Datasets \emph{mushroom} to \emph{ilpd} \label{fig:Changes-tvowel}}
\end{figure}

\begin{figure}[p]

\subfloat[\emph{ionosphere} \label{fig:ionosphere_params}]{\includegraphics[width=0.5\textwidth]{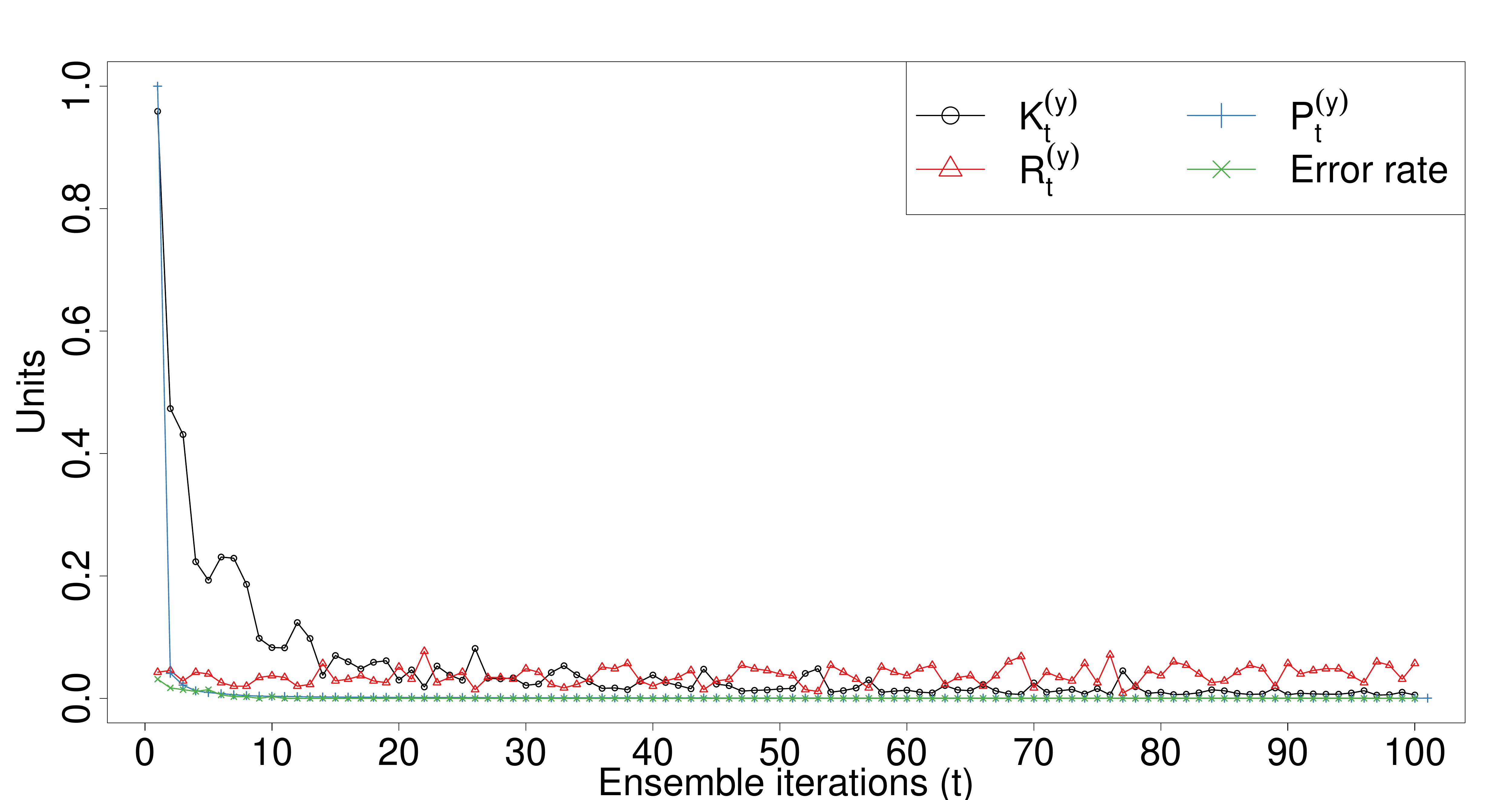}

}\subfloat[\emph{knowledge} \label{fig:knowledge_params}]{\includegraphics[width=0.5\textwidth]{plots/knowledge_plot.pdf}

}

\subfloat[\emph{vertebral} \label{fig:vertebral_params}]{\includegraphics[width=0.5\textwidth]{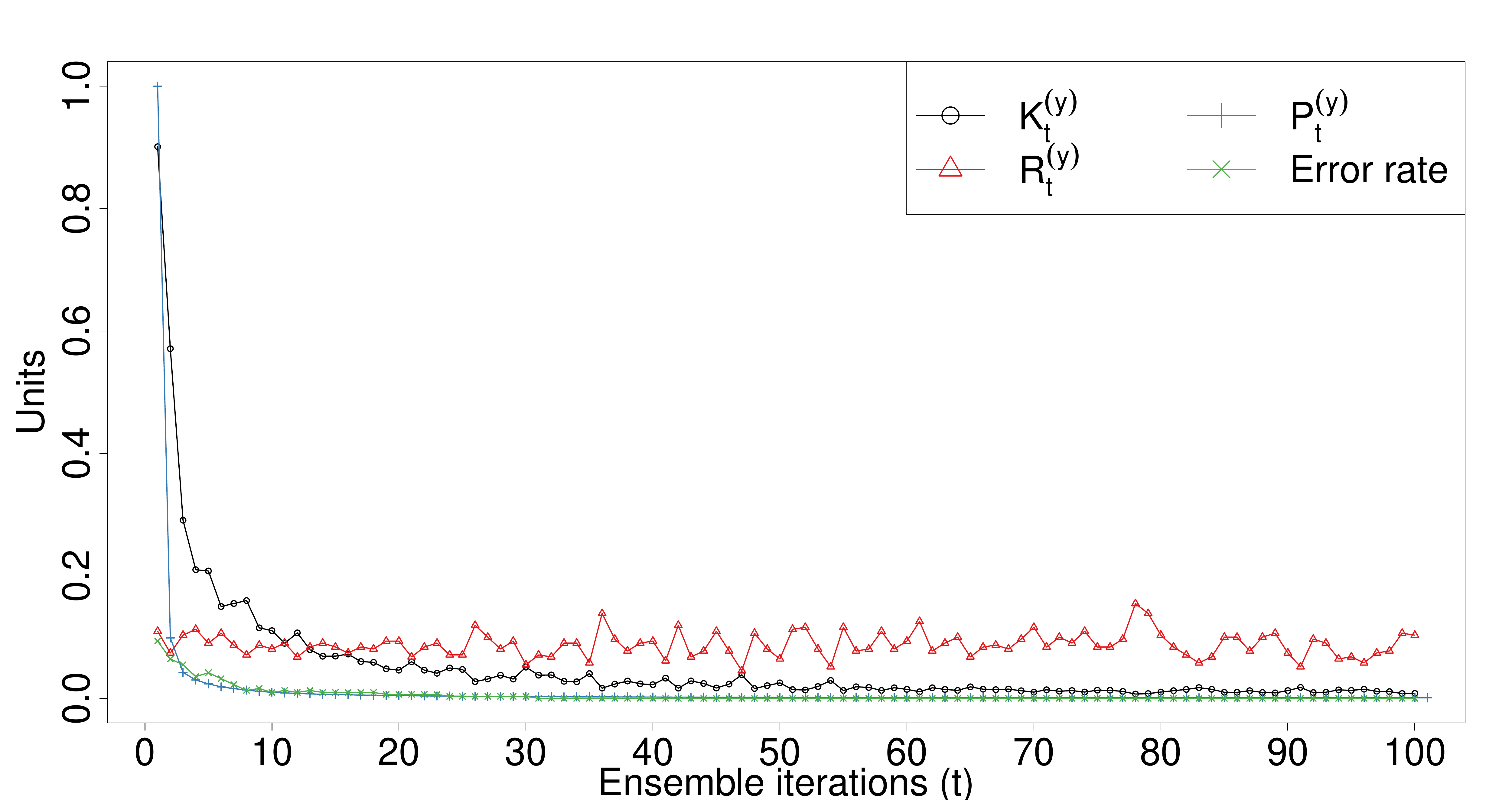}

}\subfloat[\emph{sonar} \label{fig:sonar_params}]{\includegraphics[width=0.5\textwidth]{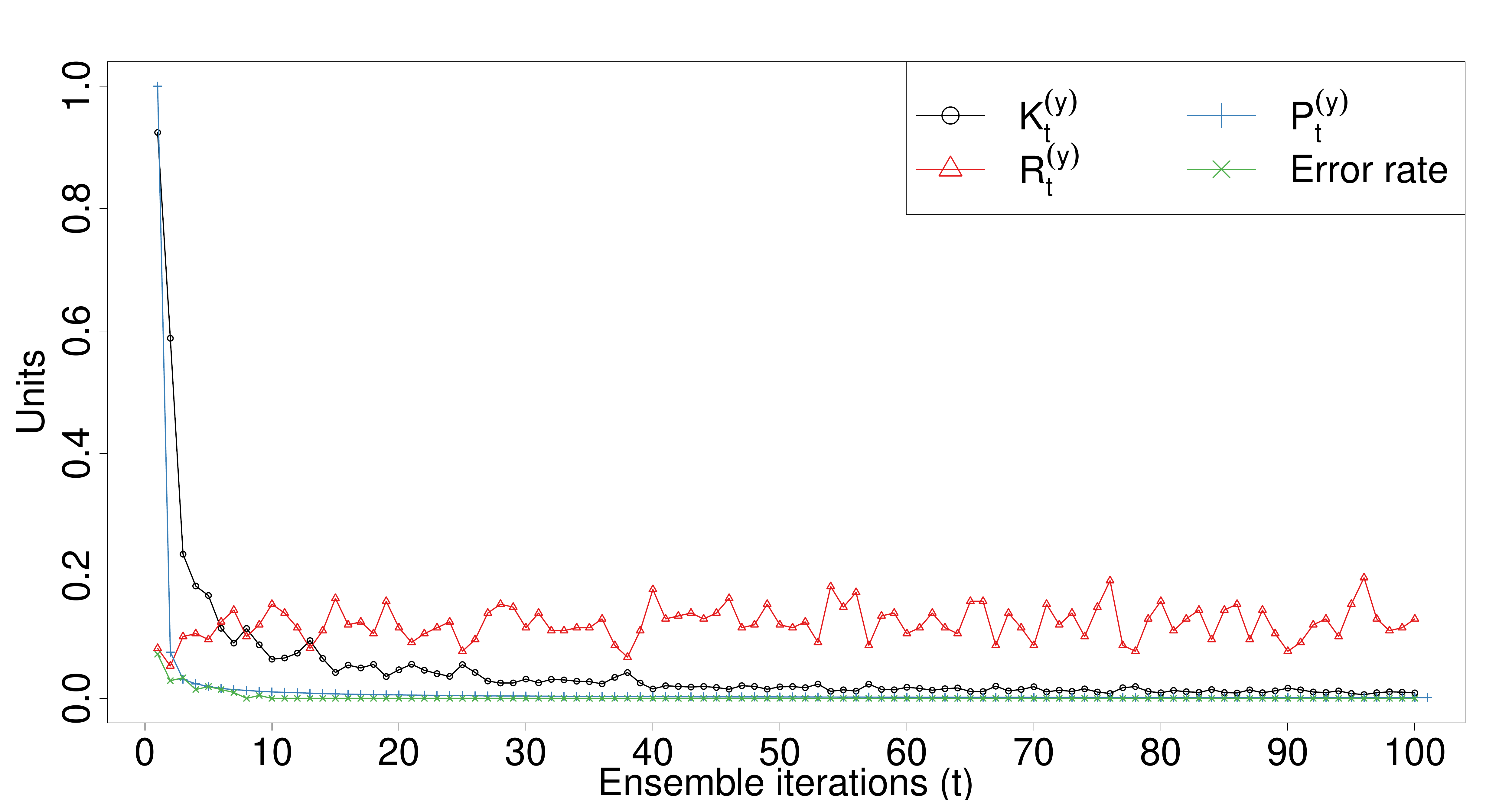}

}

\subfloat[\emph{skulls} \label{fig:skulls_params}]{\includegraphics[width=0.5\textwidth]{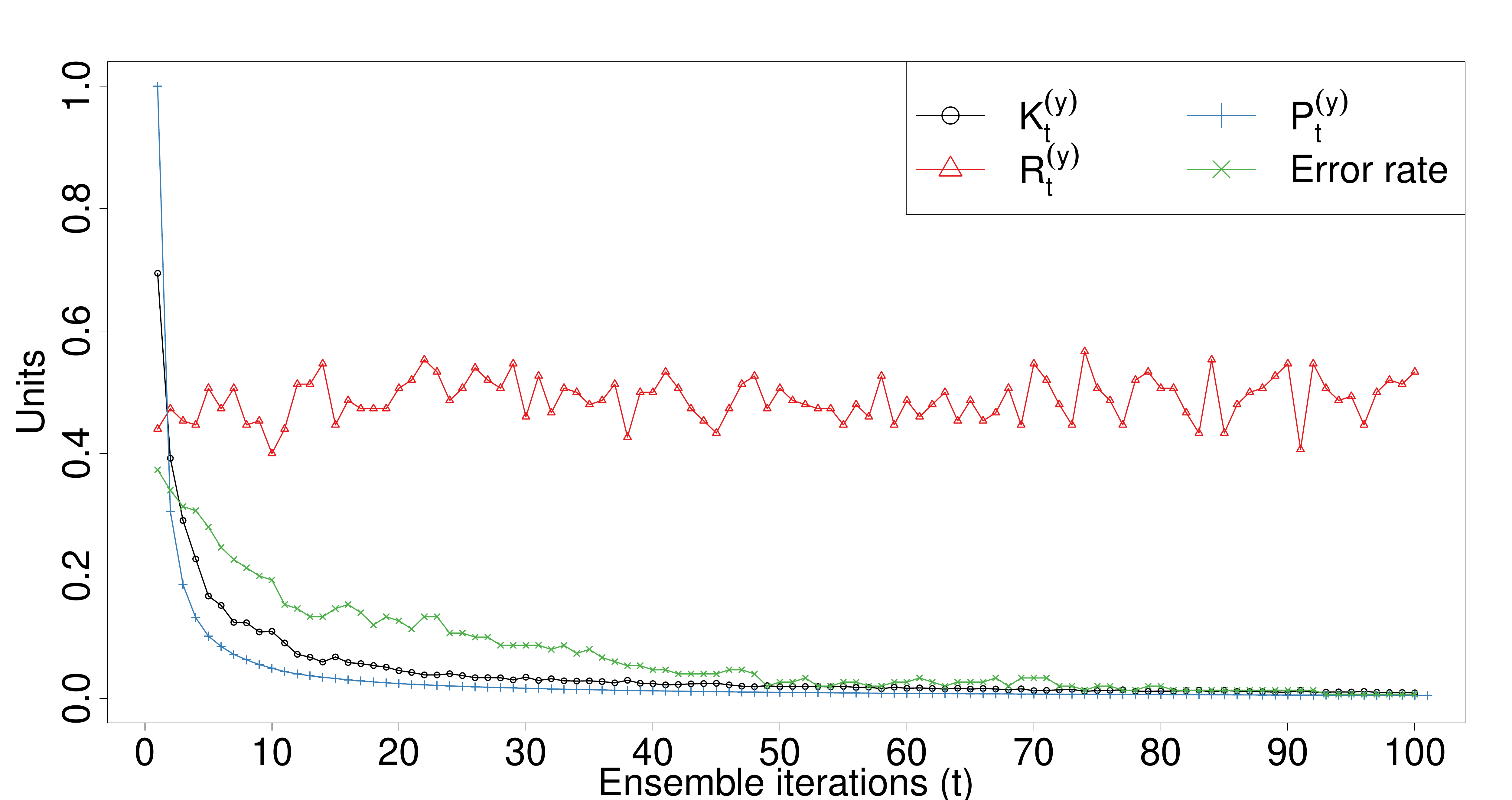}

}\subfloat[\emph{diabetes} \label{fig:diabetes_params}]{\includegraphics[width=0.5\textwidth]{plots/diabetes_plot.pdf}

}

\subfloat[\emph{physio} \label{fig:physio_params}]{\includegraphics[width=0.5\textwidth]{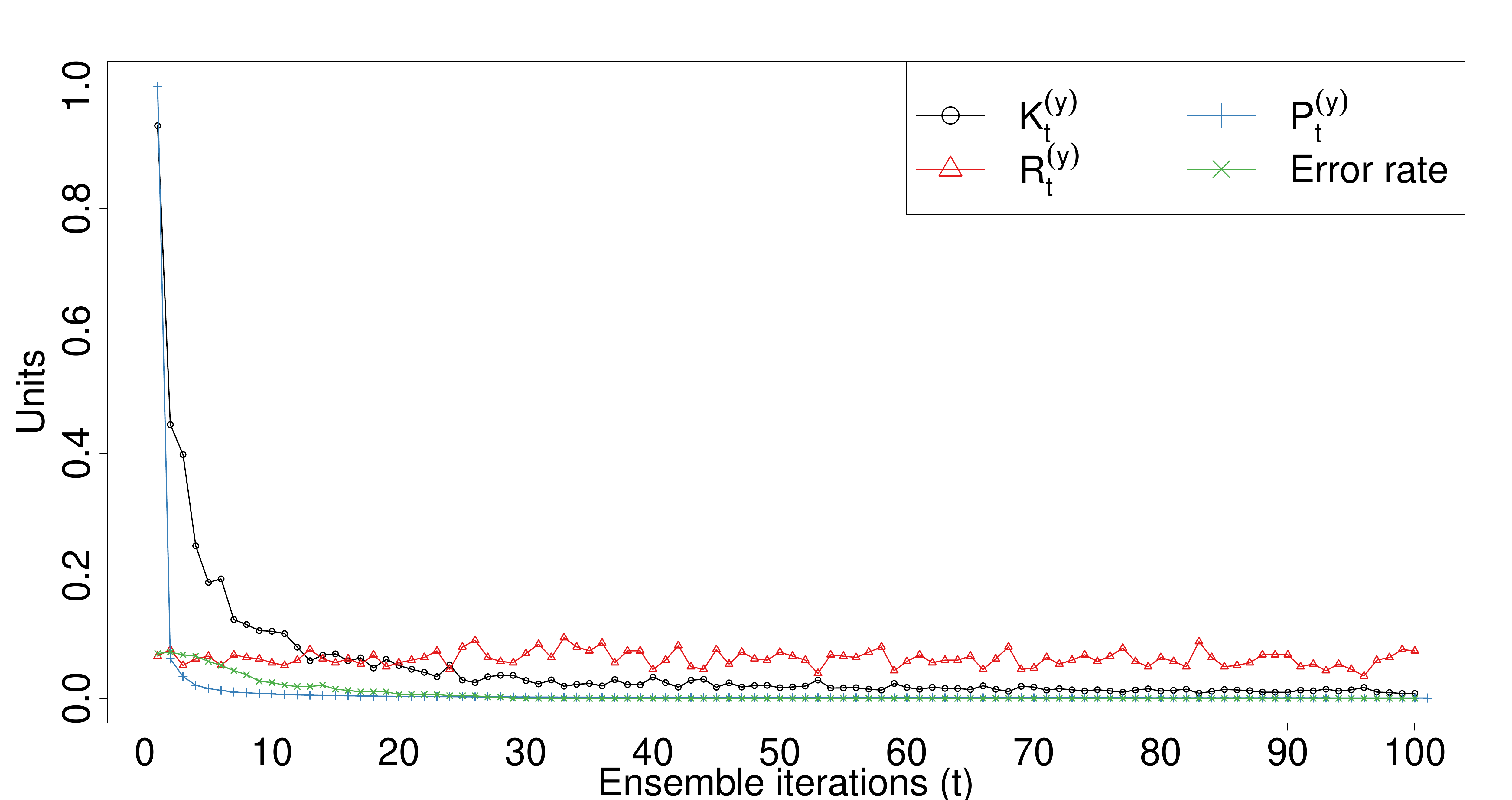}

}\subfloat[\emph{breasttissue} \label{fig:breasttissue_params}]{\includegraphics[width=0.5\textwidth]{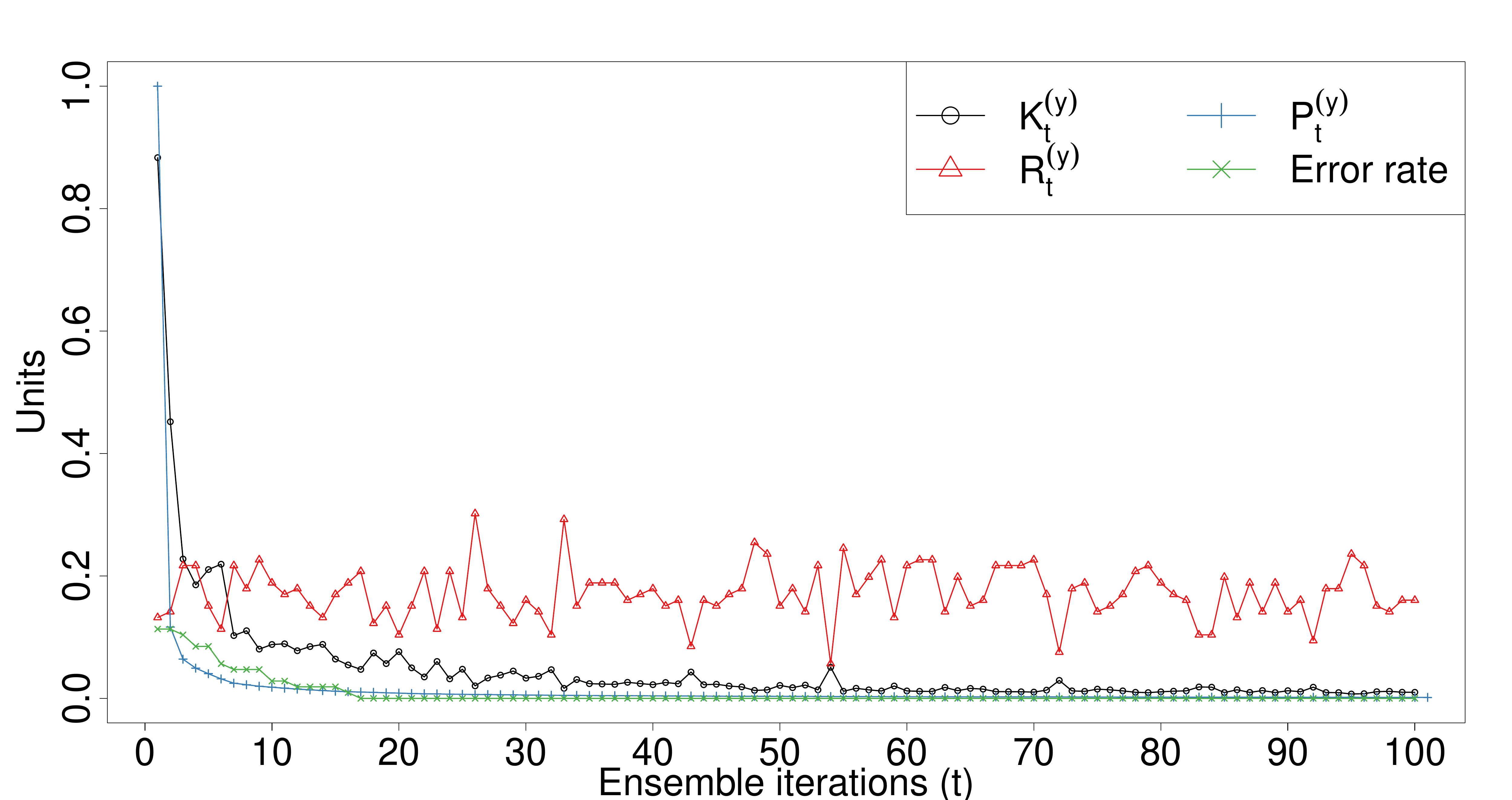}

}

\subfloat[\emph{bupa} \label{fig:bupa_params}]{\includegraphics[width=0.5\textwidth]{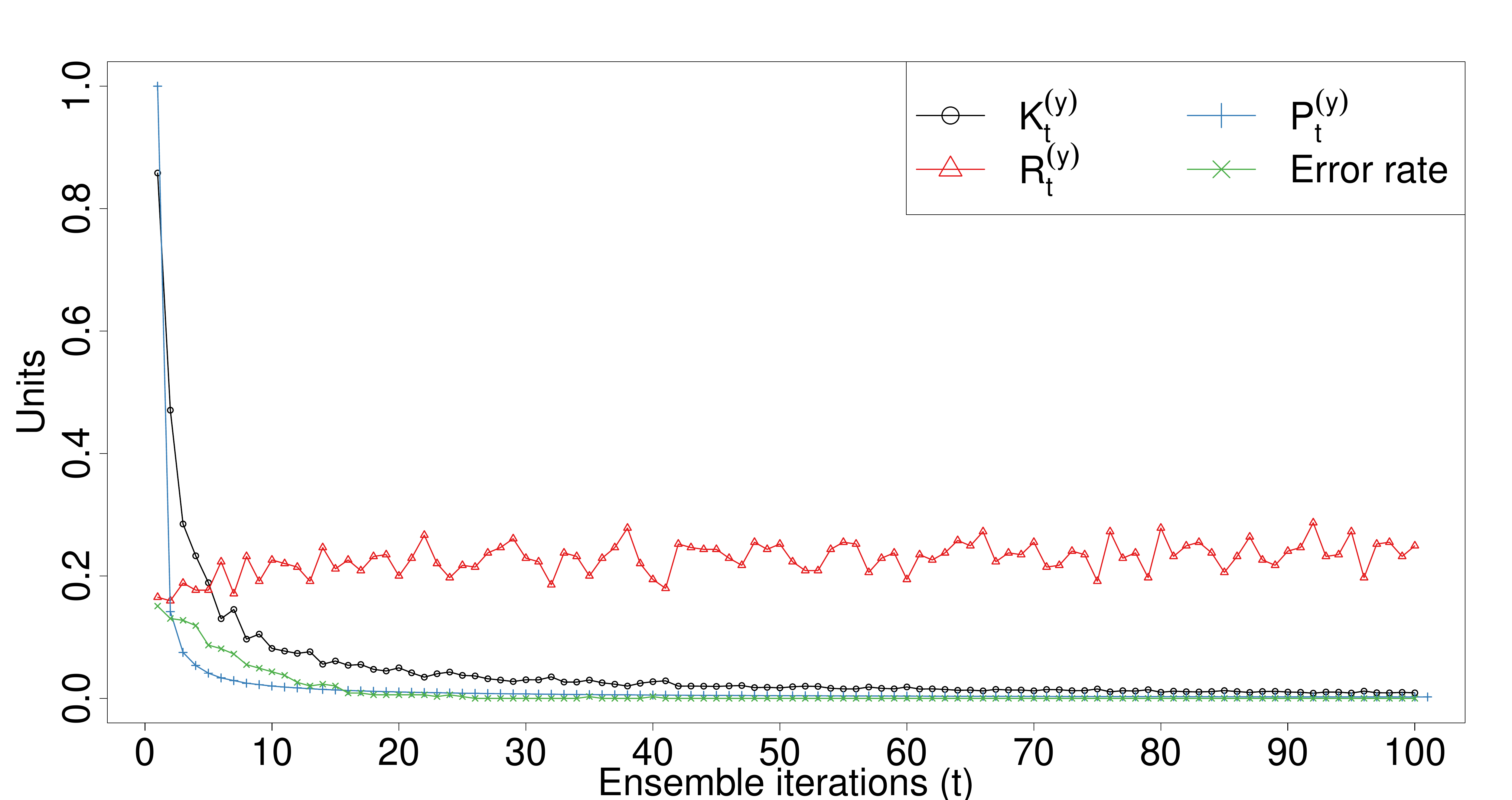}

}\subfloat[\emph{cleveland} \label{fig:cleveland_params}]{\includegraphics[width=0.5\textwidth]{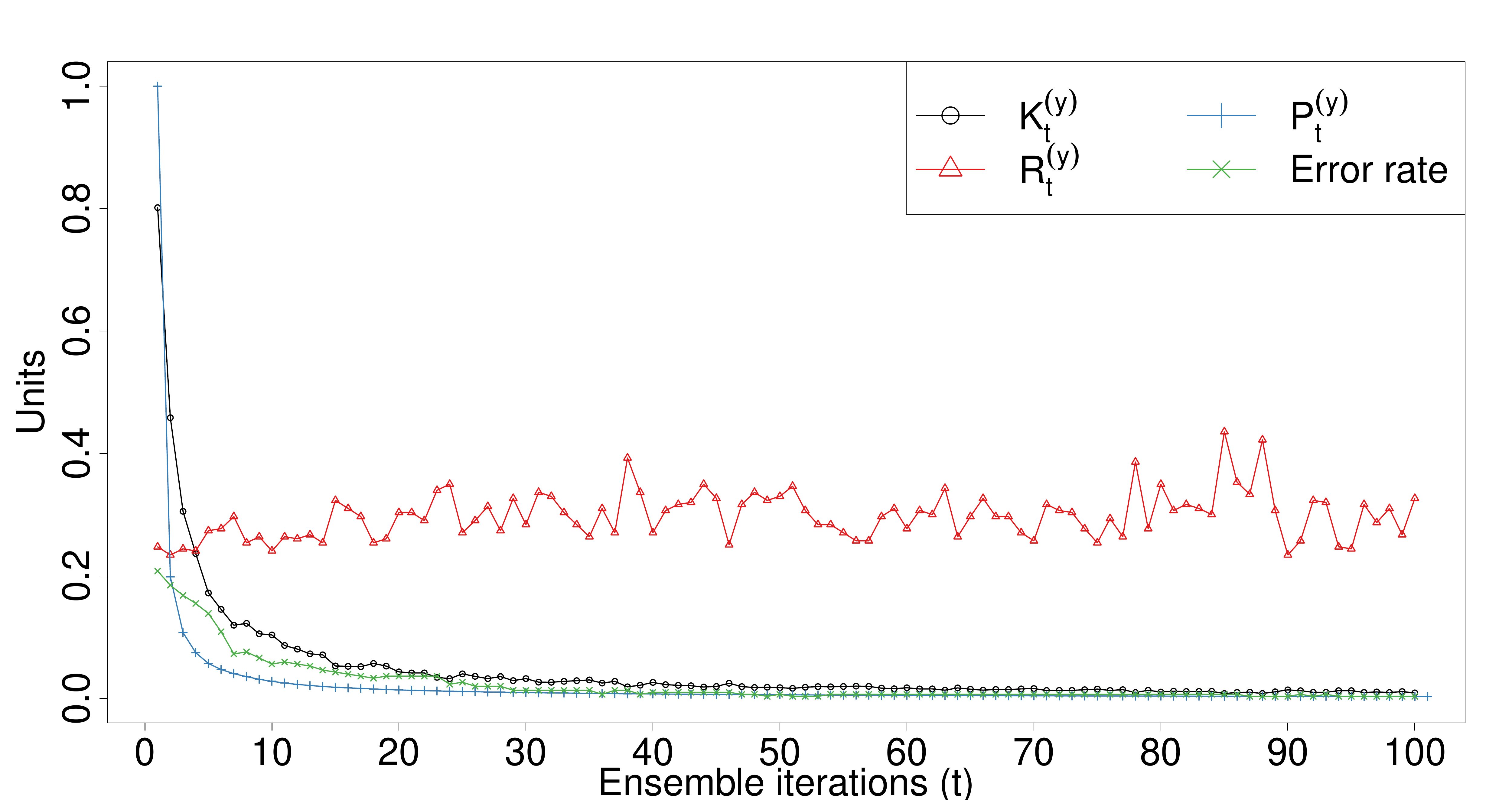}

}

\caption{Changes in the parameters and the misclassification rate for the training
sets for KFHE. Datasets \emph{ionosphere} to \emph{cleveland} \label{fig:Changes-tvowel-1}}
\end{figure}

\begin{figure}[p]
\subfloat[\emph{haberman} \label{fig:haberman_params}]{\includegraphics[width=0.5\textwidth]{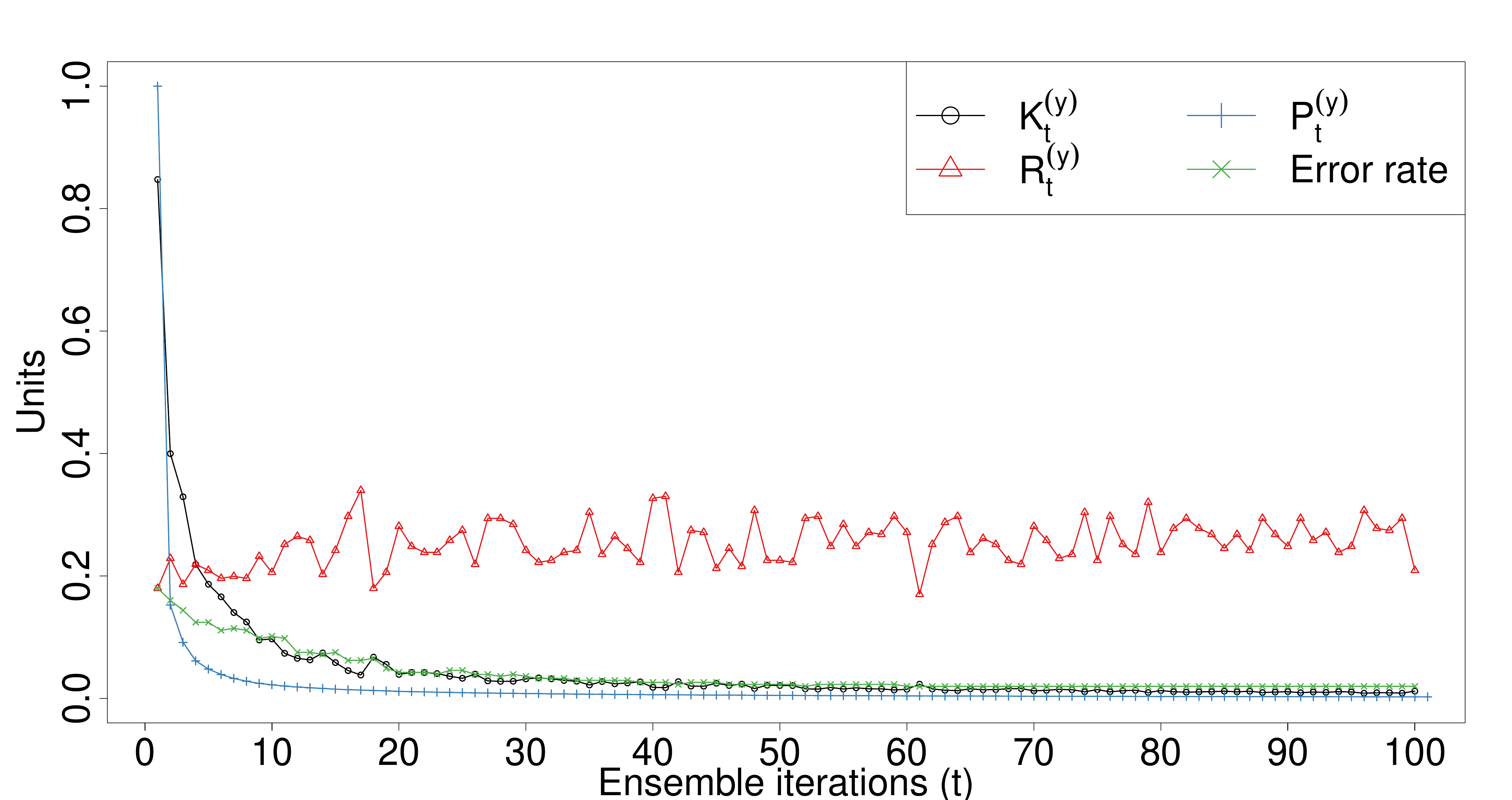}

}\subfloat[\emph{hayes\_roth} \label{fig:hayes_params}]{\includegraphics[width=0.5\textwidth]{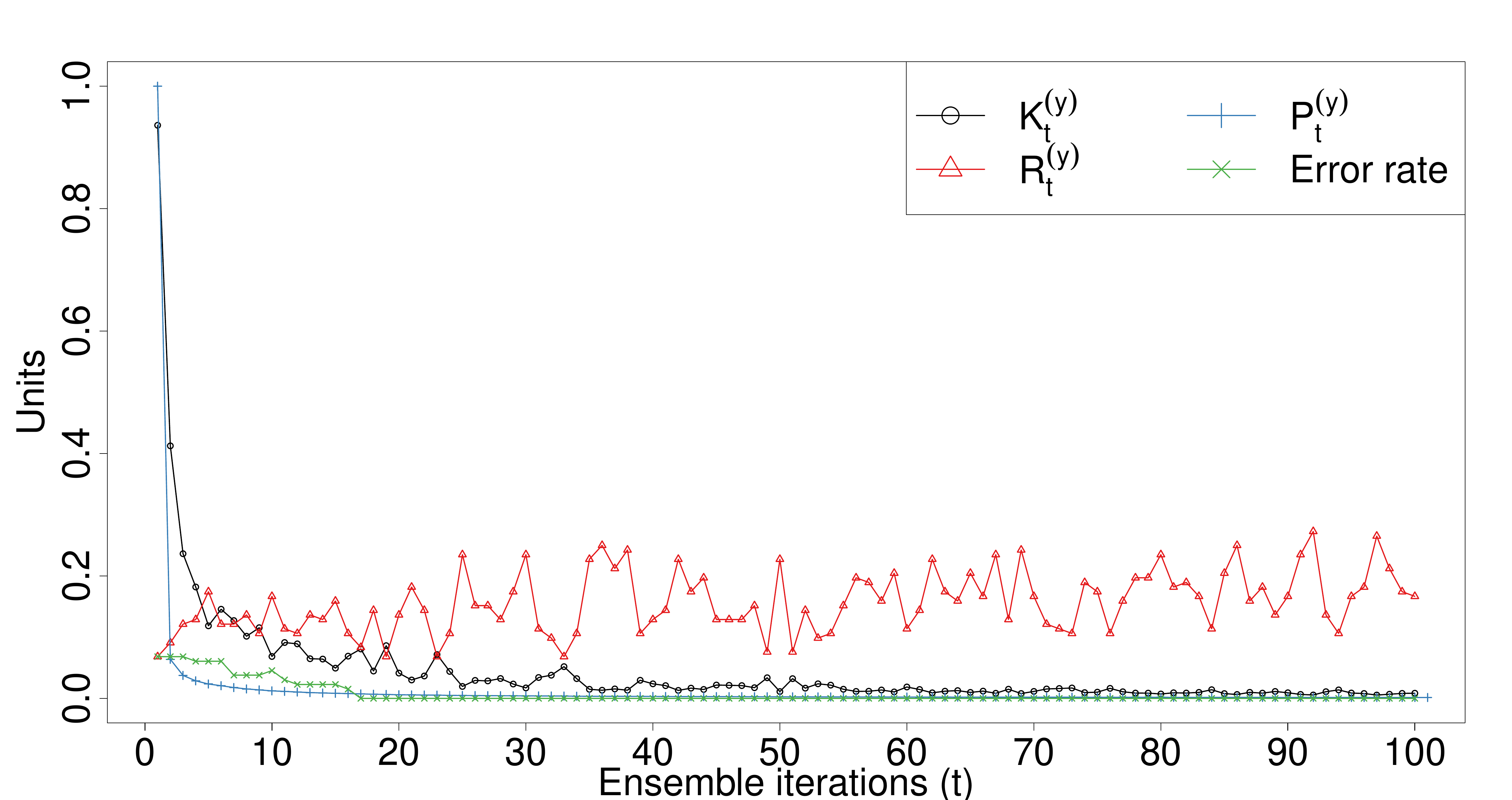}

}

\subfloat[\emph{monks} \label{fig:monks_params}]{\includegraphics[width=0.5\textwidth]{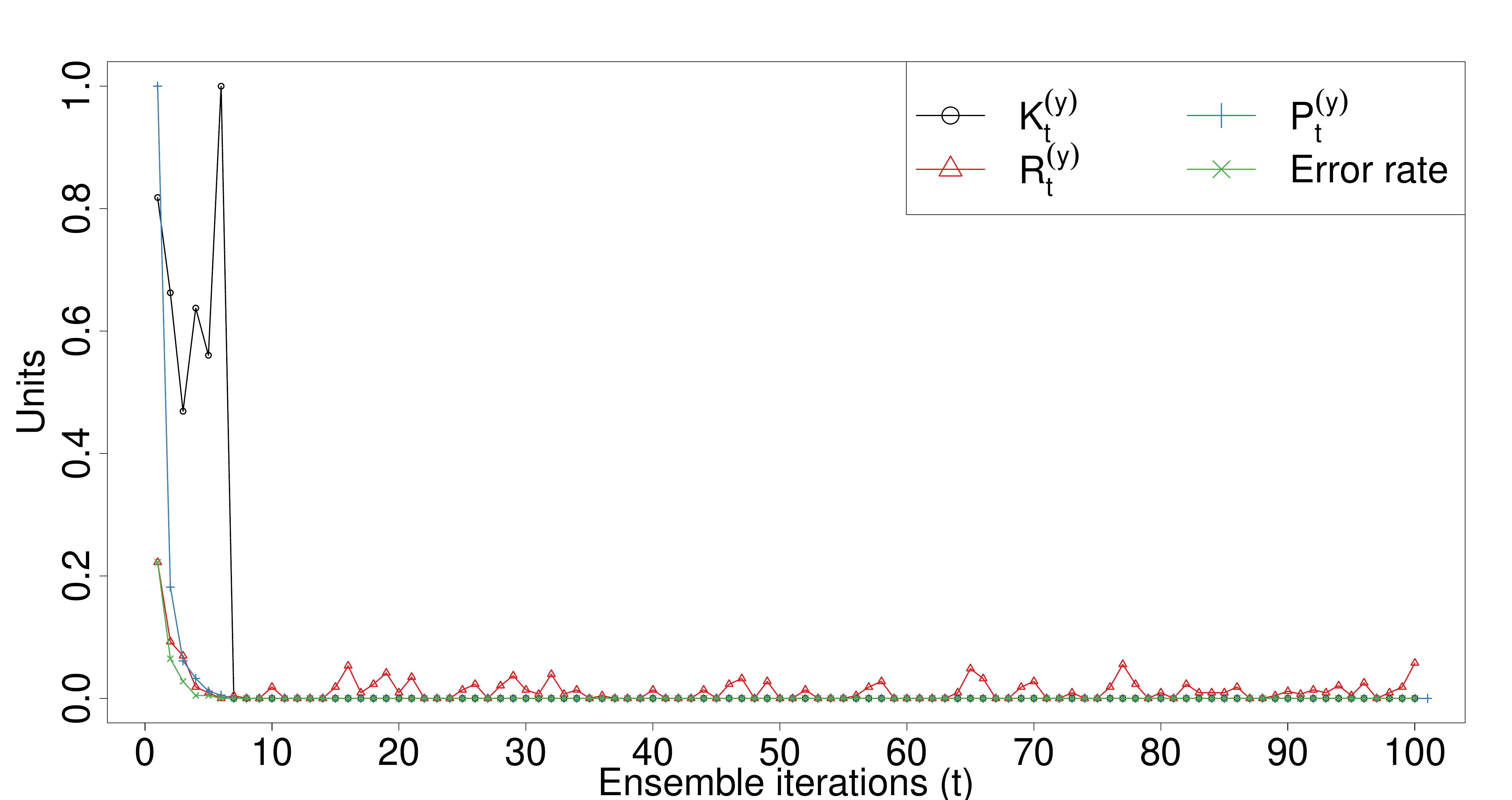}

}\subfloat[\emph{newthyroid} \label{fig:newthyroid_params}]{\includegraphics[width=0.5\textwidth]{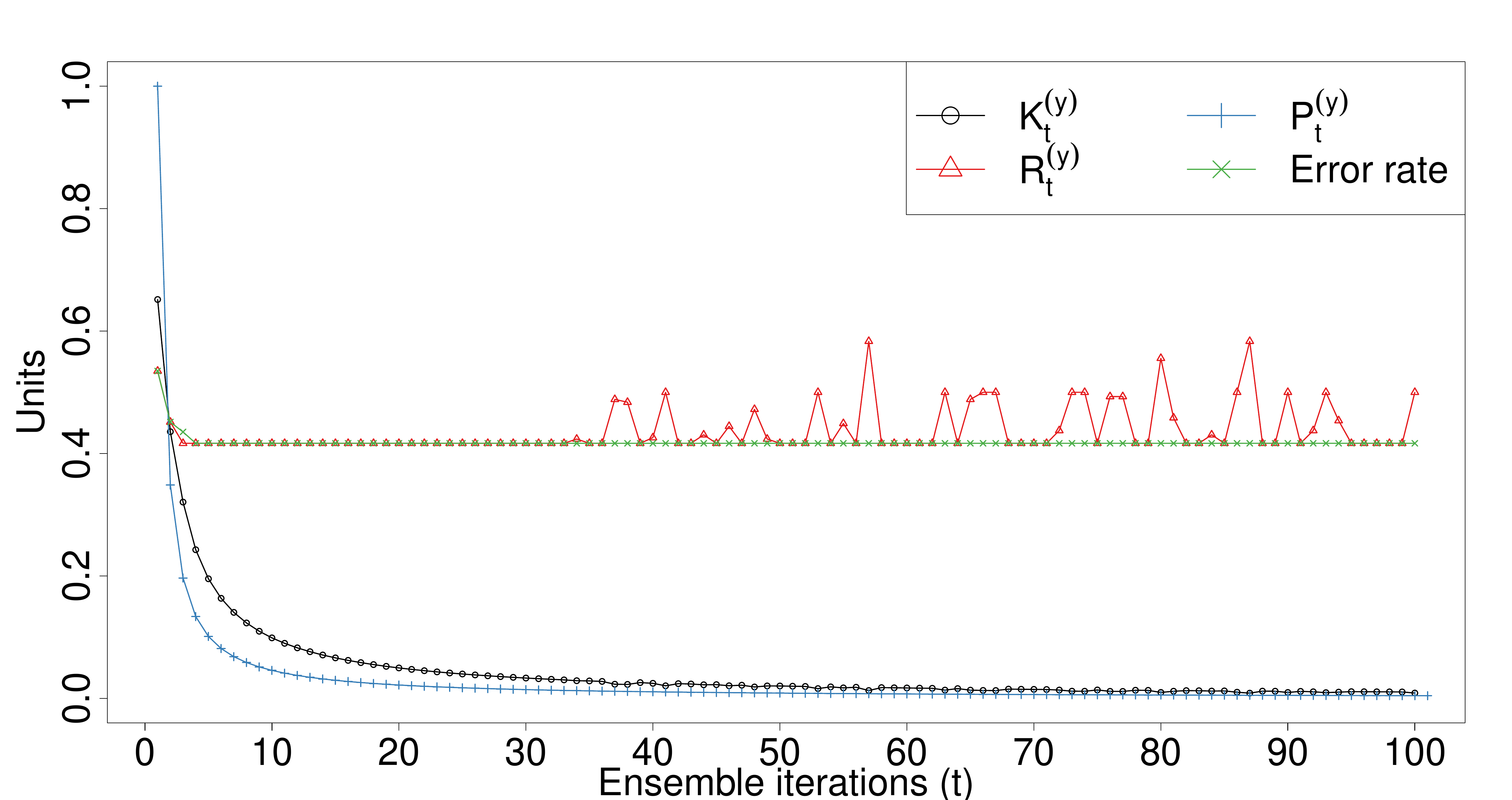}

}

\subfloat[\emph{yeast} \label{fig:yeast_params}]{\includegraphics[width=0.5\textwidth]{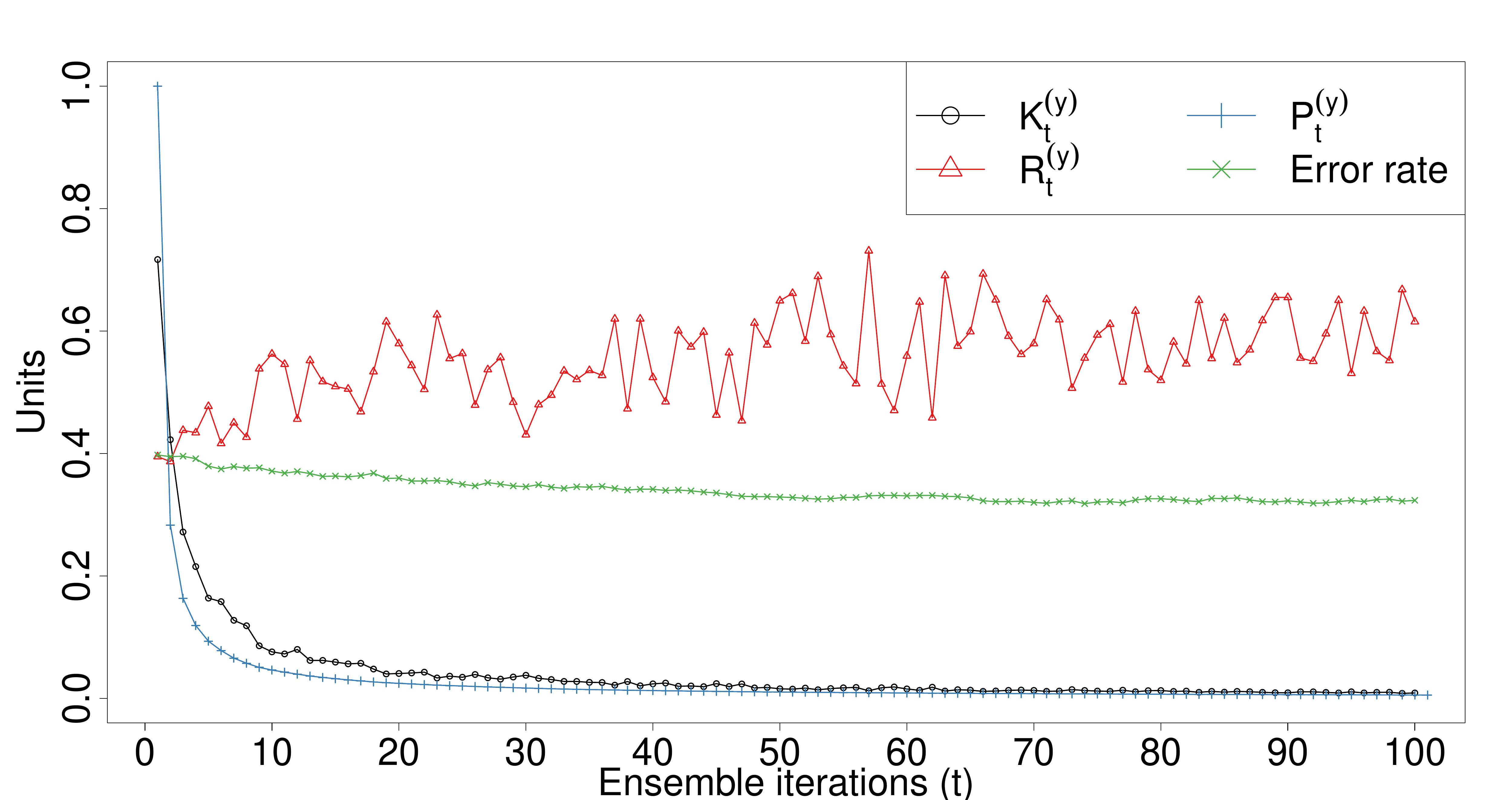}

}\subfloat[\emph{spam} \label{fig:spam_params}]{\includegraphics[width=0.5\textwidth]{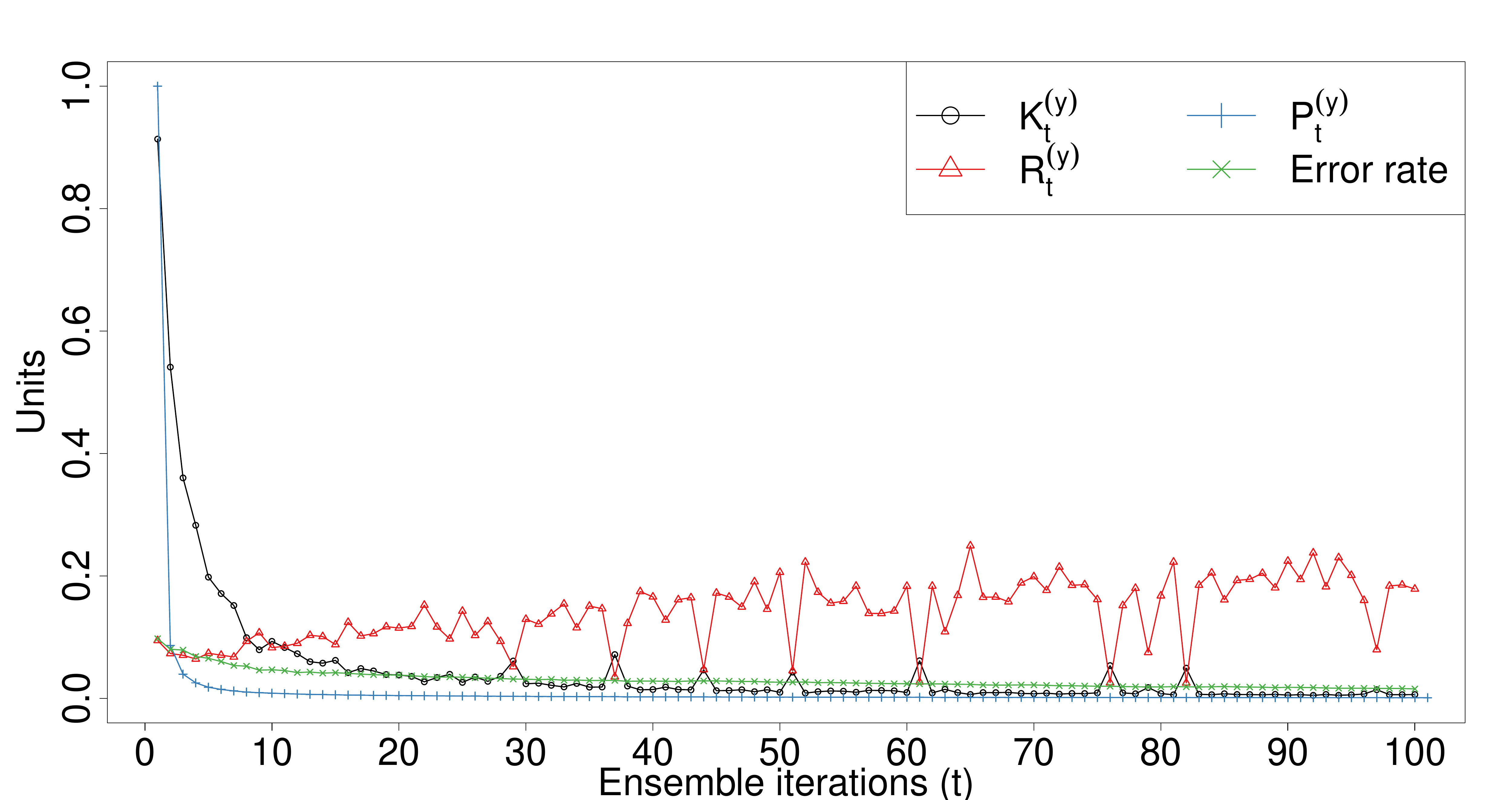}

}

\subfloat[\emph{lymphography} \label{fig:lymphograhy_params}]{\includegraphics[width=0.5\textwidth]{plots/lymphography_plot.pdf}

}\subfloat[\emph{movement\_libras} \label{fig:movement_libras_params}]{\includegraphics[width=0.5\textwidth]{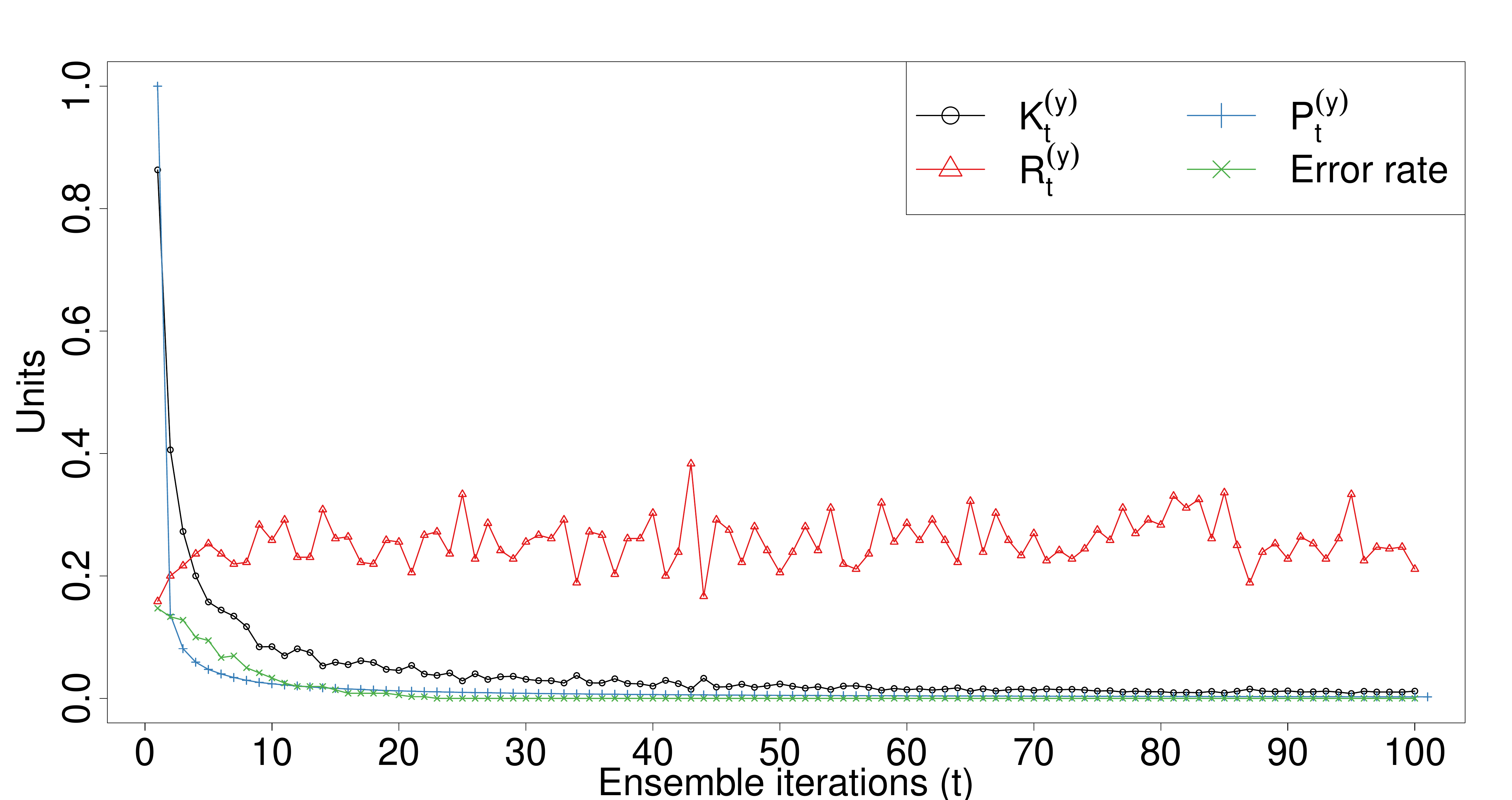}

}

\subfloat[\emph{SAheart} \label{fig:SAheart_params}]{\includegraphics[width=0.5\textwidth]{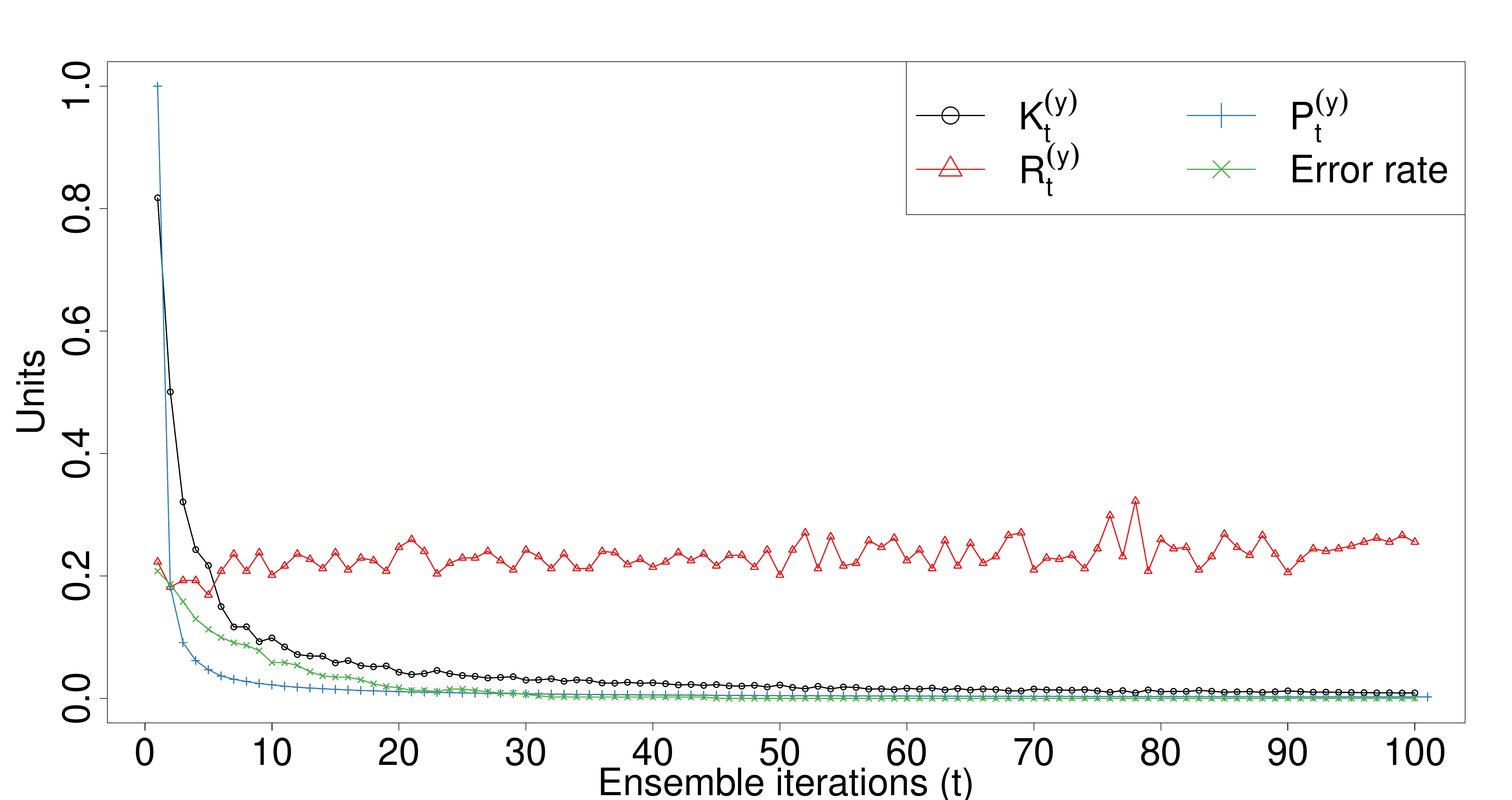}

}\subfloat[\emph{zoo} \label{fig:zoo_params}]{\includegraphics[width=0.5\textwidth]{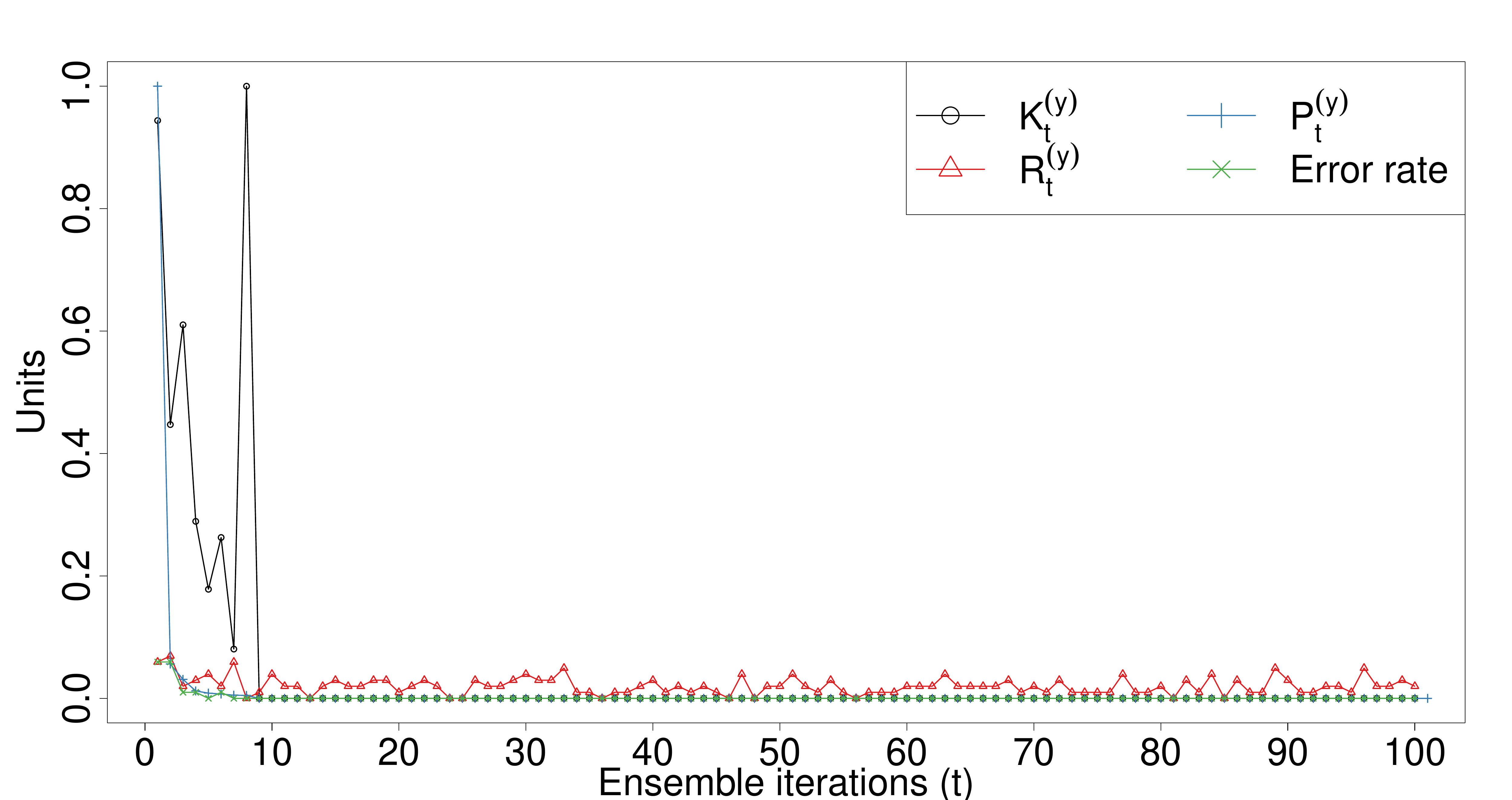}

}

\caption{Changes in the parameters and the misclassification rate for the training
sets for KFHE. Datasets \emph{haberman} to \emph{zoo }\label{fig:Changes-tvowel-2}}
\end{figure}

\clearpage{}

\subsection*{\bibliographystyle{plain}
\bibliography{kfhe_draft_ifc}

\begin{thebibliography}{10}

\bibitem{adabag}
Esteban Alfaro, Mat\'ias G\'amez, and Noelia Garc\'ia.
\newblock {adabag}: An {R} package for classification with boosting and
  bagging.
\newblock {\em Journal of Statistical Software}, 54(2):1--35, 2013.

\bibitem{UnderBagging:Barandela2003}
R.~Barandela, R.M. Valdovinos, and J.S. S{\'a}nchez.
\newblock New applications of ensembles of classifiers.
\newblock {\em Pattern Analysis {\&} Applications}, 6(3):245--256, Dec 2003.

\bibitem{NIPS2007_3321filterboost}
Joseph~K Bradley and Robert~E Schapire.
\newblock Filterboost: Regression and classification on large datasets.
\newblock In J.~C. Platt, D.~Koller, Y.~Singer, and S.~T. Roweis, editors, {\em
  Advances in Neural Information Processing Systems 20}, pages 185--192. Curran
  Associates, Inc., 2008.

\bibitem{Breiman1996}
Leo Breiman.
\newblock Bagging predictors.
\newblock {\em Machine Learning}, 24(2):123--140, Aug 1996.

\bibitem{Breiman2001rf}
Leo Breiman.
\newblock Random forests.
\newblock {\em Machine Learning}, 45(1):5--32, Oct 2001.

\bibitem{CAO20124451}
Jingjing Cao, Sam Kwong, and Ran Wang.
\newblock A noise-detection based adaboost algorithm for mislabeled data.
\newblock {\em Pattern Recognition}, 45(12):4451 -- 4465, 2012.

\bibitem{demiriz2002linear}
Ayhan Demiriz, Kristin~P Bennett, and John Shawe-Taylor.
\newblock Linear programming boosting via column generation.
\newblock {\em Machine Learning}, 46(1):225--254, 2002.

\bibitem{DietterichHSpace}
Thomas~G. Dietterich.
\newblock Ensemble methods in machine learning.
\newblock In {\em Multiple Classifier Systems}, pages 1--15, Berlin,
  Heidelberg, 2000. Springer Berlin Heidelberg.

\bibitem{Dietterich2000}
Thomas~G. Dietterich.
\newblock An experimental comparison of three methods for constructing
  ensembles of decision trees: Bagging, boosting, and randomization.
\newblock {\em Machine Learning}, 40(2):139--157, Aug 2000.

\bibitem{Domingo:2000:MMA:648299.755176_madaboost}
Carlos Domingo and Osamu Watanabe.
\newblock Madaboost: A modification of adaboost.
\newblock In {\em Proceedings of the Thirteenth Annual Conference on
  Computational Learning Theory}, COLT '00, pages 180--189, San Francisco, CA,
  USA, 2000. Morgan Kaufmann Publishers Inc.

\bibitem{evensen2003ensemble}
Geir Evensen.
\newblock The ensemble kalman filter: Theoretical formulation and practical
  implementation.
\newblock {\em Ocean dynamics}, 53(4):343--367, 2003.

\bibitem{Freund2001}
Yoav Freund.
\newblock An adaptive version of the boost by majority algorithm.
\newblock {\em Machine Learning}, 43(3):293--318, Jun 2001.

\bibitem{freund1995desicion}
Yoav Freund and Robert~E Schapire.
\newblock A desicion-theoretic generalization of on-line learning and an
  application to boosting.
\newblock In {\em European conference on computational learning theory}, pages
  23--37. Springer, 1995.

\bibitem{friedberg2018local}
Rina Friedberg, Julie Tibshirani, Susan Athey, and Stefan Wager.
\newblock Local linear forests.
\newblock {\em arXiv preprint arXiv:1807.11408}, 2018.

\bibitem{friedman2000additive}
Jerome Friedman, Trevor Hastie, Robert Tibshirani, et~al.
\newblock Additive logistic regression: a statistical view of boosting (with
  discussion and a rejoinder by the authors).
\newblock {\em The annals of statistics}, 28(2):337--407, 2000.

\bibitem{Friedman00greedyfunction}
Jerome~H. Friedman.
\newblock Greedy function approximation: A gradient boosting machine.
\newblock {\em Annals of Statistics}, 29:1189--1232, 2000.

\bibitem{FRIEDMAN2002367}
Jerome~H. Friedman.
\newblock Stochastic gradient boosting.
\newblock {\em Computational Statistics \& Data Analysis}, 38(4):367 -- 378,
  2002.
\newblock Nonlinear Methods and Data Mining.

\bibitem{GARCIA20102044}
Salvador Garc\'ia, Alberto Fern\'andez, Juli\'an Luengo, and Francisco Herrera.
\newblock Advanced nonparametric tests for multiple comparisons in the design
  of experiments in computational intelligence and data mining: Experimental
  analysis of power.
\newblock {\em Information Sciences}, 180(10):2044 -- 2064, 2010.

\bibitem{Hansen:1990:NNE:628297.628429}
L.~K. Hansen and P.~Salamon.
\newblock Neural network ensembles.
\newblock {\em IEEE Trans. Pattern Anal. Mach. Intell.}, 12(10):993--1001,
  October 1990.

\bibitem{hastie2009multi}
Trevor Hastie, Saharon Rosset, Ji~Zhu, and Hui Zou.
\newblock Multi-class adaboost.
\newblock {\em Statistics and its Interface}, 2(3):349--360, 2009.

\bibitem{hastie01statisticallearning}
Trevor Hastie, Robert Tibshirani, and Jerome Friedman.
\newblock {\em The Elements of Statistical Learning}.
\newblock Springer Series in Statistics. Springer New York Inc., New York, NY,
  USA, 2001.

\bibitem{kalman1960}
R.~E. Kalman.
\newblock A new approach to linear filtering and prediction problems.
\newblock {\em ASME Journal of Basic Engineering}, 1960.

\bibitem{kelleher2015fundamentals}
John~D Kelleher, Brian Mac~Namee, and Aoife D'Arcy.
\newblock Fundamentals of machine learning for predictive data analytics:
  algorithms, worked examples, and case studies, 2015.

\bibitem{Lichman:2013}
M.~Lichman.
\newblock {UCI} machine learning repository, 2013.

\bibitem{maybeck1982stochastic}
Peter~S Maybeck.
\newblock {\em Stochastic models, estimation, and control}, volume~3.
\newblock Academic press, 1982.

\bibitem{MENAHEM20094097_troika}
Eitan Menahem, Lior Rokach, and Yuval Elovici.
\newblock Troika - an improved stacking schema for classification tasks.
\newblock {\em Information Sciences}, 179(24):4097 -- 4122, 2009.

\bibitem{Monson04thekalman}
Christopher~K. Monson and Kevin~D. Seppi.
\newblock The kalman swarm: A new approach to particle motion in swarm
  optimization.
\newblock In {\em in Swarm Optimization, "Proceedings of the Genetic and
  Evolutionary Computation Conference (GECCO 2004)}, pages 140--150, 2004.

\bibitem{Narassiguin2016}
Anil Narassiguin, Mohamed Bibimoune, Haytham Elghazel, and Alex Aussem.
\newblock An extensive empirical comparison of ensemble learning methods for
  binary classification.
\newblock {\em Pattern Analysis and Applications}, 19(4):1093--1128, Nov 2016.

\bibitem{Opitz:1999:PEM:3013545.3013549}
David Opitz and Richard Maclin.
\newblock Popular ensemble methods: An empirical study.
\newblock {\em J. Artif. Int. Res.}, 11(1):169--198, July 1999.

\bibitem{pakrashikhka_10.1007/978-3-319-20294-5_39}
Arjun Pakrashi.
\newblock A new hybrid clustering approach based on heuristic kalman algorithm.
\newblock In {\em Swarm, Evolutionary, and Memetic Computing}, pages 445--455,
  Cham, 2015. Springer International Publishing.

\bibitem{PAKRASHI2016704}
Arjun Pakrashi and Bidyut~B. Chaudhuri.
\newblock A kalman filtering induced heuristic optimization based partitional
  data clustering.
\newblock {\em Information Sciences}, 369(Supplement C):704 -- 717, 2016.

\bibitem{Rodriguez:2006:RFN:1159167.1159358}
Juan~J. Rodriguez, Ludmila~I. Kuncheva, and Carlos~J. Alonso.
\newblock Rotation forest: A new classifier ensemble method.
\newblock {\em IEEE Trans. Pattern Anal. Mach. Intell.}, 28(10):1619--1630,
  October 2006.

\bibitem{SABZEVARI2018119}
Maryam Sabzevari, Gonzalo Mart\'inez-Mu{\~n}oz, and Alberto Su\'arez.
\newblock Vote-boosting ensembles.
\newblock {\em Pattern Recognition}, 83:119 -- 133, 2018.

\bibitem{Schapire1990}
Robert~E. Schapire.
\newblock The strength of weak learnability.
\newblock {\em Machine Learning}, 5(2):197--227, Jun 1990.

\bibitem{Seewald:2002:MSB:645531.656165}
Alexander~K. Seewald.
\newblock How to make stacking better and faster while also taking care of an
  unknown weakness.
\newblock In {\em Proceedings of the Nineteenth International Conference on
  Machine Learning}, ICML '02, pages 554--561, San Francisco, CA, USA, 2002.
  Morgan Kaufmann Publishers Inc.

\bibitem{SISWANTORO2016112}
Joko Siswantoro, Anton~Satria Prabuwono, Azizi Abdullah, and Bahari Idrus.
\newblock A linear model based on kalman filter for improving neural network
  classification performance.
\newblock {\em Expert Systems with Applications}, 49(Supplement C):112 -- 122,
  2016.

\bibitem{SUN201687}
Bo~Sun, Songcan Chen, Jiandong Wang, and Haiyan Chen.
\newblock A robust multi-class adaboost algorithm for mislabeled noisy data.
\newblock {\em Knowledge-Based Systems}, 102:87 -- 102, 2016.

\bibitem{Ting97stackedgeneralization:}
Kai~Ming Ting and Ian~H. Witten.
\newblock Stacked generalization: when does it work?
\newblock In {\em in Procs. International Joint Conference on Artificial
  Intelligence}, pages 866--871. Morgan Kaufmann, 1997.

\bibitem{TOSCANO20101955}
R.~Toscano and P.~Lyonnet.
\newblock A new heuristic approach for non-convex optimization problems.
\newblock {\em Information Sciences}, 180(10):1955 -- 1966, 2010.
\newblock Special Issue on Intelligent Distributed Information Systems.

\bibitem{Welch:1995:IKF:897831}
Greg Welch and Gary Bishop.
\newblock An introduction to the kalman filter.
\newblock Technical report, Chapel Hill, NC, USA, 1995.

\bibitem{gbm}
Greg~Ridgeway with contributions~from others.
\newblock {\em gbm: Generalized Boosted Regression Models}, 2017.
\newblock R package version 2.1.3.

\bibitem{WOLPERT1992241}
David~H. Wolpert.
\newblock Stacked generalization.
\newblock {\em Neural Networks}, 5(2):241 -- 259, 1992.

\bibitem{ZHANG20081524}
Chun-Xia Zhang and Jiang-She Zhang.
\newblock Rotboost: A technique for combining rotation forest and adaboost.
\newblock {\em Pattern Recognition Letters}, 29(10):1524 -- 1536, 2008.

\bibitem{zhou2012ensemble}
Z.H. Zhou.
\newblock {\em Ensemble Methods: Foundations and Algorithms}.
\newblock Chapman \& Hall/CRC machine learning \& pattern recognition series.
  CRC Press, 2012.

\bibitem{zhu2006multi}
Ji~Zhu, Saharon Rosset, Hui Zou, and Trevor Hastie.
\newblock Multi-class adaboost.
\newblock {\em Ann Arbor}, 1001(48109):1612, 2006.

\end{thebibliography}
 }
\end{document}